%% file: main.tex
\documentclass{article} %
\usepackage{iclr2026_conference,times}

\input{iclr2026/math_commands}

\usepackage{amssymb}
\usepackage{hyperref}
\usepackage{url}
\usepackage{graphicx}
\usepackage{subcaption}
\usepackage{booktabs}
\usepackage{multirow}
\usepackage{xcolor}
\usepackage{colortbl}
\usepackage{tcolorbox}
\tcbuselibrary{breakable}
\usepackage{fancyvrb}
\usepackage{adjustbox}
\usepackage{makecell}
\usepackage{flafter} %
\newtcolorbox{hypothesisbox}{
  colback=blue!3,
  colframe=blue!50!black,
  title=Central Hypothesis,
  fonttitle=\bfseries,
  boxrule=0.8pt,
  arc=2mm
}

\definecolor{patА}{RGB}{230,230,236}  %
\definecolor{patB}{RGB}{229,239,229}  %
\definecolor{patC}{RGB}{249,247,224}  %
\definecolor{patD}{RGB}{250,235,221}  %

\title{Language model harnesses can be \mbox{Compositional Generalizers}}

\author{Alex L. Zhang\thanks{Correspondence: \texttt{altzhang@mit.edu}. \quad Code: \texttt{\href{https://github.com/alexzhang13/harness-cg}{github.com/alexzhang13/harness-cg}}}, Souradip Chakraborty \& Omar Khattab\\[0.7em]
{\normalfont\mdseries MIT CSAIL}}

\makeatletter
\def\@maketitle{\vbox{\hsize\textwidth
  \def\@makefnmark{\hbox{\@textsuperscript{\normalfont\@thefnmark}}}%
  \setcounter{footnote}{1}%
  \centering
  {\LARGE\sc \@title\par}%
  \vskip 1.25em
  {\bf
   \begin{tabular}[t]{c}%
     \@author
   \end{tabular}\par}%
  \ificlrfinal
    \lhead{Published as a conference paper at ICLR 2026}%
  \else
    \lhead{Under review as a conference paper at ICLR 2026}%
  \fi
  \vskip 0.35in minus 0.1in}}
\makeatother

\newcommand{\qwen}{\texttt{Qwen3-30B-A3B-Instruct-2507}}
\newcommand{\qwennew}{\texttt{Qwen3.8-27B}}

\iclrfinalcopy %
\begin{document}

\maketitle

\begin{abstract}
Modern language models (LMs) are post-trained and deployed as agents inside a ``harness'', but standard harness designs accumulate context that the neural network is forced to generalize to, making training for long-horizon tasks expensive and time-consuming. We study an alternative emerging path: training inherently neurosymbolic LM harnesses, which externalize and manage context programmatically to compose existing LM capabilities. We hypothesize that training around these harnesses can offer a better return on compute by enabling \textit{compositional generalization}, as they can learn to decompose complex problems so that every observation their neural LM receives is locally in-distribution, even when the full task is novel. We find that harnesses like the Recursive Language Model (RLM) learn such decompositions readily under reinforcement learning post-training: (1) we demonstrate that training only on short tasks can generalize to tasks 2--64$\times$ longer while the base neural LM improves less, and (2) training on one domain transfers to novel domains that share latent structure. On messier, complex tasks, we find that we can target individual LLM calls within a harness by generating and training on easier, synthetically generated surrogate environments with \textit{fake tool / sub-calls}, a technique we call \textit{surrogate harness training}. From this finding, we train \textbf{RLM-27B} on synthetic data, and show it improves Qwen3.8-27B in the same RLM harness on a hard suite of long-context tasks like CTC-Bench and ExpertLongBench by 26.9\% on average while using $\sim$2$\times$ fewer tokens at inference time. Finally, we show that harnesses can compose over these newly learned decompositions using recursive sub-harness calls.

\end{abstract}

\section{Introduction} \label{sec:intro}
\input{sections/introduction}

\section{Harnesses reduce generalization burden on language models} \label{sec:motivation}

\input{sections/mgh}

\section{Composing with existing LM capabilities to generalize} \label{sec:exp}
\input{sections/experiments}

\section{Compositional generalization enables localized training} \label{sec:sim}
\input{sections/similarity}

\section{Related Work} \label{sec:related}
\input{sections/related}

\section{Conclusion} \label{sec:conclusion}
\input{sections/conclusion}

\section{Acknowledgments} \label{sec:ack}
\input{sections/acknowledgments.tex}

\bibliography{bibliography}
\bibliographystyle{iclr2026/iclr2026_conference}

\newpage

\appendix
\input{sections/appendix}

\end{document}

%% file: iclr2026/math_commands.tex
\usepackage{amsmath,amsfonts,bm}

\def\eqref#1{equation~\ref{#1}}

\def\1{\bm{1}}

\DeclareMathAlphabet{\mathsfit}{\encodingdefault}{\sfdefault}{m}{sl}
\SetMathAlphabet{\mathsfit}{bold}{\encodingdefault}{\sfdefault}{bx}{n}

%% file: sections/introduction.tex
\begin{figure}[!hbt]
    \centering
    \includegraphics[width=0.95\linewidth]{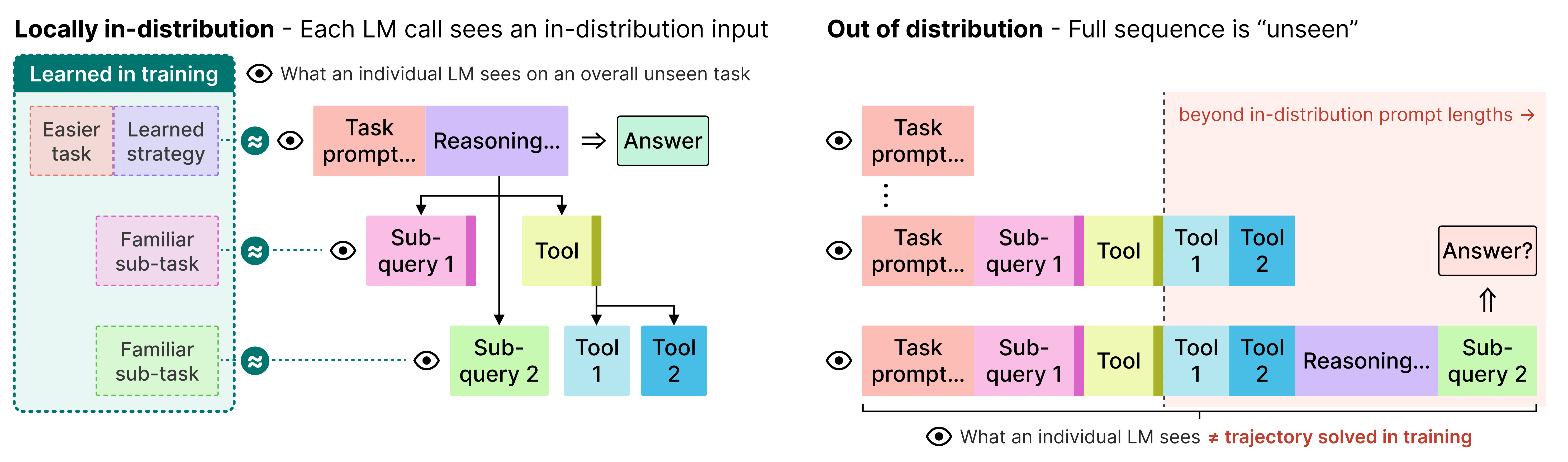}
    \caption{Harnesses like Recursive Language Models (RLM) can enable compositional generalization on unseen tasks by \textit{shaping prompts for each individual language model call} to be in-distribution, a property we call \textit{locally-in-distribution} (LID). \textit{``Trajectory-as-a-prompt''} harnesses like ReAct rely on a root language model to generalize its policy to a growing trajectory as in-context prompts, frequently leading to erroneous behavior on out-of-distribution (OOD) prompts. }
    \label{fig:difference}
\end{figure}

Many real world problems admit natural decompositions into smaller sub-problems with optimal substructure~\citep{Simon1962-SIMAOC, elmoznino2026defininggeneratingcompositionalstructure}, making compositional algorithms like MapReduce~\citep{10.1145/1327452.1327492} prevalent in computer systems. However, in modern language models (LMs), despite being able to solve superhuman tasks~\citep{OpenAIManuscript2026}, it has been empirically shown that they are still unreliable~\citep{anil2022exploring, dziri2023faith, khandelwal2026how, motwani2026longcot} at \textit{compositional generalization}: the ability to solve unseen problems by composing familiar ones. Instead, for the past few years we have tackled harder and longer tasks~\citep{jimenez2024swebenchlanguagemodelsresolve, yang2026programbenchlanguagemodelsrebuild} by \textit{scaffolding} the neural language model with multi-turn agent harnesses and tools~\citep{schick2023toolformer}. These harnesses are used in real-world problems to equip LMs with inductive biases to decompose problems into manageable subproblems~\citep{yang2024sweagent, anthropic_claude_code}, and avoid LM limitations like context rot~\citep{hong2025context}. 

However, the capabilities learned through training modern LM harnesses still do not necessarily compose well. The issue with these \textbf{\textit{``trajectory-as-a-prompt''}} language model harnesses derived from ReAct~\citep{yao2023react} -- which provide the entire agent history in-context to an \textit{orchestrator} (i.e., root) LM to produce their next action -- is that the LM within the harness still relies on generalization through the underlying Transformer~\citep{vaswani2023attentionneed}: the orchestrator LM must learn the correct policy at every harness step given the full trajectory up to that point as its input prompt. Therefore, learning the desired behavior requires curating environments using the harness and training on full rollouts -- e.g., a common practice is to post-train an LM using a coding agent~\citep{yang2024sweagent} on difficult coding benchmarks~\citep{yang2026programbenchlanguagemodelsrebuild, huang2026deepswemeasuringfrontiercoding}, which struggles to generalize to a new harness or new environment. We argue that unless our systems compose the individual lessons they learn, scaling will have slower returns than it should, as every new domain will demand its own training data to become ``in-distribution'', even when its constituent subproblems are already within the LM's capabilities (\S\ref{sec:motivation}).

In this paper, we study compositional generalization of language agents from the perspective of decomposition of information at the \textit{text-level}, rather than at the \textit{neural level}~\citep{olsson2022context, doi:10.1073/pnas.2417182122}. In particular, we study whether harnesses can shape the information propagated to each LM call so that unseen tasks reuse familiar local computations, a property we call \textit{\textbf{locally in-distribution}}\footnote{Locally in-distribution refers to the idea that each individual LM call sees a familiar, solvable input, not necessarily that it was trained on each input explicitly. LMs implicitly generalize on natural language data, so we colloquially use \textit{in-distribution} to refer to prompts that language models can reasonably respond to.} (LID; see Figure~\ref{fig:difference}). We examine this property in relation to common \textit{harness abstractions}, such as context offloading, programmatic tool calling (PTC)~\citep{wang2024executablecodeactionselicit}, and recursive sub-agent calling. For example, on dense data processing tasks~\citep{bertsch2025oolongevaluatinglongcontext}, we observe that harnesses with PTC and sub-agents can invoke batches of sub-LM calls to label short pieces of context accurately; on longer, unseen variants of the task, the harness can reuse this decomposition and repeat these sub-LM calls (\S\ref{sec:exp}). We find that training around the harness enables it to take part in generalization: the root LM reuses learned decompositions across different tasks, and its sub-calls reuse existing capabilities on local sub-tasks, even when the overall task is unseen.

Empirically, we find that reinforcement learning (RL) training~\citep{shao2024deepseekmathpushinglimitsmathematical} a language model with or without a harness improves on training, but a proper harness can better generalize across length and domain. In particular, we observe that the Recursive Language Model (RLM)~\citep{zhang2026recursivelanguagemodels} harness can generalize to unseen tasks 2--64$\times$ longer than tasks seen during training (\S\ref{subsec:length-gen}), as well as unseen tasks in novel domains that share a similar solution structure (\S\ref{subsec:strategy-gen}). We also observe that harnesses like CodeAct~\citep{wang2024executablecodeactionselicit} and Pi~\citep{zechner2026pimono} exhibit similar generalization capabilities on certain programmatic tasks over the base Transformer, but struggle when these solutions require reasoning densely~\citep{singhal2026freelunchcorpustask} over many sub-tasks.

These findings suggest a strategy for improving language model capabilities on difficult tasks with long rollouts: we can target the behavior of individual LMs within a harness, which compose with their existing capabilities to solve the real task. We call this strategy \textit{\textbf{surrogate harness training}}: to train a desired LM's behavior within a harness, we construct easier, synthetic environments (e.g. with a fake harness, tool calls, or sub-calls) that reproduce what it would observe in the harness on a real task. We select eight difficult benchmarks across different decomposition strategies (e.g. retrieval, ranking, dense MapReduce-style tasks) (\S\ref{sec:sim}), and show that training on these surrogate environments yields \textbf{RLM-27B}, improving on this suite by 26.9\% on average in the same harness while using 3$\times$ fewer turns, and 2$\times$ fewer tokens (\S\ref{sec:synth-envs}). Finally, we show that harness-level recursion enables composing over newly learned root-level orchestration behavior, so an entire learned solver becomes a reusable component of a larger task (\S\ref{subset:compgen}).

%% file: sections/mgh.tex
It is well known that the generalization capabilities of even frontier language models are limited and not well understood~\citep{anil2022exploring, dziri2023faith, liu2023transformers, wu-etal-2024-reasoning, motwani2026longcot}. Even semantically identical but token-wise different inputs can lead to vastly different output distributions~\citep{sclar2024quantifying, chatterjee-etal-2024-posix, romanou2026brittlebenchquantifyingllmrobustness}, and empirically, we observe that model behavior degrades as a function of input context length, even when the individual tasks are solvable in isolation by the model~\citep{li2026fragilitytaskscompositionimplicit}. In agentic, multi-turn settings, these issues are exacerbated in ``trajectory-as-a-prompt’’ harnesses like ReAct, which keep a running append-only context that the model is prompted with every turn. As the agent history grows and the language model is prompted with increasingly convoluted and out-of-distribution trajectory histories, we observe undesirable behavior like context rot~\citep{hong2025context}.

A simple explanation for this phenomenon is the lack of guarantees around the training process for language models, which assumes that the reward maximizing policies over our training data and evaluation environments are nearly the same. Standard post-training algorithms like RL rely on the neural network's learned representations to bridge this gap and generalize. Our hypothesis, in line with the Mismanaged Geniuses Hypothesis~\citep{zhang2026mismanaged}, is that a suitable harness can reduce this generalization burden by transforming globally out-of-distribution tasks into sets of locally in-distribution language model computations.

\begin{figure}[!hbt]
    \centering
    \includegraphics[width=\linewidth]{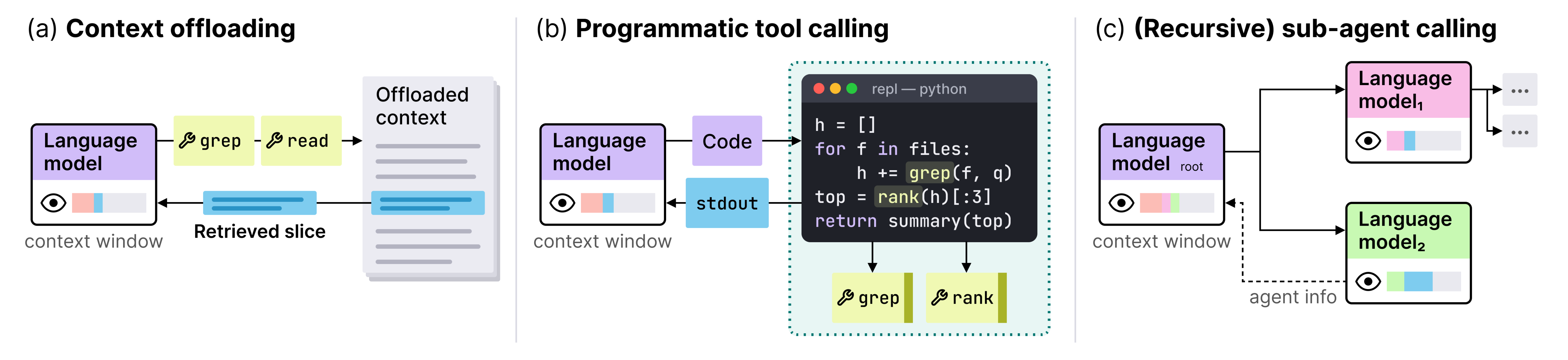}
    \caption{Several popular harness abstractions are useful for controlling what individual LM calls observe. \textbf{(a) Context offloading} keeps the full context in the environment and only allows the model to see slices. \textbf{(b) Programmatic tool calling} treats tools (e.g. sub-agents) as functions in code, which can manipulate intermediate state without the model needing to see it. \textbf{(c) Sub-agent calling} allows the model to explicitly fan out to other LMs, each with their own fresh context. These abstractions can be composed, yielding stronger task generalization effects.}
    \label{fig:harness-abstractions}
\end{figure}

We motivate and study useful design principles behind harnesses around this hypothesis. An alternate view of neural language models is to treat them as probabilistic functions, in which harnesses provide a mechanism for composing these functions around inputs and outputs they are capable of reliably solving in isolation. Under this view, recent harness abstractions in popular coding agents (Figure~\ref{fig:harness-abstractions}) significantly influence what individual LLM calls observe when solving unseen tasks.

In our experiments, we compare the implicit learning behavior of well-understood instantiations of these harness abstractions, as well as the base neural language model, to directly compare how harness design choices can lift generalization pressure off the Transformer. We choose each method based on the minimal abstractions they provide and opt out of task-specific harness designs, which are shown to yield negligible differences in overall system capability~\citep{pan2026harnesstax}. %

\textbf{Base Transformer / ``Trajectory as a Prompt''.} We choose the base Transformer model to represent the ``trajectory-as-a-prompt’’ harnesses that rely on the Transformer’s ability to generalize across both length and task structure, as they directly condition on all context to influence their actions. %

\textbf{Pi with sub-agents and context offloading.} The Pi-mono~\citep{zechner2026pimono} harness encapsulates the ReAct~\citep{yao2023react} loop with terminal, compaction, and sub-agent~\citep{zechner2026pimono}  tools used in most modern coding agents~\citep{anthropic_claude_code, openai_codex}. Additionally, we offload long context into a file that the harness interacts with.

\textbf{CodeAct / LM + REPL with context offloading.} CodeAct~\citep{wang2024executablecodeactionselicit} harnesses use programmatic tool calling (PTC) over JSON-style tool calling~\citep{schick2023toolformer}, meaning all other tools are functions within a code REPL environment. This harness is implemented as an RLM without programmatic sub-agent calling or output truncation to ablate the value of sub-calls.

\textbf{RLM.} The Recursive Language Model~\citep{zhang2026recursivelanguagemodels} is a class of harnesses that rely on context offloading and programmatic sub-agent calling to act over context. Following the original paper and other popular instantiations such as Ax~\citep{dosco2026ax} and Prime Agent~\citep{karten2026primeagentselfimprovingrlm}, we choose depth=1 sub-agents, as it is directly comparable with coding agents that invoke sub-agents, and it is equally as expressive as higher depth RLMs.

%% file: sections/experiments.tex
In long-context settings, a commonly observed problem is that language models can solve small instances of a task, but struggle on longer variants as the task length grows~\citep{li2026fragilitytaskscompositionimplicit}. In general settings, another observable issue is that models struggle to solve tasks with similar task \textit{structure} but different token-wise descriptions with respect to a task learned during training. 

We begin by studying how LM harnesses can be trained to compose their existing capabilities to induce generalization across (1) different task lengths and (2) different task domains. For the following experiments, we train Pi on both the root and the sub-agents, while for the RLM we apply loss only to the root trajectory, analogous to Recursive Agent Optimization~\citep{gandhi2026recursiveagentoptimization} with no sub-call delegation bonus. Training configuration details are in Appendix~\ref{appx:training-details}.

\input{tables/length-gen.tex}

\subsection{Length Generalization in harnesses through sub-task composition} \label{subsec:length-gen}

In \citet{anil2022exploring}, a key result is that when training on a particular task, it is often out-of-distribution for a language model to solve the same task at longer context lengths. In \citet{uzunoglu2026informationabundanceparadoxlongcontext}, the authors observe that directly training on long-context tasks can negatively impact the performance of Transformers on already learned tasks. To mitigate both findings, we are interested in whether the harness abstractions in Figure~\ref{fig:harness-abstractions} enable learning composable behavior by training on only short tasks that generalizes to longer tasks.

On a suite of tasks in Table~\ref{tab:benchmarks} where context length can be varied, we train our harnesses with \qwen{}~\citep{yang2025qwen3technicalreport} using GRPO~\citep{shao2024deepseekmathpushinglimitsmathematical} on the short splits only, and evaluate on longer splits with per-environment ranges listed in Table~\ref{tab:benchmarks} (for training details, see Appendix~\ref{appx:training-details}). For tasks where the evaluation split does not fit in the context window of the model, we apply YaRN length extension~\citep{peng2026yarnefficientcontextwindow} to the base Transformer.

\begin{figure}[htbp]
    \centering
    \includegraphics[width=1\linewidth]{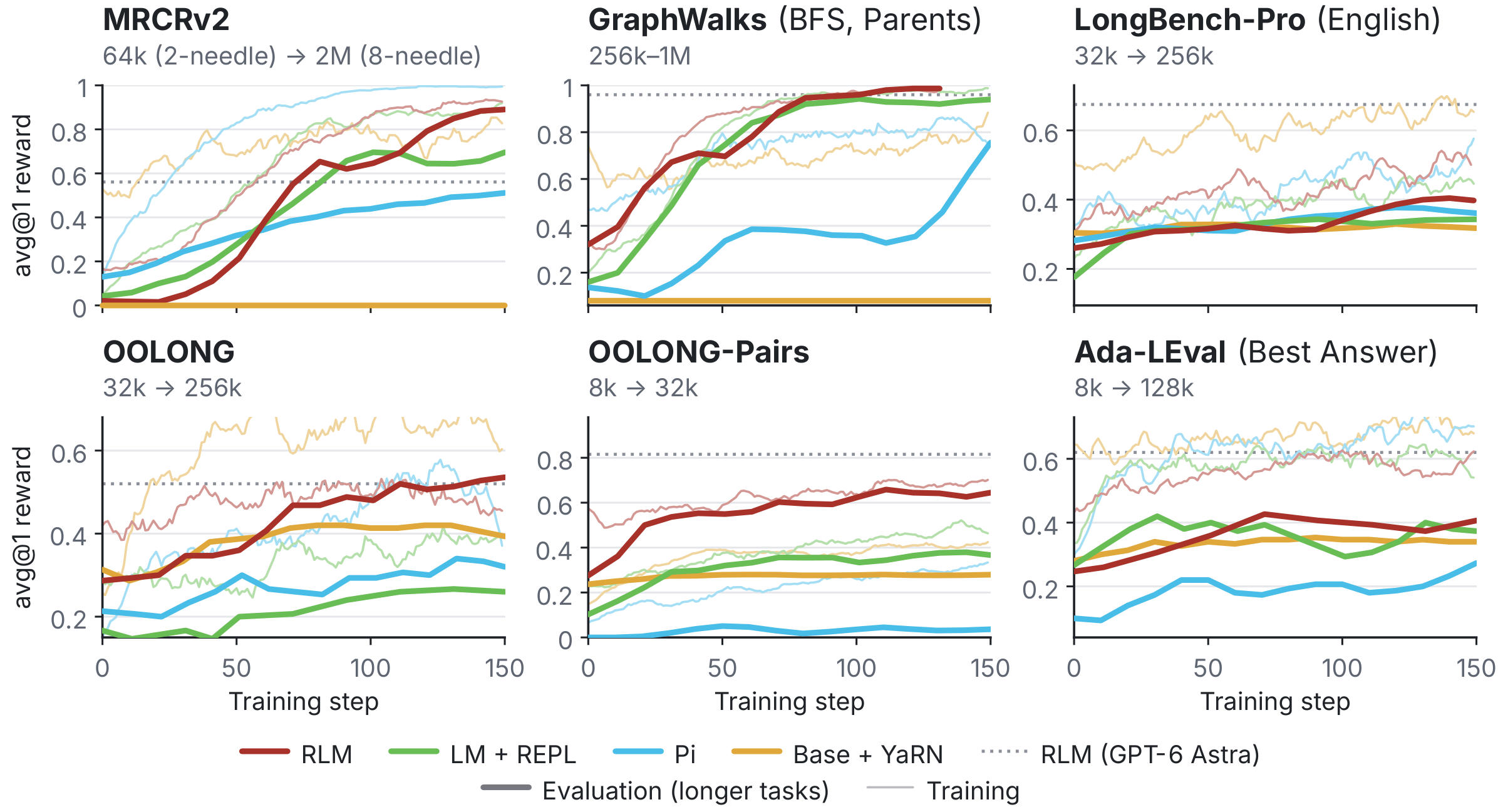}
    \caption{On six environments with controlled context lengths, we train several harnesses and the base Transformer on short tasks, and evaluate on longer variants. We plot their train reward (smoothing window=3) and evaluation reward (smoothing window=11) over training, and include a frontier model, GPT-6 Astra~\citep{openai2026gpt6astra_systemcard}, using the RLM as a reference. The evaluation reward of the RLM closely follows the trend of the training reward, while for the base Transformer it does not.}%
    \label{fig:length-gen}
\end{figure}

We report our findings in Figure~\ref{fig:length-gen}, where we find that the evaluation reward of the RLM on longer splits better matches the training reward, which often nearly solves the length-extrapolated task.

\textbf{Context offloading with the harness generalizes better to longer tasks.} Each harness often generalizes better than the base neural language model, which struggles to generalize to longer tasks by only training on shorter tasks. Notably, the train reward on short tasks is often higher for the neural language model than other harnesses, while the train reward for the RLM grows slower, but closely matches its reward on the longer evaluation tasks. Between the harness abstractions, the RLM is able to successfully lift performance or perform better on each task than the other harness variants due to the ability to fan out sub-agents over an offloaded context, and act efficiently through LM programs as actions. Furthermore, when observing the root trajectories of these harnesses on long tasks, we find that they share strong similarities to successful trajectories generated during training, reflecting that longer tasks are viewed as \textit{locally in-distribution} (see examples in Appendix~\ref{train-examples}). %

\textbf{When code is not sufficient, RLMs compose their sub-call capabilities to length generalize via LID.} We observe length generalization occurs when the harness learns a pattern at length $N$ that also applies for tasks at length $M >> N$. For example, in settings like MRCRv2 and GraphWalks, harnesses with programmatic capabilities like CodeAct generalize well over the purely neural baseline by writing the code that solves both the short and long tasks. However, on tasks like OOLONG and OOLONG-Pairs without a clear programmatic solution, harnesses must compose the capabilities of their sub-agents over in-distribution sub-tasks with their learned decomposition to generalize.

\subsection{Strategy Generalization in harnesses through sub-task composition} \label{subsec:strategy-gen}

The length generalization effect from training the RLM harness can be understood as the unseen, longer tasks being \textit{locally in-distribution} with 1) what the root LM learns during training; 2) what the LM already can do on sub-tasks invoked through the harness. %

This principle is agnostic to the axis of context length and applies generally when the root LLM in a harness learns a high-level strategy that solves tasks across different domains, by deferring domain-specific sub-tasks to sub-calls. We call this observation, \textit{strategy generalization}, in which training on one domain transfers to another with similar structure. We use the same setup as with length generalization but train for more steps, with the environments chosen in Table~\ref{tab:strategy-benchmarks}.

\input{tables/strategy-gen}

\begin{figure}[htbp]
    \centering
    \includegraphics[width=1\linewidth]{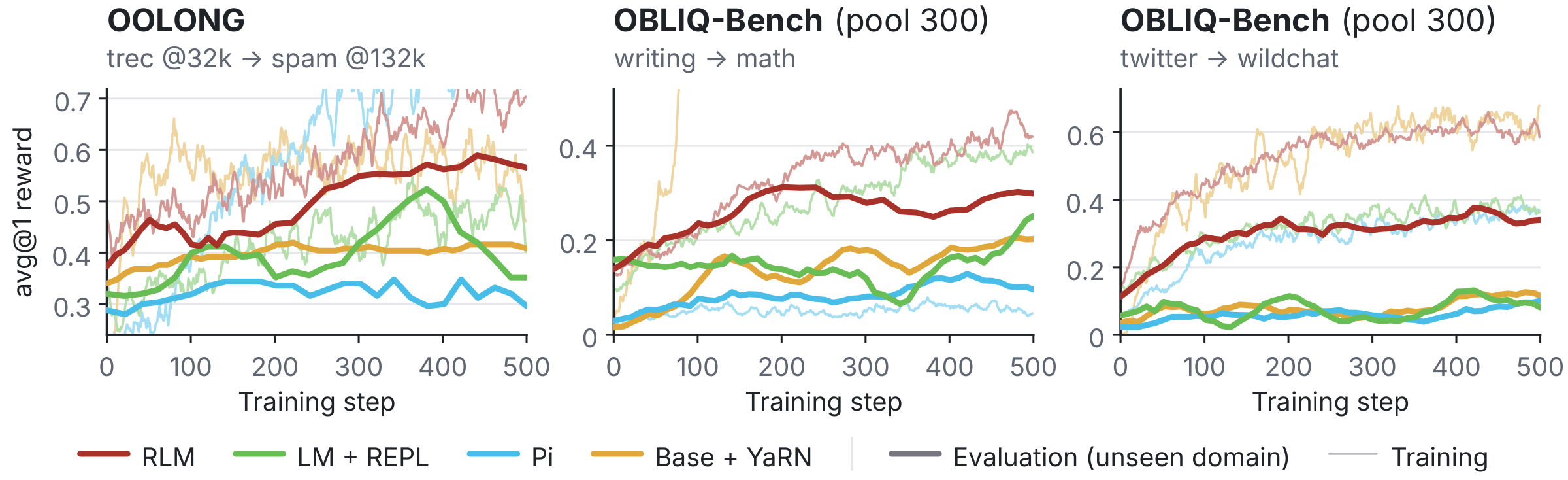}
    \caption{We choose three environments with splits that are completely disjoint in domain (e.g. math vs. essay writing questions), but share a similar decomposition structure (e.g. ranking). We train several harnesses and the base Transformer (\qwen{}) on one domain, and evaluate on the other, where the RLM generalizes the strongest among other harness abstractions and the base model.}
    \label{fig:strategy_grid}
\end{figure}

\textbf{Strategy generalization through compositional generalization follows similar harness trends to length generalization.} In each of the three environments, the token description of each task greatly differs from its evaluation counterparts, leading to poor generalization capabilities of the \textit{base Transformer} despite its train reward growing similarly to harnesses that offload context. Pi struggles on these tasks due to their dense, long-context processing requirements~\citep{singhal2026freelunchcorpustask}, with its train reward also being low on OBLIQ-Bench. The LM + REPL harness similarly struggles when context processing is dense. The RLM is able to generalize the best across all three tasks, with its growing evaluation reward trending similarly to its train reward over time.

\subsection{Characterizing Locally In-distribution Trajectories} \label{subsec:characterize-local}

The training experiments in \S\ref{subsec:length-gen}, \S\ref{subsec:strategy-gen}, and the strong evaluation rewards of certain harnesses establish that some learned harness behavior transfers to unseen tasks. In this section, we assess whether these observed generalization effects can be attributed to locally in-distribution trajectories through two complementary diagnostics: reuse of root-level orchestration trajectories and reliable execution of delegated sub-tasks. 

To characterize the reused root orchestration behavior that enabled this compositional generalization effect, we compare the root trajectories produced by each harness during training and evaluation using token-wise Weighted Jaccard similarity (see Appendix~\ref{appx:similarity-plots}). To characterize composable sub-task behavior that is ``in-distribution'', we examine whether the models solve the sub-tasks reliably, rather than tracing whether the exact sub-task prompt appears in the model's training distribution.

\textbf{Context offloading yields higher token similarity across trajectories.} In practice, many tasks share similar solutions while containing extremely different contexts, e.g., with respect to context length and task domain. We find in Figure~\ref{fig:similarity} in Appendix~\ref{appx:similarity-plots} that on the experiments in Figures~\ref{fig:length-gen} and~\ref{fig:strategy_grid}, context-offloading harnesses share significantly higher similarity across multiple token-wise metrics on most evaluation tasks to some successful trajectory seen during training.

\begin{figure}[!hbt]
    \centering
    \includegraphics[width=1\linewidth]{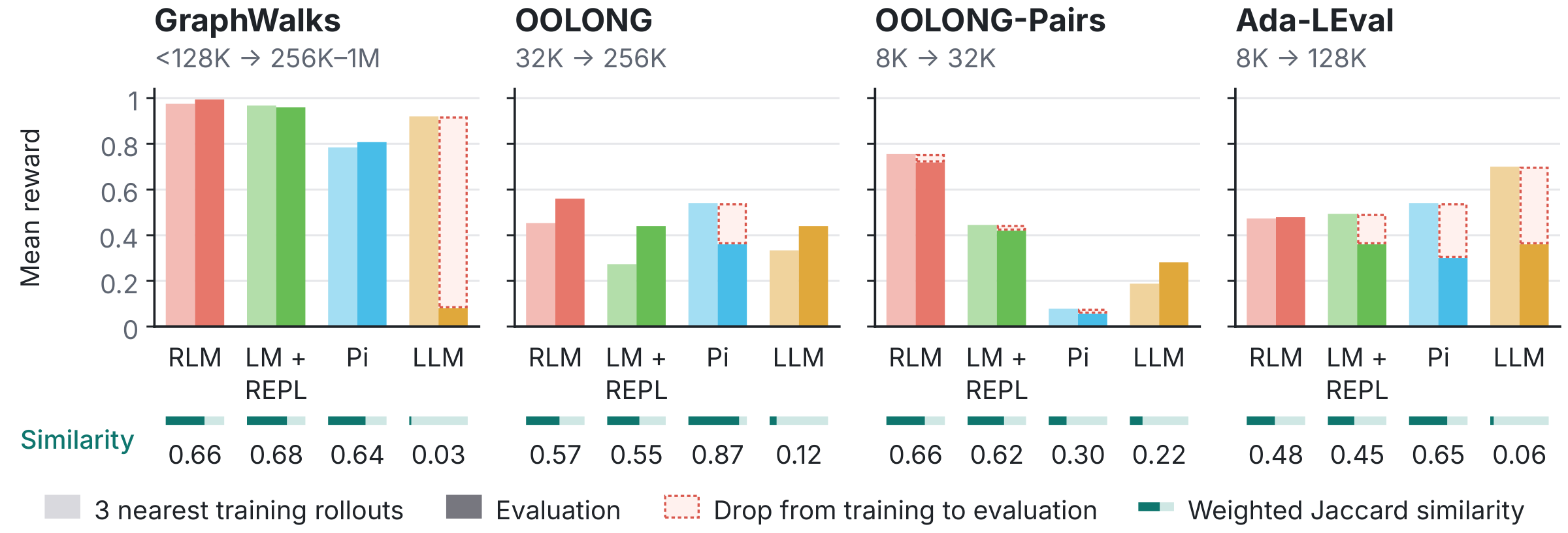}
    \caption{\textcolor{black}{For several harnesses, we train on short tasks and evaluate on their long-context variants. For each setting, we track the nearest three training trajectories by Weighted Jaccard Similarity to each long evaluation trajectory, and average their rewards. We find that context-offloading harnesses learn root-level trajectories that are re-used in longer tasks compared to ``trajectory-as-a-prompt'', but training performance does not necessarily transfer for all harnesses, like Pi on OOLONG. }} %
    \label{fig:sim-results}
\end{figure}

\textbf{Token-wise similarity of root trajectories is not enough for task transfer.} In Figure~\ref{fig:sim-results}, we observe that for the RLM and CodeAct harnesses, reward on training and evaluation tasks is similar for tasks that share similar root-level trajectories. However, for Pi, we observe significant degradation in transfer performance on non-programmatic tasks because many strategies it learns at the short length, such as reading fixed subsets of the data with sub-agents or writing out long intermediate lists by hand, do not generalize well for longer context. On programmatic tasks like GraphWalks, we find that all context-offloading harnesses produce length-invariant compact programs in Python / bash over the offloaded context that generalize. However, on tasks like OOLONG and Ada-LEval with semantic sub-tasks, the RLM exhibits the strongest transfer in reward between evaluation tasks and similar training tasks. We provide specific examples of root-level trajectories in Appendix~\ref{appx:lid-trajectories}.

\textcolor{black}{\textbf{Verifying local sub-task reliability.} The observation that these harnesses are producing locally in-distribution trajectories relies on the delegated sub-task LLM invocations being correct. For tasks that contain intermediate sub-task ground truth labels like OOLONG, we find that sub-task accuracy is similarly high across training and evaluation tasks (see Table~\ref{tab:subtask-verification}, Appendix~\ref{appx:subtask-verification}), supporting the reuse of reliable local computations as the overall task changes.}

%% file: tables/length-gen.tex
\begin{table}[!htbp]
\centering
\small
\setlength{\tabcolsep}{4pt}
\renewcommand{\arraystretch}{1.2}
\begin{tabular*}{\textwidth}{@{\extracolsep{\fill}}llll@{}}
\toprule
\textbf{Environment} & \textbf{Train Tasks} & \textbf{Evaluation Tasks} & \textbf{Length $\times$} \\
\midrule
MRCRv2~\citep{vodrahalli2024michelangelolongcontextevaluations} & 32K--64K tokens, 2-needle & 2M--4M tokens, 8-needle & 64$\times$ \\
{\color{black}GraphWalks~\citep{openai2025graphwalks}} & {\color{black}$\leq$128K tokens} & {\color{black}50 samples, 256K--1M tokens} & {\color{black}2--8$\times$} \\
LongBenchPro~\citep{chen2026longbenchprorealisticcomprehensive} & 32K tokens, English & 256K tokens, English & 8$\times$ \\
OOLONG~\citep{bertsch2025oolongevaluatinglongcontext} & 32K split, \texttt{trec-coarse} & 256K split, \texttt{trec-coarse} & 8$\times$ \\
OOLONG-Pairs~\citep{zhang2026recursivelanguagemodels} & 8K input tokens & 32K tokens & 4$\times$ \\
Ada-LEval~\citep{wang2024adalevalevaluatinglongcontextllms} & 8K tokens & 128K tokens & 16$\times$ \\
\bottomrule
\end{tabular*}
\caption{Training and evaluation environments, where the evaluation splits are longer variants of the training split within the same task domain.}
\label{tab:benchmarks}
\end{table}

%% file: tables/strategy-gen.tex
\begin{table}[!htbp]
\centering
\small
\setlength{\tabcolsep}{4pt}
\renewcommand{\arraystretch}{1.2}
\begin{tabular*}{\textwidth}{@{\extracolsep{\fill}}llll@{}}
\toprule
\textbf{Environment} & \textbf{Train Tasks} & \textbf{Evaluation Tasks} \\
\midrule
OOLONG~\citep{bertsch2025oolongevaluatinglongcontext} & trec-coarse questions, 32K tokens & spam questions, 132K tokens \\
OBLIQ-Bench~\citep{tchuindjo2026obliqbenchexposingoverlookedbottlenecks} & Writing-style analogues (300 docs) & Math analogues (300 docs) \\
OBLIQ-Bench~\citep{tchuindjo2026obliqbenchexposingoverlookedbottlenecks} & Twitter stances (300 docs) & LLM transcript errors (300 docs) \\
\bottomrule
\end{tabular*}
\caption{Training and evaluation environments that do not share domain-specific context, but share similar decomposition strategies.}
\label{tab:strategy-benchmarks}
\end{table}

%% file: sections/similarity.tex
In \S\ref{sec:exp}, we motivated how harnesses that induce locally in-distribution trajectories like the RLM can generalize across length and task strategy by composing their existing capabilities in sub-calls with their newly learned task decomposition capability. We showed that this root-level capability can be learned through RL on an easier, similar task that shares decompositional structure. In this section, we explore how this compositional generalization effect can be utilized as a data-efficient training strategy in settings where either similar, easier tasks do not exist, or the \textit{correct} decomposition strategy does not emerge naturally through RL.

We choose a selection of modern long benchmarks in Table~\ref{tab:eval-envs} that are either disjoint from or more difficult variants of environments from our prior analysis, where a certain axis of transferrable behavior is not explicit.

\input{tables/eval-envs}

\subsection{RL post-training can yield non-generalizing harness behavior} \label{sec:4.1}

Following our experiments in \S\ref{sec:exp}, we explore whether the learned behavior from RL training the RLM harness on tasks in Table~\ref{tab:benchmarks} transfers to more difficult long tasks in Table~\ref{tab:eval-envs}, none of which directly share decompositional structure with the training tasks. We post-train the \qwen{} and \qwennew{}~\citep{qwen38} models to examine the gains in learned behavior between a weaker and stronger base model. We report the Base Transformer (+ YaRN~\citep{peng2026yarnefficientcontextwindow}) and Pi as reference baselines as well.

\input{tables/main-composition}

\textbf{Harness behavior learned through RL training is not necessarily compositional.} In Table~\ref{tab:main-composition}, we find that RL over our short tasks in Table~\ref{tab:benchmarks} raises the \qwen{} average by 7 points, while for \qwennew{} it's only 2.4 points. At the same time, we find in Table~\ref{tab:main-composition-stats} that token usage decreases for the former but increases for the latter. The behavior learned via RL is highly dependent on the initial policy: the weaker \qwen{} learns to delegate less to sub-calls and rely on programmatic solutions, while the stronger \qwennew{} learns policies that fan out more reliably to sub-agents. This discrepancy is also evident in the gap between the base Transformer / Pi and the RLM, which for the weaker \qwen{}, produces similar trajectories to the other harnesses, while for the stronger \qwennew{}, it takes advantage of programmatic sub-agents. We also explore in Appendix~\ref{appx:broad-mixture-experiments} whether a broader task suite contains trajectories that compositionally generalize to our evaluation suite, but similarly find the trajectories learned in training transfer weakly.

\input{tables/main-composition-stats}

\subsection{Targeting Individual LM Calls in a Harness with Surrogate Environments} \label{sec:synth-envs}

We previously observed that harnesses like the RLM can be trained to solve long, unseen tasks by strictly learning only the trajectory of one of its LLM calls---in particular, the root---to compose with the LLM's existing capability to solve other sub-tasks. In our experiments, we focused on learning this behavior by training on easier but similar tasks that \textit{shared} the same strategy, so the root LLM would apply the same strategy on the target evaluation task, while composing with its sub-calls' ability to solve the new, but still in-distribution sub-tasks.

\begin{figure}[!hbt]
    \centering
    \includegraphics[width=1\linewidth]{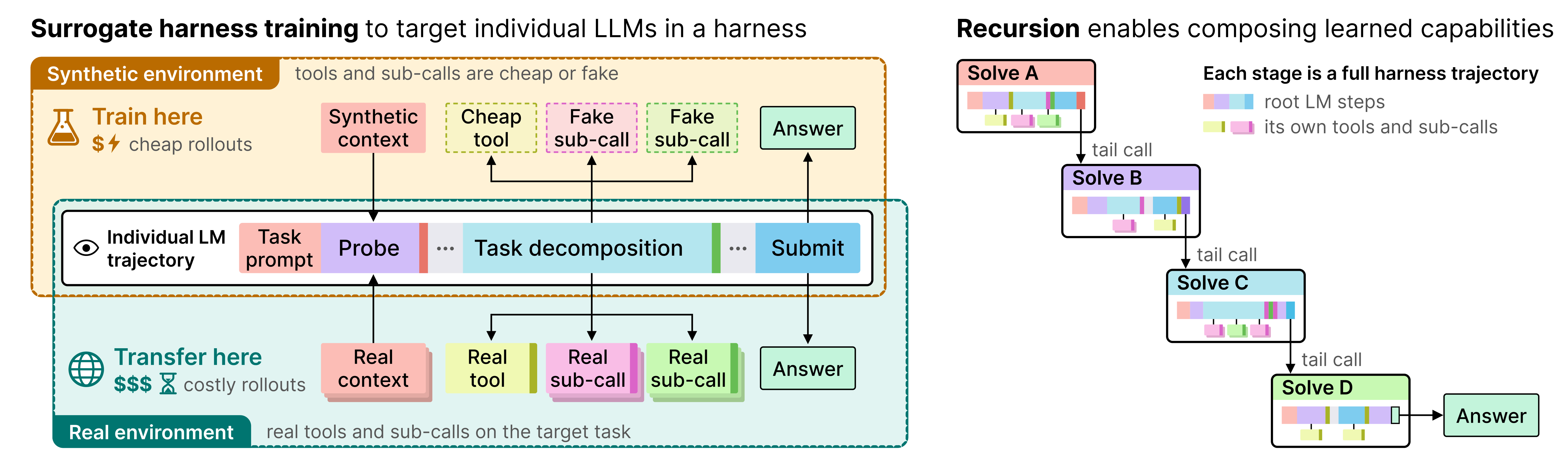}
    \caption{\textcolor{black}{Harnesses that decompose tasks to be locally in-distribution (LID) are composed of individual LLM trajectories. (a) On the left, we describe \textit{surrogate harness training}, a strategy for \textit{training on synthetic environments} to target individual LLM behavior within a harness, using a surrogate synthetic environment that transfers behavior to the real environment when tools and sub-calls are real. (b) On the right, we describe how learned decomposition strategies invoked by the harness can be composed to be LID, by enabling the harness to invoke sub-harness calls recursively.}}
    \label{fig:synth_env_training}
\end{figure}

\textbf{Surrogate harness training.} In practice, on messy and realistic tasks, such training environments do not exist. Furthermore, in \S\ref{sec:4.1}, we found that RL training on the short-task environment mixture may lead to undesirable downstream behavior that degrades performance on some settings. Instead, we use our finding that task composition allows us to target an individual LLM's behavior within a harness, and we construct synthetic training environments that do not share \textit{context} with the real environment, but share a solution that mimics the target LLM's desired behavior (see Figure~\ref{fig:synth_env_training}a). These surrogates are designed with knowledge of the target task families, preserving candidate solution strategies for an individual LLM, while changing task content and, in some cases, simulating tool calls and sub-calls.

\textbf{Generating synthetic environments for learning task decomposition strategies.} For each task environment in Table~\ref{tab:eval-envs}, we provide Opus 5.5~\citep{anthropic2026claudeopus} using Claude Code with a publicly available, separate split of the task environment and data separate from the evaluation tasks, as well as several existing RLM trajectories taken from the base model, and task it with identifying a candidate LM program that can be used to solve the task with \qwennew{}. We then task it with constructing a synthetic environment of tasks whose context does not overlap with the real task, such that this candidate LM program yields a solution to both the synthetic and the real task. We describe each synthetic environment in detail in Appendix~\ref{appx:synthetic-environments}.

\textbf{Training on synthetic environments with \textit{fake} sub-calls.}  In practice, many tasks in the wild have a natural decomposition that uses a large number of (potentially recursive) sub-calls or tool-calls with long wall-clock runtimes. Because we use synthetic environments to target root-only LLM behavior where the context does not overlap with the real task, we hypothesize that the sub-calls can be intentionally low-cost or even fake (see Table~\ref{tab:runtime-rt-rlm} for a comparison). We construct a split of synthetic environments (for BrowseComp-Plus, ExpertLongBench, and OBLIQ-Bench) that replace any invoked sub-calls over sub-tasks with pre-built oracle answers for synthetic sub-tasks.

To generate synthetic traces, we provide the proposed LM program as a hint to the RLM, and ask it to follow the hint to solve the synthetic task. We use GPT-5-mini~\citep{singh2026openaigpt5card} as a judge~\citep{zheng2023judgingllmasajudgemtbenchchatbot} along with the synthetic environment reward as a proxy for whether to accept the generated rollout. We generate a set of \textit{8,015} root-only synthetic RLM rollouts\footnote{\raisebox{-0.2\height}{\includegraphics[height=9.5pt]{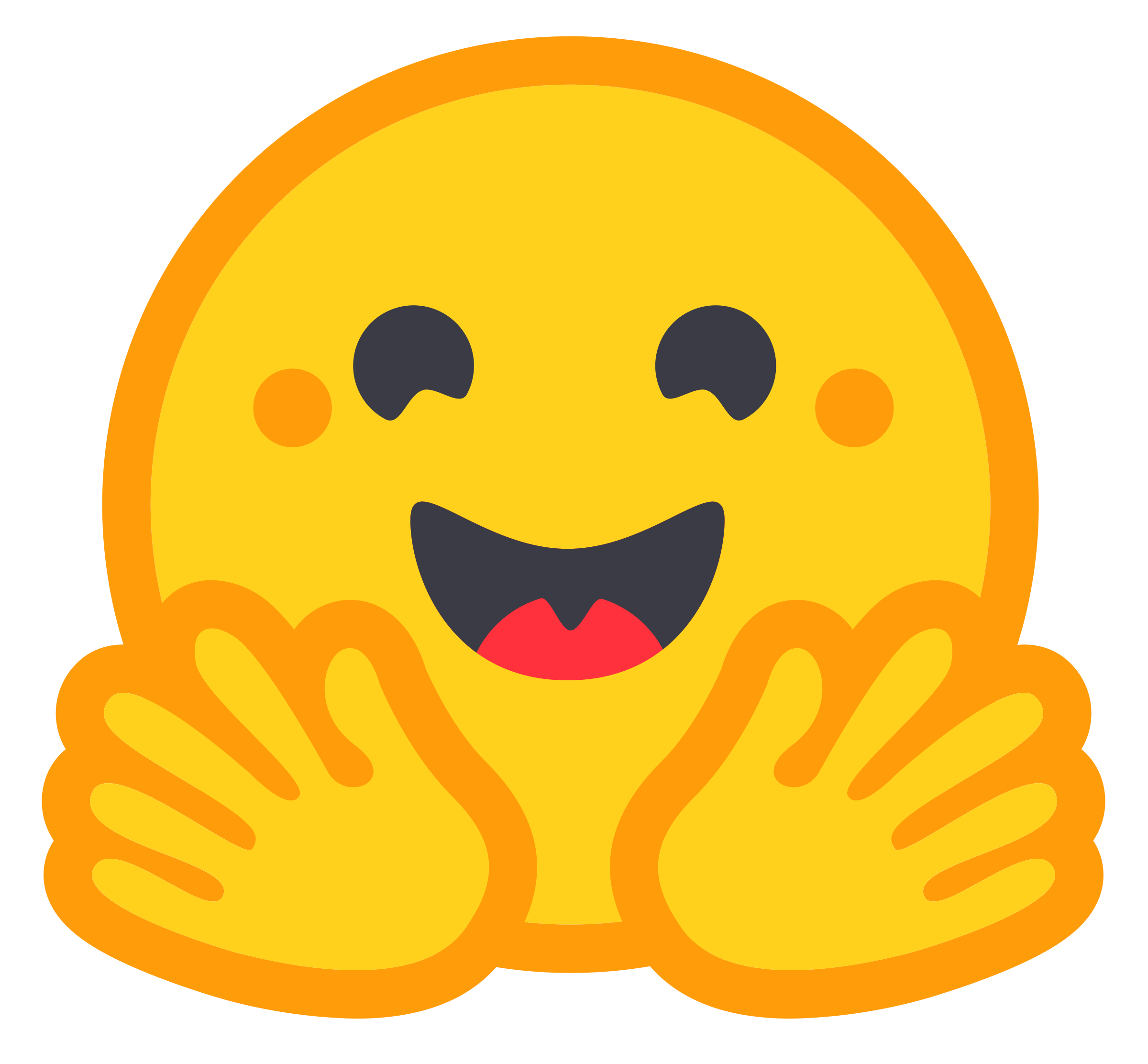}}~SFT traces: \href{https://huggingface.co/datasets/a1zhang/synthetic-rlm-sft-data}{\texttt{huggingface.co/datasets/a1zhang/synthetic-rlm-sft-data}}} (\textit{and 1,465 sub-agent rollouts on LongReason to target specific sub-task behavior}) across these eight synthetic environments and train \textbf{RLM-27B} by supervised fine-tuning (SFT) \qwennew{} on these rollouts (see Appendix~\ref{appx:synthetic-environments}). To test whether the synthetic environments also support learning without explicit hints, we directly RL post-train each model on the synthetic environments (see Appendix~\ref{appx:synth-rl}).

\textbf{Targeted training of an LM in a harness yields more performant and efficient trajectories.} In Table~\ref{tab:main-composition}, using the RLM harness, the SFT-trained \textbf{RLM-27B} improves over the \qwennew{} base on evaluation tasks by 12.4 points on average. Furthermore, in Table~\ref{tab:main-composition-stats} we find that the \textbf{RLM-27B} delegates more often, while using $3 \times$ fewer turns and $6 \times$ fewer input tokens for the root trajectories. Even on tasks like BrowseComp-Plus where raw performance gains are small, the produced trajectories use significantly fewer turns and tokens. Furthermore, we find that RL post-training over the synthetic environments also improves seven of eight tasks and raises the evaluation task average by 6.8 points, but does not yield noticeable efficiency gains, and also converges to different behavior than the SFT data over the same environments, but with teacher-generated hints. %

\subsection{Compositional generalization over learned decompositions via recursion} \label{subset:compgen}
The results in \S\ref{sec:exp}, \S\ref{sec:4.1}, and \S\ref{sec:synth-envs} show that harnesses can learn to shape tasks to be LID by generating LM programs as actions that sufficiently decompose tasks described in offloaded context. These harnesses generalize to novel tasks when the learned LM programs compose with existing LM capabilities over sub-tasks within these tasks. A natural question is whether harnesses can compose \textit{over} these generated LM programs, providing a means to compositionally generalize over existing sub-task solving, as well as newly learned decompositional capabilities. Following our original findings on how harnesses compose LLM capabilities, we can abstract the harness as an individual LLM call that can be recursively invoked by the harness, so each harness invocation solves a learned task with a fresh context. The following experiments do not impose a recursion depth limit. %

We empirically show that recursion enables composition over learned LM programs. From our results in Table~\ref{tab:main-composition}, we select a set of 87 tasks $\mathcal{S}$ where \textbf{RLM-27B} learns to solve them (97.5\% avg@3), but \qwennew{} cannot reliably solve them (57.7\% avg@3). We construct a meta-task in which the harness is given a concatenation of a random ordered set of four unique tasks $T_1,T_2,T_3,T_4$ from $\mathcal{S}$, where the model must solve task $T_{i}$ in order to receive the inputs for the next task $T_{i+1}$. The harness starts with the first task's prompt and context, and submits an answer via the \texttt{unlock(answer)} tool, which returns the new question and task-specific context when answered correctly. The harness is given at most two attempts to solve each task. We average scores over 2000 randomly sampled meta-tasks in Table~\ref{tab:recursion-composition}, and further describe the task in Figure~\ref{fig:gated-chain}.

\input{tables/recursion-composition}

By default, we observe that both \qwennew{} and the trained \textbf{RLM-27B} sequentially solve each task within the same root trajectory, relying on the LLM's ability to generalize to solving the concatenation of several tasks by training only on individual tasks. By relying on compositional generalization from the LM, it solves $55.2\%$ of the concatenated tasks.

\textbf{Learning targeted behavior that composes with existing learned targeted behavior.} In \S\ref{sec:synth-envs}, we showed that we could target individual LLM behavior using synthetic environments. We similarly apply surrogate harness training to learn to tail recurse by solving an individual task $T_{i}$, then launching a sub-RLM over $T_{>i}$ to solve the remaining tasks, which composes with the previously learned capabilities in \textbf{RLM-27B} to individually solve each $T_i$ with a fresh context (see Figure~\ref{fig:synth_env_training}b). For more details on this construction, see Appendix~\ref{appx:recursion-composition}.

\textbf{Recursion enables composition over learned root-level trajectories to be LID.} In Table~\ref{tab:recursion-composition}, we find that recursion over the harness can ease generalization burden off the LLM. However, providing a hint to tail recurse is not sufficient, and actually hurts the generalization performance relative to the LLM itself. We observe that RLM-27B with the additional recursion SFT data at depth=3 is able to solve each $T_i$ with an isolated RLM harness instance, bringing the solve rate up from $55.2\%$ to $77.5\%$ by putting less generalization pressure on the root LLM. %

%% file: tables/eval-envs.tex
 \begin{table}[htb]
  \centering
  \small
  \setlength{\tabcolsep}{4pt}
  \renewcommand{\arraystretch}{1.2}
  \begin{tabular*}{\textwidth}{@{\extracolsep{\fill}}>{\raggedright\arraybackslash}p{0.36\linewidth}>{\raggedright\arraybackslash}p{0.36\linewidth}rr@{}}
  \toprule
  \textbf{Benchmark} & \textbf{Evaluation split} &
  \textbf{Context Length} & $\boldsymbol{n}$ \\
  \midrule
  LongReason~\citep{ling2025longreasonsyntheticlongcontextreasoning}       & All \texttt{128k} items             & 118K  & 794 \\
  OOLONG-Real~\citep{bertsch2025oolongevaluatinglongcontext}      & D\&D test, 170K split. & $\sim$175K & 577 \\
  OBLIQ-Bench~\citep{tchuindjo2026obliqbenchexposingoverlookedbottlenecks}       & Math, full corpus & $\sim$278K & 151 \\
  CorpusQA~\citep{lu2026corpusqa10milliontoken}         & 1M, real estate (English)        & 451K  & 81 \\
  LooGLE-v2~\citep{ziyuanhe2026loogle}        & $\geq$500K                        & 675K  & 206 \\
  BrowseComp-Plus~\citep{chen2025browsecompplusfairtransparentevaluation}  & All queries; full corpus & $\sim$456M & 830 \\
  ExpertLongBench~\citep{ruan2026expertlongbench}  & T01, T03, T04, T06              & 31K   & 350 \\
  CTC-Bench~\citep{singhal2026freelunchcorpustask}        & QDMatch 256K, Outlier 1M, Absence 32K, Grouping 32K, Contradiction 1M               & 32K--1M & 535 \\
  \bottomrule
  \end{tabular*}
  \caption{Held-out evaluation splits. Context lengths are approximate token estimates from the input text; BrowseComp-Plus includes its full shared corpus. CTC-Bench scores are averaged over each split, which each contain a different number of entries.}
  \label{tab:eval-envs}
  \end{table}

%% file: tables/main-composition.tex
\begin{table}[htb]
\centering
\small
\captionsetup{font={color=black}}
\caption{We report average scores for several base and post-trained models using the RLM harness on long tasks in Table~\ref{tab:eval-envs}. For each model, we RL post-train on the short-context variants of the environments studied in \S\ref{subsec:length-gen}, as well as the targeted synthetic environments constructed in \S\ref{sec:synth-envs}.}
\label{tab:main-composition}
\label{fig:main-composition}
\setlength{\tabcolsep}{1.8pt}
\renewcommand{\arraystretch}{1.22}
\setlength{\aboverulesep}{0pt}
\setlength{\belowrulesep}{0pt}
\begin{tabular}{@{}>{\raggedright\arraybackslash}p{\dimexpr\linewidth-.683\linewidth-18\tabcolsep-\arrayrulewidth\relax}
>{\centering\arraybackslash}p{.108\linewidth}
>{\centering\arraybackslash}p{.08\linewidth}
>{\centering\arraybackslash}p{.06\linewidth}
>{\centering\arraybackslash}p{.06\linewidth}
>{\centering\arraybackslash}p{.091\linewidth}
>{\centering\arraybackslash}p{.094\linewidth}
>{\centering\arraybackslash}p{.07\linewidth}
>{\centering\arraybackslash}p{.06\linewidth}
|>{\centering\arraybackslash}p{.06\linewidth}@{}}
\toprule
{\fontsize{8}{9.5}\selectfont\textbf{Method}} & {\fontsize{8}{9.5}\selectfont\textbf{LongReason}} & {\fontsize{8}{9.5}\selectfont\textbf{LooGLE}} & {\fontsize{8}{9.5}\selectfont\textbf{CTC}} & {\fontsize{8}{9.5}\selectfont\textbf{BC+}} & {\fontsize{8}{9.5}\selectfont\textbf{OOLONG}} & {\fontsize{8}{9.5}\selectfont\textbf{CorpusQA}} & {\fontsize{8}{9.5}\selectfont\textbf{OBLIQ}} & {\fontsize{8}{9.5}\selectfont\textbf{ELB}} & {\fontsize{8}{9.5}\selectfont\textbf{Avg.}} \\
\midrule
\multicolumn{10}{@{}l}{\textbf{Qwen3-30B-A3B-Instruct-2507}} \\
\midrule
\textcolor{black}{Base Transformer} & \makebox[2em][r]{63.1}\makebox[.45em][l]{$^{}$} & \makebox[2em][r]{\textbf{34.5}}\makebox[.45em][l]{$^{}$} & \makebox[2em][r]{0.2}\makebox[.45em][l]{$^{}$} & \makebox[2em][r]{0.0}\makebox[.45em][l]{$^{}$} & \makebox[2em][r]{18.7}\makebox[.45em][l]{$^{}$} & \makebox[2em][r]{11.7}\makebox[.45em][l]{$^{}$} & \makebox[2em][r]{0.5}\makebox[.45em][l]{$^{}$} & \makebox[2em][r]{\textbf{52.2}}\makebox[.45em][l]{$^{}$} & \makebox[2em][r]{22.6}\makebox[.45em][l]{$^{}$} \\
Pi (base) & \makebox[2em][r]{72.2}\makebox[.45em][l]{$^{}$} & \makebox[2em][r]{26.2}\makebox[.45em][l]{$^{}$} & \makebox[2em][r]{0.2}\makebox[.45em][l]{$^{}$} & \makebox[2em][r]{3.5}\makebox[.45em][l]{$^{}$} & \makebox[2em][r]{13.6}\makebox[.45em][l]{$^{}$} & \makebox[2em][r]{8.1}\makebox[.45em][l]{$^{}$} & \makebox[2em][r]{0.4}\makebox[.45em][l]{$^{}$} & \makebox[2em][r]{27.6}\makebox[.45em][l]{$^{}$} & \makebox[2em][r]{19.0}\makebox[.45em][l]{$^{}$} \\
\textcolor{black}{LM + REPL} & \makebox[2em][r]{44.7}\makebox[.45em][l]{$^{}$} & \makebox[2em][r]{24.7}\makebox[.45em][l]{$^{}$} & \makebox[2em][r]{1.0}\makebox[.45em][l]{$^{}$} & \makebox[2em][r]{0.0}\makebox[.45em][l]{$^{}$} & \makebox[2em][r]{13.6}\makebox[.45em][l]{$^{}$} & \makebox[2em][r]{7.3}\makebox[.45em][l]{$^{}$} & \makebox[2em][r]{4.7}\makebox[.45em][l]{$^{}$} & \makebox[2em][r]{33.1}\makebox[.45em][l]{$^{}$} & \makebox[2em][r]{16.1}\makebox[.45em][l]{$^{}$} \\
RLM (base) & \makebox[2em][r]{44.6}\makebox[.45em][l]{$^{}$} & \makebox[2em][r]{18.3}\makebox[.45em][l]{$^{}$} & \makebox[2em][r]{1.1}\makebox[.45em][l]{$^{}$} & \makebox[2em][r]{2.1}\makebox[.45em][l]{$^{}$} & \makebox[2em][r]{15.1}\makebox[.45em][l]{$^{}$} & \makebox[2em][r]{11.2}\makebox[.45em][l]{$^{}$} & \makebox[2em][r]{4.5}\makebox[.45em][l]{$^{}$} & \makebox[2em][r]{25.6}\makebox[.45em][l]{$^{}$} & \makebox[2em][r]{15.3}\makebox[.45em][l]{$^{}$} \\
\rowcolor{black!3}[\tabcolsep][\tabcolsep]
+ RL (short tasks) & \makebox[2em][r]{\textbf{74.9}}\makebox[.45em][l]{$^{}$} & \makebox[2em][r]{21.7}\makebox[.45em][l]{$^{}$} & \makebox[2em][r]{\textbf{2.2}}\makebox[.45em][l]{$^{}$} & \makebox[2em][r]{\textbf{8.3}}\makebox[.45em][l]{$^{}$} & \makebox[2em][r]{\textbf{19.3}}\makebox[.45em][l]{$^{}$} & \makebox[2em][r]{17.8}\makebox[.45em][l]{$^{}$} & \makebox[2em][r]{\textbf{7.3}}\makebox[.45em][l]{$^{}$} & \makebox[2em][r]{26.8}\makebox[.45em][l]{$^{}$} & \makebox[2em][r]{22.3}\makebox[.45em][l]{$^{}$} \\
\rowcolor{black!3}[\tabcolsep][\tabcolsep]
+ RL (synthetic tasks) & \makebox[2em][r]{69.6}\makebox[.45em][l]{$^{}$} & \makebox[2em][r]{28.2}\makebox[.45em][l]{$^{}$} & \makebox[2em][r]{1.6}\makebox[.45em][l]{$^{}$} & \makebox[2em][r]{6.8}\makebox[.45em][l]{$^{}$} & \makebox[2em][r]{18.6}\makebox[.45em][l]{$^{}$} & \makebox[2em][r]{\textbf{23.7}}\makebox[.45em][l]{$^{}$} & \makebox[2em][r]{7.2}\makebox[.45em][l]{$^{}$} & \makebox[2em][r]{44.0}\makebox[.45em][l]{$^{}$} & \makebox[2em][r]{\textbf{25.0}}\makebox[.45em][l]{$^{}$} \\
\midrule\midrule
\multicolumn{10}{@{}l}{\textbf{Qwen3.8-27B}} \\
\midrule
\textcolor{black}{Base Transformer} & \makebox[2em][r]{69.1}\makebox[.45em][l]{$^{}$} & \makebox[2em][r]{44.1}\makebox[.45em][l]{$^{}$} & \makebox[2em][r]{7.9}\makebox[.45em][l]{$^{}$} & \makebox[2em][r]{0.0}\makebox[.45em][l]{$^{}$} & \makebox[2em][r]{20.7}\makebox[.45em][l]{$^{}$} & \makebox[2em][r]{28.2}\makebox[.45em][l]{$^{}$} & \makebox[2em][r]{15.6}\makebox[.45em][l]{$^{}$} & \makebox[2em][r]{58.2}\makebox[.45em][l]{$^{}$} & \makebox[2em][r]{30.5}\makebox[.45em][l]{$^{}$} \\
Pi (base) & \makebox[2em][r]{80.7}\makebox[.45em][l]{$^{}$} & \makebox[2em][r]{61.6}\makebox[.45em][l]{$^{}$} & \makebox[2em][r]{1.0}\makebox[.45em][l]{$^{}$} & \makebox[2em][r]{54.0}\makebox[.45em][l]{$^{}$} & \makebox[2em][r]{16.8}\makebox[.45em][l]{$^{}$} & \makebox[2em][r]{26.2}\makebox[.45em][l]{$^{}$} & \makebox[2em][r]{0.7}\makebox[.45em][l]{$^{}$} & \makebox[2em][r]{38.2}\makebox[.45em][l]{$^{}$} & \makebox[2em][r]{34.9}\makebox[.45em][l]{$^{}$} \\
\textcolor{black}{LM + REPL} & \makebox[2em][r]{79.6}\makebox[.45em][l]{$^{}$} & \makebox[2em][r]{51.0}\makebox[.45em][l]{$^{}$} & \makebox[2em][r]{12.4}\makebox[.45em][l]{$^{}$} & \makebox[2em][r]{34.7}\makebox[.45em][l]{$^{}$} & \makebox[2em][r]{20.6}\makebox[.45em][l]{$^{}$} & \makebox[2em][r]{49.6}\makebox[.45em][l]{$^{}$} & \makebox[2em][r]{14.3}\makebox[.45em][l]{$^{}$} & \makebox[2em][r]{29.8}\makebox[.45em][l]{$^{}$} & \makebox[2em][r]{36.5}\makebox[.45em][l]{$^{}$} \\
RLM (base) & \makebox[2em][r]{85.3}\makebox[.45em][l]{$^{}$} & \makebox[2em][r]{51.4}\makebox[.45em][l]{$^{}$} & \makebox[2em][r]{19.1}\makebox[.45em][l]{$^{}$} & \makebox[2em][r]{64.8}\makebox[.45em][l]{$^{}$} & \makebox[2em][r]{33.3}\makebox[.45em][l]{$^{}$} & \makebox[2em][r]{48.4}\makebox[.45em][l]{$^{}$} & \makebox[2em][r]{13.5}\makebox[.45em][l]{$^{}$} & \makebox[2em][r]{53.1}\makebox[.45em][l]{$^{}$} & \makebox[2em][r]{46.1}\makebox[.45em][l]{$^{}$} \\
\rowcolor{black!3}[\tabcolsep][\tabcolsep]
+ RL (short tasks) & \makebox[2em][r]{87.7}\makebox[.45em][l]{$^{}$} & \makebox[2em][r]{52.4}\makebox[.45em][l]{$^{}$} & \makebox[2em][r]{18.7}\makebox[.45em][l]{$^{}$} & \makebox[2em][r]{64.6}\makebox[.45em][l]{$^{}$} & \makebox[2em][r]{46.6}\makebox[.45em][l]{$^{}$} & \makebox[2em][r]{45.5}\makebox[.45em][l]{$^{}$} & \makebox[2em][r]{18.7}\makebox[.45em][l]{$^{}$} & \makebox[2em][r]{53.8}\makebox[.45em][l]{$^{}$} & \makebox[2em][r]{48.5}\makebox[.45em][l]{$^{}$} \\
\rowcolor{black!3}[\tabcolsep][\tabcolsep]
+ RL (synthetic tasks) & \makebox[2em][r]{\textbf{90.7}}\makebox[.45em][l]{$^{}$} & \makebox[2em][r]{60.1}\makebox[.45em][l]{$^{}$} & \makebox[2em][r]{20.8}\makebox[.45em][l]{$^{}$} & \makebox[2em][r]{\textbf{67.5}}\makebox[.45em][l]{$^{}$} & \makebox[2em][r]{\textbf{49.5}}\makebox[.45em][l]{$^{}$} & \makebox[2em][r]{58.1}\makebox[.45em][l]{$^{}$} & \makebox[2em][r]{\textbf{20.0}}\makebox[.45em][l]{$^{}$} & \makebox[2em][r]{56.4}\makebox[.45em][l]{$^{}$} & \makebox[2em][r]{52.9}\makebox[.45em][l]{$^{}$} \\
\rowcolor{black!6}[\tabcolsep][\tabcolsep]
RLM-27B (synthetic) & \makebox[2em][r]{90.1}\makebox[.45em][l]{$^{}$} & \makebox[2em][r]{\textbf{82.0}}\makebox[.45em][l]{$^{}$} & \makebox[2em][r]{\textbf{39.8}}\makebox[.45em][l]{$^{}$} & \makebox[2em][r]{64.9}\makebox[.45em][l]{$^{}$} & \makebox[2em][r]{43.2}\makebox[.45em][l]{$^{}$} & \makebox[2em][r]{\textbf{61.6}}\makebox[.45em][l]{$^{}$} & \makebox[2em][r]{15.2}\makebox[.45em][l]{$^{}$} & \makebox[2em][r]{\textbf{71.2}}\makebox[.45em][l]{$^{}$} & \makebox[2em][r]{\textbf{58.5}}\makebox[.45em][l]{$^{}$} \\
\bottomrule
\end{tabular}
\par
\end{table}

%% file: tables/main-composition-stats.tex
\begin{table}[htb]
\small
\begingroup
\captionsetup{font={color=black}}
\caption{We report average harness statistics for the evaluation trajectories in Table~\ref{tab:main-composition}. \textit{Delegates} reports the percentage of trajectories with at least one sub-call. \textit{Turn Cap} reports the percentage of trajectories that hit the 40-turn harness limit (harnesses can still submit an answer at the turn cap). Expanded per-environment statistics are reported in Appendix~\ref{appx:composition-statistics}.}
\label{tab:main-composition-stats}
\centering
\setlength{\tabcolsep}{2.3pt}
\renewcommand{\arraystretch}{1.12}
\begin{tabular}{@{}lccc@{\hspace{7pt}}c p{4.4pt} r p{4.4pt} r@{}}
\toprule
{\fontsize{8}{9.5}\selectfont\textbf{Method}} & {\fontsize{8}{9.5}\selectfont \# Turns} & {\fontsize{8}{9.5}\selectfont \# Sub-calls} & {\fontsize{8}{9.5}\selectfont Delegates} & {\fontsize{8}{9.5}\selectfont Turn Cap} & & \multicolumn{1}{c}{{\fontsize{7}{8}\selectfont\makecell{Root tokens\\(in / out)}}} & & \multicolumn{1}{c}{{\fontsize{7}{8}\selectfont\makecell{Sub-call tokens\\(in / out)}}} \\
\midrule
\multicolumn{9}{@{}l}{\textbf{Qwen3-30B-A3B-Instruct-2507}} \\
\midrule
RLM (base) & \makebox[2.4em][r]{16.9} & \makebox[2.4em][r]{697.0} & \makebox[3.1em][r]{49.5\%} & \makebox[3.1em][r]{35.4\%} & & \makebox[5.4em][r]{\makebox[2.5em][r]{321K}\,/\,\makebox[2em][r]{16K}} & & \makebox[6em][r]{\makebox[2.5em][r]{979K}\,/\,\makebox[2.5em][r]{219K}} \\
\rowcolor{black!6}[4pt][4pt]
+ RL (short tasks) & \makebox[2.4em][r]{13.1} & \makebox[2.4em][r]{416.5} & \makebox[3.1em][r]{35.8\%} & \makebox[3.1em][r]{16.6\%} & & \makebox[5.4em][r]{\makebox[2.5em][r]{215K}\,/\,\makebox[2em][r]{16K}} & & \makebox[6em][r]{\makebox[2.5em][r]{619K}\,/\,\makebox[2.5em][r]{50K}} \\
\rowcolor{black!6}[4pt][4pt]
+ RL (synthetic tasks) & \makebox[2.4em][r]{13.1} & \makebox[2.4em][r]{1031.1} & \makebox[3.1em][r]{55.9\%} & \makebox[3.1em][r]{4.6\%} & & \makebox[5.4em][r]{\makebox[2.5em][r]{260K}\,/\,\makebox[2em][r]{19K}} & & \makebox[6em][r]{\makebox[2.5em][r]{577K}\,/\,\makebox[2.5em][r]{50K}} \\
\midrule\midrule
\multicolumn{9}{@{}l}{\textbf{Qwen3.8-27B}} \\
\midrule
RLM (base) & \makebox[2.4em][r]{18.4} & \makebox[2.4em][r]{87.3} & \makebox[3.1em][r]{43.2\%} & \makebox[3.1em][r]{32.9\%} & & \makebox[5.4em][r]{\makebox[2.5em][r]{263K}\,/\,\makebox[2em][r]{5K}} & & \makebox[6em][r]{\makebox[2.5em][r]{443K}\,/\,\makebox[2.5em][r]{10K}} \\
\rowcolor{black!6}[4pt][4pt]
+ RL (short tasks) & \makebox[2.4em][r]{16.9} & \makebox[2.4em][r]{214.1} & \makebox[3.1em][r]{42.4\%} & \makebox[3.1em][r]{25.4\%} & & \makebox[5.4em][r]{\makebox[2.5em][r]{292K}\,/\,\makebox[2em][r]{7K}} & & \makebox[6em][r]{\makebox[2.5em][r]{743K}\,/\,\makebox[2.5em][r]{25K}} \\
\rowcolor{black!6}[4pt][4pt]
+ RL (synthetic tasks) & \makebox[2.4em][r]{16.1} & \makebox[2.4em][r]{27.4} & \makebox[3.1em][r]{45.2\%} & \makebox[3.1em][r]{19.0\%} & & \makebox[5.4em][r]{\makebox[2.5em][r]{392K}\,/\,\makebox[2em][r]{10K}} & & \makebox[6em][r]{\makebox[2.5em][r]{383K}\,/\,\makebox[2.5em][r]{29K}} \\
\rowcolor{black!6}[4pt][4pt]
RLM-27B (synthetic) & \makebox[2.4em][r]{6.5} & \makebox[2.4em][r]{98.5} & \makebox[3.1em][r]{97.4\%} & \makebox[3.1em][r]{1.1\%} & & \makebox[5.4em][r]{\makebox[2.5em][r]{44K}\,/\,\makebox[2em][r]{3K}} & & \makebox[6em][r]{\makebox[2.5em][r]{276K}\,/\,\makebox[2.5em][r]{62K}} \\
\bottomrule
\end{tabular}
\par
\endgroup
\end{table}

%% file: tables/recursion-composition.tex
\begin{table}[htb]
\centering
\small
\captionsetup{font={color=black}}
\caption{We randomly draw ordered sets of four tasks from Table~\ref{tab:eval-envs}, and form a new meta-task where a harness must solve each task in sequence to find the task prompt for the next task, and eventually to find the final answer. The partial solve rate is the percentage of individual tasks whose answers are correct. The full solve rate is the percentage of meta-tasks where all four task answers are correct. We report 95\% normal confidence intervals, using meta-tasks as the independent units.}
\label{tab:recursion-composition}
\setlength{\tabcolsep}{0pt}
\renewcommand{\arraystretch}{1.22}
\begin{tabular}{@{}p{.142\linewidth}*{3}{p{.04\linewidth}*{2}{>{\centering\arraybackslash}p{.123\linewidth}}}@{}}
\toprule
{\fontsize{8}{9.5}\selectfont\textbf{Model}} & & \multicolumn{2}{c}{{\fontsize{8}{9.5}\selectfont\textbf{RLM Harness}}} & & \multicolumn{2}{c}{{\fontsize{8}{9.5}\selectfont\textbf{+ recursion hint}}} & & \multicolumn{2}{c}{{\fontsize{8}{9.5}\selectfont\textbf{+ SFT}} {\fontsize{7}{8}\selectfont(recursion synthetic data)}} \\
\cmidrule{3-4}\cmidrule{6-7}\cmidrule{9-10}
{\fontsize{8}{9.5}\selectfont Solve rate (\%)} & & {\fontsize{8}{9.5}\selectfont Partial} & {\fontsize{8}{9.5}\selectfont Full} & & {\fontsize{8}{9.5}\selectfont Partial} & {\fontsize{8}{9.5}\selectfont Full} & & {\fontsize{8}{9.5}\selectfont Partial} & {\fontsize{8}{9.5}\selectfont Full} \\
\midrule
Qwen3.8-27B & & \textcolor{black}{38.4}{\,\textcolor{black!55}{\scriptsize$\pm$1.7}} & \textcolor{black}{19.3}{\,\textcolor{black!55}{\scriptsize$\pm$1.7}} & & 32.5{\,\textcolor{black!55}{\scriptsize$\pm$1.5}} & 12.4{\,\textcolor{black!55}{\scriptsize$\pm$1.4}} & & 32.1{\,\textcolor{black!55}{\scriptsize$\pm$1.4}} & 9.9{\,\textcolor{black!55}{\scriptsize$\pm$1.3}} \\
\rowcolor{black!6}[3pt][3pt]
RLM-27B & & \textcolor{black}{74.5}{\,\textcolor{black!55}{\scriptsize$\pm$1.4}} & \textcolor{black}{55.2}{\,\textcolor{black!55}{\scriptsize$\pm$2.2}} & & 63.3{\,\textcolor{black!55}{\scriptsize$\pm$1.7}} & 47.2{\,\textcolor{black!55}{\scriptsize$\pm$2.2}} & & \textbf{84.4}{\,\textcolor{black!55}{\scriptsize$\pm$1.4}} & \textbf{77.5}{\,\textcolor{black!55}{\scriptsize$\pm$1.8}} \\
\bottomrule
\end{tabular}
\par
\end{table}

%% file: sections/related.tex
\subsection{Programmatic sub-agent calling systems.}
Language model systems that invoke tool calls or sub-agents have traditionally done so by parsing a language model's output and invoking APIs when appropriate~\citep{schick2023toolformer, yao2023react}. 
Recently, CodeAct~\citep{wang2024executablecodeactionselicit} and Recursive Language Models (RLMs)~\citep{zhang2026recursivelanguagemodels} have proposed using a code REPL as the primary tool call, with all other tools and sub-agents being invoked as functions in code. Modern harnesses such as Claude Code~\citep{anthropic_claude_code} and Codex~\citep{openai_codex} have adopted these features as well, such as in Dynamic Workflows~\citep{anthropic2026dynamicworkflows}. Prime Agent~\citep{karten2026primeagentselfimprovingrlm} adopts the RLM as a central design. 

Training these programmatic sub-agent and tool-calling systems has also been explored recently. Recursive Agent Optimization (RAO)~\citep{gandhi2026recursiveagentoptimization} proposes a training strategy that assigns partial rewards to recursive sub-agent trajectories during RL training, and is applied directly to RLMs. However, they do not study what leads to generalization in their results. In this paper, we explore how naive training over these harness abstractions elicits compositional generalization.

\subsection{Generalization in language models and harnesses.}
Understanding generalization in language model systems can provide recipes for data-efficient scaling of LLM capabilities. Prior works have studied generalization from the data and model architecture perspective~\citep{elhage2021mathematical, chan2022data, doi:10.1073/pnas.2417182122}, including isolated studies on composing learned primitives~\citep{pmlr-v80-lake18a,keysers2020measuring,kim-linzen-2020-cogs, lippl2026algotrace}. Related studies investigate how LMs struggle to generalize to longer inputs and more complex computations~\citep{anil2022exploring,dziri2023faith, motwani2026longcot}, while others highlight how training distributions and computational inductive biases can improve compositional generalization~\citep{repec:nat:nature:v:623:y:2023:i:7985:d:10.1038_s41586-023-06668-3,zhou2024what,yuan2026from}.

Harnesses are a symbolic layer on top of the language model that can directly compose language model calls to provide a mechanism for generalization through composition. Early harness strategies like Least-to-Most Prompting~\citep{zhou2023leasttomost} and Decomposed Prompting~\citep{khot2023decomposed} show that fixed language models can improve easy-to-hard and length generalization through individual language model calls solving isolated, local sub-tasks. Recursive strategies like Recursion of Thought~\citep{lee-kim-2023-recursion} and Re-Tuning~\citep{pasewark-etal-2024-tuning} show similar properties when task decomposition is arbitrarily deep. We directly study whether training around general language model harnesses on unsolved training tasks can generalize to a class of unsolved evaluation tasks through composition. Lastly, alternate context management abstractions like Context Language Models~\citep{shao2026contextlanguagemodels} that treat the model's context as directly editable rather than an offloaded context are not explored in this paper, but may similarly yield LID behavior through training.

%% file: sections/conclusion.tex
Large language models have historically been brittle when composing learned capabilities as a means of generalization. We propose and show that well-designed neurosymbolic LM harnesses can induce composition over existing LM capabilities by shaping the inputs to individual LM calls to be in-distribution. This property enables the harness trajectory to be \textit{locally in-distribution} on an otherwise out-of-distribution task for the LM. We provide empirical evidence of this phenomenon where existing Transformers struggle: RLMs can learn to generalize across context length, and learn latent ``strategies'' that transfer to a new domain with a different token-level description.  In practical settings where LM agents are deployed in extremely long and complex settings, we also show that this effect can be used as a data efficient strategy to target and train specific individual LLM behavior within a harness without training on the full task. We post-train \textbf{RLM-27B} primarily on root trajectories, and train entirely on easier synthetically generated environments, some of which include fake tool / sub-calls. Finally, we provide evidence that the compositional generalization effect of harnesses applies recursively, by applying surrogate harness training to learn tail recursion strategies. We are excited by future work that informs the design of new harness abstractions and efficient training strategies around LM harnesses for solving increasingly long and expensive tasks.

%% file: sections/acknowledgments.tex
This material is based upon work supported by the U.S. National Science Foundation Graduate Research Fellowship Program under Grant No. 2141064. Any opinions, findings, and conclusions or recommendations expressed in this material are those of the author(s) and do not necessarily reflect the views of the National Science Foundation. This research is also supported directly by the Laude Institute Residency and Slingshot programs, Prime Intellect, and Thinking Machines Lab Grant. We thank Jayden Teoh for reviewing the paper and providing helpful feedback on framing. We thank Zhuohan (Joshua) Gu, Noah Ziems, Jacob Li, and the rest of MIT OASYS for helpful discussions throughout this project.

%% file: sections/appendix.tex
\newtcolorbox{appprompt}[1]{
  breakable,
  colback=black!2, colframe=black!45, colbacktitle=black!7,
  coltitle=black, title={#1}, fonttitle=\normalfont\small\bfseries,
  fontupper=\fontencoding{T1}\fontfamily{lmtt}\fontsize{8}{10}\selectfont,
  boxrule=0.45pt, arc=1mm, left=9pt, right=9pt, top=7pt, bottom=7pt,
  before skip=10pt, after skip=10pt,
  before upper={\raggedright\setlength{\parindent}{0pt}\setlength{\parskip}{0pt}}
}
\newcommand{\apppromptrole}[1]{{\normalfont\footnotesize\bfseries #1}\par\smallskip}

\input{appendix/training_details}

\input{appendix/synthetic_environments}

\input{appendix/composition_statistics}
\input{appendix/lid_trajectories}

%% file: appendix/training_details.tex
\clearpage
\section{Training Details}
\label{appx:training-details}
\normalfont\normalsize\raggedbottom

All experiments in the length generalization (Figure~\ref{fig:length-gen}) and strategy generalization plots (Figure~\ref{fig:strategy_grid}) in \S\ref{sec:exp} were trained with Prime Intellect \texttt{prime-rl} on a single 8xH100 node, with four training and four inference GPUs; inference TP=2, DP=2; memory utilization 0.8. We provide the controlled training configuration in Table~\ref{tab:training-config}, and specify any changes for specific settings below. Runs are seeded with $\text{seed}=0$ (default \texttt{prime-rl} setting).

\begin{center}
\captionof{table}{Default training configuration used in \texttt{prime-rl} for all RL post-training experiments, unless otherwise specified.}
\label{tab:training-config}
\small\renewcommand{\arraystretch}{1.16}\setlength{\tabcolsep}{8pt}
\begin{tabular}{@{}>{\raggedright\arraybackslash}p{0.34\linewidth}>{\raggedright\arraybackslash}p{\dimexpr0.66\linewidth-2\tabcolsep\relax}@{}}
\toprule
\textbf{Setting} & \textbf{Value} \\
\midrule
\rowcolor{black!5}\multicolumn{2}{@{}l}{\textbf{Model and optimization}} \\
Base model & \texttt{Qwen/Qwen3-30B-A3B-Instruct-2507} \\
RL algorithm & GRPO \\
Optimizer & AdamW, $\beta=(0.9,0.999)$ \\
Learning rate & $5\times10^{-5}$, constant \\
Weight decay & $0.01$ \\
Gradient norm cap & $1$ \\
Precision & BF16 (optimization and reduction) \\
Activation checkpointing & Full \\
\addlinespace[3pt]
\rowcolor{black!5}\multicolumn{2}{@{}l}{\textbf{LoRA} (Qwen3 only; Qwen3.8 uses full finetuning)} \\
Rank & 32 \\
Scaling parameter $\alpha$ & 64 \\
Dropout & 0 \\
Attention targets & \texttt{q\_proj, k\_proj, v\_proj, o\_proj} \\
MLP targets & \texttt{gate\_proj, up\_proj, down\_proj} \\
\addlinespace[3pt]
\rowcolor{black!5}\multicolumn{2}{@{}l}{\textbf{Rollouts and loss}} \\
Batch size & 64 \\
Rollouts per example & 4 \\
Training sequence length & 8,192 tokens (i.e., sequences trained in one batch) \\
Loss targets & Root assistant tokens; sub-call tokens excluded \\
Loss settings & \texttt{dppo\_mask\_low/high}$=0.2$, \texttt{adv\_tau}$=1$, \texttt{kl\_tau}$=0.001$ \\
Training temperature & $1$ \\
Root / sub-call output cap & 8,192 tokens per completion \\
Thinking & Disabled \\
Evaluation sampling & One rollout per example; includes step 0 \\
Max off-policy steps & 8 \\
\bottomrule
\end{tabular}
\end{center}

On Ada-LEval we increase the training sequence length to 20,480, and on OBLIQ Twitter$\to$WildChat to 24,576 because we observe frequent context window limits. On validation runs, we increase the context window size to 131,072 for both the root and sub-calls.

\textbf{Filters.} Repetition filter: window 3,000, threshold 0.99, enforced.
Zero-advantage filter enforced except Ada-LEval; gibberish monitored only.

\textbf{SFT config for RLM-27B.} The RLM-27B model (Qwen3.8-27B) was not trained with RL, and instead was SFT'd (full-parameter fine-tuning) for one epoch over packed sequences of rollouts on synthetic data of length 32,768. We use AdamW~\citep{loshchilov2019decoupledweightdecayregularization} with learning rate $5\times10^{-6}$ and cosine decay to $1\times10^{-6}$ after 3 warmup steps, zero weight decay, gradient-norm clipping at 1, FP32 optimization and gradient reduction, and global batch size 8, trained on an 8xH100.

\subsection{Prompts used for harnesses and baselines}
\label{appx:nudge}
We provide the prompts used for each harness below.

\textbf{RLM.} We focus on depth-1 sub-agents with a persistent REPL; root and sub-calls share the served policy.
We set a REPL output cap: 20,000 characters per code block, meaning anything beyond is truncated. 
\input{appendix/prompts/rlm-system}
\input{appendix/prompts/rlm-messages}
{For the RLM, later turn messages contain only \texttt{Turn <turn number>/<turn limit>:}; these
messages persist in history. Depth=1 sub-calls receive the initial instruction defined by the root LM as a user message, with no fixed system prompt.}
\input{appendix/prompts/nudge}

\textcolor{black}{The orchestration prompt is appended as part of the system prompt given to the RLM. We provided it because these models are not familiar with this harness design, but we also provide a reference where we do not include it below. We re-ran Figure~\ref{fig:length-gen} below in Figure~\ref{fig:length-gen-no-orchestration}, where we excluded this orchestration prompt addendum. We find that it yields little difference in the harness and overall learned compositional behavior.}
\begin{figure}[tbp]
\centering
\includegraphics[width=\linewidth]{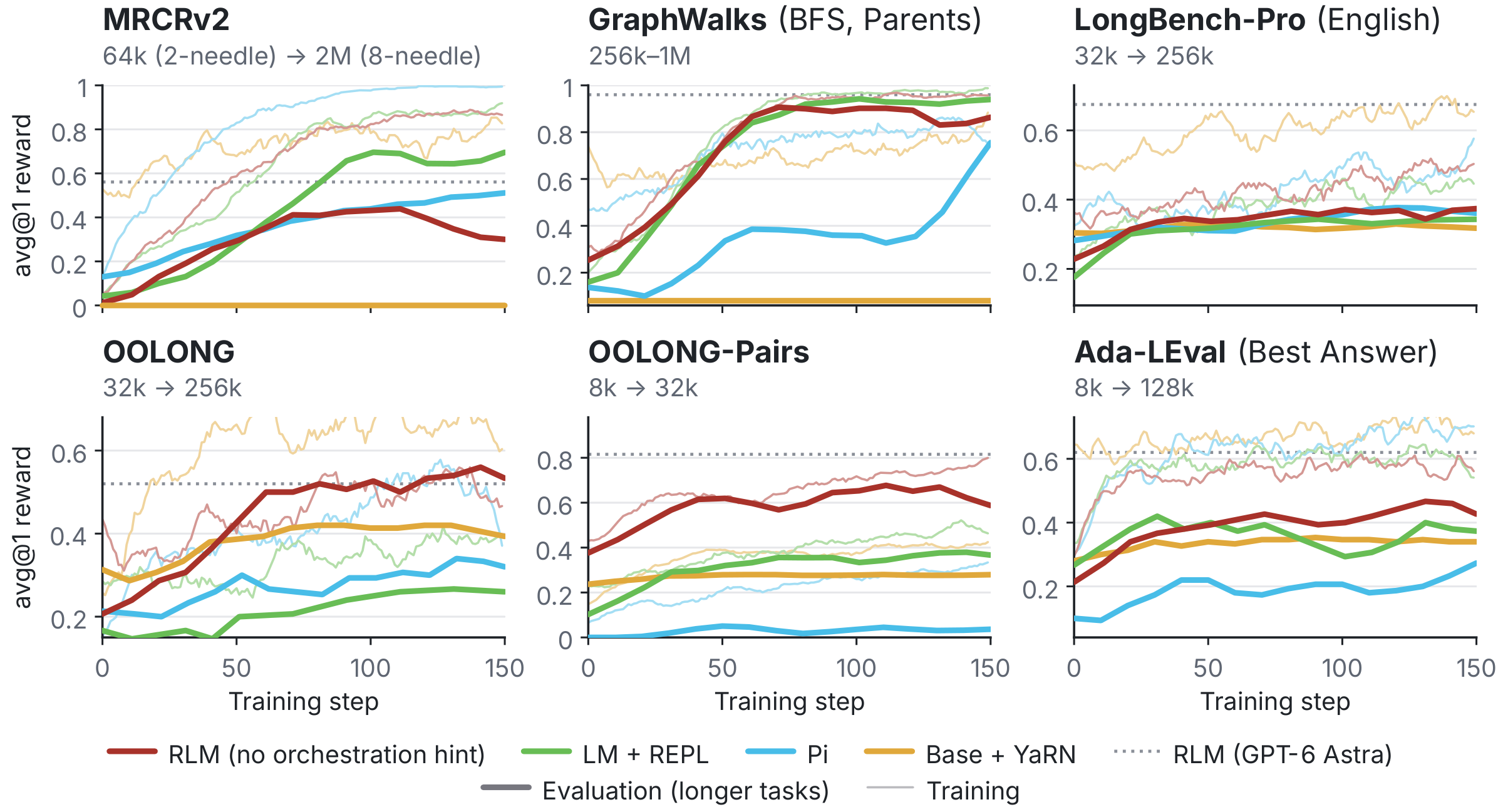}
\caption{\textcolor{black}{We reproduced Figure~\ref{fig:length-gen} but swapped out the RLM implementation with a version that does not use the orchestration hint. The model roughly follows the same trends, except on MRCRv2, where it learns a non-generalizable solution that has context window issues.}}
\label{fig:length-gen-no-orchestration}
\end{figure}

\noindent\begin{minipage}{\linewidth}
\textbf{LM + REPL.} We implement this CodeAct~\citep{wang2024executablecodeactionselicit} variant, where sub-LLM calls are disabled, as a modification of the depth=1 RLM harness.
There is no longer a sub-LLM budget sentence in the initial task message, and also REPL outputs are no longer truncated.
\input{appendix/prompts/repl-system}
\end{minipage}\par

\begin{samepage}
\textbf{Transformer baseline.} This is effectively a model query, where the full context is provided as the user message.
The transformer baseline is evaluated at temperature 0.
When required, we use YaRN~\citep{peng2026yarnefficientcontextwindow} with a window factor $\max(1,\texttt{max\_model\_len}/262144)$,
with 262,144 original positions. 
\input{appendix/prompts/transformer-system}
\end{samepage}

\textbf{Pi 0.73.1 (Node 22.14.0)} Contexts are initially offloaded to \texttt{context.txt}.
The root system prompt is the \texttt{verifiers}~\citep{brown_verifiers_2025} (v0.1.14) wrapper default, passed with
\texttt{--system-prompt}: a short instruction listed below to complete the task with the available tools, plus the runtime date and working directory.
Reader and worker instructions are appended with \texttt{--append-system-prompt}.
Tool feedback supplies a read-cap notice and a soft-turn-cap stop directive. We sample at the \qwen{} defaults: temperature $0.7$, top-$p$ $0.8$,
top-$k$ $20$, and min-$p$ $0$. We use a default context window of 65,536 tokens and the training sequence length
is 8,192; server-side rendering is used, and the zero-advantage filter is monitored only.  

\textbf{Extensions.} We use the \texttt{readcap} (commit: \texttt{f3a226e}) extension to cap tool output at 6,000 bytes, with a cap of 100 tool results per agent unless stated otherwise. We use the \texttt{compaction} (commit: c880157) extension to compact when the root hits the 65,536 token window limit, but we observe in practice that this does not occur during training.
We use the \texttt{subagent} (commit: 9bb2c7f) extension to provide sub-agents as a tool with up to 10 turns per sub-agent, and unlike the RLM harness, we train \textbf{both the root and sub-agents}.

\input{appendix/prompts/pi-system}
\input{appendix/prompts/pi-user}
\input{appendix/prompts/pi-preamble-short}
The user message joins this preamble, the task instructions and the required answer format with blank lines. The parent generates each helper's task message.
\input{appendix/prompts/pi-reader}
\input{appendix/prompts/pi-worker}
\input{appendix/prompts/pi-feedback}

\input{appendix/training/experiments}

\subsection{Environment mixture training details}
\label{appx:mixture-training}
Following the findings in \S\ref{sec:exp}, we explore in \S\ref{sec:4.1} how well our generalization experiments transfer when training on a mixture of tasks, and how well they transfer to a larger, unrelated suite of tasks. We provide details for these training runs in this subsection.

\subsubsection{RL training on short tasks in the length generalization experiments}
\label{appx:short-mixture}
We first confirm that the length generalization effect of Figure~\ref{fig:length-gen} holds when mixed, and also examine how each environment is related. Table~\ref{tab:generalization} evaluates checkpoints trained on individual short splits of each length-generalization environment, including single-environment runs and one mixture of those short tasks across each harness method. We follow the training configuration in Table~\ref{tab:training-config}, train for 1,000 steps, and select the best checkpoint for each.

\input{tables/main}

We find that training on each split performs the best compared to training on other splits across each task, including the unseen tasks used for Figure~\ref{fig:strategy_grid}. Another important finding is that training on individual tasks transfers unevenly to unseen other length-generalization tasks, indicating that the model is not just learning to use the harness better, but is actually learning task-specific behavior that transfers.

\subsubsection{Short-mixture RL on held-out long tasks}
\label{appx:heldout-short-rl}
The \emph{+ RL (short tasks)} rows in Table~\ref{tab:main-composition} are the \S\ref{sec:4.1} runs over training a harness around real environments in Table~\ref{tab:benchmarks}, while the \emph{+ RL (synthetic tasks)} rows are the \S\ref{sec:synth-envs} runs over training a harness around synthetic environments. %

\textbf{RL on short tasks from Table~\ref{tab:benchmarks}.} Each RL run on \qwen{} uses the training configuration in Table~\ref{tab:training-config} (including the LoRA config) on equal-weight short splits of each of the six environments, trained for 1,000 steps, and validated every 20 steps. We select the best checkpoint by mean validation reward. Our validation split was 25 random tasks from each of CorpusQA 128K (education and finance), CTC-Bench and OOLONG-real. The validation splits are each chosen to be disjoint from the evaluation splits.

\qwennew{} was post-trained with the same defaults, except it was full fine-tuning with a lower learning rate $5\times10^{-6}$ and batch size 96 on the same six short-task streams. Validation ran on the same split as the other RL runs.

\textbf{RL on synthetic tasks from \S\ref{sec:synth-envs}.} {Both models \qwen{} and \qwennew{} were post-trained with the same configuration as on the short tasks, but now on eight synthetic environments (BrowseComp-Plus, ExpertLongBench, OBLIQ, CTC-Bench, LooGLE, OOLONG, LongReason, and CorpusQA at 128K/1M), with equal sampling weight per environment and hints disabled. The first three used environment-specific fake sub-call responses; the others used real model sub-calls. Validation ran on the same split as the other RL runs.}

\label{appx:synth-rl}

\input{appendix/training/broad_mix_experiments}

%% file: appendix/prompts/rlm-system.tex
\begin{appprompt}{RLM | System prompt}
You are a Recursive Language Model (RLM): a language model with a prompt, and a very important context stored in a Python REPL related to that prompt. You can iteratively interact with the Python REPL, which has access to LLM calls as a function. You will be queried turn-by-turn until you have an answer to the query.\par
\smallskip
{\hangindent=1.3em\hangafter=1 To use the REPL, you need to write code in \char96{}\char96{}\char96{}repl\char96{}\char96{}\char96{} blocks; the REPL persists across turns. Available in the REPL:\par}
{\hangindent=1.3em\hangafter=1 - \char96{}context\char96{}: the important, potentially very long information related to the prompt (typically \char96{}str\char96{} or \char96{}list[str]\char96{}).\par}
{\hangindent=1.3em\hangafter=1 - \char96{}llm\_\allowbreak{}query(prompt: str, model: str | None = None) -\textgreater{} str\char96{}: a single sub-LLM completion. Use for extraction, summarization, or Q\&A over a chunk of text. Sub-LLM context window \(\approx\) 500K chars.\par}
{\hangindent=1.3em\hangafter=1 - \char96{}llm\_\allowbreak{}query\_\allowbreak{}batched(prompts: list[str], model=None) -\textgreater{} list[str]\char96{}: concurrently call several LLM calls in parallel over a list of prompts; same order out as in.\par}
{\hangindent=1.3em\hangafter=1 - \char96{}rlm\_\allowbreak{}query(prompt, model=None)\char96{} / \char96{}rlm\_\allowbreak{}query\_\allowbreak{}batched(prompts, model=None)\char96{}: recursive RLM sub-calls. Fall back to \char96{}llm\_\allowbreak{}query\char96{} / \char96{}llm\_\allowbreak{}query\_\allowbreak{}batched\char96{} when recursion is disabled.\par}
{\hangindent=1.3em\hangafter=1 - \char96{}SHOW\_\allowbreak{}VARS() -\textgreater{} str\char96{}: list every variable currently in the REPL.\par}
{\hangindent=1.3em\hangafter=1 - \char96{}answer\char96{}: dict initialized to \char96{}\{\char34{}content\char34{}: \char34{}\char34{}, \char34{}ready\char34{}: False\}\char96{}. To submit, set \char96{}answer[\char34{}content\char34{}]\char96{} to the final answer and \char96{}answer[\char34{}ready\char34{}] = True\char96{} inside a \char96{}\char96{}\char96{}repl\char96{}\char96{}\char96{} block.\par}
\smallskip
REPL outputs over \textasciitilde{}20K characters are truncated, so for longer payloads slice \char96{}context\char96{} and pass slices through \char96{}llm\_\allowbreak{}query\char96{} rather than \char96{}print\char96{}-ing them whole. The REPL is NOT a Jupyter cell \textemdash{} only \char96{}print(...)\char96{} output (stdout) is shown back to you between turns; a bare expression on the last line is silently discarded. Always wrap inspections in \char96{}print(...)\char96{}.\par
\smallskip
As a general strategy, you should start by probing your context to understand it better (e.g. print a few lines, count them, etc.). Then, use the REPL to build up an answer to the query.\par
\smallskip

Plan in prose, then execute one \char96{}\char96{}\char96{}repl\char96{}\char96{}\char96{} block every turn, get feedback from the output, then continue on the next turn. Do not flip \char96{}answer[\char34{}ready\char34{}] = True\char96{} on turn 1 without first inspecting \char96{}context\char96{}.\par
\smallskip
\smallskip
\end{appprompt}

%% file: appendix/prompts/rlm-messages.tex
\begin{appprompt}{RLM / LM + REPL | User-message templates}
\apppromptrole{Initial task message}
Answer the following: \textless{}task root prompt\textgreater{}\par
\smallskip
Your context is a \textless{}context type\textgreater{} of \textless{}character count\textgreater{} total characters. Each sub-LLM call can handle roughly \textasciitilde{}100k tokens at once.\par
\medskip
\apppromptrole{First turn}
You have not interacted with the REPL environment or seen your prompt / context yet. Look at the context first; do not provide a final answer yet.\par
\smallskip
Turn 1/\textless{}turn limit\textgreater{}:\par
\end{appprompt}

%% file: appendix/prompts/nudge.tex
\begin{appprompt}{RLM | Orchestration prompt}
{
As an RLM, you should act as an orchestrator, not a solver.\par
\smallskip
Directly after you probe the \char96{}context\char96{} and understand your task, pause and plan: state explicitly how the task decomposes into sub-LLM / REPL steps, and sketch the concrete sequence of turns \textemdash{} what each turn computes and which sub-LLM call (if any) it issues \textemdash{} like a condensed trajectory, before you execute them. Then execute one turn at a time: after each step \char96{}print\char96{} a small sample of the result, verify it looks right, and only flip \char96{}answer[\char34{}ready\char34{}] = True\char96{} once you have actually printed the candidate answer. If you are running out of turns without a confirmed answer, submit your best inference rather than letting the rollout terminate unsubmitted.\par
\smallskip
Your own context window is small. Push every long-context operation that would not fit comfortably in your own working window \textemdash{} reading, summarizing, classifying, verifying, answering sub-questions, even recapping your own progress \textemdash{} into \char96{}llm\_\allowbreak{}query\char96{} / \char96{}llm\_\allowbreak{}query\_\allowbreak{}batched\char96{} calls instead of pulling that text into your own message stream. (Conversely: if a Python keyword / regex search over \char96{}context\char96{} would already pin the answer, or if a single visible passage already contains it, just read it directly \textemdash{} sub-LMs are for when the raw text won\char39{}t fit or the question needs semantic interpretation.) Long REPL stdout pollutes history the same way raw \char96{}context\char96{} does: if you want a recap, ask \char96{}llm\_\allowbreak{}query\char96{} for a 1\textendash{}2 sentence summary and \char96{}print\char96{} only that. Aggregate the small results back in the REPL.\par
\smallskip
Sub-LLMs have no REPL; they only see the prompt and the \char96{}context\char96{} slice you pass them. Hand them clean, focused inputs and ask for terse, structured outputs you can manipulate programmatically.\par
\smallskip
Sub-call budget is finite on two independent axes, and \char96{}llm\_\allowbreak{}query\_\allowbreak{}batched\char96{} only parallelizes \textemdash{} it does not relax either. (1) Per-prompt capacity: a single sub-call answers well only when its input stays modestly sized \textemdash{} a useful rough ceiling is \textasciitilde{}100K characters per prompt, less when the text is dense. Pack each prompt close to that capacity (a chunk of many items, a whole document) so one call accomplishes a lot of work. (2) Per-batch fan-out: \char96{}llm\_\allowbreak{}query\_\allowbreak{}batched\char96{} concurrency is bounded too \textemdash{} a useful rough ceiling is \textasciitilde{}20 prompts per batch. Tiny-prompt mega-batches (hundreds or thousands of single-item prompts) are the anti-pattern; fat-prompt small batches are correct. For many independent units, use several \textasciitilde{}20-wide batches of full-capacity prompts in sequence, not one mega-batch of tiny prompts. When the work can be expressed either as a sequential loop of \char96{}llm\_\allowbreak{}query\char96{}s or as one comparably-sized batched call, prefer batched \textemdash{} same total work, far fewer turns burned. After Python-side filtering has narrowed the candidate set, batch-extract the survivors rather than reading them by hand. If the raw workload exceeds both budgets at once (e.g. a context far larger than \textasciitilde{}20 \(\times\) 100K chars), don\char39{}t brute-force it: filter aggressively in Python first to a tractable subset, or stage the task \textemdash{} a cheap coarse pass narrows candidates, then a targeted second pass extracts from the survivors.\par
\smallskip
Reserve your own tokens for high-level decisions: what to ask next, how to combine sub-LM outputs, when to finalize. Delegate everything else.\par
}
\end{appprompt}

%% file: appendix/prompts/repl-system.tex
\begin{appprompt}{LM + REPL | System prompt}
You are a language model with a prompt, and a very important context stored in a Python REPL related to that prompt. You can iteratively interact with a Python REPL to inspect, transform, and analyze the context. You will be queried turn-by-turn until you have an answer to the query.\par
\smallskip
{\hangindent=1.3em\hangafter=1 To use the REPL, you need to write code in \char96{}\char96{}\char96{}repl\char96{}\char96{}\char96{} blocks; the REPL persists across turns. Available in the REPL:\par}
{\hangindent=1.3em\hangafter=1 - \char96{}context\char96{}: the important, potentially very long information related to the prompt (typically \char96{}str\char96{} or \char96{}list[str]\char96{}).\par}
{\hangindent=1.3em\hangafter=1 - \char96{}SHOW\_\allowbreak{}VARS() -\textgreater{} str\char96{}: list every variable currently in the REPL.\par}
{\hangindent=1.3em\hangafter=1 - \char96{}answer\char96{}: dict initialized to \char96{}\{\char34{}content\char34{}: \char34{}\char34{}, \char34{}ready\char34{}: False\}\char96{}. To submit, set \char96{}answer[\char34{}content\char34{}]\char96{} to the final answer and \char96{}answer[\char34{}ready\char34{}] = True\char96{} inside a \char96{}\char96{}\char96{}repl\char96{}\char96{}\char96{} block.\par}
\smallskip
The REPL is a plain Python sandbox (stdlib only, no network and no LLM calls) \textemdash{} solve the task with string operations, slicing, regex, counting, and any pure-Python computation over \char96{}context\char96{}.\par
\smallskip
REPL output is NOT truncated: you may \char96{}print(...)\char96{} as much as you need \textemdash{} including large slices, or the whole of \char96{}context\char96{} \textemdash{} and the full stdout is returned to you. The REPL is NOT a Jupyter cell \textemdash{} only \char96{}print(...)\char96{} output (stdout) is shown back to you between turns; a bare expression on the last line is silently discarded. Always wrap inspections in \char96{}print(...)\char96{}.\par
\smallskip
As a general strategy, you should start by probing your context to understand it better (e.g. print a few lines, count them, inspect its type/length). Then, use the REPL to build up an answer to the query.\par
\smallskip
Plan in prose, then execute one \char96{}\char96{}\char96{}repl\char96{}\char96{}\char96{} block every turn, get feedback from the output, then continue on the next turn. Do not flip \char96{}answer[\char34{}ready\char34{}] = True\char96{} on turn 1 without first inspecting \char96{}context\char96{}.\par
\end{appprompt}

%% file: appendix/prompts/transformer-system.tex
\begin{appprompt}{Transformer | System prompt}
You are a careful assistant. You are given a long document or problem followed by a question or task. Read all of it and answer directly in the exact format requested. Do not ask for tools or external help \textemdash{} produce the final answer yourself.\par
\end{appprompt}

%% file: appendix/prompts/pi-system.tex
\begin{appprompt}{Pi | Default verifiers wrapper system prompt}
Complete the user\char39{}s task using the available tools.\par
Current date: \textless{}date\textgreater{}\par
Current working directory: \textless{}sandbox directory\textgreater{}\par
\end{appprompt}

%% file: appendix/prompts/pi-user.tex
\begin{appprompt}{Pi | Custom user-message instructions}
\apppromptrole{File-reading preamble}
IMPORTANT \textemdash{} READ THIS FIRST. The data for this task is in a file named \char96{}context.txt\char96{} in your working directory, and it is almost certainly LARGER THAN YOUR ENTIRE CONTEXT WINDOW \textemdash{} it will NOT fit. If you read it in full (\char96{}cat context.txt\char96{}, opening it in an editor, or any unbounded read), the request will FAIL with a context-overflow error and you will score ZERO. Do not do this. Never load the whole file \textemdash{} or a large chunk of it \textemdash{} into your context.\par
\smallskip
{\hangindent=1.3em\hangafter=1 Work carefully within your window instead:\par}
{\hangindent=1.3em\hangafter=1 1. First check the file\char39{}s size: \char96{}wc -l context.txt\char96{} and \char96{}wc -c context.txt\char96{}.\par}
{\hangindent=1.3em\hangafter=1 2. Find what you need with \char96{}grep -n\char96{} / \char96{}rg\char96{} (search by keyword), and read only small ranges with \char96{}sed -n \char39{}START,ENDp\char39{} context.txt\char96{}, \char96{}head\char96{}, \char96{}tail\char96{}, or \char96{}awk\char96{} \textemdash{} never more than a couple thousand lines at a time.\par}
{\hangindent=1.3em\hangafter=1 3. For counting or aggregation, compute the answer with a shell command (\char96{}grep -c\char96{}, \char96{}awk\char96{}, \char96{}sort | uniq -c\char96{}) rather than reading raw data into your reply.\par}
\smallskip
You are an autonomous coding agent running NON-INTERACTIVELY: there is no human to talk to. Do not greet, do not ask for clarification, do not wait for input \textemdash{} begin working immediately and finish the task entirely on your own.\par
\smallskip
When finished, put your final answer on its own line at the very end, in the exact OUTPUT FORMAT stated below.\par
\medskip
\apppromptrole{Helper-agent announcement}
You also have a \char96{}subagent\char96{} tool. It hands a sub-task to a helper agent that has its OWN separate context window, and returns only that agent\char39{}s short reply to you \textemdash{} so the raw text the helper reads never enters your window. Two helpers are available: \char96{}reader\char96{} (read/search a slice of the file and report back what it found) and \char96{}worker\char96{} (carry out a self-contained sub-task, e.g. score or aggregate a batch). Use it when it helps; you can also keep using the shell tools directly.\par
\end{appprompt}

%% file: appendix/prompts/pi-preamble-short.tex
\begin{appprompt}{Pi: user-message preamble (hard-cap reruns)}
You are an autonomous agent running non-interactively: no human will reply, so never ask questions. The task data is in \char96{}context.txt\char96{} in your working directory. It is far larger than your context window, so never read it whole: check its size, search it with \char96{}grep\char96{}, read small ranges with \char96{}sed -n\char96{}, \char96{}head\char96{} or \char96{}awk\char96{}, and compute counts and aggregates with shell commands.\par
\smallskip
Hard limit: you have at most 30 turns (model calls) in total. When they are used up the run ends with no answer and you score zero, so give your final answer before then.\par
\smallskip
You also have a \char96{}subagent\char96{} tool. Each call runs a helper agent on a sub-task in its own separate context window and returns only its reply. A helper gets at most 10 turns, and you have only 100 sub-agent calls in total for the whole task. Use them wisely: delegate sub-tasks that are worth a helper (for example reading or scoring a slice of the file), put several items into one call instead of making many small calls, and use the shell directly whenever it is enough.\par
\smallskip
Put your final answer at the very end, in the exact OUTPUT FORMAT given below.\par
\end{appprompt}

%% file: appendix/prompts/pi-reader.tex
\begin{appprompt}{Pi reader | Custom system-prompt addition}
You are a reader sub-agent with your own private context window.\par
\smallskip
You are given ONE narrow question about a large local file (usually \char96{}context.txt\char96{}) that the agent delegating to you has NOT read and cannot afford to read in full. Your job is to look at only the part you need and hand back a short, faithful answer.\par
\smallskip
{\hangindent=1.3em\hangafter=1 How to work:\par}
{\hangindent=1.3em\hangafter=1 1. Use \char96{}bash\char96{} to locate what matters \textemdash{} \char96{}grep -n\char96{}, \char96{}wc -l\char96{}, \char96{}sed -n \char39{}A,Bp\char39{} context.txt\char96{}. Never cat the whole file; it is far larger than your window.\par}
{\hangindent=1.3em\hangafter=1 2. Read only the lines you actually need.\par}
{\hangindent=1.3em\hangafter=1 3. Stop as soon as you can answer. You have a limited turn budget.\par}
\smallskip
{\hangindent=1.3em\hangafter=1 How to answer:\par}
{\hangindent=1.3em\hangafter=1 - Lead with the direct answer to the question you were asked, in the first line.\par}
{\hangindent=1.3em\hangafter=1 - Then, only if useful, add the concrete supporting evidence (ids, labels, counts, short verbatim quotes, line numbers).\par}
{\hangindent=1.3em\hangafter=1 - Quote exactly. Never invent an id, label, count, or quote. If the answer is not present in what you read, say so plainly rather than guessing.\par}
{\hangindent=1.3em\hangafter=1 - Be terse. Your entire reply is pasted back into another agent\char39{}s context window, so every wasted token costs the parent. No preamble, no restating the task.\par}
\end{appprompt}

%% file: appendix/prompts/pi-worker.tex
\begin{appprompt}{Pi worker | Custom system-prompt addition}
You are a worker sub-agent with your own private context window.\par
\smallskip
You are given ONE self-contained sub-task by a parent agent that is coordinating a larger job over a big local file (usually \char96{}context.txt\char96{}). Do the sub-task and return only what the parent needs to carry on.\par
\smallskip
{\hangindent=1.3em\hangafter=1 How to work:\par}
{\hangindent=1.3em\hangafter=1 1. Read only what the sub-task requires. \char96{}grep -n\char96{}, \char96{}wc -l\char96{}, and \char96{}sed -n \char39{}A,Bp\char39{} context.txt\char96{} are your main tools; the file is much larger than your context window, so never read it whole.\par}
{\hangindent=1.3em\hangafter=1 2. Do the actual work you were asked to do \textemdash{} compare, count, rank, extract, judge.\par}
{\hangindent=1.3em\hangafter=1 3. Stop as soon as the sub-task is done. You have a limited turn budget.\par}
\smallskip
{\hangindent=1.3em\hangafter=1 How to answer:\par}
{\hangindent=1.3em\hangafter=1 - Give the result first, in the exact shape the parent asked for (a list of ids, a count, a label, a short ranking). If the parent specified an output format, follow it exactly.\par}
{\hangindent=1.3em\hangafter=1 - Ground every claim in what you actually read. Never invent ids, labels, counts, or quotes. If you could not determine something, say so explicitly instead of guessing \textemdash{} a wrong confident answer is worse than a flagged gap.\par}
{\hangindent=1.3em\hangafter=1 - Be terse. Your reply is pasted straight back into the parent\char39{}s context window.\par}
\end{appprompt}

%% file: appendix/prompts/pi-feedback.tex
\begin{appprompt}{Pi | Custom tool-feedback notices}
\apppromptrole{Read-cap notice}
...[read-cap: \textless{}omitted bytes\textgreater{} bytes omitted. Your context window is small \textemdash{} do NOT read large chunks. Narrow the search: grep a more specific pattern, or read a small line range with \char96{}sed -n \char39{}A,Bp\char39{} context.txt\char96{}.]...\par
\medskip
\apppromptrole{Soft-cap replacement}
\(\ominus\) TURN LIMIT REACHED (\textless{}turn cap\textgreater{} tool-using turns). Do NOT call any more tools. Using only what you have already gathered, output your FINAL ANSWER now as plain text, in the exact required OUTPUT FORMAT.\par
\end{appprompt}

%% file: appendix/training/experiments.tex
\newcommand{\envfield}[2]{\item\textbf{#1:} #2}
\newenvironment{envdetails}{%
  \mbox{}\par
  \begin{itemize}\setlength{\itemsep}{5pt}\setlength{\parsep}{0pt}\setlength{\topsep}{4pt}\setlength{\partopsep}{0pt}\setlength{\parskip}{0pt}%
}{\end{itemize}}

\subsection{Length generalization environment details}
\label{appx:length-gen}

We provide details for each of the six environments used in our length generalization experiments in Figure~\ref{fig:length-gen}.
Unless otherwise specified, we train for up to 150 steps with the configuration in Table~\ref{tab:training-config}, evaluate every 10 steps, and use a maximum of 20 turns.
Each environment lists its split, score, and any hint, timeout, or harness window that departs from that configuration.

We observed in many cases that the default settings led to context overflow issues in some settings (e.g. in an earlier run the Pi harness on MRCRv2 attempted to fill the model context window with the entire offloaded context and had no bound on its number of turns, which created noisy dips in evaluation performance and frequent sandbox timeouts; the final Pi runs use the hard turn, sub-agent and wall-clock caps listed below) that did not reflect a fair baseline comparison, so we slightly tuned each setting when appropriate to properly trade off training speed (most settings lead to long-tailed, slow-decode periods) and evaluation accuracy (avoiding dropping rollouts due to a context constraint within the window of the base model).

Most harness-specific changes are to avoid long timeout issues or trivial context window issues, while balancing training efficiency. We detail each below, if they deviated from the default settings.

\paragraph{MRCRv2~\citep{vodrahalli2024michelangelolongcontextevaluations}.}
\begin{envdetails}
\envfield{Training split}{2 needles, \texttt{32k--64k}, 121 examples.}
\envfield{Evaluation split}{8 needles, \texttt{2m--4m}, 50 examples.}
\envfield{Transformer}{Training sequence and inference window 73,728 tokens. The long evaluation caps at 4M and uses YaRN, although samples beyond 2M all score zero in our observation.}
\envfield{LM + REPL}{Selected rerun evaluation window 262,144.}
\envfield{Pi}{Hard caps per rollout: 30 root model calls, 100 sub-agent calls and 10 model calls per sub-agent; a rollout that reaches a cap ends with no answer and scores zero. Training window 65,536 tokens (4,096 completion tokens per call) and evaluation window 262,144; a single file read is capped at 6\,KB; compaction is off. Wall clock 600\,s for the root and 120\,s per sub-agent in training, 1,800\,s and 300\,s in evaluation (sandbox command timeout 660\,s in training, 1,860\,s in evaluation).}
\end{envdetails}

\paragraph{GraphWalks~\citep{openai2025graphwalks}.}
\begin{envdetails}
\envfield{Training split}{\textcolor{black}{\texttt{128k\_and\_shorter} tokens, 512 examples, BFS and Parents.}}
\envfield{Evaluation split}{\textcolor{black}{\texttt{256k\_to\_1mil} tokens, 50 examples.}}
\envfield{Prompt}{The archived prompt says ``at most D hops.'' The current loader says exactly depth D and selects the last Operation after the worked example.}
\envfield{Transformer}{Training sequence and inference window 49,152 tokens. Long evaluation uses the separate offline adapter evaluation, with a long evaluation cap of 2M and YaRN.}
\end{envdetails}

\paragraph{LongBench-Pro~\citep{chen2026longbenchprorealisticcomprehensive}.}
\begin{envdetails}
\envfield{Training split}{English 32K, all 115 examples.}
\envfield{Evaluation split}{English 256K, 115 examples; 23 non-summary secondary tasks, five examples each. Archived questions and task metadata are in \path{appendix/longbenchpro_queries.json}.}
\envfield{RLM / LM + REPL}{Root prompt is the dataset question verbatim.}
\envfield{Pi}{Similar to the setup on MRCRv2 (hard caps of 30 root and 10 sub-agent model calls, 100 sub-agent calls; windows 65,536 and 262,144), with two additions for this task only: the prompt tells the agent to read the file with \texttt{wc}, \texttt{grep -n} and \texttt{sed -n} instead of the 6\,KB \texttt{read} tool and names the \texttt{reader} and \texttt{worker} helpers, and the last 6 root turns (3 for a sub-agent) carry a turn-budget notice. The caps are unchanged. Without the additions, 54\% of evaluation rollouts hit the turn cap and scored zero.}
\envfield{Transformer}{Training sequence and inference window 40,960 tokens; long evaluation is offline. User message: \texttt{<context>}, two newlines, then the same question.}
\end{envdetails}

\paragraph{OOLONG~\citep{bertsch2025oolongevaluatinglongcontext}.}
\begin{envdetails}
\envfield{Training split}{\texttt{trec\_coarse}, \texttt{min\_ctx=max\_ctx=32768}, all 50 examples.}
\envfield{Evaluation split}{Same task at 262,144 tokens, 50 examples; \texttt{exclude\_numeric=false}.}
\envfield{Transformer}{Training sequence and inference window 40,960 tokens. Corrected offline evaluation: window 262,144, completion cap 8,192, temperature 0, answer-only system-prompt override. User message: \texttt{<context>}, two newlines, then the task root prompt.}
\end{envdetails}

\paragraph{OOLONG-Pairs~\citep{zhang2026recursivelanguagemodels}.}
\begin{envdetails}
\envfield{Training split}{8,192-token context, all 20 pair questions.}
\envfield{Evaluation split}{32,768 tokens, 20 questions.}
\envfield{RLM / LM + REPL}{Increased cap to 30 total turns. Sub-call cap 1,000 tokens (\texttt{clean\_judge\_max\_sub\_calls}).}
\envfield{Pi}{As for MRCRv2 (hard caps of 30 root and 10 sub-agent model calls, 100 sub-agent calls; windows 65,536 and 262,144), with additions for this task only: the prompt describes the task (classify the records, compute the pairs, write the complete list), the last 6 root turns carry a turn-budget notice, and sub-agents may read up to 96\,KB per tool call (the root 6\,KB). The output budget per model call is 12,288 tokens in training and 131,072 in evaluation (4,096 for the other tasks), the trainer sequence length is 32,768 (8,192 elsewhere), and the evaluation wall clocks are 3,600\,s (root) and 600\,s (sub-agent). A rollout that reaches the 3,720\,s backstop counts as reward 0.}
\envfield{Transformer}{Training sequence is 40,960 tokens, and inference context window is 262,000.}
\end{envdetails}

\paragraph{Ada-LEval---BestAnswer~\citep{wang2024adalevalevaluatinglongcontextllms}.}
\begin{envdetails}
\envfield{Training split}{8K, 798 question IDs, disjoint from evaluation. Separate \texttt{train\_long} pool: 149 IDs.}
\envfield{Evaluation split}{128K, 50 question IDs.}
\envfield{Environment hint}{The additional user message is used for hinted RLM and omitted from the regular RLM run.}
\envfield{LM + REPL}{Evaluation inference window 262,144.}
\envfield{Pi}{As for MRCRv2 (hard caps of 30 root and 10 sub-agent model calls, 100 sub-agent calls; windows 65,536 and 262,144), with additions for this task only: the prompt describes the candidate format and tells the agent to hand slices of the list to \texttt{reader} helpers, the last 6 root turns carry a turn-budget notice, sub-agents may read up to 96\,KB per tool call (the root 6\,KB), and the trainer sequence length is 32,768 (8,192 elsewhere). With the plain prompt, evaluation rollouts reached the turn cap after reading about 20\% of the 573\,KB candidate list.}
\envfield{Transformer}{Training sequence 16,384 and inference window 139,264.}
\end{envdetails}

\subsection{Strategy generalization environment details}
\label{appx:strategy-gen}

We provide details for each environment in the strategy generalization experiments in Figure~\ref{fig:strategy_grid}, in the same form as Section~\ref{appx:length-gen}.
Unless otherwise specified, we train for up to 500 steps with the configuration in Table~\ref{tab:training-config}, evaluate every 10 steps, and use a maximum of 20 turns.

\paragraph{OOLONG~\citep{bertsch2025oolongevaluatinglongcontext}: TREC-coarse $\to$ spam.}
\begin{envdetails}
\envfield{Training split}{TREC-coarse, 32,768 tokens, 50 examples.}
\envfield{Evaluation split}{Spam, 131,072 tokens, 50 examples. Domain and length both change.}
\envfield{Environment hint}{Training and evaluation use TREC and spam task framing, respectively. Transformer retains the neutral system prompt.}
\envfield{Transformer}{Training sequence 40,960 and inference window 147,456.}
\end{envdetails}

\paragraph{OBLIQ~\citep{tchuindjo2026obliqbenchexposingoverlookedbottlenecks}: writing $\to$ math.}
\begin{envdetails}
\envfield{Training split}{All 512 writing queries, using 300 examples guided by the OBLIQ loader.}
\envfield{Evaluation split}{50 math queries, using 300 examples guided by the OBLIQ loader.}
\envfield{Environment hint}{Domain-specific task framing; the same additional user hint in both domains.}
\envfield{Transformer}{Training sequence and inference window 172,032 tokens.}
\end{envdetails}

\paragraph{OBLIQ~\citep{tchuindjo2026obliqbenchexposingoverlookedbottlenecks}: Twitter $\to$ WildChat.}
\begin{envdetails}
\envfield{Training split}{All 281 Twitter queries, using 300 examples guided by the OBLIQ loader.}
\envfield{Evaluation split}{35 WildChat queries, using 300 examples guided by the OBLIQ loader.}
\envfield{Environment hint}{Domain-specific task framing; the same additional user hint in both domains.}
\envfield{Transformer}{Training sequence 65,536 and inference window 172,032.}
\end{envdetails}

%% file: tables/main.tex
\begin{table}[hbt!]
\centering
\caption{We explore how training on the full suite of short task environments influences harness behavior. For several harnesses and the base neural language model, we compare the base model and one trained on only short tasks from Table~\ref{tab:benchmarks}. For the RLM, we also isolate checkpoints that only trained on one short task to compare to the mix.}
\label{tab:generalization}

\vspace{2pt}
\definecolor{baserow}{gray}{0.92}
\footnotesize
\setlength{\tabcolsep}{4.5pt}
\renewcommand{\arraystretch}{1.15}

\begin{adjustbox}{max width=\linewidth}
\begin{tabular}{lccccccccc}
\toprule
& \multicolumn{6}{c}{\textsc{Length generalization}} & \multicolumn{3}{c}{\textsc{Strategy generalization}} \\
\cmidrule(lr){2-7}\cmidrule(lr){8-10}
Method
& MRCRv2 & GraphWalks & \makecell{LongBench\\Pro} & OOLONG & \makecell{OOLONG\\Pairs} & \makecell{Ada\\LEval}
& \makecell{OOLONG\\spam} & \makecell{OBLIQ\\math} & \makecell{OBLIQ\\wildchat} \\
\midrule

\rowcolor{baserow}[\tabcolsep][\tabcolsep]
Pi (with sub-agents)
  & 0.13 & 0.17 & 0.29 & 0.20 & 0.00 & 0.06 & 0.26 & 0.02 & 0.03 \\
\hspace{1em}\textit{trained on short tasks}
  & 0.33   & 0.12   & 0.20   & 0.20   & 0.01   & 0.10   & 0.46   & 0.06   & 0.04   \\
\addlinespace[3pt]

\rowcolor{baserow}[\tabcolsep][\tabcolsep]
Base Transformer + YaRN
  & 0.00 & 0.08 & 0.30 & 0.32 & 0.24 & 0.28 & 0.32 & 0.22 & 0.04 \\
\hspace{1em}\textit{trained on short tasks}
  & 0.01 & 0.08 & 0.37 & 0.32 & 0.28 & 0.36 & 0.36 & 0.23 & 0.08 \\
\addlinespace[3pt]

\rowcolor{baserow}[\tabcolsep][\tabcolsep]
LM + REPL (no sub-agents, depth $=0$)
  & 0.00 & 0.14 & 0.11 & 0.14 & 0.10 & 0.16 & 0.36 & 0.15 & 0.04 \\
\hspace{1em}\textit{trained on short tasks}
  & 0.56 & 0.84 & 0.28 & 0.00 & 0.34 & 0.32 & 0.04 & 0.14 & 0.07 \\
\addlinespace[3pt]

\rowcolor{baserow}[\tabcolsep][\tabcolsep]
RLM
  & 0.02 & 0.30 & 0.26 & 0.32 & 0.18 & 0.24 & 0.36 & 0.12 & 0.07 \\
\rowcolor{baserow}[\tabcolsep][\tabcolsep]
\hspace{1em}\textit{trained on GraphWalks (short only)}
  & 0.15 & $0.96^{\dag}$ & 0.34 & 0.32 & 0.39 & 0.20 & 0.44 & 0.10 & 0.04 \\
\rowcolor{baserow}[\tabcolsep][\tabcolsep]
\hspace{1em}\textit{trained on LongBench-Pro (short only)}
  & 0.14 & 0.67 & $0.39^{\dag}$ & 0.40 & 0.39 & 0.42 & 0.50 & 0.11 & 0.15 \\
\rowcolor{baserow}[\tabcolsep][\tabcolsep]
\hspace{1em}\textit{trained on OOLONG-Pairs (short only)}
  & 0.00 & 0.50 & 0.19 & 0.36 & $0.72^{\dag}$ & 0.26 & 0.44 & 0.26 & 0.06 \\
\hspace{1em}\textit{trained on short tasks}
  & \textbf{0.88} & \textbf{0.98} & \textbf{0.40} & \textbf{0.60} & \textbf{0.75} & \textbf{0.46} & \textbf{0.56} & \textbf{0.30} & \textbf{0.19} \\
\bottomrule
\end{tabular}
\end{adjustbox}
\end{table}

%% file: appendix/training/broad_mix_experiments.tex
\subsubsection{Training on a broader environment mixture}
\label{appx:broad-mixture-experiments}

In \S\ref{sec:4.1}, we found that training on our mixture of short length-generalization tasks weakly transfers to the unseen, long tasks in Table~\ref{tab:eval-envs}, while in \S\ref{sec:synth-envs} we found that targeting specific behavior with synthetic environments provides a strategy for learning desirable behaviors. However, when training general harness behavior, it is not always known \textit{what} desirable behavior we want, so a natural question is whether training on a large mixture of real tasks would yield stronger transfer. A plausible explanation for why training on the real environments does not provide large performance gains is that behavior learned over these six environments does not transfer meaningfully to these harder tasks, so we explore in this Appendix section whether a wider mixture of tasks yields transfer.

We post-train \qwen{} on a wider mixture of tasks that include the six length-generalization tasks. We follow the training configuration in Table~\ref{tab:training-config}, train for 1,000 steps, and select the best checkpoint. We validate on a held-out set of 20 random disjoint validation tasks from each environment with a validation split in Table~\ref{tab:broad-mix}. We do not include environments that share tasks with our training set, so we exclude BrowseComp-Plus and OBLIQ-Bench in our reported benchmark results in Table~\ref{tab:broad-ablation-results}.

\begin{figure}[htbp]
\centering
\includegraphics[width=\linewidth]{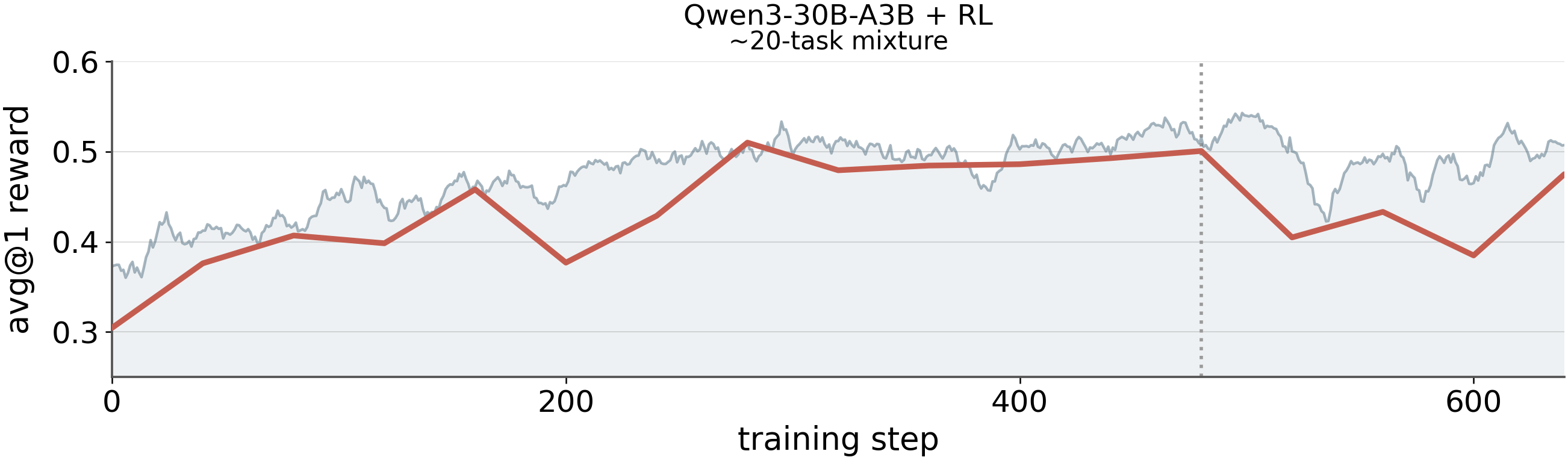}
\caption{We report the average training reward in gray and the held-out reward (red) for \qwen{} on the broader training mixture in Table~\ref{tab:broad-mix}. The dotted line is the checkpoint selected for evaluation.}
\label{fig:broad-mixture-rl}
\end{figure}

Figure~\ref{fig:broad-mixture-rl} shows training and held-out reward, and Table~\ref{tab:broad-ablation-stats} reports trajectory statistics. We report scores for this mixture compared to other trained artifacts we used in Table~\ref{tab:broad-ablation-results}. We find that the broader mixture scores slightly higher on average than the short-task mixture and the synthetic task mixture, despite the synthetic task mixture yielding significant gains over other splits on the \qwennew{} model. We generally find that RL training on any tasks with the harness yields significant improvements for the \qwen{} model, particularly because it uses sub-calls more effectively or not at all, and submits more reliably rather than looping forever.

\makeatletter
\begingroup
\setlength{\@fptop}{0pt}
\setlength{\@fpsep}{12pt plus 2pt minus 2pt}
\setlength{\@fpbot}{0pt plus 1fil}
\makeatother

\begin{table}[htbp]
\centering
\setlength{\tabcolsep}{3.5pt}
\renewcommand{\arraystretch}{1.05}
\caption{Training and held-out streams in the broad RL mixture, ordered by first public release. A dash means that family has no held-out stream in this run.}
\label{tab:broad-mix}
\scriptsize
\begin{tabular*}{\linewidth}{@{\extracolsep{\fill}}lll@{}}
\toprule
\textbf{Task family} & \textbf{Training split(s)} & \textbf{Held-out validation split(s)} \\
\midrule
FanOutQA~\citep{zhu-etal-2024-fanoutqa} & Dev evidence, contexts up to 400K characters & Dev evidence, contexts $\geq$800K characters \\
Ada-LEval~\citep{wang2024adalevalevaluatinglongcontextllms} & Best Answer, 8K-32K train split & -- \\
BABILong~\citep{kuratov2024babilong} & QA3, 32K/64K/128K & QA1, 128K \\
NoCha~\citep{karpinska2024thousandpairsnovelchallenge} & -- & NoCha evaluation stream (20 examples) \\
{\color{black}GraphWalks~\citep{openai2025graphwalks}} & {\color{black}128K \textcolor{black}{tokens} and shorter} & {\color{black}256K--1M \textcolor{black}{tokens}} \\
MRCRv2~\citep{vodrahalli2024michelangelolongcontextevaluations} & Two needles, 32K--64K & Eight needles, 2M--4M \\
LongProc~\citep{ye2025longprocbenchmarkinglongcontextlanguage} & TOM tracking, 2K and 8K & -- \\
PrefEval~\citep{zhao2025do} & Implicit persona, 10 intervening turns & Implicit persona, 150 turns; explicit, 300 turns \\
PersonaMem~\citep{jiang2025know} & v2, 32K, all question types; v3 (128-event cap) & v2 benchmark, 128K \\
ManyIH~\citep{jaroslawicz2025instructionsllmsfollowonce} & Private instruction-following, at most 5 tags & Evaluation examples with $\geq 15$ tags \\
BrowseComp-Plus~\citep{chen2025browsecompplusfairtransparentevaluation} & Train split, 50 documents/query & Test split, 200 documents/query \\
LoongRL~\citep{wang2026loongrl} & MuSiQue, plain, train & MuSiQue, keychain \\
OOLONG~\citep{bertsch2025oolongevaluatinglongcontext} & TREC-Coarse and spam, 8K--64K & TREC, 100K--131K; spam, 128K \\
OOLONG-Pairs~\citep{zhang2026recursivelanguagemodels} & 4K and 8K & -- \\
LongBench-Pro~\citep{chen2026longbenchprorealisticcomprehensive} & English, 16K/32K/64K & English, 256K \\
LongRLVR~\citep{chen2026longrlvr} & Llama, 16K/32K, train & Qwen2.5-7B subset, test \\
OBLIQ-Bench~\citep{tchuindjo2026obliqbenchexposingoverlookedbottlenecks} & Writing, pool of 300 & Congress, pool of 200 \\
GoLongRL~\citep{lv2026golongrlcapabilityorientedlongcontext} & English table ($\geq$20K); English summaries & -- \\
\bottomrule
\end{tabular*}
\end{table}

\begin{table}[htbp]
\centering
\small
\setlength{\tabcolsep}{3.2pt}
\renewcommand{\arraystretch}{1.12}
\caption{Average benchmark results for the broad-mixture RL ablation and main experiment models. BrowseComp-Plus and OBLIQ-Bench are excluded from the average because they are included in the larger training mix.}
\label{tab:broad-ablation-results}
\resizebox{\linewidth}{!}{%
\begin{tabular}{@{}l*{6}{c}|c@{}}
\toprule
\textbf{Model} & \textbf{LongReason} & \textbf{LooGLE} & \textbf{CTC} & \textbf{OOLONG-Real} & \textbf{CorpusQA} & \textbf{ELB} & \textbf{Avg.} \\
\midrule
Qwen3-30B-A3B (base) & 44.6 & 18.3 & 1.1 & 15.1 & 11.2 & 25.6 & 19.3 \\
+ RL (short tasks) & 74.9 & 21.7 & \textbf{2.2} & 19.3 & 17.8 & 26.8 & 27.1 \\
+ RL (synthetic tasks) & 69.6 & 28.2 & 1.6 & 18.6 & \textbf{23.7} & \textbf{44.0} & 31.0 \\
\rowcolor{black!6}[\tabcolsep][\tabcolsep]
+ RL (broad task suite) & \textbf{83.1} & \textbf{28.5} & 2.0 & \textbf{19.7} & 20.4 & 35.8 & \textbf{31.6} \\
\bottomrule
\end{tabular}%
}
\end{table}

\begin{table}[htbp]
\centering
\small
\setlength{\tabcolsep}{3pt}
\renewcommand{\arraystretch}{1.1}
\caption{Average trajectory statistics for the models in Table~\ref{tab:broad-ablation-results}, averaged equally across its six benchmarks (the Table~\ref{tab:eval-envs} environments excluding BrowseComp-Plus and OBLIQ-Bench).}
\label{tab:broad-ablation-stats}
\begin{tabular}{@{}lrrcc@{}}
\toprule
\textbf{Model} & \textbf{Root turns} & \textbf{Sub-calls} & \textbf{Root tokens} & \textbf{Sub-call tokens} \\
\midrule
Base & 17.0 & 438.8 & 229K / 14K & 572K / 24K \\
\rowcolor{black!6}[\tabcolsep][\tabcolsep]
+ RL (short tasks) & 12.8 & 179.4 & 170K / 16K & 285K / 7K \\
\rowcolor{black!6}[\tabcolsep][\tabcolsep]
+ RL (synthetic tasks) & 12.6 & 774.6 & 230K / 20K & 429K / 23K \\
\rowcolor{black!6}[\tabcolsep][\tabcolsep]
+ RL (broad task suite) & 13.3 & 422.6 & 245K / 19K & 250K / 30K \\
\bottomrule
\end{tabular}
\end{table}
\clearpage
\endgroup

%% file: appendix/synthetic_environments.tex
\clearpage
\section{Details on synthetic environments to target LM harness behavior}
\label{appx:synthetic-environments}

\definecolor{synthcodeline}{RGB}{204,214,224}
\definecolor{synthcodetitle}{RGB}{231,240,248}
\definecolor{synthcodeink}{RGB}{36,69,99}
\definecolor{synthcodecomment}{RGB}{105,116,127}
\makeatletter
\def\PY@reset{\let\PY@it=\relax \let\PY@bf=\relax%
    \let\PY@ul=\relax \let\PY@tc=\relax%
    \let\PY@bc=\relax \let\PY@ff=\relax}
\def\PY@tok#1{\csname PY@tok@#1\endcsname}
\def\PY@toks#1+{\ifx\relax#1\empty\else%
    \PY@tok{#1}\expandafter\PY@toks\fi}
\def\PY@do#1{\PY@bc{\PY@tc{\PY@ul{%
    \PY@it{\PY@bf{\PY@ff{#1}}}}}}}
\def\PY#1#2{\PY@reset\PY@toks#1+\relax+\PY@do{#2}}

\@namedef{PY@tok@c}{\def\PY@tc##1{\textcolor[rgb]{0.00,0.50,0.00}{##1}}}
\@namedef{PY@tok@cp}{\def\PY@tc##1{\textcolor[rgb]{0.00,0.00,1.00}{##1}}}
\@namedef{PY@tok@k}{\def\PY@tc##1{\textcolor[rgb]{0.00,0.00,1.00}{##1}}}
\@namedef{PY@tok@ow}{\def\PY@tc##1{\textcolor[rgb]{0.00,0.00,1.00}{##1}}}
\@namedef{PY@tok@kt}{\def\PY@tc##1{\textcolor[rgb]{0.17,0.57,0.69}{##1}}}
\@namedef{PY@tok@nc}{\def\PY@tc##1{\textcolor[rgb]{0.17,0.57,0.69}{##1}}}
\@namedef{PY@tok@s}{\def\PY@tc##1{\textcolor[rgb]{0.64,0.08,0.08}{##1}}}
\@namedef{PY@tok@gh}{\let\PY@bf=\textbf}
\@namedef{PY@tok@gu}{\let\PY@bf=\textbf}
\@namedef{PY@tok@ge}{\let\PY@it=\textit}
\@namedef{PY@tok@gs}{\let\PY@bf=\textbf}
\@namedef{PY@tok@ges}{\let\PY@bf=\textbf\let\PY@it=\textit}
\@namedef{PY@tok@gp}{\let\PY@bf=\textbf}
\@namedef{PY@tok@err}{\def\PY@bc##1{{\setlength{\fboxsep}{\string -\fboxrule}\fcolorbox[rgb]{1.00,0.00,0.00}{1,1,1}{\strut ##1}}}}
\@namedef{PY@tok@kc}{\def\PY@tc##1{\textcolor[rgb]{0.00,0.00,1.00}{##1}}}
\@namedef{PY@tok@kd}{\def\PY@tc##1{\textcolor[rgb]{0.00,0.00,1.00}{##1}}}
\@namedef{PY@tok@kn}{\def\PY@tc##1{\textcolor[rgb]{0.00,0.00,1.00}{##1}}}
\@namedef{PY@tok@kp}{\def\PY@tc##1{\textcolor[rgb]{0.00,0.00,1.00}{##1}}}
\@namedef{PY@tok@kr}{\def\PY@tc##1{\textcolor[rgb]{0.00,0.00,1.00}{##1}}}
\@namedef{PY@tok@sa}{\def\PY@tc##1{\textcolor[rgb]{0.64,0.08,0.08}{##1}}}
\@namedef{PY@tok@sb}{\def\PY@tc##1{\textcolor[rgb]{0.64,0.08,0.08}{##1}}}
\@namedef{PY@tok@sc}{\def\PY@tc##1{\textcolor[rgb]{0.64,0.08,0.08}{##1}}}
\@namedef{PY@tok@dl}{\def\PY@tc##1{\textcolor[rgb]{0.64,0.08,0.08}{##1}}}
\@namedef{PY@tok@sd}{\def\PY@tc##1{\textcolor[rgb]{0.64,0.08,0.08}{##1}}}
\@namedef{PY@tok@s2}{\def\PY@tc##1{\textcolor[rgb]{0.64,0.08,0.08}{##1}}}
\@namedef{PY@tok@se}{\def\PY@tc##1{\textcolor[rgb]{0.64,0.08,0.08}{##1}}}
\@namedef{PY@tok@sh}{\def\PY@tc##1{\textcolor[rgb]{0.64,0.08,0.08}{##1}}}
\@namedef{PY@tok@si}{\def\PY@tc##1{\textcolor[rgb]{0.64,0.08,0.08}{##1}}}
\@namedef{PY@tok@sx}{\def\PY@tc##1{\textcolor[rgb]{0.64,0.08,0.08}{##1}}}
\@namedef{PY@tok@sr}{\def\PY@tc##1{\textcolor[rgb]{0.64,0.08,0.08}{##1}}}
\@namedef{PY@tok@s1}{\def\PY@tc##1{\textcolor[rgb]{0.64,0.08,0.08}{##1}}}
\@namedef{PY@tok@ss}{\def\PY@tc##1{\textcolor[rgb]{0.64,0.08,0.08}{##1}}}
\@namedef{PY@tok@ch}{\def\PY@tc##1{\textcolor[rgb]{0.00,0.50,0.00}{##1}}}
\@namedef{PY@tok@cm}{\def\PY@tc##1{\textcolor[rgb]{0.00,0.50,0.00}{##1}}}
\@namedef{PY@tok@cpf}{\def\PY@tc##1{\textcolor[rgb]{0.00,0.50,0.00}{##1}}}
\@namedef{PY@tok@c1}{\def\PY@tc##1{\textcolor[rgb]{0.00,0.50,0.00}{##1}}}
\@namedef{PY@tok@cs}{\def\PY@tc##1{\textcolor[rgb]{0.00,0.50,0.00}{##1}}}

\def\PYZbs{\char`\\}
\def\PYZus{\char`\_}
\def\PYZob{\char`\{}
\def\PYZcb{\char`\}}
\def\PYZca{\char`\^}
\def\PYZam{\char`\&}
\def\PYZlt{\char`\<}
\def\PYZgt{\char`\>}
\def\PYZsh{\char`\#}
\def\PYZpc{\char`\%}
\def\PYZdl{\char`\$}
\def\PYZhy{\char`\-}
\def\PYZsq{\char`\'}
\def\PYZdq{\char`\"}
\def\PYZti{\char`\~}
\def\PYZat{@}
\def\PYZlb{[}
\def\PYZrb{]}
\makeatother
\newcommand{\synthprogramtitle}[1]{\par\smallskip\noindent
  \colorbox{synthcodetitle}{\parbox{\dimexpr\linewidth-2\fboxsep\relax}{%
    \sffamily\small\bfseries\color{synthcodeink} LM program: #1}}\par}
\DefineVerbatimEnvironment{synthprogram}{Verbatim}{
  fontsize=\small, formatcom=\color{black!88},
  frame=single, framerule=0.5pt, framesep=4pt,
  rulecolor=\color{synthcodeline}, numbers=left,
  numbersep=6pt,
  xleftmargin=1em, xrightmargin=0.2em, tabsize=4,
  commandchars=\\\{\}
}
\newtcolorbox{synthpairbox}[1]{
  colback=blue!2, colframe=synthcodeink, colbacktitle=synthcodetitle,
  coltitle=synthcodeink, coltext=black!88, title={#1},
  fonttitle=\footnotesize\sffamily\bfseries, fontupper=\footnotesize,
  boxrule=0.8pt, arc=1.5mm, left=7pt, right=7pt, top=6pt, bottom=6pt,
  before skip=7pt, after skip=7pt}
\newcommand{\synthpairheading}[1]{\noindent\colorbox{synthcodetitle}{\parbox{\dimexpr\linewidth-2\fboxsep\relax}{\sffamily\small\bfseries\color{synthcodeink}#1}}\par\smallskip}
\newcommand{\synthpairlabel}[1]{\colorbox{blue!8}{\scriptsize\sffamily\bfseries\color{synthcodeink}#1}}

In \S\ref{sec:synth-envs}, we proposed a viable strategy to efficiently train LM harness behavior by exploiting composition. Assuming we have a long and difficult task that the harness can decompose into solvable sub-problems, we show that we can target an individual LM's behavior over a synthetic environment such that it mimics the actual behavior on a real environment. But within this synthetic environment, we can modify the tools and sub-calls to be fake, or even the task to be much easier for training.

We created paired synthetic environments for each task in Table~\ref{tab:eval-envs}, and post-trained our models on them via RL. Some environments use \textit{fake sub-calls} that route to oracle answers, while others use real sub-calls over simpler sub-tasks. We also argued that these environments were a proxy to learn desirable behavior via hints, as the goal was to distill a certain trajectory into the model. We generated an SFT dataset over these environments that we used to train \textbf{RLM-27B}. 

This section of the appendix provides details for each of these environments, as well as more information on the trained SFT corpus. We start by detailing each synthetic environment, how it was generated based on the target behavior, and how the environment is structured.

\subsection{BrowseComp-Plus: a generated encyclopedia}
\label{appx:synth-browsecomp}
\synthprogramtitle{BrowseComp-Plus / retrieve and verify}
\begin{synthprogram}
\PY{c+c1}{\PYZsh{} Search clue combinations and pivot on linked entities.}
\PY{c+c1}{\PYZsh{} Verify the final answer against every clue.}
\PY{k}{def} \PY{n+nf}{solve}\PY{p}{(}\PY{n}{context}\PY{p}{,} \PY{n}{clues}\PY{p}{)}\PY{p}{:}
    \PY{n}{docs} \PY{o}{=} \PY{n}{split\PYZus{}documents}\PY{p}{(}\PY{n}{context}\PY{p}{)}
    \PY{n}{index} \PY{o}{=} \PY{n}{build\PYZus{}inverted\PYZus{}index}\PY{p}{(}\PY{n}{docs}\PY{p}{)}
    \PY{n}{hits} \PY{o}{=} \PY{n}{search\PYZus{}clue\PYZus{}combinations}\PY{p}{(}\PY{n}{index}\PY{p}{,} \PY{n}{clues}\PY{p}{)}
    \PY{n}{facts} \PY{o}{=} \PY{p}{[}
        \PY{n}{llm\PYZus{}query}\PY{p}{(}\PY{n}{format\PYZus{}screen\PYZus{}prompt}\PY{p}{(}\PY{n}{clues}\PY{p}{,} \PY{n}{doc}\PY{p}{)}\PY{p}{)}
        \PY{k}{for} \PY{n}{doc} \PY{o+ow}{in} \PY{n}{hits}\PY{p}{[}\PY{p}{:}\PY{n}{MAX\PYZus{}DOCS}\PY{p}{]}
    \PY{p}{]}
    \PY{n}{bridges} \PY{o}{=} \PY{n}{extract\PYZus{}entities}\PY{p}{(}\PY{n}{facts}\PY{p}{)}
    \PY{n}{pivots} \PY{o}{=} \PY{n}{search\PYZus{}entity\PYZus{}links}\PY{p}{(}\PY{n}{index}\PY{p}{,} \PY{n}{bridges}\PY{p}{)}
    \PY{n}{checks} \PY{o}{=} \PY{p}{[}
        \PY{n}{llm\PYZus{}query}\PY{p}{(}\PY{n}{format\PYZus{}check\PYZus{}prompt}\PY{p}{(}\PY{n}{clues}\PY{p}{,} \PY{n}{doc}\PY{p}{)}\PY{p}{)}
        \PY{k}{for} \PY{n}{doc} \PY{o+ow}{in} \PY{n}{pivots}\PY{p}{[}\PY{p}{:}\PY{n}{MAX\PYZus{}DOCS}\PY{p}{]}
    \PY{p}{]}
    \PY{k}{return} \PY{n}{submit}\PY{p}{(}\PY{n}{verify\PYZus{}all\PYZus{}clues}\PY{p}{(}\PY{n}{clues}\PY{p}{,} \PY{n}{facts} \PY{o}{+} \PY{n}{checks}\PY{p}{)}\PY{p}{)}
\end{synthprogram}
{\textbf{Real task and current split.} We target multi-hop retrieval tasks in
\href{https://huggingface.co/datasets/Tevatron/browsecomp-plus}{BrowseComp-Plus}, which act over the shared
\href{https://huggingface.co/datasets/Tevatron/browsecomp-plus-corpus}{100,195-document web corpus}. The synthetic environment is designed around fake multi-hop retrieval where sub-calling information over a document provides instantaneous oracle answers.}

{\textbf{Generation and source.} The
\href{https://huggingface.co/datasets/a1zhang/synthetic-rlm-data-clean\#browsecomp_plus}{1,835 exported trajectories}
use a \emph{new procedurally-generated corpus}, disjoint from documents sampled from BrowseComp-Plus or another external corpus. The training worktree constructs a random graph with 50 regions, 1,000 towns, 625 organizations, about 5,900 people and 7,000 works. Each document is generated from templates and padded with extra text, and the multi-hop relationships are determined by the graph. The environment generates 2,000 tasks.}

{\textbf{How fake sub-calling was implemented.} The assumption on BrowseComp-Plus and other multi-hop retrieval is that the program will quickly filter then apply sub-LLM calls over raw documents to find information. We can quickly simulate this behavior by replacing sub-call invocations with a function that returns the summarized or relevant information when a document is pasted as input into that query. We add additional realistic constraints to mimic context window limitations and context rot such as a 64K-character input limit, skipped documents beyond eight per call, and roughly 3--13\% load-dependent judgment errors. The environment also caps the sub-call budget at 120 to prevent expensive trajectories from being accepted.}
\input{appendix/synthetic_examples/browsecomp_plus}

\subsection{CorpusQA: generated statistical releases}
\label{appx:synth-corpusqa}
\synthprogramtitle{CorpusQA / reconcile and reduce}
\begin{synthprogram}
\PY{c+c1}{\PYZsh{} Infer table layout and units.}
\PY{c+c1}{\PYZsh{} Keep only the newest report for each month.}
\PY{k}{def} \PY{n+nf}{solve}\PY{p}{(}\PY{n}{context}\PY{p}{,} \PY{n}{question}\PY{p}{)}\PY{p}{:}
    \PY{n}{sample} \PY{o}{=} \PY{n}{sample\PYZus{}releases}\PY{p}{(}\PY{n}{context}\PY{p}{)}
        \PY{n}{schema} \PY{o}{=} \PY{n}{llm\PYZus{}query}\PY{p}{(}\PY{n}{infer\PYZus{}schema}\PY{p}{(}\PY{n}{question}\PY{p}{,} \PY{n}{sample}\PY{p}{)}\PY{p}{)}
    \PY{n}{rows} \PY{o}{=} \PY{p}{[}\PY{p}{]}
    \PY{k}{for} \PY{n}{release} \PY{o+ow}{in} \PY{n}{split\PYZus{}releases}\PY{p}{(}\PY{n}{context}\PY{p}{)}\PY{p}{:}
        \PY{n}{prompt} \PY{o}{=} \PY{n}{format\PYZus{}row\PYZus{}prompt}\PY{p}{(}\PY{n}{schema}\PY{p}{,} \PY{n}{question}\PY{p}{,} \PY{n}{release}\PY{p}{)}
        \PY{n}{rows}\PY{o}{.}\PY{n}{extend}\PY{p}{(}\PY{n}{parse\PYZus{}rows}\PY{p}{(}\PY{n}{llm\PYZus{}query}\PY{p}{(}\PY{n}{prompt}\PY{p}{)}\PY{p}{)}\PY{p}{)}
    \PY{n}{rows} \PY{o}{=} \PY{n}{convert\PYZus{}units}\PY{p}{(}\PY{n}{rows}\PY{p}{,} \PY{n}{schema}\PY{p}{)}
    \PY{n}{rows} \PY{o}{=} \PY{n}{newest\PYZus{}report\PYZus{}per\PYZus{}month}\PY{p}{(}\PY{n}{rows}\PY{p}{)}
    \PY{k}{return} \PY{n}{calculate\PYZus{}statistic}\PY{p}{(}\PY{n}{rows}\PY{p}{,} \PY{n}{question}\PY{p}{)}
\end{synthprogram}
{\textbf{Real benchmark.} The evaluation split is CorpusQA's long tasks (1m-4m) in \texttt{real\_estate\_en}: 81 questions over 34 U.S. residential-construction releases. Each question combines values across reports, sometimes requiring unit conversion and use of a value revised in a later release.}

{\textbf{Synthetic corpus.} The \href{https://huggingface.co/datasets/a1zhang/synthetic-rlm-data-clean\#corpusqa}{853 exported trajectories} use documents generated from AI models, and are disjoint from the real CorpusQA corpus. In the default mix, 60\% of tasks use generated monthly series with trends, seasonality and revisions; each series is rendered as 4-90 fictional statistical releases. The other 40\% use fictional company and school reports with generated financial or enrollment values.  Questions and gold answers are computed from the generated data. The generator borrows the real task's structure, but the documents and records are new.}

{\textbf{Sub-calls.} In these exported trajectories, Qwen3.8-27B reads selected passages from the generated documents. The root extracts values across releases, reconciles revisions and units, and computes the requested statistic in code. This retains the real task's division of work while using entirely generated source documents.}
\input{appendix/synthetic_examples/corpusqa}

\subsection{CTC-Bench: new compositions of external source items}
\label{appx:synth-ctc}
\synthprogramtitle{CTC grouping example, label then cluster}
\begin{synthprogram}
\PY{c+c1}{\PYZsh{} The sub\PYZhy{}call names a topic. It does not see labels or assign groups.}
\PY{c+c1}{\PYZsh{} The root clusters those names, then a second sub\PYZhy{}call checks.}
\PY{k}{def} \PY{n+nf}{solve}\PY{p}{(}\PY{n}{context}\PY{p}{,} \PY{n}{question}\PY{p}{)}\PY{p}{:}
    \PY{n}{abstracts} \PY{o}{=} \PY{n}{split\PYZus{}abstracts}\PY{p}{(}\PY{n}{context}\PY{p}{)}
    \PY{n}{k} \PY{o}{=} \PY{n}{parse\PYZus{}num\PYZus{}groups}\PY{p}{(}\PY{n}{question}\PY{p}{)}
    \PY{n}{topics} \PY{o}{=} \PY{p}{[}\PY{p}{]}
    \PY{k}{for} \PY{n}{batch} \PY{o+ow}{in} \PY{n}{batches}\PY{p}{(}\PY{n}{abstracts}\PY{p}{,} \PY{n}{BATCH\PYZus{}SIZE}\PY{p}{)}\PY{p}{:}
        \PY{c+c1}{\PYZsh{} \PYZdq{}Name each abstract\PYZsq{}s topic in one sentence.\PYZdq{}}
        \PY{n}{prompt} \PY{o}{=} \PY{n}{topic\PYZus{}prompt}\PY{p}{(}\PY{n}{batch}\PY{p}{)}
        \PY{n}{topics}\PY{o}{.}\PY{n}{extend}\PY{p}{(}\PY{n}{parse\PYZus{}topics}\PY{p}{(}
            \PY{n}{llm\PYZus{}query}\PY{p}{(}\PY{n}{prompt}\PY{p}{)}\PY{p}{)}\PY{p}{)}
    \PY{n}{draft} \PY{o}{=} \PY{n}{cluster\PYZus{}by\PYZus{}topic}\PY{p}{(}\PY{n}{topics}\PY{p}{,} \PY{n}{k}\PY{p}{)}
    \PY{c+c1}{\PYZsh{} \PYZdq{}Move any abstract whose topic does not match its group.\PYZdq{}}
    \PY{n}{prompt} \PY{o}{=} \PY{n}{check\PYZus{}prompt}\PY{p}{(}\PY{n}{k}\PY{p}{,} \PY{n}{draft}\PY{p}{)}
    \PY{n}{checked} \PY{o}{=} \PY{n}{parse\PYZus{}groups}\PY{p}{(}\PY{n}{llm\PYZus{}query}\PY{p}{(}\PY{n}{prompt}\PY{p}{)}\PY{p}{)}
    \PY{k}{return} \PY{n}{validate\PYZus{}partition}\PY{p}{(}\PY{n}{checked}\PY{p}{,} \PY{n}{k}\PY{p}{,} \PY{n}{abstracts}\PY{p}{)}
\end{synthprogram}
{\textbf{Real task and current split.} We target five splits of
\href{https://huggingface.co/datasets/PrasannSinghal/ctc-suite-eval}{CTC-Bench}: QDMatch--HotpotQA at 256K ($O_{T}(N^{2})$), paraphrase absence at 32K ($O_{T}(N^{2})$), contradiction at 1M ($O_{T}(N^{2})$), outlier at 1M ($O_{T}(N^{2})$) and grouping at 32K ($O_{T}(NM)$). These tasks are long, and fall under the superlinear complexity class.}

{\textbf{Generation and source.} The
\href{https://huggingface.co/datasets/a1zhang/synthetic-rlm-data-clean\#ctcbench}{1,799 exported trajectories}
are new items from the benchmark generators over the
\href{https://huggingface.co/datasets/PrasannSinghal/ctc-seed-pools}{CTC seed pools}.
Each cell keeps the original question and replaces the eval items. Pools are filtered against exact text and 13-grams from all 23 published evaluation families at every rung.}
{\color{black}
\begin{itemize}\setlength{\itemsep}{3pt}\setlength{\parsep}{0pt}
\item \textbf{QDMatch--HotpotQA.} The eval split is the published HotpotQA question--document with different documents. Synthetic rows recombine HotpotQA~\citep{yang2018hotpotqadatasetdiverseexplainable} \emph{train} questions, passages, and distractors into that same layout.
\item \textbf{Paraphrase absence.} The eval split is cross-corpus absence with paraphrased twins: two shuffled PubMed lists, where the answer is the sentences present in only one. The paper's exact-copy version uses an identical twin; this split replaces that twin with a paraphrase. Synthetic rows pair PubMed abstract sentences with paraphrases written offline by \qwennew{}, then drop selected counterparts, so the missing sentences are new.
\item \textbf{Contradiction.} The eval split is PubMed claim sentences with a few contradictory twins hidden among fillers, and the answer is every contradicting pair. Synthetic rows mix PubMed claims, edited contradictory twins, and filler claims from abstracts outside the real split.
\item \textbf{Outlier.} The eval split is the Wikipedia scale-\(M\) ladder, not the fixed-\(M\) control and not the Amazon-review outlier: chunks from several articles, and the answer is the chunks from the rarest article. Synthetic rows mix Wikipedia windows from several articles around a different minority article.
\item \textbf{Grouping.} The eval split is OpenAlex~\citep{priem2022openalexfullyopenindexscholarly} abstracts partitioned by a concept in the OpenAlex hierarchy; the question states \(k\), and the concept is not printed in the abstract. Synthetic rows sample non-overlapping OpenAlex abstracts and take the target partition from their concept labels, which stay hidden the same way.
\end{itemize}

\input{appendix/synthetic_examples/ctcbench}

\subsection{ExpertLongBench: generated case files and forms}
\label{appx:synth-elb}
\synthprogramtitle{ExpertLongBench / fill an evidence-backed form}
\begin{synthprogram}
\PY{c+c1}{\PYZsh{} Extract evidence for each checklist field.}
\PY{c+c1}{\PYZsh{} Leave unsupported fields empty.}
\PY{k}{def} \PY{n+nf}{solve}\PY{p}{(}\PY{n}{context}\PY{p}{,} \PY{n}{question}\PY{p}{)}\PY{p}{:}
    \PY{n}{kind} \PY{o}{=} \PY{n}{llm\PYZus{}query}\PY{p}{(}\PY{n}{probe\PYZus{}case}\PY{p}{(}\PY{n}{question}\PY{p}{,} \PY{n}{context}\PY{p}{[}\PY{p}{:}\PY{n}{PROBE\PYZus{}CHARS}\PY{p}{]}\PY{p}{)}\PY{p}{)}
    \PY{k}{if} \PY{n}{kind} \PY{o}{==} \PY{l+s+s2}{\PYZdq{}}\PY{l+s+s2}{recipe}\PY{l+s+s2}{\PYZdq{}}\PY{p}{:}
        \PY{n}{evidence} \PY{o}{=} \PY{n}{extract\PYZus{}short\PYZus{}record}\PY{p}{(}\PY{n}{context}\PY{p}{)}
    \PY{k}{else}\PY{p}{:}
        \PY{n}{evidence} \PY{o}{=} \PY{p}{[}\PY{p}{]}
        \PY{k}{for} \PY{n}{part} \PY{o+ow}{in} \PY{n}{split\PYZus{}case\PYZus{}parts}\PY{p}{(}\PY{n}{context}\PY{p}{)}\PY{p}{:}
            \PY{n}{prompt} \PY{o}{=} \PY{n}{format\PYZus{}evidence\PYZus{}prompt}\PY{p}{(}\PY{n}{question}\PY{p}{,} \PY{n}{part}\PY{p}{)}
            \PY{n}{evidence}\PY{o}{.}\PY{n}{extend}\PY{p}{(}\PY{n}{parse\PYZus{}evidence}\PY{p}{(}\PY{n}{llm\PYZus{}query}\PY{p}{(}\PY{n}{prompt}\PY{p}{)}\PY{p}{)}\PY{p}{)}
    \PY{n}{fields} \PY{o}{=} \PY{n}{fill\PYZus{}checklist}\PY{p}{(}\PY{n}{question}\PY{p}{,} \PY{n}{evidence}\PY{p}{)}
    \PY{k}{return} \PY{n}{remove\PYZus{}unsupported\PYZus{}values}\PY{p}{(}\PY{n}{fields}\PY{p}{,} \PY{n}{evidence}\PY{p}{)}
\end{synthprogram}
{\textbf{Real task and current split.} We target several splits of
\href{https://huggingface.co/datasets/launch/ExpertLongBench}{ExpertLongBench}:
T01LegalMDS, T03MaterialSEG, T04EduPAE and T06HealthCNG, which differ in their source domains. The synthetic environment is designed around following the same evidence extraction decompositional program, where we use oracle sub-calls that return the correct field values from the supplied case text.}

{\textbf{Generation and source.} The
\href{https://huggingface.co/datasets/a1zhang/synthetic-rlm-data-clean\#expertlongbench}{1,306 exported trajectories}
contain 377 dockets, 352 tutoring dialogues, 299 visits and 278 recipes. We used Python templates to generate all of the source content. Dockets range from 108K to 1.2M characters and mix dated filings, amendments, parties and fields the record does not support. Tutor dialogues follow a hidden sequence of correct, partial and incorrect responses about invented entities and facts. Visits sample clinical facts together with tempting unsupported additions.}

{\textbf{How fake sub-calling was implemented.} The target decompositional behavior on ExpertLongBench is to split a long evidence file, then sub-query LMs to find checklist field values corresponding to evidence. We simulate an oracle reader that, when provided with a case-file text, answers from the supplied evidence and injects omission, misreading and confabulation. The source content in all settings is entirely synthetic.}
\input{appendix/synthetic_examples/expertlongbench}

\subsection{LooGLE-v2: procedural filings, repositories and logs}
\label{appx:synth-loogle}
\synthprogramtitle{LooGLE-v2 / inspect and compute}
\begin{synthprogram}
\PY{c+c1}{\PYZsh{} Probe the source type and retrieve relevant evidence.}
\PY{c+c1}{\PYZsh{} Run the analysis that matches the source.}
\PY{k}{def} \PY{n+nf}{solve}\PY{p}{(}\PY{n}{context}\PY{p}{,} \PY{n}{question}\PY{p}{)}\PY{p}{:}
        \PY{n}{plan} \PY{o}{=} \PY{n}{llm\PYZus{}query}\PY{p}{(}\PY{n}{probe}\PY{p}{(}\PY{n}{question}\PY{p}{,} \PY{n}{context}\PY{p}{)}\PY{p}{)}
    \PY{n}{excerpts} \PY{o}{=} \PY{n}{search\PYZus{}context}\PY{p}{(}\PY{n}{context}\PY{p}{,} \PY{n}{plan}\PY{o}{.}\PY{n}{search\PYZus{}terms}\PY{p}{)}
    \PY{n}{facts} \PY{o}{=} \PY{n}{llm\PYZus{}query}\PY{p}{(}\PY{n}{format\PYZus{}extraction\PYZus{}prompt}\PY{p}{(}\PY{n}{question}\PY{p}{,} \PY{n}{excerpts}\PY{p}{)}\PY{p}{)}
    \PY{k}{if} \PY{n}{plan}\PY{o}{.}\PY{n}{source} \PY{o}{==} \PY{l+s+s2}{\PYZdq{}}\PY{l+s+s2}{filings}\PY{l+s+s2}{\PYZdq{}}\PY{p}{:}
        \PY{n}{answer} \PY{o}{=} \PY{n}{calculate\PYZus{}financial\PYZus{}values}\PY{p}{(}\PY{n}{facts}\PY{p}{,} \PY{n}{question}\PY{p}{)}
    \PY{k}{elif} \PY{n}{plan}\PY{o}{.}\PY{n}{source} \PY{o}{==} \PY{l+s+s2}{\PYZdq{}}\PY{l+s+s2}{repository}\PY{l+s+s2}{\PYZdq{}}\PY{p}{:}
        \PY{n}{answer} \PY{o}{=} \PY{n}{traverse\PYZus{}call\PYZus{}graph}\PY{p}{(}\PY{n}{facts}\PY{p}{,} \PY{n}{question}\PY{p}{)}
    \PY{k}{elif} \PY{n}{plan}\PY{o}{.}\PY{n}{source} \PY{o}{==} \PY{l+s+s2}{\PYZdq{}}\PY{l+s+s2}{event\PYZus{}log}\PY{l+s+s2}{\PYZdq{}}\PY{p}{:}
        \PY{n}{answer} \PY{o}{=} \PY{n}{count\PYZus{}or\PYZus{}compare\PYZus{}events}\PY{p}{(}\PY{n}{facts}\PY{p}{,} \PY{n}{question}\PY{p}{)}
    \PY{k}{else}\PY{p}{:}
        \PY{n}{answer} \PY{o}{=} \PY{n}{compare\PYZus{}file\PYZus{}versions}\PY{p}{(}\PY{n}{facts}\PY{p}{,} \PY{n}{question}\PY{p}{)}
    \PY{k}{return} \PY{n}{verify\PYZus{}against\PYZus{}context}\PY{p}{(}\PY{n}{answer}\PY{p}{,} \PY{n}{excerpts}\PY{p}{)}
\end{synthprogram}

{\textbf{Real task and current split.} We target all
\href{https://huggingface.co/datasets/MuLabPKU/LooGLE-v2}{LooGLE-v2}
tasks with $\geq$500K-token contexts. The synthetic environment constructs an environment that can be solved with the same inspect-and-compute program over procedural source text, with real sub-calls doing the local reads.}

{\textbf{Generation and source.} The
\href{https://huggingface.co/datasets/a1zhang/synthetic-rlm-data-clean\#loogle}{397 exported trajectories}
use \texttt{loogle-strategy-synth}. We generate the source text procedurally. Finance instances generate 3--9 fictional 10-Ks with random values, unit traps, label variants and prior-year-first columns. Code instances plant a call graph in a fictional Python repository, including same-name functions and misleading strings and comments. Game instances render random CS-style events in a log grammar close to the benchmark. Version instances insert, delete or modify files across a pair of generated trees. Contexts run from about 0.28M to 8.5M characters, and gold follows the planted numbers, graph, event ledger or file edits.}
\input{appendix/synthetic_examples/loogle}

\subsection{LongReason: expanded questions from external reasoning corpora}
\label{appx:synth-longreason}
\synthprogramtitle{LongReason / retrieve the full scenario}
\begin{synthprogram}
\PY{c+c1}{\PYZsh{} Find the expanded problem among filler.}
\PY{c+c1}{\PYZsh{} Delegate the reasoning once it is recovered.}
\PY{k}{def} \PY{n+nf}{solve}\PY{p}{(}\PY{n}{context}\PY{p}{,} \PY{n}{question}\PY{p}{,} \PY{n}{options}\PY{p}{)}\PY{p}{:}
    \PY{n}{terms} \PY{o}{=} \PY{n}{llm\PYZus{}query}\PY{p}{(}\PY{n}{search\PYZus{}prompt}\PY{p}{(}\PY{n}{question}\PY{p}{,} \PY{n}{context}\PY{p}{[}\PY{p}{:}\PY{n}{PROBE\PYZus{}CHARS}\PY{p}{]}\PY{p}{)}\PY{p}{)}
    \PY{n}{paragraphs} \PY{o}{=} \PY{n}{search\PYZus{}context}\PY{p}{(}\PY{n}{context}\PY{p}{,} \PY{n}{terms}\PY{p}{)}
    \PY{n}{scenario} \PY{o}{=} \PY{n}{reject\PYZus{}filler}\PY{p}{(}\PY{n}{paragraphs}\PY{p}{,} \PY{n}{terms}\PY{p}{)}
    \PY{n}{prompt} \PY{o}{=} \PY{n}{format\PYZus{}reasoning\PYZus{}prompt}\PY{p}{(}\PY{n}{question}\PY{p}{,} \PY{n}{options}\PY{p}{,} \PY{n}{scenario}\PY{p}{)}
    \PY{n}{answer} \PY{o}{=} \PY{n}{llm\PYZus{}query}\PY{p}{(}\PY{n}{prompt}\PY{p}{)}
    \PY{k}{return} \PY{n}{validate\PYZus{}option}\PY{p}{(}\PY{n}{answer}\PY{p}{,} \PY{n}{options}\PY{p}{)}
\end{synthprogram}
{\textbf{Real task and current split.} We target the \href{https://huggingface.co/datasets/lz1bytedance/LongReason}{LongReason}
\texttt{128k} split, which involves a short reasoning problem expanded into several paragraphs and buried in a long background. The synthetic environment is designed around the same retrieve-then-reason LM program: the root finds the expanded scenario, and in tail-recursion fashion, invokes a sub-call to solve the recovered question in isolation.}

{\textbf{Generation and source.} The
\href{https://huggingface.co/datasets/a1zhang/synthetic-rlm-data-clean\#longreason}{1,821 exported records}
comprise 356 root trajectories and 1,465 reasoning sub-calls. Source multiple-choice questions come from
\href{https://github.com/yuweihao/reclor}{ReClor}~\citep{yu2020reclor} train and validation,
\href{https://github.com/csitfun/LogiQA2.0}{LogiQA 2.0 MRC}~\citep{liu2023logiqa} train, dev and test,
and \href{https://github.com/google-deepmind/AQuA}{AQuA-RAT}~\citep{ling-etal-2017-program} train.
We use \qwennew{} to add a fictional setting, rewrite each source sentence or sentence pair into a 90--140-word paragraph, and reword questions in these data sources while keeping the options and the source label. To emulate the scattering of the real question / reasoning within a larger corpus of text, we add 
\href{https://huggingface.co/datasets/mlfoundations/dclm-baseline-1.0}{DCLM-baseline}~\citep{li2024datacomplm} web-text chunks to the context but maintain the overall order of the paragraphs within the context.}
\input{appendix/synthetic_examples/longreason}

\subsection{OBLIQ: generated problems with hidden reasoning techniques}
\label{appx:synth-obliq}
\synthprogramtitle{OBLIQ / rank reasoning techniques}
\begin{synthprogram}
\PY{c+c1}{\PYZsh{} Ask readers to describe each candidate\PYZsq{}s technique.}
\PY{c+c1}{\PYZsh{} Rank descriptions, then flag a shortlist.}
\PY{k}{def} \PY{n+nf}{solve}\PY{p}{(}\PY{n}{context}\PY{p}{,} \PY{n}{question}\PY{p}{)}\PY{p}{:}
    \PY{n}{descriptions} \PY{o}{=} \PY{p}{[}\PY{p}{]}
    \PY{k}{for} \PY{n}{batch} \PY{o+ow}{in} \PY{n}{candidate\PYZus{}batches}\PY{p}{(}\PY{n}{context}\PY{p}{,} \PY{n}{BATCH\PYZus{}SIZE}\PY{p}{)}\PY{p}{:}
        \PY{n}{prompt} \PY{o}{=} \PY{n}{format\PYZus{}technique\PYZus{}prompt}\PY{p}{(}\PY{n}{question}\PY{p}{,} \PY{n}{batch}\PY{p}{)}
        \PY{n}{descriptions}\PY{o}{.}\PY{n}{extend}\PY{p}{(}\PY{n}{parse\PYZus{}descriptions}\PY{p}{(}\PY{n}{llm\PYZus{}query}\PY{p}{(}\PY{n}{prompt}\PY{p}{)}\PY{p}{)}\PY{p}{)}
    \PY{n}{ranked} \PY{o}{=} \PY{n}{rank\PYZus{}by\PYZus{}technique}\PY{p}{(}\PY{n}{question}\PY{p}{,} \PY{n}{descriptions}\PY{p}{)}
    \PY{n}{shortlist} \PY{o}{=} \PY{n}{ranked}\PY{p}{[}\PY{p}{:}\PY{n}{SHORTLIST\PYZus{}SIZE}\PY{p}{]}
    \PY{n}{flags} \PY{o}{=} \PY{n}{llm\PYZus{}query}\PY{p}{(}\PY{n}{format\PYZus{}flag\PYZus{}prompt}\PY{p}{(}\PY{n}{question}\PY{p}{,} \PY{n}{shortlist}\PY{p}{)}\PY{p}{)}
    \PY{k}{return} \PY{n}{complete\PYZus{}ranking}\PY{p}{(}\PY{n}{ranked}\PY{p}{,} \PY{n}{parse\PYZus{}flags}\PY{p}{(}\PY{n}{flags}\PY{p}{)}\PY{p}{)}
\end{synthprogram}
{\textbf{Real task and current split.} We target
\href{https://huggingface.co/datasets/dianetc/OBLIQ-Bench}{OBLIQ-Bench math} with the entire 3,508-entry math corpus rather than a subset, scored with nDCG@50. The synthetic environment is designed around the generic program of ranking candidates using language models, and provides an oracle to rank correctly under the sub-calls invoked by this program.}

{\textbf{Generation and source.} The
\href{https://huggingface.co/datasets/a1zhang/synthetic-rlm-data-clean\#obliq}{474 exported trajectories}
use fake documents with a fake ranking. Using \qwennew{}, we generate 125 techniques in 12 families from generic seed lists, about 25 stories per technique with a one-sentence solution sketch, rewrites of those stories in distant settings, and same-topic decoys that require a different technique. The pool has 5,054 problems and averages 229 characters. A query is one sampled problem. Its candidate pool holds 1--30 positives from the same technique plus decoys and random problems, roughly 2,500--3,100 entries, under opaque IDs.}

{\textbf{How fake sub-calling was implemented.} The assumption on OBLIQ is that the program batches candidates, asks a reader to describe each one's technique, ranks those descriptions, and then flags a shortlist. We replace sub-calls with a programmatic oracle that returns this information given a batch of candidate questions from the task. To simulate context rot, description accuracy falls as the batch grows. The simulator also injects incomplete descriptions, format drift and flags written as prose.}
\input{appendix/synthetic_examples/obliq}

\subsection{OOLONG-Real: generated tabletop sessions}
\label{appx:synth-oolong-real}
\synthprogramtitle{OOLONG-Real / extract and aggregate events}
\begin{synthprogram}
\PY{c+c1}{\PYZsh{} Probe transcript structure and extract source\PYZhy{}linked events.}
\PY{c+c1}{\PYZsh{} Reconcile actors before aggregating.}
\PY{k}{def} \PY{n+nf}{solve}\PY{p}{(}\PY{n}{context}\PY{p}{,} \PY{n}{question}\PY{p}{)}\PY{p}{:}
        \PY{n}{schema} \PY{o}{=} \PY{n}{llm\PYZus{}query}\PY{p}{(}\PY{n}{probe}\PY{p}{(}\PY{n}{question}\PY{p}{,} \PY{n}{context}\PY{p}{)}\PY{p}{)}
    \PY{n}{events} \PY{o}{=} \PY{p}{[}\PY{p}{]}
    \PY{k}{for} \PY{n}{chunk} \PY{o+ow}{in} \PY{n}{transcript\PYZus{}chunks}\PY{p}{(}\PY{n}{context}\PY{p}{,} \PY{n}{schema}\PY{p}{)}\PY{p}{:}
        \PY{n}{prompt} \PY{o}{=} \PY{n}{format\PYZus{}event\PYZus{}prompt}\PY{p}{(}\PY{n}{question}\PY{p}{,} \PY{n}{schema}\PY{p}{,} \PY{n}{chunk}\PY{p}{)}
        \PY{n}{events}\PY{o}{.}\PY{n}{extend}\PY{p}{(}\PY{n}{parse\PYZus{}events}\PY{p}{(}\PY{n}{llm\PYZus{}query}\PY{p}{(}\PY{n}{prompt}\PY{p}{)}\PY{p}{)}\PY{p}{)}
    \PY{n}{events} \PY{o}{=} \PY{n}{reconcile\PYZus{}actors\PYZus{}and\PYZus{}duplicates}\PY{p}{(}\PY{n}{events}\PY{p}{,} \PY{n}{schema}\PY{p}{)}
    \PY{k}{return} \PY{n}{aggregate\PYZus{}events}\PY{p}{(}\PY{n}{events}\PY{p}{,} \PY{n}{question}\PY{p}{)}
\end{synthprogram}
{\textbf{Real task and current split.} We target a split of 
\href{https://huggingface.co/datasets/oolongbench/oolong-real}{OOLONG-Real} (\texttt{dnd/test}), containing 577 questions over ten three-episode windows at 175K tokens per window. The synthetic environment keeps the extract-and-aggregate program over generated tabletop sessions, with real sub-calls reading the synthetic transcripts.}

{\textbf{Generation and source.} The
\href{https://huggingface.co/datasets/a1zhang/synthetic-rlm-data-clean\#oolong_real}{995 exported trajectories} use a synthetically generated set of transcripts. Python samples an invented cast and one to three episodes of 700--1,400 lines, then renders scenes, banter, combat, dice outcomes and spell actions from hidden event tables. Similar to the real task, the queries ask for roll counts, criticals, totals, modes, spell order and counts, upcasts or cantrips. Episode delimiters, speaker notation, cast-list styles and date formats vary. Guest appearances, similar player names and stale player-to-character mappings force the program to inspect the source.}
\input{appendix/synthetic_examples/oolong_real}

\input{appendix/synthetic/synthetic_overlap}

\input{appendix/synthetic/runtime_comparison}

\input{appendix/synthetic/recursion_composition}

%% file: appendix/synthetic_examples/browsecomp_plus.tex
\begin{synthpairbox}{BrowseComp-Plus and its synthetic variant}
\begin{minipage}[t]{0.477\linewidth}\raggedright\synthpairheading{BrowseComp-Plus}
\synthpairlabel{Corpus} Search the shared corpus of 100,195 real web documents. The harness must retrieve pages and connect evidence across them.\par\smallskip
\synthpairlabel{Question} An author: -	published their first work sometime between the years 2000 and 2010 -	lived in Granada and Berlin -	started a business in face painting -	hosted a festival on literature. The same author published a book about a dog. What is the name of this author?
\par\smallskip\synthpairlabel{Corpus trail} The example cites \emph{The Loudest Sound and Nothing} (2008-08-21), which identifies its author, Clare Wigfall. The harness searches other pages in the corpus to check the remaining clues about her.\par\smallskip
\par\smallskip\synthpairlabel{Gold} Clare Wigfall
\end{minipage}\hfill\begin{minipage}[t]{0.477\linewidth}\raggedright\synthpairheading{Synthetic variant}
\synthpairlabel{Corpus} First generate an entity graph, then render it as about 21K documents. The harness searches these generated pages; the fake reader answers only from text pasted into a sub-call.\par\smallskip
\synthpairlabel{Question} I am looking for a musical work that fits the following clues.
- It was not well received at first.
- The composer was born in the month of June.
- It was completed between 1946 and 1953.
- It centres on a celestial event.
- The composer was educated at the Naroumont Academy for Girls.
- The composer's father was a person who worked with horses.
Please provide the title.
\par\smallskip\synthpairlabel{Corpus trail (generated pages)}\par
Doc A, \emph{The Lantern of Seloukov} (work article): a symphony composed by Brigid Maridale, completed in 1947.\par
Doc B, \emph{Brigid Maridale} (person article): supplies the composer clues, including her school.\par
Graph link: work $\xrightarrow{\text{composed by}}$ composer $\xrightarrow{\text{educated at}}$ school.
\par\smallskip\synthpairlabel{Recorded target} The Lantern of Seloukov
\end{minipage}
\end{synthpairbox}

%% file: appendix/synthetic_examples/corpusqa.tex
\begin{synthpairbox}{CorpusQA and its synthetic variant}
\color{black}
\begin{minipage}[t]{0.477\linewidth}\raggedright\synthpairheading{CorpusQA}
\synthpairlabel{Corpus} Thirty-four monthly U.S.\ residential-construction releases. Each release reprints a window of months, and a later release can revise an earlier month.\par\smallskip
\synthpairlabel{Question} From 2022/08 to 2025/05, how long is the longest continuous rise in the seasonally adjusted total of U.S.\ housing units under construction? If a month is printed more than once, the newest release prevails.\par\smallskip
\synthpairlabel{Trace} Read that series from every release in the window. Where a month is reprinted, keep the later value, then count consecutive month-over-month increases. The longest run has length 3.\par\smallskip
\synthpairlabel{Gold} 3
\end{minipage}\hfill\begin{minipage}[t]{0.477\linewidth}\raggedright\synthpairheading{Synthetic variant}
\synthpairlabel{Corpus} Nothing is copied from those releases. Python draws a monthly series with a trend, a seasonal pattern and revisions, then renders each month as its own fictional release. This example is a retail-shaped series retitled \emph{Monthly Network Traffic}, with the measure named enrolments, the category All and the adjustment smoothed: 80 releases.\par\smallskip
\synthpairlabel{Question} From 2018/08 to 2025/03, how long is the longest continuous rise in smoothed enrolments for All? Three rises in a row count as 3, and the newest release prevails.\par\smallskip
\synthpairlabel{Trace} Read the smoothed All panel from each release and keep the newest print of every month. All 80 months are present. The longest rise is 2021-11 through 2022-02 (442{,}564, 492{,}945, 502{,}004, 512{,}694): three increases, and no longer run.\par\smallskip
\synthpairlabel{Recorded target} 3.0
\end{minipage}
\end{synthpairbox}

%% file: appendix/synthetic_examples/ctcbench.tex
\begin{synthpairbox}{Grouping, on the eval cell and on a new OpenAlex sample}
\color{black}
\begin{minipage}[t]{0.477\linewidth}\raggedright\synthpairheading{CTC grouping}
\synthpairlabel{Context} The 32K OpenAlex cell. One published item has 20 abstracts and asks for 15 groups. The concept label is not in the text. Two of the abstracts are a paper on electrocatalytic nitrate-to-ammonia conversion and a review of zinc in disease.\par\smallskip
\synthpairlabel{Sub-call} The root pastes a batch and asks for one topic sentence per id. For those two, the reader can return ``electrocatalytic conversion of nitrate to ammonia'' and ``the role of zinc in health and disease.'' It is not asked for groups, and the label is not there to copy.\par\smallskip
\synthpairlabel{Root} It clusters matching topic sentences, so the ammonia abstracts share a group and the zinc review does not join them. A second sub-call reads that draft and can move an id.\par\smallskip
\synthpairlabel{Gold} The OpenAlex partition: the ammonia abstracts are one group, and the zinc abstract is in another.
\end{minipage}\hfill\begin{minipage}[t]{0.477\linewidth}\raggedright\synthpairheading{Synthetic variant}
\synthpairlabel{Context} A new draw of OpenAlex abstracts. An abstract is dropped if its exact text or any 13-gram appears in a published eval family. The concept label is still withheld. Ids are often strings, and the instruction often sits below the list rather than above it.\par\smallskip
\synthpairlabel{Sub-call} The same topic question, but the pasted text is these new abstracts. The reader again returns a topic sentence per id. It still cannot see a label, and it never sees the eval abstracts.\par\smallskip
\synthpairlabel{Root} The same cluster step, then the same check sub-call, with this draw's \(k\). Nothing in the program refers to the eval item.\par\smallskip
\synthpairlabel{Recorded target} The concept partition of this draw. The groups are different abstracts from the eval item; the two \texttt{llm\_query} questions are the same.
\end{minipage}
\end{synthpairbox}

%% file: appendix/synthetic_examples/expertlongbench.tex
\begin{synthpairbox}{ExpertLongBench and its synthetic variant}
\color{black}
\begin{minipage}[t]{0.477\linewidth}\raggedright\synthpairheading{ExpertLongBench}
\synthpairlabel{Corpus} A real tutoring transcript. The checklist scores the student's latest turn: whether it is correct, whether the turn before was also wrong, and what the tutor should say next.\par\smallskip
\synthpairlabel{Question} A rare plant grows only on one mountaintop. How does it produce viable seeds? The student first says it reproduces by spores. The tutor says it reproduces by seeds. The student then says it needs water and sunlight for photosynthesis.\par\smallskip
\synthpairlabel{Trace} The spore answer is rejected. Water and sunlight are real growth conditions, but they are not the conditions for producing seeds at altitude, so the last turn is only partial. The previous turn was also not accepted, so the mistake is repeated and the correction is direct.\par\smallskip
\synthpairlabel{Gold} Partially correct; repeated mistake: Yes
\end{minipage}\hfill\begin{minipage}[t]{0.477\linewidth}\raggedright\synthpairheading{Synthetic variant}
\synthpairlabel{Corpus} Python invents a table of cycle and diet for four fictional species, then renders it as a tutor--student dialogue. No real transcript is used. A fake reader returns a turn-by-turn ledger, and the root fills the checklist from that ledger.\par\smallskip
\synthpairlabel{Question} The tutor has established that the Selzoron has a 9-day cycle and eats seeds, and asks for the cycle and diet of the Elirena. The student says: maybe a 3-day cycle, and water plants for the diet.\par\smallskip
\synthpairlabel{Trace} The hidden table says the Elirena's cycle is 3 days and its diet is algae. ``Water plants'' is only a vague name for algae, so the cycle is right and the diet is not: partial. Earlier Selzoron answers were also only partial, so the mistake is repeated and the correction states the diet.\par\smallskip
\synthpairlabel{Recorded target} Partially correct; repeated mistake: Yes
\end{minipage}
\end{synthpairbox}

%% file: appendix/synthetic_examples/loogle.tex
\begin{synthpairbox}{LooGLE-v2 and its synthetic variant}
\color{black}
\begin{minipage}[t]{0.477\linewidth}\raggedright\synthpairheading{LooGLE-v2}
\synthpairlabel{Corpus} Real Form 10-K filings, one file per company-year, concatenated as \texttt{<FILE:>} blocks.\par\smallskip
\synthpairlabel{Question} Among Eli Lilly, Exxon Mobil, Intuitive Surgical, Coca-Cola and Abbott, which has the highest working capital in FY2023? Working capital is current assets minus current liabilities. Answer with the company's list number.\par\smallskip
\synthpairlabel{Trace} Split out the five FY2023 filings and read current assets and current liabilities from each consolidated balance sheet. Subtracting, Exxon Mobil is the largest.\par\smallskip
\synthpairlabel{Gold} 2 (Exxon Mobil)
\end{minipage}\hfill\begin{minipage}[t]{0.477\linewidth}\raggedright\synthpairheading{Synthetic variant}
\synthpairlabel{Corpus} Python builds five fictional companies and writes a 10-K for each, with statements that add up, a chosen unit, and a chosen column order. The names and figures are not taken from real filings.\par\smallskip
\synthpairlabel{Question} Rank Varden Industries, Westcott Aerospace, Dunmore Financial, Oakhurst Health and Foxhall Beverages by FY2024 long-term debt, highest first. What is Westcott's rank?\par\smallskip
\synthpairlabel{Trace} Read the FY2024 long-term-debt line in each generated filing: Varden 256{,}078, Westcott 203{,}620, Dunmore 98{,}927.4, Oakhurst 16{,}394, Foxhall 9{,}283.4. Sorting those figures places Westcott second.\par\smallskip
\synthpairlabel{Recorded target} 2
\end{minipage}
\end{synthpairbox}

%% file: appendix/synthetic_examples/longreason.tex
\begin{synthpairbox}{LongReason and its synthetic variant}
\color{black}
\begin{minipage}[t]{0.477\linewidth}\raggedright\synthpairheading{LongReason}
\synthpairlabel{Corpus} A short puzzle is expanded into several paragraphs and buried in a long unrelated background. What remains is a multiple-choice question.\par\smallskip
\synthpairlabel{Question} In Green Town, Jack, Emily and Tom are each honest or a liar. Jack says Emily and Tom are both liars. Emily says she is honest. Tom says Emily is a liar. How many are liars?\par\smallskip
\synthpairlabel{Trace} If Jack were honest, both others would be liars, but then Tom's claim that Emily is a liar would be true. So Jack is a liar, and at least one of the others is honest. Emily and Tom contradict each other, so exactly one of them is a liar. That is two liars.\par\smallskip
\synthpairlabel{Gold} C (2)
\end{minipage}\hfill\begin{minipage}[t]{0.477\linewidth}\raggedright\synthpairheading{Synthetic variant}
\synthpairlabel{Corpus} A LogiQA item is rewritten into two paragraphs set at the Corvane Institute of Logical Studies. The options stay the same. The paragraphs are scattered, in order, through DCLM web-text chunks, about 462{,}000 characters.\par\smallskip
\synthpairlabel{Question} Which option exemplifies the accidental fallacy as those seminars define it? Option B: speeding is illegal, so ambulances should not exceed the speed limit.\par\smallskip
\synthpairlabel{Trace} Search the filler for Corvane and ``accidental fallacy.'' The recovered definition is a general rule used to deny a legitimate exception. Option B does that. Option A flips a cause, option C treats seahorses as horses, and option D generalizes from a few rainy days.\par\smallskip
\synthpairlabel{Recorded target} B
\end{minipage}
\end{synthpairbox}

%% file: appendix/synthetic_examples/obliq.tex
\begin{synthpairbox}{OBLIQ and its synthetic variant}
\color{black}
\begin{minipage}[t]{0.477\linewidth}\raggedright\synthpairheading{OBLIQ}
\synthpairlabel{Corpus} A pool of 3{,}508 math problems. A relevant problem shares a reasoning technique with the query; the technique need not show up in the topic or the vocabulary.\par\smallskip
\synthpairlabel{Question} Show that any finite integral domain is a field. Rank the pool by that technique.\par\smallskip
\synthpairlabel{Trace} The two relevance-2 documents are Kyiv undergraduate problems 2013.2 and 2013.10. Both ask whether a finite ring can exist in which every element---2013.10 restricts this to every nonzero element---is the square of some other element. Those are the documents to rank first.\par\smallskip
\synthpairlabel{Gold} Kyiv 2013.2 and 2013.10 (both relevance 2)
\end{minipage}\hfill\begin{minipage}[t]{0.477\linewidth}\raggedright\synthpairheading{Synthetic variant}
\synthpairlabel{Corpus} The base model writes 125 techniques, about 25 stories for each, rewrites in distant settings, and same-topic decoys that need a different technique. No OBLIQ problem is shown to the generator. This query is one sampled story, ranked against 3{,}053 candidates under opaque ids.\par\smallskip
\synthpairlabel{Question} A stage designer arranges three kinds of props and must prove that total visual complexity---sizes plus pairwise interference---is minimized when two kinds have the same size.\par\smallskip
\synthpairlabel{Trace} The hidden technique (UVW) rewrites a symmetric inequality in elementary symmetric polynomials and reduces it to the case where two variables are equal. Document \texttt{d59efaf629b} is a three-variable variance story with that same reduction. Stories that only share the surface use other techniques. A fake reader returns a technique sketch, and the root ranks by the sketch.\par\smallskip
\synthpairlabel{Recorded target} \texttt{d59efaf629b}
\end{minipage}
\end{synthpairbox}

%% file: appendix/synthetic_examples/oolong_real.tex
\begin{synthpairbox}{OOLONG-Real and its synthetic variant}
\color{black}
\begin{minipage}[t]{0.477\linewidth}\raggedright\synthpairheading{OOLONG-Real}
\synthpairlabel{Corpus} Transcripts of Critical Role campaign 1. This question uses one three-episode window: episodes 2, 16 and 64.\par\smallskip
\synthpairlabel{Question} How many Investigation rolls are there across all three episodes?\par\smallskip
\synthpairlabel{Trace} Tie each speaker to a character, then count rolls whose type is Investigation. One of them is Matt calling for an investigation roll. The window contains 9.\par\smallskip
\synthpairlabel{Gold} 9
\end{minipage}\hfill\begin{minipage}[t]{0.477\linewidth}\raggedright\synthpairheading{Synthetic variant}
\synthpairlabel{Corpus} Python samples an invented cast and renders one episode from a hidden event table: scenes, banter, combat and dice. No Critical Role transcript is used, and real cast names are rejected. Here Cassius is the DM and Greer plays Torquinth.\par\smallskip
\synthpairlabel{Question} How many Wisdom Save rolls are there in this episode?\par\smallskip
\synthpairlabel{Trace} Readers list every roll with its speaker and type. Eight are Wisdom Saves, including Torquinth's 21 after Cassius calls for a wisdom saving throw. A second read, shifted by a few lines, adds other rolls but no further Wisdom Saves.\par\smallskip
\synthpairlabel{Recorded target} 8
\end{minipage}
\end{synthpairbox}

%% file: appendix/synthetic/synthetic_overlap.tex
\subsection{Measured differences in task descriptions}
\label{appx:synthetic-overlap}
\textbf{Comparing synthetically generated tasks with real tasks.} We measure the similarity between the synthetic and real tasks used in Table~\ref{tab:synthetic-text-overlap}. Some instances (e.g. BrowseComp-Plus) use entirely different questions and context, while others that use more general aggregation queries like CTC-Bench share similar questions and context, but shuffled and otherwise modified to be different.

\textbf{Metrics.} Let $G_n(x)$ be the set of contiguous word $n$-grams in description $x$, $S$ the training descriptions and $Q$ the real questions. $J_3$ averages
$\max_{s\in S}|G_3(q)\cap G_3(s)|/|G_3(q)\cup G_3(s)|$
over $q\in Q$. $H_{13}$ counts real questions sharing at least one exact 13-word sequence with any training description. \textcolor{black}{It is a question-level hit count, not the number of matching 13-gram sequences.}

\begin{table}[!htbp]
\centering\small\color{black}
\setlength{\tabcolsep}{6pt}\renewcommand{\arraystretch}{1.12}
\begin{tabular}{@{}lrrrr@{}}
\toprule
Environment & Real $|Q|$ & Train $|S|$ & $J_3$ & $H_{13}$ \\
\midrule
BrowseComp-Plus & 830 & 1,322 & 0.021 & 0 \\
CorpusQA & 81 & 853 & 0.315 & 81 \\
CTC: QDMatch 256K & 125 & 136 & 1.000 & 125 \\
CTC: absence 32K & 100 & 516 & 1.000 & 100 \\
CTC: contradiction 1M & 125 & 151 & 1.000 & 125 \\
CTC: outlier 1M & 125 & 209 & 1.000 & 125 \\
CTC: grouping 32K & 60 & 185 & 0.965 & 60 \\
ExpertLongBench & 350 & 678 & 0.748 & 350 \\
LooGLE-v2 & 206 & 327 & 0.479 & 197 \\
LongReason & 794 & 256 & 0.022 & 0 \\
OBLIQ math & 151 & 443 & 0.011 & 0 \\
OOLONG-Real & 577 & 383 & 0.810 & 232 \\
\bottomrule
\end{tabular}
\caption{\textcolor{black}{Task-description overlap between the clean training
export and updated real evaluation splits. $|S|$ counts distinct descriptions,
not trajectories. Larger $J_3$ means more shared wording.} \textcolor{black}{$H_{13}$ counts the number of real questions with at least one exact 13-word sequence shared with any training description; it does not count all matching sequences and does not by itself measure contamination.}}
\label{tab:synthetic-text-overlap}
\end{table}

\textbf{What differs.} BrowseComp-Plus, LongReason and generated OBLIQ-Bench have low description overlap with their synthetic environment counterparts. CorpusQA retains the revision rule; ExpertLongBench retains the checklist instructions; LooGLE and OOLONG retain question templates. For four CTC-Bench splits, the native task specification is identical despite different item collections. The high similarity on CTC-Bench is due to re-used queries (they are generic queries over aggregation patterns), while the context being investigated differs.

%% file: appendix/synthetic/runtime_comparison.tex
\subsection{Runtime comparison of synthetic environments}
\label{appx:synth-runtime}
\definecolor{rtgood}{RGB}{84,128,98}
\definecolor{rtbad}{RGB}{160,88,88}
In Tables~\ref{tab:runtime-rt-rlm} and \ref{tab:runtime-rt-base}, we compare wall-clock rollout time on each real benchmark with its synthetic counterpart, for RLM-27B and \qwennew{}, using up to $n=100$ random samples per benchmark (all if fewer). Rollouts use training sampling (thinking off, 6,000 output tokens, temperature 1.0) at a concurrency of 8 per vLLM server, and we report p50, p99 and mean seconds per rollout. Parentheses give the synthetic time as a multiple of the real time (green: faster, red: not faster). $^\dagger$ marks benchmarks whose synthetic variant answers sub-calls with a programmatic oracle; all other synthetic environments call the real sub-LM. 

\begin{table}[h]
\centering\small
\caption{Seconds per rollout for RLM-27B, real benchmark vs.\ synthetic environment (lower is bold; the parenthesis is synthetic/real).}
\label{tab:runtime-rt-rlm}
\setlength{\tabcolsep}{4pt}
\begin{tabular}{@{}l rr||rr||rr@{}}
\toprule
& \multicolumn{2}{c||}{p50} & \multicolumn{2}{c||}{p99} & \multicolumn{2}{c}{Mean} \\
Benchmark & Real & Synth. & Real & Synth. & Real & Synth. \\
\midrule
CTC-Bench & 219 & \textbf{145}\,{\scriptsize\textcolor{rtgood}{(0.66$\times$)}} & \textbf{1021} & 1639\,{\scriptsize\textcolor{rtbad}{(1.61$\times$)}} & 276 & \textbf{222}\,{\scriptsize\textcolor{rtgood}{(0.81$\times$)}} \\
OBLIQ$^\dagger$ & 213 & \textbf{47}\,{\scriptsize\textcolor{rtgood}{(0.22$\times$)}} & 521 & \textbf{78}\,{\scriptsize\textcolor{rtgood}{(0.15$\times$)}} & 229 & \textbf{50}\,{\scriptsize\textcolor{rtgood}{(0.22$\times$)}} \\
BrowseComp-Plus$^\dagger$ & 195 & \textbf{79}\,{\scriptsize\textcolor{rtgood}{(0.40$\times$)}} & 698 & \textbf{394}\,{\scriptsize\textcolor{rtgood}{(0.56$\times$)}} & 240 & \textbf{96}\,{\scriptsize\textcolor{rtgood}{(0.40$\times$)}} \\
ExpertLongBench$^\dagger$ & 45 & \textbf{12}\,{\scriptsize\textcolor{rtgood}{(0.27$\times$)}} & 167 & \textbf{33}\,{\scriptsize\textcolor{rtgood}{(0.20$\times$)}} & 61 & \textbf{17}\,{\scriptsize\textcolor{rtgood}{(0.27$\times$)}} \\
LooGLE-v2 & 75 & \textbf{54}\,{\scriptsize\textcolor{rtgood}{(0.72$\times$)}} & 253 & \textbf{180}\,{\scriptsize\textcolor{rtgood}{(0.71$\times$)}} & 84 & \textbf{61}\,{\scriptsize\textcolor{rtgood}{(0.73$\times$)}} \\
OOLONG-Real & \textbf{147} & 180\,{\scriptsize\textcolor{rtbad}{(1.22$\times$)}} & 482 & \textbf{429}\,{\scriptsize\textcolor{rtgood}{(0.89$\times$)}} & 203 & \textbf{187}\,{\scriptsize\textcolor{rtgood}{(0.92$\times$)}} \\
CorpusQA & 86 & \textbf{63}\,{\scriptsize\textcolor{rtgood}{(0.73$\times$)}} & 446 & \textbf{161}\,{\scriptsize\textcolor{rtgood}{(0.36$\times$)}} & 110 & \textbf{67}\,{\scriptsize\textcolor{rtgood}{(0.61$\times$)}} \\
LongReason & \textbf{32} & 35\,{\scriptsize\textcolor{rtbad}{(1.10$\times$)}} & \textbf{84} & 110\,{\scriptsize\textcolor{rtbad}{(1.31$\times$)}} & \textbf{36} & 39\,{\scriptsize\textcolor{rtbad}{(1.08$\times$)}} \\
\bottomrule
\end{tabular}
\end{table}

\begin{table}[h]
\centering\small
\caption{Seconds per rollout for \qwennew{}, real benchmark vs.\ synthetic environment (lower is bold; the parenthesis is synthetic/real).}
\label{tab:runtime-rt-base}
\setlength{\tabcolsep}{4pt}
\begin{tabular}{@{}l rr||rr||rr@{}}
\toprule
& \multicolumn{2}{c||}{p50} & \multicolumn{2}{c||}{p99} & \multicolumn{2}{c}{Mean} \\
Benchmark & Real & Synth. & Real & Synth. & Real & Synth. \\
\midrule
CTC-Bench & 157 & \textbf{90}\,{\scriptsize\textcolor{rtgood}{(0.57$\times$)}} & 1009 & \textbf{601}\,{\scriptsize\textcolor{rtgood}{(0.60$\times$)}} & 219 & \textbf{154}\,{\scriptsize\textcolor{rtgood}{(0.70$\times$)}} \\
OBLIQ$^\dagger$ & 144 & \textbf{119}\,{\scriptsize\textcolor{rtgood}{(0.83$\times$)}} & 487 & \textbf{233}\,{\scriptsize\textcolor{rtgood}{(0.48$\times$)}} & 158 & \textbf{120}\,{\scriptsize\textcolor{rtgood}{(0.76$\times$)}} \\
BrowseComp-Plus$^\dagger$ & 87 & \textbf{56}\,{\scriptsize\textcolor{rtgood}{(0.64$\times$)}} & 572 & \textbf{370}\,{\scriptsize\textcolor{rtgood}{(0.65$\times$)}} & 135 & \textbf{73}\,{\scriptsize\textcolor{rtgood}{(0.54$\times$)}} \\
ExpertLongBench$^\dagger$ & 28 & \textbf{20}\,{\scriptsize\textcolor{rtgood}{(0.72$\times$)}} & \textbf{182} & 219\,{\scriptsize\textcolor{rtbad}{(1.21$\times$)}} & \textbf{37} & 47\,{\scriptsize\textcolor{rtbad}{(1.27$\times$)}} \\
LooGLE-v2 & 48 & \textbf{40}\,{\scriptsize\textcolor{rtgood}{(0.83$\times$)}} & \textbf{275} & 313\,{\scriptsize\textcolor{rtbad}{(1.14$\times$)}} & 82 & \textbf{56}\,{\scriptsize\textcolor{rtgood}{(0.68$\times$)}} \\
OOLONG-Real & 86 & \textbf{68}\,{\scriptsize\textcolor{rtgood}{(0.79$\times$)}} & \textbf{394} & 579\,{\scriptsize\textcolor{rtbad}{(1.47$\times$)}} & \textbf{107} & 120\,{\scriptsize\textcolor{rtbad}{(1.12$\times$)}} \\
CorpusQA & \textbf{56} & 88\,{\scriptsize\textcolor{rtbad}{(1.58$\times$)}} & \textbf{142} & 263\,{\scriptsize\textcolor{rtbad}{(1.85$\times$)}} & \textbf{65} & 98\,{\scriptsize\textcolor{rtbad}{(1.52$\times$)}} \\
LongReason & \textbf{19} & 20\,{\scriptsize\textcolor{rtbad}{(1.11$\times$)}} & 152 & \textbf{94}\,{\scriptsize\textcolor{rtgood}{(0.62$\times$)}} & 29 & \textbf{26}\,{\scriptsize\textcolor{rtgood}{(0.89$\times$)}} \\
\bottomrule
\end{tabular}
\end{table}

%% file: appendix/synthetic/recursion_composition.tex
\subsection{Concatenated task chain environment - details}
\label{appx:recursion-composition}
\begin{figure}[!htbp]
    \centering
    \captionsetup{font={color=black}}
    \includegraphics[width=1\linewidth]{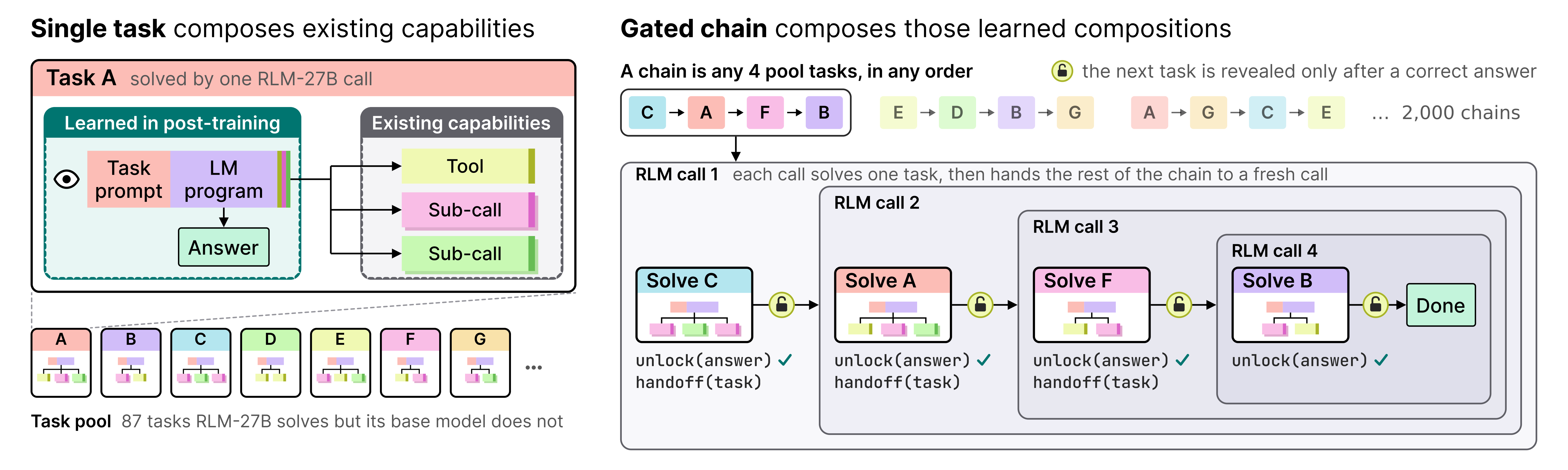}
    \caption{We evaluate 2000 meta-tasks, each sampling (without replacement) four tasks that must be solved in order. \textit{Left:} The selected tasks are drawn from a pool of tasks that the \textbf{RLM-27B} was trained to solve, whose root LM learned in post-training how to decompose the task over existing tools and sub-calls. \textit{Right:} a chain is any four pool tasks in any order. Only the first task is visible, and must be solved. The \texttt{unlock} tool reveals the next task after a correct answer and \texttt{handoff} starts a fresh RLM call on it, so solving a chain composes the learned compositions by recursion.}
    \label{fig:gated-chain}
\end{figure}

\color{black}
In \S\ref{subset:compgen}, we described a task setup in which we took concatenated chains of individual tasks that the \textbf{RLM-27B} learned to solve in \S\ref{sec:synth-envs}, and formed new tasks in which the harness had to solve each task in sequence. We visualize this setup in Figure~\ref{fig:gated-chain}, where a chain consists of four tasks that must be solved in sequence.
\color{black}

\color{black}
\textbf{Task pool and benchmark construction.} We selected 87 evaluation tasks (54 from LooGLE-v2, 29 from LongReason, 4 from CorpusQA) on which, over three runs each, \textbf{RLM-27B} averages at least 0.7 and \qwennew{} at most 0.34. Under the following sampling settings (temperature 0, sub-calls with thinking), \textbf{RLM-27B} scores 97.5\% on these tasks, while \qwennew{} scores 57.7\%. We randomly sample 2000 distinct meta-tasks, each an ordered selection of four distinct pool tasks, and evaluate every condition on this same set.

The model / harness is initially given the first task only. It is prompted to provide an answer via the \texttt{unlock(answer)} tool. If the answer passes, it returns the next task, or \texttt{DONE} after the last one. Otherwise, the tool returns \texttt{LOCKED}, and the model has one more attempt. An answer passes at full score, or at an F1 of at least 0.8 for CorpusQA. We use \texttt{handoff} as the tool to invoke a new RLM call on the returned task, so a solved chain is four RLM calls nested by tail recursion, i.e., an RLM with depth=3. The harnesses we evaluated used the following settings:
\begin{enumerate}
    \item \textcolor{black}{In the \textit{RLM harness} setting, the interface provides \texttt{unlock} but no \texttt{handoff}; therefore, the entire chain is solved within a single root trajectory. The harness receives a short protocol prompt before the first question, listed in the box below, that explains the gated sequence and shows the \texttt{unlock} loop. The call has a budget of 80 turns for the entire chain (4 tasks $\times$ 20 turns), and there is no progress line.} Each harness is given 80 turns, while the \textbf{+ SFT} variant is given 20 per RLM.
    \item In the \textit{+ recursion hint} setting, the harness is given an optimized prompt hint that we list below, and has both the \texttt{unlock} and \texttt{handoff} tools, as well as 80 turns per RLM call.
    \item In the \textit{+ SFT} setting it has both tools, no hint, and 20 turns per RLM call. One line after the question states how many tasks the chain has and which are done.
\end{enumerate}

\textcolor{black}{The first box gives the full prompt of the default \textit{RLM harness} setting. It states the protocol and gives the exact \texttt{unlock} loop, but does not mention recursion or \texttt{handoff}, and it reveals nothing about the tasks.}

\begin{synthpairbox}{Prompt for the default RLM harness setting}
\color{black}
\synthpairlabel{Before the question}\par\smallskip
{\ttfamily\scriptsize\raggedright\frenchspacing\setlength{\parindent}{0pt}
{\hangindent=1.2em\hangafter=1 [MANDATORY PROTOCOL] Your answer is NOT accepted through \char96{}answer[\char34{}content\char34{}]\char96{} until the whole sequence is done. You must solve several tasks one after another in THIS session; the later tasks stay hidden until the current one is answered correctly. Work out your answer to the task below as you normally would, then check it with \char96{}res = unlock(my\char95{}answer)\char96{} (my\char95{}answer = a string in the format the question asks). If \char96{}res\char96{} is a dict, it is the NEXT task: \char96{}res[\char39{}question\char39{}]\char96{} is its question and \char96{}res[\char39{}context\char39{}]\char96{} its data (bind \char96{}context\char95{}0 = res[\char39{}context\char39{}]; context = context\char95{}0\char96{}). Solve it the same way and call \char96{}unlock\char96{} again. If \char96{}res\char96{} starts with \char39{}LOCKED\char39{}, your answer was wrong: recompute and call \char96{}unlock\char96{} once more. When \char96{}res\char96{} == \char39{}DONE\char39{} you are finished: set \char96{}answer[\char34{}content\char34{}] = my\char95{}answer; answer[\char34{}ready\char34{}] = True\char96{}. Do not set \char96{}answer[\char34{}ready\char34{}]\char96{} before \char96{}unlock\char96{} has returned \char39{}DONE\char39{}: that ends the session with the remaining tasks unsolved.\par}
{\hangindent=1.2em\hangafter=1 Use exactly this after you have worked out your answer to a task:\par}
{\hangindent=1.2em\hangafter=1 \char96{}\char96{}\char96{}repl\par}
{\hangindent=1.2em\hangafter=1 my\char95{}answer = \char60{}your final answer to the CURRENT task, as a string in the format its question asks\char62{}\par}
{\hangindent=1.2em\hangafter=1 res = unlock(my\char95{}answer)\par}
{\hangindent=1.2em\hangafter=1 if isinstance(res, dict):\ \ \ \ \ \# the next task\par}
{\hangindent=1.2em\hangafter=1 \ \ \ \ print(res[\char39{}stage\char39{}], res[\char39{}of\char39{}]); print(res[\char39{}question\char39{}])\par}
{\hangindent=1.2em\hangafter=1 \ \ \ \ context\char95{}0 = res[\char39{}context\char39{}]; context = context\char95{}0\ \ \ \ \ \# data for the next task\par}
{\hangindent=1.2em\hangafter=1 else:\par}
{\hangindent=1.2em\hangafter=1 \ \ \ \ print(res)\par}
{\hangindent=1.2em\hangafter=1 \char96{}\char96{}\char96{}\par}
}
\end{synthpairbox}

\color{black}
The second box gives the full prompt of the \textit{+ recursion hint} setting: the text placed before the question, the reminder placed after it, and the descriptions of the two tools, which replace the default ones.

\begin{synthpairbox}{Prompt for the + recursion hint setting}
\color{black}
\synthpairlabel{Before the question}\par\smallskip
{\ttfamily\scriptsize\raggedright\frenchspacing\setlength{\parindent}{0pt}
{\hangindent=1.2em\hangafter=1 [MANDATORY PROTOCOL - read before starting] You are solving a gated sequence of tasks. Your answer is NOT accepted by submitting it in \char96{}answer[\char34{}content\char34{}]\char96{}. Work out your answer as you normally would, then, in a repl cell, run EXACTLY this (fill in the answer):\par}
{\hangindent=1.2em\hangafter=1 \char96{}\char96{}\char96{}repl\par}
{\hangindent=1.2em\hangafter=1 my\char95{}answer = \char60{}your final answer, as a string in the format the question asks\char62{}\par}
{\hangindent=1.2em\hangafter=1 pkg = unlock(my\char95{}answer)\par}
{\hangindent=1.2em\hangafter=1 print(pkg if not isinstance(pkg, dict) else \char123{}k: (v if k in (\char39{}stage\char39{}, \char39{}of\char39{}) else \char39{}...\char39{}) for k, v in pkg.items()\char125{})\par}
{\hangindent=1.2em\hangafter=1 \char96{}\char96{}\char96{}\par}
{\hangindent=1.2em\hangafter=1 Case 1: \char96{}pkg\char96{} is a dict with a \char39{}question\char39{} and \char39{}context\char39{}: the next tasks are in it. Do NOT solve them yourself. Run \char96{}down = handoff(pkg)\char96{} in the next cell. A fresh level solves the rest.\par}
{\hangindent=1.2em\hangafter=1 Case 2: \char96{}pkg\char96{} starts with \char39{}LOCKED\char39{}: your answer was wrong; recompute it and call unlock again (one more attempt).\par}
{\hangindent=1.2em\hangafter=1 Case 3: \char96{}pkg\char96{} == \char39{}DONE\char39{}: nothing is left.\par}
{\hangindent=1.2em\hangafter=1 Only AFTER unlock (and handoff, if there was a pkg) has been called, finish with \char96{}answer[\char34{}content\char34{}] = my\char95{}answer; answer[\char34{}ready\char34{}] = True\char96{}. Never set \char96{}answer[\char34{}ready\char34{}]\char96{} before calling \char96{}unlock\char96{}.\par}
}\smallskip
\synthpairlabel{After the question}\par\smallskip
{\ttfamily\scriptsize\raggedright\frenchspacing\setlength{\parindent}{0pt}
{\hangindent=1.2em\hangafter=1 [REMINDER] Before you finish: you must have run \char96{}pkg = unlock(my\char95{}answer)\char96{} and, if pkg is a dict, \char96{}down = handoff(pkg)\char96{}. Submitting without calling \char96{}unlock\char96{} leaves the remaining tasks unsolved and scores zero.\par}
}\smallskip
\synthpairlabel{Description of \texttt{unlock} in the tool list}\par\smallskip
{\ttfamily\scriptsize\raggedright\frenchspacing\setlength{\parindent}{0pt}
{\hangindent=1.2em\hangafter=1 REQUIRED. unlock(answer): you MUST call this with your final answer (a string in the format the question asks) BEFORE submitting \char96{}answer[\char39{}ready\char39{}]\char96{}. Returns a dict \char123{}\char39{}stage\char39{},\char39{}of\char39{},\char39{}question\char39{},\char39{}context\char39{}\char125{} (the next task: pass it to handoff, do not solve it yourself), \char39{}DONE\char39{}, or \char39{}LOCKED: ...\char39{} (wrong; 2 attempts per task).\par}
}\smallskip
\synthpairlabel{Description of \texttt{handoff} in the tool list}\par\smallskip
{\ttfamily\scriptsize\raggedright\frenchspacing\setlength{\parindent}{0pt}
{\hangindent=1.2em\hangafter=1 REQUIRED whenever unlock returned a dict. handoff(pkg): pass that dict; a fresh RLM level solves the remaining tasks. Never solve the next task yourself.\par}
}
\end{synthpairbox}

\color{black}

\textbf{Training recursion on synthetic traces with surrogate harness training.} The desired behavior of the RLM harness is to launch tail-recursive sub-RLM queries over each successive task in the task chain. We explicitly want to compose this behavior with the learned programs for each of the individual tasks, such that any permutation of tasks in this setup would succeed.

Our synthetic environment therefore focuses on teaching the RLM to solve an arbitrary task, then tail-recurse another RLM over the new task it has to solve. The arbitrary tasks here are synthetic and unrelated to the tasks selected for the real setup. The sub-calls over these tasks are also fake, and return programmatic answers.

\textbf{Teacher trajectories.} \textbf{RLM-27B} at temperature 0.3 solves chains of one to four fake tasks. Its own turns are kept until it submits an answer. We rewrite that submission to store the answer in a variable and call \texttt{unlock} on it. If \texttt{unlock} returns \texttt{LOCKED} with an attempt left, the model continues and revises its answer on its own. Once the answer passes, we append two scripted turns: one that calls \texttt{handoff} on the returned task, and one that submits the stored answer after \texttt{handoff} returns. On 59\% of RLM calls the teacher also receives a note asking it to write the solution as one program. This note is removed from the saved prompts.

\textbf{Training data.} The teacher produced 2{,}809 RLM calls. The final training export contains 2{,}479 distinct full dialogues: 2{,}277 calls whose answer passed the gate and 202 that stopped correctly after failing both attempts. Calls that submitted before calling \texttt{unlock}, never submitted, repeated the same answer on both attempts, or could not be rewritten automatically are excluded. We oversample full dialogues: the 2{,}196 first-attempt successes appear once, the 81 second-attempt successes twice, and the 202 clean stops three times. Thus there are $2196 + 2\times81 + 3\times202 = 2964$ full-dialogue samples. We also add prefixes ending right after an \texttt{unlock} or \texttt{handoff} turn so that these turns are final turns more often; one additional successful teacher call contributes only two \texttt{unlock} prefixes. This gives 7{,}705 training samples: 2{,}964 full dialogues (including repeated copies), 3{,}147 ending at an \texttt{unlock} turn, and 1{,}594 ending at a \texttt{handoff} turn. We fine-tune all weights of \textbf{RLM-27B} and, separately, of \qwennew{} for 100 steps with batch size 4 and learning rate $6\times10^{-6}$. Data is here: \href{https://huggingface.co/datasets/a1zhang/rlm-tail-recursion-sft-data/}{https://huggingface.co/datasets/a1zhang/rlm-tail-recursion-sft-data/}.

\textbf{Example.} The box below shows a training chain of two fake tasks. Turns before the submission are omitted.
\color{black}

\begin{synthpairbox}{A training chain of two fake tasks (abridged)}
\color{black}
\synthpairheading{RLM call 1: task A (ledger)}
\synthpairlabel{Question} The material for this task is stored in \texttt{context}. What is the total value (quantity times unit cost) of all records for `Fujoyo bolt' at site Bahexe? Answer with the number alone (two decimals if it is not a whole number).\par\smallskip
\synthpairlabel{Progress line} \texttt{[SEQUENCE]} You need to solve 2 tasks in order: A, then B. So far you have solved: none. The task above is task A.\par\smallskip
\synthpairlabel{Context} A list of 205 records such as \texttt{TX-6140; Fujoyo bolt; Dopisu; 55 pcs @ 4.26}.\par\smallskip
\synthpairlabel{Turn 2, written by the teacher} The last three lines replace its original submission.
\begin{Verbatim}[fontsize=\scriptsize, xleftmargin=0.6em]
for line in context:
    parts = [p.strip() for p in line.split(';')]
    [...]    # keep 'Fujoyo bolt' at 'Bahexe', add quantity * unit cost
print('count', count, 'total', total)    # prints: count 33 total 37676.08
my_answer = f"{total:.2f}"
result = unlock(my_answer)
print(result if isinstance(result, str)
      else f"unlocked: next task received (stage {result['stage']} of {result['of']})")
\end{Verbatim}
\synthpairlabel{Output} \texttt{unlocked: next task received (stage 2 of 2)}\par\smallskip
\synthpairlabel{Turn 3, scripted}
\begin{Verbatim}[fontsize=\scriptsize, xleftmargin=0.6em]
down = handoff(result)
print('handoff returned', len(down), 'result(s)')
\end{Verbatim}
\synthpairlabel{Turn 4, scripted}
\begin{Verbatim}[fontsize=\scriptsize, xleftmargin=0.6em]
answer["content"] = my_answer
answer["ready"] = True
\end{Verbatim}
\smallskip
\synthpairheading{RLM call 2, started by \texttt{handoff}: task B (reporting chain)}
\synthpairlabel{Question} You are given some data in the variable \texttt{context}. Follow the chain of managers upward starting from Resuvo. Who is at the very top? Reply with just the name.\par\smallskip
\synthpairlabel{Progress line} \texttt{[SEQUENCE]} You need to solve 2 tasks in order: A, then B. So far you have solved: A. The task above is task B.\par\smallskip
\synthpairlabel{Context} A string of 4{,}934 characters with sentences such as ``Pixere's manager is Tavomi.'' and ``Babata reports to Tavomi.''\par\smallskip
\synthpairlabel{Turn 2, written by the teacher} The last four lines replace its original submission.
\begin{Verbatim}[fontsize=\scriptsize, xleftmargin=0.6em]
[...]    # parse the three sentence patterns into edges[person] = manager
while cur in edges:
    [...]    # step to the manager, stop on a cycle
print("top:", chain[-1])    # prints: top: Tavomi
my_answer = chain[-1]
result = unlock(my_answer)
print(result if isinstance(result, str)
      else f"unlocked: next task received (stage {result['stage']} of {result['of']})")
\end{Verbatim}
\synthpairlabel{Output} \texttt{DONE}\par\smallskip
\synthpairlabel{Turn 3, scripted}
\begin{Verbatim}[fontsize=\scriptsize, xleftmargin=0.6em]
answer["content"] = my_answer
answer["ready"] = True
\end{Verbatim}
\end{synthpairbox}

\textbf{Run statistics.} Table~\ref{tab:recursion-composition-stats} summarizes the same 2000 chains per condition as Table~\ref{tab:recursion-composition}. A chain solved by tail recursion needs four successful \texttt{unlock} calls and three \texttt{handoff} calls; retries add \texttt{unlock} calls. Each task passing the gate counts as one solve, whether it passes on the first or second attempt. In the RLM harness setting, each chain uses one root trajectory and no \texttt{handoff}.
\begin{enumerate}
    \item \textbf{RLM-27B} makes no \texttt{unlock} call in 18 chains, passes the first task in 1914, and averages 3.87 \texttt{unlock} calls and 2.98 passed tasks per chain.
    \item With the tail recursion hint, \textbf{RLM-27B} calls \texttt{unlock} in 1914 chains and passes the first task in 1664, with 3.20 \texttt{unlock} and 2.06 \texttt{handoff} calls per chain.
    \item After SFT, \textbf{RLM-27B} averages 4.40 \texttt{unlock} and 2.63 \texttt{handoff} calls and passes 3.38 tasks per chain; 6.8\% of passed tasks need the second attempt.
    \item \qwennew{} makes no \texttt{unlock} call in 426 chains, passes the first task in 1244, and averages 2.57 \texttt{unlock} calls and 1.53 passed tasks per chain.
    \item \qwennew{} with the hint makes no \texttt{unlock} call in 412 chains and passes the first task in 1234, with 2.20 \texttt{unlock} and 1.13 \texttt{handoff} calls per chain.
    \item \qwennew{} after SFT averages 3.10 \texttt{unlock} calls and 1.28 passed tasks per chain; 27.2\% of passed tasks need the second attempt. Of its 1802 chains that are not fully solved, 799 stop after an incorrect attempt on the first task not passed, and 1003 make no \texttt{unlock} call on that task.
\end{enumerate}

The final columns describe RLM calls with saved trajectories. Failed calls can lack a saved trajectory, so the recorded call count can average less than one per chain even when a root call was started. Across all three conditions, \textbf{RLM-27B} uses fewer turns per recorded RLM call and needs a second attempt on a smaller share of passed tasks than \qwennew{}.
\color{black}

\begin{table}[!htbp]
\centering\small\color{black}
\captionsetup{font={color=black}}
\caption{Statistics of the gated-chain runs in Table~\ref{tab:recursion-composition}, over 2000 chains each. \textit{Chains by tasks passed} counts chains by the number of tasks whose answer passed the gate. The \textit{Per chain} columns are averages over chains: tasks passed, \texttt{unlock} and \texttt{handoff} calls, and RLM calls with saved trajectories. The \textit{Per RLM call} columns are averages over RLM calls with saved trajectories; failed calls without a saved trajectory are excluded from these call-level averages. \textit{2nd att.} is the share of passed tasks that passed on the second attempt. The RLM harness has an 80-turn budget for the whole chain, the \textbf{+ recursion hint} setting has 80 turns per RLM call, and the + SFT setting has 20.}
\label{tab:recursion-composition-stats}
\setlength{\tabcolsep}{3.5pt}\renewcommand{\arraystretch}{1.15}
\begin{tabular}{@{}lrrrrrrrrrrrr@{}}
\toprule
 & \multicolumn{5}{c}{{\fontsize{8}{9.5}\selectfont\textbf{Chains by tasks passed}}} & \multicolumn{4}{c}{{\fontsize{8}{9.5}\selectfont\textbf{Per chain}}} & \multicolumn{2}{c}{{\fontsize{8}{9.5}\selectfont\textbf{Per RLM call}}} & {\fontsize{8}{9.5}\selectfont\textbf{2nd att.}} \\
\cmidrule(lr){2-6}\cmidrule(lr){7-10}\cmidrule(lr){11-12}
{\fontsize{8}{9.5}\selectfont\textbf{Setting}} & 0 & 1 & 2 & 3 & 4 & {\fontsize{8}{9.5}\selectfont Passed} & {\fontsize{8}{9.5}\selectfont\texttt{unlock}} & {\fontsize{8}{9.5}\selectfont\texttt{handoff}} & {\fontsize{8}{9.5}\selectfont RLM calls} & {\fontsize{8}{9.5}\selectfont Turns} & {\fontsize{8}{9.5}\selectfont Sub-calls} & {\fontsize{8}{9.5}\selectfont\textbf{(\%)}} \\
\midrule
\multicolumn{13}{@{}l}{\textbf{Qwen3.8-27B}} \\
\quad RLM harness & \textcolor{black}{756} & \textcolor{black}{382} & \textcolor{black}{285} & \textcolor{black}{191} & \textcolor{black}{386} & \textcolor{black}{1.53} & \textcolor{black}{2.57} & \textcolor{black}{0.00} & \textcolor{black}{0.84} & \textcolor{black}{60.8} & \textcolor{black}{0.75} & \textcolor{black}{21.5} \\
\quad + recursion hint & 766 & 523 & 306 & 156 & 249 & 1.30 & 2.20 & 1.13 & 1.92 & 31.6 & 0.41 & 21.5 \\
\quad + SFT & 646 & 693 & 306 & 157 & 198 & 1.28 & 3.10 & 0.93 & 1.92 & 15.1 & 0.91 & 27.2 \\
\midrule
\multicolumn{13}{@{}l}{\textbf{RLM-27B}} \\
\quad RLM harness & \textcolor{black}{86} & \textcolor{black}{311} & \textcolor{black}{266} & \textcolor{black}{232} & \textcolor{black}{1105} & \textcolor{black}{2.98} & \textcolor{black}{3.87} & \textcolor{black}{0.00} & \textcolor{black}{0.84} & \textcolor{black}{23.2} & \textcolor{black}{5.05} & \textcolor{black}{13.0} \\
\quad + recursion hint & 336 & 325 & 226 & 168 & 945 & 2.53 & 3.20 & 2.06 & 2.80 & 6.6 & 1.84 & 6.0 \\
\quad + SFT & 114 & 193 & 67 & 76 & 1550 & 3.38 & 4.40 & 2.63 & 3.62 & 8.0 & 1.99 & 6.8 \\
\bottomrule
\end{tabular}
\end{table}

%% file: appendix/composition_statistics.tex
\clearpage
\section{Benchmark-level trajectory statistics}
\label{appx:composition-statistics}

\textcolor{black}{Tables~\ref{tab:composition-turns}, \ref{tab:composition-subcalls}, and \ref{tab:composition-tokens} provide expanded per-environment statistics for the statistics reported in Table~\ref{tab:main-composition-stats} across the eight evaluation benchmarks. We observe that each environment exhibits distinct patterns in terms of token and sub-call use, and we expand on these results here.}

\begin{table}[!htbp]
\centering
\caption{Average number of root turns per rollout, by benchmark.}\label{tab:composition-turns}
\begingroup
\fontsize{8}{10}\selectfont
\setlength{\tabcolsep}{2pt}
\renewcommand{\arraystretch}{1.18}
\setlength{\aboverulesep}{0pt}
\setlength{\belowrulesep}{0pt}
\begin{tabular}{@{}l*{8}{r}|r@{}}
\toprule
\textbf{Method} & LongReason & LooGLE & CTC & BC+ & OOLONG-Real & CorpusQA & OBLIQ & ELB & Avg. \\
\midrule
\multicolumn{10}{@{}l}{\textbf{Qwen3-30B-A3B-Instruct-2507}} \\
\midrule
RLM (base) & 21.0 & 10.9 & 16.6 & 21.2 & 17.8 & 25.4 & 11.8 & 10.3 & 16.9 \\
\rowcolor{black!6}[\tabcolsep][\tabcolsep]
+ RL (6 short tasks) & 8.0 & 11.0 & 13.5 & 14.1 & 15.3 & 22.4 & 13.6 & 6.8 & 13.1 \\
\rowcolor{black!6}[\tabcolsep][\tabcolsep]
+ RL ($\sim$20 tasks) & 6.0 & 11.1 & 14.6 & 18.1 & 16.3 & 22.7 & 14.4 & 9.2 & 14.1 \\
\rowcolor{black!6}[\tabcolsep][\tabcolsep]
+ RL (synthetic tasks) & 11.7 & 10.5 & 10.1 & 18.8 & 19.7 & 17.4 & 10.0 & 6.2 & 13.1 \\
\midrule\midrule
\multicolumn{10}{@{}l}{\textbf{Qwen3.8-27B}} \\
\midrule
RLM (base) & 7.9 & 9.8 & 26.1 & 24.7 & 26.2 & 21.3 & 19.1 & 11.8 & 18.4 \\
\rowcolor{black!6}[\tabcolsep][\tabcolsep]
+ RL (6 short tasks) & 7.2 & 9.9 & 25.6 & 24.0 & 18.3 & 20.9 & 18.3 & 11.3 & 16.9 \\
\rowcolor{black!6}[\tabcolsep][\tabcolsep]
+ RL (synthetic data) & 5.3 & 9.1 & 24.5 & 22.8 & 19.0 & 19.2 & 18.6 & 10.3 & 16.1 \\
\rowcolor{black!6}[\tabcolsep][\tabcolsep]
RLM-27B (synthetic data) & 2.2 & 5.0 & 9.7 & 17.9 & 4.8 & 3.5 & 6.4 & 2.3 & 6.5 \\
\bottomrule
\end{tabular}
\endgroup
\end{table}

\begin{table}[!htbp]
\centering
\caption{Average number of sub-calls per rollout, by benchmark.}\label{tab:composition-subcalls}
\begingroup
\fontsize{8}{10}\selectfont
\setlength{\tabcolsep}{2pt}
\renewcommand{\arraystretch}{1.18}
\setlength{\aboverulesep}{0pt}
\setlength{\belowrulesep}{0pt}
\begin{tabular}{@{}l*{8}{r}|r@{}}
\toprule
\textbf{Method} & LongReason & LooGLE & CTC & BC+ & OOLONG-Real & CorpusQA & OBLIQ & ELB & Avg. \\
\midrule
\multicolumn{10}{@{}l}{\textbf{Qwen3-30B-A3B-Instruct-2507}} \\
\midrule
RLM (base) & 3.4 & 3.2 & 2610.8 & 113.0 & 0.6 & 10.1 & 2830.3 & 4.8 & 697.0 \\
\rowcolor{black!6}[\tabcolsep][\tabcolsep]
+ RL (6 short tasks) & 0.0 & 0.5 & 1064.1 & 15.5 & 0.2 & 9.5 & 2240.5 & 2.0 & 416.5 \\
\rowcolor{black!6}[\tabcolsep][\tabcolsep]
+ RL ($\sim$20 tasks) & 0.0 & 0.2 & 2532.0 & 91.9 & 0.0 & 0.4 & 1756.8 & 2.8 & 548.0 \\
\rowcolor{black!6}[\tabcolsep][\tabcolsep]
+ RL (synthetic tasks) & 0.0 & 3.5 & 4584.1 & 207.1 & 8.1 & 48.0 & 3394.7 & 3.8 & 1031.1 \\
\midrule\midrule
\multicolumn{10}{@{}l}{\textbf{Qwen3.8-27B}} \\
\midrule
RLM (base) & 0.1 & 0.2 & 644.1 & 6.6 & 6.8 & 0.9 & 35.9 & 3.5 & 87.3 \\
\rowcolor{black!6}[\tabcolsep][\tabcolsep]
+ RL (6 short tasks) & 0.1 & 0.1 & 1629.9 & 5.7 & 18.8 & 0.9 & 54.9 & 2.3 & 214.1 \\
\rowcolor{black!6}[\tabcolsep][\tabcolsep]
+ RL (synthetic data) & 0.1 & 0.3 & 166.6 & 3.6 & 23.9 & 0.5 & 22.1 & 1.8 & 27.4 \\
\rowcolor{black!6}[\tabcolsep][\tabcolsep]
RLM-27B (synthetic data) & 1.0 & 3.9 & 614.2 & 15.3 & 50.7 & 1.0 & 93.7 & 8.3 & 98.5 \\
\bottomrule
\end{tabular}
\endgroup
\end{table}

\newsavebox{\comptoksubbox}
\makeatletter\setlength{\@fptop}{0pt}\setlength{\@fpbot}{0pt plus 1fil}\makeatother
\begin{table}[p]
\centering
\caption{Average root and sub-call input / output tokens in thousands (K = 1,000), rounded to the nearest K.}
\label{tab:composition-tokens}
\sbox{\comptoksubbox}{%
\fontsize{6.5}{8}\selectfont
\setlength{\tabcolsep}{2pt}%
\renewcommand{\arraystretch}{1.2}%
\begin{tabular*}{\textwidth}{@{\extracolsep{\fill}}l*{8}{r}@{}}
\toprule
& \multicolumn{4}{c}{Qwen3-30B-A3B} & \multicolumn{4}{c}{Qwen3.8-27B} \\
\cmidrule(lr){2-5}\cmidrule(l){6-9}
Benchmark & Base & + RL (6 short) & + RL ($\sim$20) & + RL (synth.) & Base & + RL (6 short) & + RL (synth.) & RLM-27B \\
\midrule
LongReason & 12K / 0K & 0K / 0K & 0K / 0K & 1K / 0K & 0K / 0K & 0K / 0K & 0K / 1K & 1K / 0K \\
LooGLE & 46K / 1K & 12K / 0K & 0K / 0K & 120K / 1K & 12K / 0K & 2K / 0K & 5K / 0K & 14K / 0K \\
CTC & 3219K / 142K & 1445K / 37K & 1498K / 179K & 2019K / 134K & 3220K / 57K & 5431K / 112K & 2511K / 114K & 1526K / 77K \\
BC+ & 615K / 18K & 29K / 1K & 191K / 10K & 1156K / 23K & 30K / 5K & 41K / 6K & 13K / 20K & 29K / 10K \\
OOLONG-Real & 5K / 0K & 10K / 0K & 0K / 0K & 14K / 0K & 96K / 3K & 264K / 59K & 267K / 75K & 269K / 18K \\
CorpusQA & 134K / 0K & 218K / 4K & 1K / 0K & 371K / 4K & 11K / 0K & 5K / 0K & 2K / 0K & 5K / 0K \\
OBLIQ & 3786K / 1592K & 3219K / 355K & 3126K / 1100K & 885K / 237K & 141K / 15K & 178K / 19K & 239K / 22K & 324K / 389K \\
ELB & 18K / 1K & 23K / 1K & 3K / 1K & 50K / 1K & 33K / 4K & 27K / 4K & 30K / 3K & 39K / 3K \\
\midrule
Avg. & 979K / 219K & 619K / 50K & 602K / 161K & 577K / 50K & 443K / 10K & 743K / 25K & 383K / 29K & 276K / 62K \\
\bottomrule
\end{tabular*}}%
\textbf{Root tokens}\par\smallskip\begingroup
\fontsize{6.5}{8}\selectfont
\setlength{\tabcolsep}{2pt}
\renewcommand{\arraystretch}{1.2}
\begin{tabular*}{\textwidth}{@{\extracolsep{\fill}}l*{8}{r}@{}}
\toprule
& \multicolumn{4}{c}{Qwen3-30B-A3B} & \multicolumn{4}{c}{Qwen3.8-27B} \\
\cmidrule(lr){2-5}\cmidrule(l){6-9}
Benchmark & Base & + RL (6 short) & + RL ($\sim$20) & + RL (synth.) & Base & + RL (6 short) & + RL (synth.) & RLM-27B \\
\midrule
LongReason & 227K / 4K & 53K / 3K & 35K / 3K & 327K / 8K & 31K / 1K & 29K / 1K & 14K / 1K & 6K / 2K \\
LooGLE & 48K / 3K & 45K / 3K & 76K / 7K & 77K / 5K & 46K / 2K & 50K / 1K & 54K / 2K & 19K / 3K \\
CTC & 170K / 43K & 95K / 49K & 238K / 61K & 140K / 65K & 400K / 10K & 428K / 10K & 496K / 14K & 60K / 5K \\
BC+ & 572K / 20K & 191K / 7K & 353K / 16K & 603K / 24K & 495K / 6K & 658K / 6K & 1036K / 13K & 203K / 4K \\
OOLONG-Real & 178K / 7K & 107K / 6K & 275K / 13K & 315K / 13K & 545K / 7K & 321K / 23K & 395K / 23K & 20K / 4K \\
CorpusQA & 656K / 20K & 685K / 31K & 742K / 21K & 447K / 21K & 200K / 5K & 279K / 5K & 355K / 8K & 14K / 3K \\
OBLIQ & 620K / 29K & 507K / 24K & 771K / 41K & 99K / 8K & 209K / 6K & 291K / 7K & 501K / 15K & 23K / 2K \\
ELB & 92K / 4K & 36K / 2K & 104K / 6K & 71K / 5K & 181K / 4K & 281K / 4K & 285K / 4K & 5K / 1K \\
\midrule
Avg. & 321K / 16K & 215K / 16K & 324K / 21K & 260K / 19K & 263K / 5K & 292K / 7K & 392K / 10K & 44K / 3K \\
\bottomrule
\end{tabular*}
\endgroup
\par\medskip
\textbf{Sub-call tokens}\par\smallskip
\usebox{\comptoksubbox}
\par\medskip
\end{table}

We generally observe the post-trained models exhibit very different behavior from each other and the base model variants, and when trained correctly, often use fewer turns, fewer tokens, and fewer sub-calls than the base model variants. However, the SFT'd model on synthetic traces targets trajectory behavior that is more compact, while maintaining similar or better performance on the evaluation benchmarks. In some cases, efficiency improvements are not useful, such as the \emph{+ RL (synthetic data)} variant on CTC-Bench, which uses far fewer sub-calls than the SFT variant but performs much worse.

\clearpage

%% file: appendix/lid_trajectories.tex
\section{Characterizing locally in-distribution trajectories}
\label{appx:lid-trajectories}

In this section, we discuss why locally in-distribution trajectories are a desirable property, and how they induce generalization at the neurosymbolic level, over the existing capabilities of language models. We first motivate how certain harness abstractions can exploit the locally in-distribution property to yield compositional generalization, and provide brief mathematical intuition on how harnesses like RLMs do so (\S\ref{appx:gen_gap}-\S\ref{appx:one-step}). In \S\ref{appx:similarity-plots}, we then provide empirical evidence for, at a token-level, what kind of root-level trajectories are produced by each harness in the experiments in \S\ref{sec:exp}. Finally, in \S\ref{train-examples}, we provide several examples of trajectories on unseen evaluations that match similar training trajectories through the orchestrator LM.

\subsection{Generalization Gap under Neural Network Parametrization} \label{appx:gen_gap}

Consider a training task distribution $p(x)$ and a test task distribution $q(x)$. Standard RL post-training learns a parameterized policy $\pi_\theta$ by maximizing expected reward on the training distribution:
\begin{align}
\pi_{\theta_p^*} = \arg\max_{\pi_\theta} \mathbb{E}_{x\sim p,\;y\sim\pi_\theta(\cdot\mid x)} [R(x,y)].
\end{align}
At test time, this learned policy is evaluated on $q$, yielding
\begin{align}
J_q(\pi_{\theta_p^*}) = \mathbb{E}_{x\sim q,\;y\sim\pi_{\theta_p^*}(\cdot\mid x)} [R(x,y)].
\end{align}
In general, the optimal performance achievable under the test distribution within the same policy class on $q$ is $J_q^{*} \triangleq J_q(\pi_{\theta_q^*})$, where $\pi_{\theta_q^*}=\arg\max_{\pi_\theta}J_q(\pi_\theta)$. The ideal case in RL post-training is that the reward-maximizing policy over the training distribution is also the reward maximizing policy over the test distribution, i.e., $J_q(\pi_{\theta^{*}_p}) \approx J_q^*$, but we do not have this guarantee. The resulting generalization gap therefore is defined as:
\begin{align}
\boxed{
\begin{aligned}
\Delta_{\mathrm{gen}}(p,q)
&=J_q^*-J_q(\pi_{\theta_p^*})\\
&=\mathbb{E}_{x\sim q}
\left[
\mathbb{E}_{y\sim\pi_{\theta_q^*}}R(x,y)
-
\mathbb{E}_{y\sim\pi_{\theta_p^*}}R(x,y)
\right].
\end{aligned}
}
\end{align}
For bounded rewards $R(x,y)\in[0,R_{\max}]$, this expression satisfies
\begin{align}
\boxed{
0\leq \Delta_{\mathrm{gen}}(p,q)
\leq
R_{\max}
\mathbb{E}_{x\sim q}
\left[
D_{\mathrm{TV}}\left(
\pi_{\theta_p^*}(\cdot\mid x),
\pi_{\theta_q^*}(\cdot\mid x)
\right)
\right].
}
\end{align}
The fundamental challenge is that RL optimization on $p$ provides no general guarantee that $\pi_{\theta_p^*}$ will behave similarly to $\pi_{\theta_q^*}$ on unseen tasks. When $p$ and $q$ differ substantially, the generalization gap can be as large as the full reward range.

In conventional \textit{``trajectory-as-a-prompt''} approaches, generalization therefore relies entirely on the neural network's learned representations: the parameters $\theta_p^*$, optimized on training trajectories, must implicitly capture the structure necessary to solve previously unseen trajectories. Without an explicit mechanism for composing previously learned capabilities, there is no guarantee that this extrapolation will succeed, even when the individual subtasks are already solvable by the model.

This motivates our central question: can a harness reduce this generalization burden by transforming globally out-of-distribution tasks into locally in-distribution computations?

\subsection{Generalization through RLM decomposition} \label{appx:one-step}
Rather than learning a single policy $\pi_\theta(y\mid x)$, an RLM decomposes the computation into two components:
\begin{itemize}
    \item Root controller: $z\sim\pi_H(\cdot\mid x)$, where $z$ is the root's action (e.g., an LM program).
    \item REPL execution: $y\sim\pi_L(\cdot\mid z,c)$, where $c$ is the offloaded context, and $y$ is the result of executing the action, potentially using sub-LM calls.
\end{itemize}

\paragraph{Step 1. Expected reward on training and test tasks}
For a training task $(x,c)\sim p$, define
\begin{align}
J_p=\mathbb E_{(x,c)\sim p}
\mathbb E_{z\sim\pi_H(\cdot\mid x)}
\underbrace{
\mathbb E_{y\sim\pi_L(\cdot\mid z,c)}
[R(x,y)]}_{Q_x(z,c)}.
\end{align}
Similarly, for a test task $(x',c')\sim q$, $J_q
=\mathbb E_{(x',c')\sim q}
\mathbb E_{z\sim\pi_H(\cdot\mid x')}
\left[Q_{x'}(z,c')\right]$. Here $Q_x(z,c)$ measures the expected reward of executing root action $z$ on task $x$ with external context $c$.

\paragraph{Step 2. Decomposing the generalization gap}
For clarity, compare a training task $(x,c)$ and a test task $(x',c')$. We have
\begin{align}
J_1-J_2
= \mathbb E_{z\sim\pi_H(\cdot\mid x)}[Q_x(z,c)]
- \mathbb E_{z\sim\pi_H(\cdot\mid x')}[Q_{x'}(z,c')].
\end{align}
We then add and subtract
$\mathbb E_{z\sim\pi_H(\cdot\mid x')}[Q_x(z,c)]$
to decompose the difference into the root-controller shift
and the execution/sub-call shift.
\begin{align}
\begin{aligned}
J_1-J_2
={}&
\underbrace{
\left(
\mathbb E_{z\sim\pi_H(\cdot\mid x)}Q_x(z,c)
-
\mathbb E_{z\sim\pi_H(\cdot\mid x')}Q_x(z,c)
\right)
}_{\text{(A) Root controller shift}}
\\[6pt]
&+
\underbrace{
\mathbb E_{z\sim\pi_H(\cdot\mid x')}
\left[
Q_x(z,c)-Q_{x'}(z,c')
\right]
}_{\text{(B) Execution / sub-call shift}}.
\end{aligned}
\end{align}
Assuming $R\in[0,R_{\max}]$, we obtain
\begin{align}
\boxed{
\begin{aligned}
|J_1-J_2|
\leq{}&
R_{\max}
D_{\mathrm{TV}}\left(
\pi_H(\cdot\mid x),
\pi_H(\cdot\mid x')
\right)
\\
&+
\mathbb E_{z\sim\pi_H(\cdot\mid x')}
\left|
Q_x(z,c)-Q_{x'}(z,c')
\right|.
\end{aligned}
}
\end{align}
The same bound applies to $|J_p-J_q|$ after averaging over paired training and test tasks under any coupling of $p$ and $q$.

\textbf{Theoretical Insight}: The key observation is that the full task distributions $p, q$ need not be similar. Instead, RLMs can learn higher-level abstractions that decompose different tasks into similar root-level actions, whose execution through programmatic REPL operations and sub-LLM calls remains effective across both distributions.

We characterize this through two conditions.

\textbf{Condition 1 --- Root-level abstraction transfers.}
\begin{align}
\pi_H(\cdot\mid x)\approx\pi_H(\cdot\mid x').
\end{align}
The root controller learns a shared decomposition strategy, producing similar high-level actions $z$ even when the underlying tasks $x$ and $x'$ are substantially different.

\textbf{Condition 2 --- The learned abstraction remains reward-effective.}
\begin{align}
\boxed{
Q_x(z,c)\approx Q_{x'}(z,c')\approx R_{\max}.
}
\end{align}
For the root actions selected with high probability, the corresponding executions must achieve high reward on both training and test tasks. This requires that the underlying programmatic operations and sub-LM capabilities remain effective on the unseen context. Importantly, matching root actions is insufficient if their execution fails on the test task.

\paragraph{Generalization guarantee}
Suppose the root-controller shift and execution shift are bounded by $\epsilon_H$ and $\epsilon_L$, respectively. From our previous decomposition,
\begin{align}
|J_p-J_q|\leq R_{\max}\epsilon_H+\epsilon_L.
\end{align}
Let $\epsilon_{\mathrm{train}} := R_{\max}-J_p \geq 0$
denote the training suboptimality gap of the learned RLM,
measuring how far its expected training reward is from the
maximum possible reward. Thus
\begin{align}
J_p\geq R_{\max}-\epsilon_{\mathrm{train}},
\end{align}
Its test performance therefore satisfies
\begin{align}
\boxed{
J_q\geq
R_{\max}
-\underbrace{\epsilon_{\mathrm{train}}}_{\text{Training error}}
-\underbrace{R_{\max}\epsilon_H}_{\text{Root abstraction shift}}
-\underbrace{\epsilon_L}_{\text{Execution shift}}.
}
\end{align}

\textbf{Key takeaway:} RLMs can generalize across globally different task distributions by learning a shared high-level decomposition, provided that the resulting actions compose existing capabilities that remain successful on unseen tasks. This captures the essential distinction: LID generalization requires not only similar decompositions, but also preservation of their reward-generating capabilities. 

\paragraph{Main takeaway}
Unlike conventional RL, where generalization relies on a single neural policy extrapolating from $p$ to $q$, an RLM can transfer performance by preserving a learned root-level decomposition and composing it with existing sub-call capabilities.
The task can therefore be globally OOD while the individual computations remain locally in-distribution.

One important subtlety is that matching root actions alone is insufficient. We also need their sub-call execution steps (i.e., $y \sim \pi_L(\cdot | z, c)$) to remain effective on the new tasks. This is precisely why the second term matters, especially for tasks involving semantic subcalls rather than deterministic programs.

The above analysis considers a single root action followed by REPL execution. The extension to multi-step trajectories follows the same principle (with an added persistent REPL state), with the generalization gap bounded by the accumulation of root-controller policy shifts and REPL execution errors across steps, accounting for their dependence on preceding interactions.

\subsection{Token-level trajectory similarity of harnesses} \label{appx:similarity-plots}

\input{appendix/trajectories/similarity_metrics}

\begin{figure}[!htbp]
    \centering
    \makebox[\linewidth][c]{\includegraphics[width=0.9\linewidth]{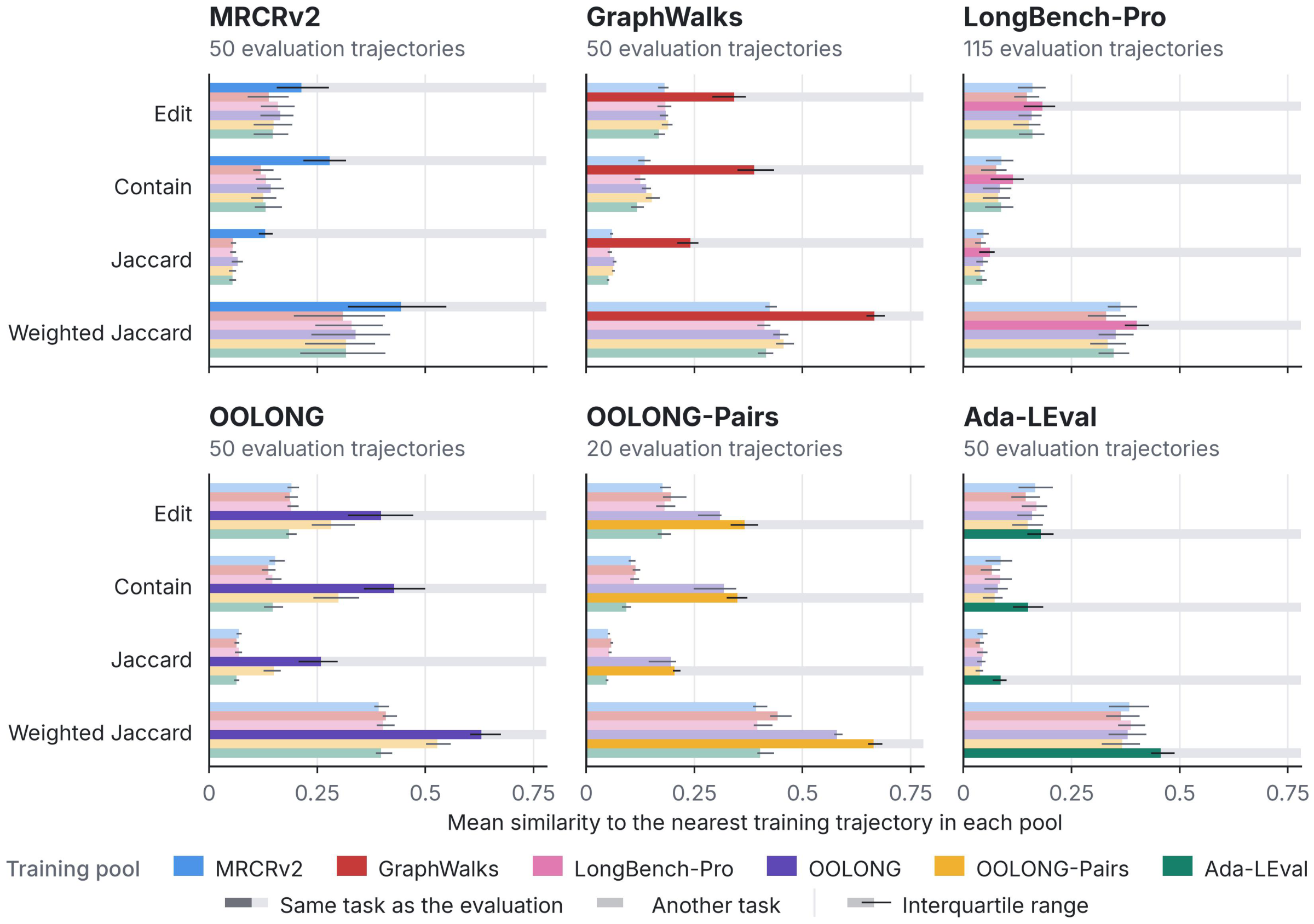}}
    \caption{\textbf{Cross-task harness trajectory similarity.} For each evaluation environment in the length generalization experiments in Figure~\ref{fig:length-gen}, we also check how similar RLM trajectories on evaluation tasks are to unrelated tasks. We look at cross-task trajectory similarity: we plot the mean maximum similarity of trajectories in each long evaluation setting, to its closest training trajectory in each training environment, even those unrelated to the target evaluation task. Across all tasks, there is a higher similarity for harness trajectories transferring from a shorter variant to the longer variant of the same environment versus shorter variants of different tasks.}
    \label{fig:rlm-cross-task-similarity}
\end{figure}

Figure~\ref{fig:rlm-cross-task-similarity} extends this comparison to RLM trajectories from different training environments.

\begin{figure}[!hbtp]
    \centering
    \includegraphics[width=0.9\linewidth]{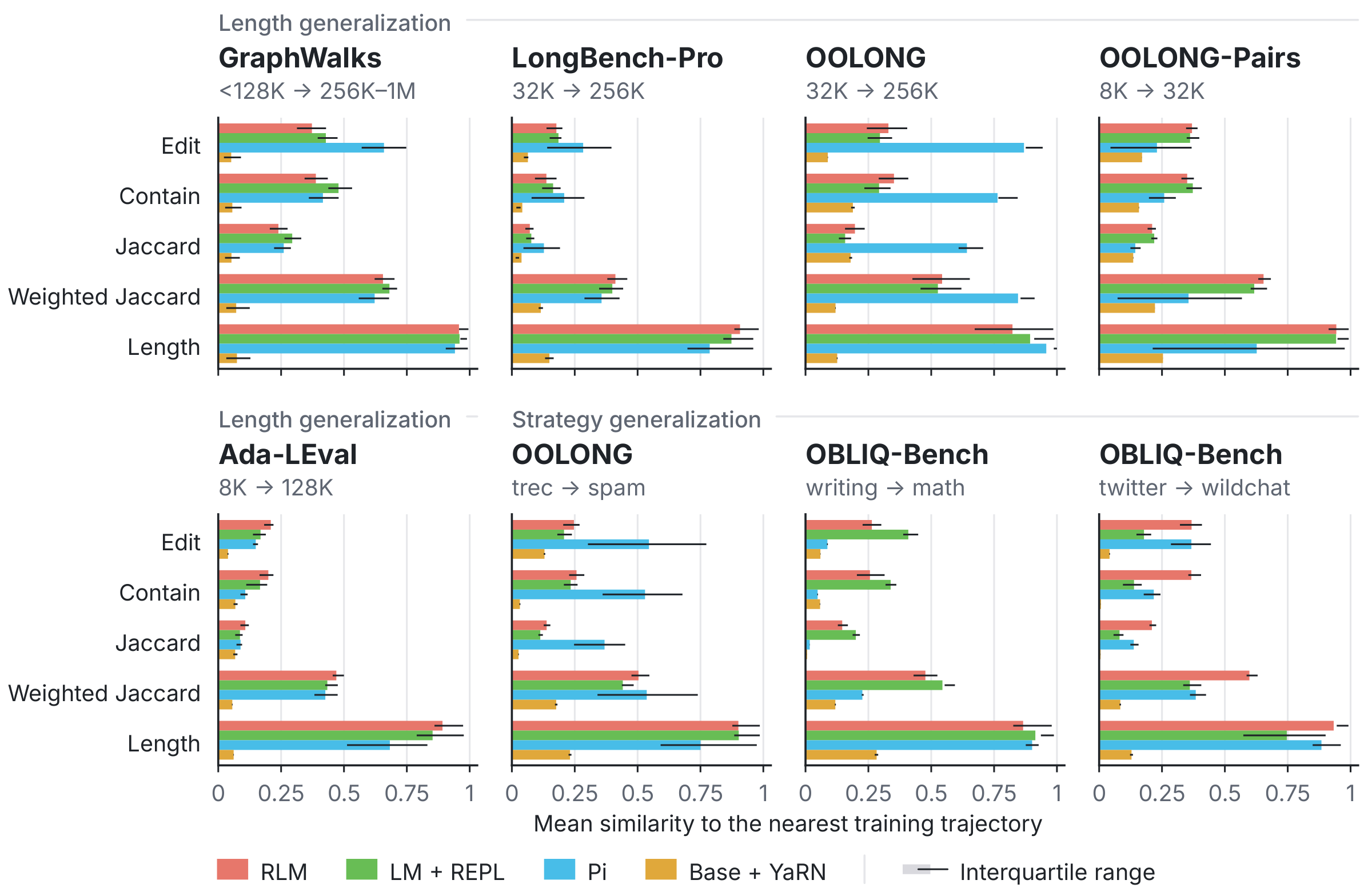}
    \caption{\textcolor{black}{For each evaluation environment in the length and strategy generalization experiments in Figures~\ref{fig:length-gen},~\ref{fig:strategy_grid}, we plot the mean maximum similarity to a successful training trajectory under each metric. Higher values indicate higher similarity.}}
    \label{fig:similarity}
\end{figure}

\clearpage
\color{black}

\subsection{Trajectory comparisons and examples}
\label{train-examples}

\input{appendix/trajectories/trajectory_examples_intro}

\input{appendix/trajectories/trajectory_examples}

\subsection{Verifying local sub-task execution}
\label{appx:subtask-verification}

In harnesses that use a root controller, like the ones we study in \S\ref{sec:exp}, the learned decompositions that invoke sub-calls rely on sub-call correctness to be locally in-distribution. In Table~\ref{tab:subtask-verification}, we verify for the RLM (which invokes the most sub-calls) that the sub-tasks its sub-calls are solving are reliably correct.

We verify whether there is a discrepancy between the sub-calls on the shorter train tasks and the longer evaluation tasks, using the rollouts from the length and strategy generalization experiments (Figures~\ref{fig:length-gen},~\ref{fig:strategy_grid}). Many of these sub-call prompts were programmatically generated by the REPL used. 

\textcolor{black}{Not all benchmarks contain oracle sub-call ground truth labels (e.g. LongBench-Pro), and the RLM rarely invokes sub-calls on some benchmarks (e.g. MRCRv2). Table~\ref{tab:subtask-verification} reports delegated-subtask scores for the base model and best checkpoint on the training and evaluation splits. Label-accuracy rows score by correct scored labels divided by all scored labels pooled over the split. GraphWalks exact match is the number of exact adjacency parses divided by scored parse calls. Balanced accuracy is computed from pooled candidate-verdict counts as $(\mathrm{TPR}+\mathrm{TNR})/2$. For OBLIQ-Bench, AUC$\times100$ is ROC AUC over all numeric document scores pooled across trajectories, with relevant documents as positives and ties contributing 0.5.}

\begin{table}[!htbp]
\centering
\caption{\textcolor{black}{Delegated-subtask scores (\%) for the base model $\to$ best checkpoint, on the training and evaluation splits. Metric-specific aggregation is defined in the text.}}
\label{tab:subtask-verification}
\begingroup
\fontsize{8}{10}\selectfont
\renewcommand{\arraystretch}{1.2}
\setlength{\tabcolsep}{2.5pt}
\resizebox{\linewidth}{!}{%
\begin{tabular}{@{}llllcc@{}}
\toprule
\textbf{Environment} & \textbf{Delegated sub-task} & \textbf{Ground truth} & \textbf{Metric} & \textbf{Train} & \textbf{Eval} \\
\midrule
\rowcolor{black!6}[\tabcolsep][\tabcolsep]
\multicolumn{6}{@{}l}{\textbf{Length generalization}} \\
OOLONG & Label one question line (6 classes) & TREC label & Accuracy & 86.1$\to$85.6 & 81.7$\to$83.7 \\
OOLONG-Pairs & Label one question line (6 classes) & TREC label & Accuracy & 88.5$\to$90.3 & 83.9$\to$89.3 \\
Ada-LEval & Is this candidate answer relevant? & Accepted answer vs.\ distractors & Balanced acc. & 82.5$\to$93.4 & 78.5$\to$83.0 \\
GraphWalks & Parse edges into an adjacency dict & Edges in the prompt & Exact match & -- & 99.8$\to$(--) \\
LongBench-Pro & 25 task types & \multicolumn{4}{l}{No per-call ground truth; not scored} \\
MRCRv2 & 0--2\% of rollouts delegate; 0.0 sub-calls on average & \multicolumn{4}{l}{No per-call ground truth; not scored.} \\
\midrule
\rowcolor{black!6}[\tabcolsep][\tabcolsep]
\multicolumn{6}{@{}l}{\textbf{Strategy generalization}} \\
OOLONG TREC$\to$spam & Label one line (TREC / spam) & Per-line label & Accuracy & 87.7$\to$85.9 & 70.6$\to$87.3 \\
OBLIQ Twitter$\to$\allowbreak WildChat & Score one document for the query & qrels positive & AUC$\times$100 & 76.4$\to$77.3 & 83.6$\to$88.1 \\
OBLIQ Writing$\to$Math & Score one document for the query & qrels positive & AUC$\times$100 & 69.3$\to$79.3 & 70.0$\to$67.3 \\
\bottomrule
\end{tabular}%
}
\par\vspace{4pt}
\begin{minipage}{0.97\linewidth}
\end{minipage}
\endgroup
\end{table}

%% file: appendix/trajectories/similarity_metrics.tex
In \S\ref{subsec:characterize-local}, we argued that certain harness abstractions enable their individual LM calls to see similar trajectories on different tasks that share some structure. We showed that under a token-wise weighted Jaccard similarity metric, the root controller trajectory of context-offloading harnesses was similar, where we could see a trend between trajectory similarity and transferred performance to novel tasks. In this section, we explore other distance metrics.

Let $a$ and $b$ be the token sequences of a training and evaluation trajectory, with lengths $n$ and $m$. Tokens use the analysis code's lowercased word/code tokenizer, not the model's BPE tokenizer. Let $G_3(a)$ be the set of distinct consecutive token triples, $c_a(t)$ the count of token $t$, and $\operatorname{Lev}(a,b)$ the token-level Levenshtein distance. The plots in Figure~\ref{fig:similarity} report similarity $s\in[0,1]$; the corresponding distance is $1-s$.

\begin{center}
\small\renewcommand{\arraystretch}{1.22}\setlength{\tabcolsep}{6pt}
\begin{tabular}{@{}>{\raggedright\arraybackslash}p{0.16\linewidth}>{\raggedright\arraybackslash}p{0.35\linewidth}>{\raggedright\arraybackslash}p{\dimexpr0.49\linewidth-4\tabcolsep\relax}@{}}
\toprule
\textbf{Metric} & \textbf{Similarity $s(a,b)$} & \textbf{What it measures} \\
\midrule
Edit & $1-\dfrac{\operatorname{Lev}(a,b)}{\max(n,m)}$ &
One minus the minimum number of token insertions, deletions and substitutions,
divided by the longer length. Sensitive to order. \\
\addlinespace[4pt]
Contain & $\dfrac{|G_3(a)\cap G_3(b)|}{|G_3(b)|}$ &
Fraction of evaluation triples also present in training. Asymmetric: extra
training text is not penalized. \\
\addlinespace[4pt]
Jaccard & $\dfrac{|G_3(a)\cap G_3(b)|}{|G_3(a)\cup G_3(b)|}$ &
Shared triples relative to all distinct triples. Symmetric; captures local
phrasing but ignores global order and repeated occurrences. \\
\addlinespace[4pt]
Weighted Jaccard & $\dfrac{\sum_t\min(c_a(t),c_b(t))}{\sum_t\max(c_a(t),c_b(t))}$ &
Overlap in token counts. Repetition and length imbalance matter; token order
does not. Uses unigrams in this implementation. \\
\addlinespace[4pt]
Length & $\dfrac{\min(n,m)}{\max(n,m)}$ &
Length-only reference. Equal-length trajectories score 1 even when their
contents differ completely. \\
\bottomrule
\end{tabular}
\end{center}

%% file: appendix/trajectories/trajectory_examples_intro.tex
We show six pairs of root trajectories, listed in the table below. Each pair
contains one evaluation trajectory from the checkpoint used in
Figure~\ref{fig:sim-results} and one training trajectory from the same run. The
training trajectory is one of the nearest neighbors of the evaluation trajectory
under weighted Jaccard similarity, among all training trajectories collected up
to that checkpoint. The examples illustrate the behaviors discussed in the main
text. They are not a random sample.

\textbf{Reading the examples.}
Each example starts on a new page with the two tasks, their rewards, and a
short description of what we observe. The two trajectories follow, with
training on the left and evaluation on the right. Turns that match are placed
side by side, and a blank cell means the other trajectory has no matching turn.
The background color separates the text the model writes, the code or tool
calls it issues, and the output it gets back. Lines that send a request to a
sub-call or sub-agent are marked in amber.

\textbf{What is not shown.}
We show only the root trajectory. The initial prompt, the long context, and the
content of the sub-calls are left out. Long blocks are shortened, and each
omission states how many lines were removed. Similarities are computed on the
full trajectories, before any shortening.

%% file: appendix/trajectories/trajectory_examples.tex
\begingroup
\definecolor{trtrain}{RGB}{24,101,87}
\definecolor{treval}{RGB}{43,86,139}
\definecolor{trfail}{RGB}{159,54,48}
\definecolor{trink}{RGB}{38,46,54}
\definecolor{trkeyword}{RGB}{115,60,155}
\definecolor{trstring}{RGB}{30,105,79}
\definecolor{trnumber}{RGB}{160,76,33}
\definecolor{trbuiltin}{RGB}{38,83,145}
\definecolor{trcomment}{RGB}{112,121,130}
\definecolor{trmuted}{RGB}{96,102,108}
\definecolor{trcode}{RGB}{240,244,249}
\definecolor{trout}{RGB}{226,228,231}
\definecolor{trsub}{RGB}{255,238,190}
\newcommand{\trcodefont}{\fontencoding{T1}\fontfamily{lmtt}\fontsize{6}{7}\selectfont}
\newcommand{\trsans}{\fontencoding{T1}\fontfamily{lmss}\fontsize{7}{8.5}\selectfont}
\newcommand{\trcell}[2]{\colorbox{#1}{\makebox[.492\linewidth][l]{\hspace{7pt}\rule[-1.6pt]{0pt}{7pt}#2}}}
\newcommand{\trrow}[4]{\hbox to\linewidth{\trcodefont\trcell{#1}{#2}\hfill\trcell{#3}{#4}}}
\newcommand{\trturn}[2]{\colorbox{#1!14}{\makebox[.492\linewidth][l]{\hspace{7pt}\rule[-2.4pt]{0pt}{10pt}\trsans\bfseries\color{#1}#2}}}
\newcommand{\trnoturn}{\makebox[.492\linewidth]{}}
\newcommand{\trhead}[2]{\vskip 3pt\hbox to\linewidth{#1\hfill#2}\nobreak}
\newcommand{\trmore}{\makebox[0pt][r]{\color{black!35}\tiny$\hookrightarrow$\hspace{1pt}}}
\newcommand{\trkey}[2]{{\setlength{\fboxsep}{0pt}\setlength{\fboxrule}{.3pt}\fcolorbox{black!45}{#1}{\rule[-1.5pt]{0pt}{7pt}\hspace{9pt}}}~#2}
\ifcsname c@trexample\endcsname\else\newcounter{trexample}\fi
\setcounter{trexample}{0}
\setlength{\fboxsep}{0pt}
\setlength{\parindent}{0pt}

\begin{center}\small\setlength{\tabcolsep}{5pt}\renewcommand{\arraystretch}{1.15}
\begin{tabular}{@{}cllccc@{}}\toprule
\textbf{Ex.} & \textbf{Harness} & \textbf{Environment (train $\to$ eval)} & \textbf{Train} & \textbf{Eval} & \textbf{Similarity} \\\midrule
\ref{ex:rlm-graphwalks} & RLM & GraphWalks (256K--1M) & 1.00 & 1.00 & 0.63 \\
\ref{ex:rlm-oolong} & RLM & OOLONG (787 $\to$ 6{,}374 questions) & 1.00 & 1.00 & 0.60 \\
\ref{ex:pi-graphwalks} & Pi & GraphWalks (256K--1M) & 1.00 & 1.00 & 0.71 \\
\ref{ex:pi-oolong-pairs} & Pi & OOLONG-Pairs (188 $\to$ 787 questions) & 0.03 & 0.004 & 0.67 \\
\ref{ex:pi-oolong} & Pi & OOLONG (787 $\to$ 6{,}374 questions) & 1.00 & 0.00 & 0.92 \\
\ref{ex:rlm-obliq} & RLM & OBLIQ (Twitter $\to$ WildChat) & 0.91 & 0.77 & 0.71 \\
\bottomrule\end{tabular}\end{center}
\begin{center}\small\trkey{white}{model text}\quad\trkey{trcode}{code or tool call}\quad\trkey{trsub}{sub-call request}\quad\trkey{trout}{returned output}\end{center}
\clearpage
\refstepcounter{trexample}\label{ex:rlm-graphwalks}
{\normalsize\bfseries Example \thetrexample. RLM on GraphWalks: a programmatic task solved directly\par}\vspace{5pt}
\begin{minipage}[t]{.492\linewidth}\footnotesize\raggedright\setlength{\parskip}{1.5pt}
{\trsans\bfseries\color{trtrain}TRAIN}\hfill{\trsans\bfseries\color{trtrain}F1 1.00}\par
\textbf{Context.} Edge list, 110,213 characters.\par
\textbf{Task.} Find the parents of node 25b2822c2f.\par
\textbf{Gold.} 2 parent nodes.\par
\textbf{Trajectory.} 3 root turns, 0 sub-calls\par
\end{minipage}\hfill
\begin{minipage}[t]{.492\linewidth}\footnotesize\raggedright\setlength{\parskip}{1.5pt}
{\trsans\bfseries\color{treval}EVAL}\hfill{\trsans\bfseries\color{treval}F1 1.00}\par
\textbf{Context.} Edge list, 437,413 characters.\par
\textbf{Task.} Find the parents of node a96b65a721.\par
\textbf{Gold.} 10 parent nodes.\par
\textbf{Trajectory.} 3 root turns, 0 sub-calls\par
\end{minipage}\par
\vspace{5pt}{\footnotesize\textbf{Similarity.} Weighted Jaccard 0.63, the nearest of 8,448 training trajectories in this run (checkpoint at step 131).\par}
\vspace{4pt}{\small\color{black}\textbf{What we observe.} The task is programmatic, so the root solves it directly by printing the start and end of the context, extracting the answer with one line of string processing, and submitting this answer. Neither trajectory makes a sub-call. The evaluation context is 4x longer, the root program is length-invariant, so the root strategy generalizes without needing to compose with sub-call capabilities.\par}
\vspace{3pt}\nobreak
\begingroup\offinterlineskip\vbadness=10000
\trhead{\trturn{trtrain}{TRAIN \textperiodcentered{} turn 1}}{\trturn{treval}{EVAL \textperiodcentered{} turn 1}}
\trrow{white}{\textcolor{trink}{Understood.\ I\ will\ begin\ by\ inspecting\ the\ \char96{}context\char96{}\ to\ }}{white}{\textcolor{trink}{Understood.\ I\ will\ begin\ by\ inspecting\ the\ \char96{}context\char96{}\ to\ }}\nobreak
\trrow{white}{\textcolor{trink}{understand\ its\ structure.\ Since\ \char96{}context\char96{}\ is\ a\ string\ of\ }}{white}{\textcolor{trink}{understand\ its\ structure,{}\ particularly\ the\ edge\ list\ and\ }}
\trrow{white}{\textcolor{trink}{110,{}284\ characters\ containing\ a\ directed\ graph\ in\ edge\ }}{white}{\textcolor{trink}{the\ operation\ specification.}}
\trrow{white}{\textcolor{trmuted}{[...\ 4\ lines\ omitted\ ...]}}{white}{\textcolor{trmuted}{[...\ 5\ lines\ omitted\ ...]}}
\trrow{white}{\textcolor{trink}{few\ lines\ of\ \char96{}context\char96{}\ to\ inspect\ the\ format.}}{white}{\textcolor{trink}{last\ 500\ characters\ of\ \char96{}context\char96{}\ to\ examine\ the\ structure.}}
\trrow{trcode}{\textcolor{trbuiltin}{print}\textcolor{trink}{(context[:}\textcolor{trnumber}{500}\textcolor{trink}{])}}{trcode}{\textcolor{trbuiltin}{print}\textcolor{trink}{(context[:}\textcolor{trnumber}{500}\textcolor{trink}{])}}
\trrow{trcode}{\textcolor{trbuiltin}{print}\textcolor{trink}{(}\textcolor{trstring}{\char34{}\textbackslash{}n...\ (truncated)\textbackslash{}n\char34{}}\textcolor{trink}{)}}{trcode}{\textcolor{trbuiltin}{print}\textcolor{trink}{(}\textcolor{trstring}{\char34{}\textbackslash{}n...\ (middle\ omitted)\ ...\textbackslash{}n\char34{}}\textcolor{trink}{)}}
\trrow{trcode}{\textcolor{trbuiltin}{print}\textcolor{trink}{(context[-{}}\textcolor{trnumber}{500}\textcolor{trink}{:])}}{trcode}{\textcolor{trbuiltin}{print}\textcolor{trink}{(context[-{}}\textcolor{trnumber}{500}\textcolor{trink}{:])}}
\trrow{white}{\textcolor{trink}{This\ will\ help\ me\ confirm:}}{white}{}
\trrow{white}{\textcolor{trink}{1.\ The\ format\ of\ the\ edge\ list\ (one\ \char96{}A\ -{}\textgreater{}\ B\char96{}\ per\ line).}}{white}{}
\trrow{white}{\textcolor{trink}{2.\ The\ presence\ and\ format\ of\ the\ \char96{}Operation:\char96{}\ line.}}{white}{}
\trrow{white}{\textcolor{trmuted}{[...\ 2\ lines\ omitted\ ...]}}{white}{}
\trrow{white}{\textcolor{trink}{Let\char39{}s\ begin.}}{white}{}
\trrow{trout}{\textcolor{trmuted}{You\ will\ be\ given\ a\ graph\ as\ a\ list\ of\ directed\ edges.\ All\ }}{trout}{\textcolor{trmuted}{You\ will\ be\ given\ a\ graph\ as\ a\ list\ of\ directed\ edges.\ All\ }}
\trrow{trout}{\trmore \textcolor{trmuted}{nodes\ are\ at\ least\ degree\ 1.\ You\ will\ also\ get\ a\ }}{trout}{\trmore \textcolor{trmuted}{nodes\ are\ at\ least\ degree\ 1.\ You\ will\ also\ get\ a\ }}
\trrow{trout}{\trmore \textcolor{trmuted}{description\ of\ an\ operation\ to\ perform\ on\ the\ graph.}}{trout}{\trmore \textcolor{trmuted}{description\ of\ an\ operation\ to\ perform\ on\ the\ graph.}}\nobreak
\trrow{trout}{\textcolor{trmuted}{[...\ 19\ lines\ omitted\ ...]}}{trout}{\textcolor{trmuted}{[...\ 19\ lines\ omitted\ ...]}}\nobreak
\trrow{trout}{\textcolor{trmuted}{Final\ Answer:\ []}}{trout}{\textcolor{trmuted}{Final\ Answer:\ []}}
\trhead{\trturn{trtrain}{TRAIN \textperiodcentered{} turn 2}}{\trturn{treval}{EVAL \textperiodcentered{} turn 2}}
\trrow{white}{\textcolor{trink}{The\ \char96{}context\char96{}\ has\ been\ successfully\ inspected.\ Here\char39{}s\ what\ }}{white}{\textcolor{trink}{The\ context\ has\ been\ successfully\ inspected.\ From\ the\ }}\nobreak
\trrow{white}{\textcolor{trink}{we\ know:}}{white}{\textcolor{trink}{output:}}
\trrow{white}{\textcolor{trink}{-{}\ The\ graph\ is\ provided\ as\ a\ list\ of\ directed\ edges\ (one\ }}{white}{\textcolor{trink}{-{}\ The\ graph\ is\ given\ as\ a\ list\ of\ directed\ edges,{}\ one\ per\ }}
\trrow{white}{\textcolor{trmuted}{[...\ 28\ lines\ omitted\ ...]}}{white}{\textcolor{trmuted}{[...\ 31\ lines\ omitted\ ...]}}
\trrow{white}{\textcolor{trink}{destination\ \char96{}25b2822c2f\char96{}.}}{white}{\textcolor{trink}{Let\char39{}s\ do\ that\ now.}}
\trrow{trcode}{\textcolor{trcomment}{\#\ Split\ context\ into\ lines}}{trcode}{\textcolor{trkeyword}{import}\textcolor{trink}{\ re}}
\trrow{trcode}{\textcolor{trink}{lines\ =\ context.splitlines()}}{trcode}{\textcolor{trcomment}{\#\ Find\ all\ lines\ where\ a96b65a721\ is\ the\ target\ }}
\trrow{trcode}{\textcolor{trcomment}{\#\ Filter\ lines\ where\ the\ edge\ ends\ with\ \char39{}25b2822c2f\char39{}}}{trcode}{\trmore \textcolor{trcomment}{(right-{}hand\ side)}}
\trrow{trcode}{\textcolor{trink}{parents\ =\ [}}{trcode}{\textcolor{trink}{pattern\ =\ }\textcolor{trstring}{r\char39{}(.*)\ -{}\textgreater{}\ a96b65a721\char39{}}}
\trrow{trcode}{\textcolor{trink}{\ \ \ \ line.split(}\textcolor{trstring}{\char34{}\ -{}\textgreater{}\ \char34{}}\textcolor{trink}{)[}\textcolor{trnumber}{0}\textcolor{trink}{]}}{trcode}{\textcolor{trink}{matches\ =\ re.findall(pattern,{}\ context)}}
\trrow{trcode}{\textcolor{trink}{\ \ \ \ }\textcolor{trkeyword}{for}\textcolor{trink}{\ line\ in\ lines}}{trcode}{\textcolor{trcomment}{\#\ The\ matches\ are\ the\ source\ nodes}}
\trrow{trcode}{\textcolor{trink}{\ \ \ \ }\textcolor{trkeyword}{if}\textcolor{trink}{\ }\textcolor{trstring}{\char34{}\ -{}\textgreater{}\ \char34{}}\textcolor{trink}{\ in\ line}}{trcode}{\textcolor{trbuiltin}{print}\textcolor{trink}{(}}
\trrow{trcode}{\textcolor{trink}{\ \ \ \ and\ line.split(}\textcolor{trstring}{\char34{}\ -{}\textgreater{}\ \char34{}}\textcolor{trink}{)[}\textcolor{trnumber}{1}\textcolor{trink}{].strip()\ ==\ }\textcolor{trstring}{\char34{}25b2822c2f\char34{}}}{trcode}{\textcolor{trink}{\ \ \ \ }\textcolor{trstring}{f\char34{}Found\ \{}\textcolor{trbuiltin}{len}\textcolor{trink}{(matches)}\textcolor{trstring}{\}\ edges\ pointing\ to\ a96b65a721.\char34{}}}
\trrow{trcode}{\textcolor{trink}{]}}{trcode}{\textcolor{trink}{)}}
\trrow{trcode}{\textcolor{trcomment}{\#\ Print\ a\ sample\ to\ verify}}{trcode}{\textcolor{trbuiltin}{print}\textcolor{trink}{(}\textcolor{trstring}{\char34{}Sample\ matches:\char34{}}\textcolor{trink}{,{}\ matches[:}\textcolor{trnumber}{10}\textcolor{trink}{])\ \ }\textcolor{trcomment}{\#\ Show\ first\ 10}}
\trrow{trcode}{\textcolor{trbuiltin}{print}\textcolor{trink}{(}}{white}{}
\trrow{trcode}{\textcolor{trink}{\ \ \ \ }\textcolor{trstring}{f\char34{}Found\ \{}\textcolor{trbuiltin}{len}\textcolor{trink}{(parents)}\textcolor{trstring}{\}\ parent\ nodes.\ First\ 5:\ }}{white}{}
\trrow{trcode}{\trmore \textcolor{trstring}{\{}\textcolor{trink}{parents[:}\textcolor{trnumber}{5}\textcolor{trink}{]}\textcolor{trstring}{\}\char34{}}}{white}{}
\trrow{trcode}{\textcolor{trink}{)}}{white}{}
\trrow{white}{\textcolor{trink}{This\ will\ give\ us\ the\ list\ of\ nodes\ that\ point\ to\ }}{white}{}
\trrow{white}{\textcolor{trink}{\char96{}25b2822c2f\char96{}.\ We\char39{}ll\ then\ submit\ them\ as\ the\ final\ answer.}}{white}{}
\trrow{white}{\textcolor{trink}{Proceeding.}}{white}{}
\trrow{trout}{\textcolor{trmuted}{Found\ 2\ parent\ nodes.\ First\ 5:\ [\char39{}959a557f5f\char39{},{}\ \char39{}45f31d16b1\char39{}]}}{trout}{\textcolor{trmuted}{Found\ 10\ edges\ pointing\ to\ a96b65a721.}}
\trrow{white}{}{trout}{\textcolor{trmuted}{Sample\ matches:\ [\char39{}fa7cdfad1a\char39{},{}\ \char39{}eefc9e10eb\char39{},{}\ \char39{}0d0871f080\char39{},{}\ }}\nobreak
\trrow{white}{}{trout}{\trmore \textcolor{trmuted}{\char39{}d94e18a8ad\char39{},{}\ \char39{}17e23e50be\char39{},{}\ \char39{}d1e946f4e6\char39{},{}\ \char39{}77369e37b2\char39{},{}\ }}\nobreak
\trrow{white}{}{trout}{\trmore \textcolor{trmuted}{\char39{}c61fbef63d\char39{},{}\ \char39{}e37b08dd30\char39{},{}\ \char39{}b2004314aa\char39{}]}}
\trhead{\trturn{trtrain}{TRAIN \textperiodcentered{} turn 3}}{\trturn{treval}{EVAL \textperiodcentered{} turn 3}}
\trrow{white}{\textcolor{trink}{We\ successfully\ extracted\ the\ parents\ of\ node\ \char96{}25b2822c2f\char96{}\ }}{white}{\textcolor{trink}{The\ regex\ search\ successfully\ found\ 10\ edges\ where\ }}\nobreak
\trrow{white}{\textcolor{trink}{by\ filtering\ the\ edge\ list\ for\ lines\ where\ the\ destination\ }}{white}{\textcolor{trink}{\char96{}a96b65a721\char96{}\ is\ the\ target\ node.\ The\ source\ nodes\ (i.e.,{}\ }}
\trrow{white}{\textcolor{trink}{is\ \char96{}25b2822c2f\char96{}.\ The\ result\ is:}}{white}{\textcolor{trink}{the\ parents)\ are\ correctly\ extracted.}}
\trrow{white}{\textcolor{trmuted}{[...\ 5\ lines\ omitted\ ...]}}{white}{\textcolor{trmuted}{[...\ 5\ lines\ omitted\ ...]}}
\trrow{white}{\textcolor{trink}{list\ of\ node\ IDs.}}{white}{\textcolor{trink}{We\ can\ now\ submit\ the\ result\ as\ a\ Python\ list.}}
\trrow{trcode}{\textcolor{trink}{answer[}\textcolor{trstring}{\char34{}content\char34{}}\textcolor{trink}{]\ =\ parents}}{trcode}{\textcolor{trink}{answer[}\textcolor{trstring}{\char34{}content\char34{}}\textcolor{trink}{]\ =\ matches}}
\trrow{trcode}{\textcolor{trink}{answer[}\textcolor{trstring}{\char34{}ready\char34{}}\textcolor{trink}{]\ =\ }\textcolor{trkeyword}{True}}{trcode}{\textcolor{trink}{answer[}\textcolor{trstring}{\char34{}ready\char34{}}\textcolor{trink}{]\ =\ }\textcolor{trkeyword}{True}}
\trrow{white}{}{trcode}{\textcolor{trbuiltin}{print}\textcolor{trink}{(}\textcolor{trstring}{\char34{}Final\ answer\ submitted:\char34{}}\textcolor{trink}{,{}\ answer[}\textcolor{trstring}{\char34{}content\char34{}}\textcolor{trink}{])}}
\trrow{trout}{\textcolor{trmuted}{[no\ output]}}{trout}{\textcolor{trmuted}{Final\ answer\ submitted:\ [\char39{}fa7cdfad1a\char39{},{}\ \char39{}eefc9e10eb\char39{},{}\ }}\nobreak
\trrow{white}{}{trout}{\trmore \textcolor{trmuted}{\char39{}0d0871f080\char39{},{}\ \char39{}d94e18a8ad\char39{},{}\ \char39{}17e23e50be\char39{},{}\ \char39{}d1e946f4e6\char39{},{}\ }}\nobreak
\trrow{white}{}{trout}{\trmore \textcolor{trmuted}{\char39{}77369e37b2\char39{},{}\ \char39{}c61fbef63d\char39{},{}\ \char39{}e37b08dd30\char39{},{}\ \char39{}b2004314aa\char39{}]}}
\endgroup
\clearpage
\refstepcounter{trexample}\label{ex:rlm-oolong}
{\normalsize\bfseries Example \thetrexample. RLM on OOLONG: the same root program, with more sub-calls\par}\vspace{5pt}
\begin{minipage}[t]{.492\linewidth}\footnotesize\raggedright\setlength{\parskip}{1.5pt}
{\trsans\bfseries\color{trtrain}TRAIN}\hfill{\trsans\bfseries\color{trtrain}Score 1.00}\par
\textbf{Context.} 787 questions, 78,138 characters.\par
\textbf{Task.} Is label `location' more or less common than `human being'?\par
\textbf{Gold.} less common than.\par
\textbf{Trajectory.} 8 root turns, 41 sub-calls\par
\end{minipage}\hfill
\begin{minipage}[t]{.492\linewidth}\footnotesize\raggedright\setlength{\parskip}{1.5pt}
{\trsans\bfseries\color{treval}EVAL}\hfill{\trsans\bfseries\color{treval}Score 1.00}\par
\textbf{Context.} 6,374 questions, 618,252 characters.\par
\textbf{Task.} Is label `human being' more or less common than `entity'?\par
\textbf{Gold.} more common than.\par
\textbf{Trajectory.} 7 root turns, 321 sub-calls\par
\end{minipage}\par
\vspace{5pt}{\footnotesize\textbf{Similarity.} Weighted Jaccard 0.60, the second-nearest of 6,528 training trajectories in this run (checkpoint at step 101). The nearest (0.60) runs the same program with batches of 30 and one failed cell.\par}
\vspace{4pt}{\small\color{black}\textbf{What we observe.} The root LLM writes the same LM program in both trajectories: split the context into questions, send them to sub-calls in batches of 20 for labeling, count the returned labels, and compare. The 8x longer evaluation context changes how many sub-calls run (41 vs. 321), but the overall program strategy is the same. The actual work each sub-call performs is also unique and different, but it relies on the existing capability for the LLM to solve these sub-tasks without additional training. \par}
\vspace{3pt}\nobreak
\begingroup\offinterlineskip\vbadness=10000
\trhead{\trturn{trtrain}{TRAIN \textperiodcentered{} turn 1}}{\trturn{treval}{EVAL \textperiodcentered{} turn 1}}
\trrow{white}{\textcolor{trink}{Plan:}}{white}{\textcolor{trink}{Plan:}}\nobreak
\trrow{white}{\textcolor{trink}{1.\ First,{}\ inspect\ the\ \char96{}context\char96{}\ to\ understand\ its\ }}{white}{\textcolor{trink}{1.\ First,{}\ inspect\ the\ \char96{}context\char96{}\ to\ understand\ its\ }}
\trrow{white}{\textcolor{trink}{structure.\ Since\ it\char39{}s\ a\ string\ with\ thousands\ of\ lines,{}\ }}{white}{\textcolor{trink}{structure.\ Since\ it\ contains\ thousands\ of\ }}
\trrow{white}{\textcolor{trmuted}{[...\ 21\ lines\ omitted\ ...]}}{white}{\textcolor{trmuted}{[...\ 20\ lines\ omitted\ ...]}}
\trrow{white}{\textcolor{trink}{its\ format.}}{white}{\textcolor{trink}{Turn\ 1:\ Probe\ the\ context\ to\ understand\ its\ structure.}}
\trrow{trcode}{\textcolor{trbuiltin}{print}\textcolor{trink}{(context[:}\textcolor{trnumber}{500}\textcolor{trink}{])\ \ }\textcolor{trcomment}{\#\ First\ 500\ chars\ for\ inspection}}{trcode}{\textcolor{trbuiltin}{print}\textcolor{trink}{(}}
\trrow{white}{}{trcode}{\textcolor{trink}{\ \ \ \ context[:}\textcolor{trnumber}{500}\textcolor{trink}{]}}
\trrow{white}{}{trcode}{\textcolor{trink}{)\ \ }\textcolor{trcomment}{\#\ Show\ first\ 500\ characters\ to\ inspect\ format}}
\trrow{trout}{\textcolor{trmuted}{The\ following\ lines\ contain\ 787\ general-{}knowledge\ }}{trout}{\textcolor{trmuted}{The\ following\ lines\ contain\ 6374\ general-{}knowledge\ }}
\trrow{trout}{\trmore \textcolor{trmuted}{questions,{}\ one\ per\ line.\ Each\ question\ has\ an\ answer\ that\ }}{trout}{\trmore \textcolor{trmuted}{questions,{}\ one\ per\ line.\ Each\ question\ has\ an\ answer\ that\ }}
\trrow{trout}{\trmore \textcolor{trmuted}{can\ be\ described\ as\ one\ of\ 6\ categories:\ \char39{}location\char39{},{}\ }}{trout}{\trmore \textcolor{trmuted}{can\ be\ described\ as\ one\ of\ 6\ categories:\ \char39{}numeric\ value\char39{},{}\ }}\nobreak
\trrow{trout}{\textcolor{trmuted}{[...\ 6\ lines\ omitted\ ...]}}{trout}{\textcolor{trmuted}{[...\ 6\ lines\ omitted\ ...]}}\nobreak
\trrow{trout}{\textcolor{trmuted}{Date:\ Apr\ 18,{}\ 2022}}{trout}{\textcolor{trmuted}{Date:\ May\ 21,{}\ 20}}
\trhead{\trturn{trtrain}{TRAIN \textperiodcentered{} turn 2}}{\trturn{treval}{EVAL \textperiodcentered{} turn 2}}
\trrow{white}{\textcolor{trink}{The\ context\ has\ been\ revealed:\ it\char39{}s\ a\ dataset\ of\ 787\ }}{white}{\textcolor{trink}{The\ context\ starts\ with\ a\ header,{}\ then\ contains\ 6374\ lines\ }}\nobreak
\trrow{white}{\textcolor{trink}{general-{}knowledge\ questions,{}\ each\ with\ an\ associated\ label\ }}{white}{\textcolor{trink}{of\ data,{}\ each\ with\ a\ User\ ID\ and\ a\ question.\ The\ header\ }}
\trrow{white}{\textcolor{trink}{among\ six\ categories.\ However,{}\ the\ actual\ data\ (the\ }}{white}{\textcolor{trink}{says\ \char34{}The\ following\ lines\ contain\ 6374\ general-{}knowledge\ }}
\trrow{white}{\textcolor{trmuted}{[...\ 19\ lines\ omitted\ ...]}}{white}{\textcolor{trmuted}{[...\ 11\ lines\ omitted\ ...]}}
\trrow{white}{\textcolor{trink}{lines.}}{white}{\textcolor{trink}{sample\ 10\ lines\ to\ verify\ parsing.}}
\trrow{trcode}{\textcolor{trink}{lines\ =\ context.strip().split(}\textcolor{trstring}{\char39{}\textbackslash{}n\char39{}}\textcolor{trink}{)}}{trcode}{\textcolor{trcomment}{\#\ Split\ context\ by\ lines}}
\trrow{trcode}{\textcolor{trcomment}{\#\ Find\ where\ the\ actual\ data\ starts\ -{}-{}\ skip\ metadata\ lines}}{trcode}{\textcolor{trink}{lines\ =\ context.strip().split(}\textcolor{trstring}{\char39{}\textbackslash{}n\char39{}}\textcolor{trink}{)}}
\trrow{trcode}{\textcolor{trcomment}{\#\ We\ know\ the\ first\ few\ lines\ are\ header.\ Let\char39{}s\ search\ for\ }}{trcode}{\textcolor{trcomment}{\#\ Find\ the\ first\ line\ that\ is\ not\ part\ of\ the\ header\ -{}-{}\ it\ }}
\trrow{trcode}{\trmore \textcolor{trcomment}{the\ first\ occurrence\ of\ a\ line\ starting\ with\ a\ digit\ or\ }}{trcode}{\trmore \textcolor{trcomment}{should\ start\ with\ a\ user\ ID}}
\trrow{trcode}{\trmore \textcolor{trcomment}{\char39{}User\ ID\char39{}}}{trcode}{\textcolor{trcomment}{\#\ Assume\ header\ ends\ when\ we\ see\ a\ line\ that\ looks\ like:\ }}
\trrow{trcode}{\textcolor{trcomment}{\#\ Or\ look\ for\ expected\ structure:\ e.g.,{}\ \char34{}123456789\textbackslash{}tWhat\ }}{trcode}{\trmore \textcolor{trcomment}{\char34{}user123456:\ What\ is\ the\ capital\ of\ France?\char34{}}}
\trrow{trcode}{\trmore \textcolor{trcomment}{is\ the\ capital\ of\ France?\char34{}}}{trcode}{\textcolor{trcomment}{\#\ So\ we\ can\ skip\ lines\ until\ we\ find\ one\ that\ has\ a\ colon\ }}
\trrow{trcode}{\textcolor{trink}{start\_idx\ =\ }\textcolor{trnumber}{0}}{trcode}{\trmore \textcolor{trcomment}{and\ a\ question}}
\trrow{trcode}{\textcolor{trkeyword}{for}\textcolor{trink}{\ i,{}\ line\ in\ }\textcolor{trbuiltin}{enumerate}\textcolor{trink}{(lines):}}{trcode}{\textcolor{trink}{data\_lines\ =\ []}}
\trrow{trcode}{\textcolor{trink}{\ \ \ \ }\textcolor{trkeyword}{if}\textcolor{trink}{\ }\textcolor{trbuiltin}{len}\textcolor{trink}{(line.strip())\ \textgreater{}\ }\textcolor{trnumber}{0}\textcolor{trink}{\ and\ (}}{trcode}{\textcolor{trkeyword}{for}\textcolor{trink}{\ line\ in\ lines:}}
\trrow{trcode}{\textcolor{trink}{\ \ \ \ \ \ \ \ line.strip()}}{trcode}{\textcolor{trink}{\ \ \ \ }\textcolor{trkeyword}{if}\textcolor{trink}{\ }\textcolor{trstring}{\char39{}:\char39{}}\textcolor{trink}{\ in\ line\ and\ }\textcolor{trbuiltin}{len}\textcolor{trink}{(line.split(}\textcolor{trstring}{\char39{}:\char39{}}\textcolor{trink}{))\ \textgreater{}=\ }\textcolor{trnumber}{2}\textcolor{trink}{:}}
\trrow{trcode}{\textcolor{trink}{\ \ \ \ \ \ \ \ .split(}\textcolor{trstring}{\char39{}\textbackslash{}t\char39{}}\textcolor{trink}{)[}\textcolor{trnumber}{0}\textcolor{trink}{]}}{trcode}{\textcolor{trink}{\ \ \ \ \ \ \ \ data\_lines.append(line.strip())}}
\trrow{trcode}{\textcolor{trink}{\ \ \ \ \ \ \ \ .replace(}\textcolor{trstring}{\char39{}\ \char39{}}\textcolor{trink}{,{}\ }\textcolor{trstring}{\char39{}\char39{}}\textcolor{trink}{)}}{trcode}{\textcolor{trcomment}{\#\ Sample\ 10\ lines}}
\trrow{trcode}{\textcolor{trink}{\ \ \ \ \ \ \ \ .isdigit()}}{trcode}{\textcolor{trbuiltin}{print}\textcolor{trink}{(data\_lines[:}\textcolor{trnumber}{10}\textcolor{trink}{])}}
\trrow{trcode}{\textcolor{trink}{\ \ \ \ \ \ \ \ or\ line.startswith(}\textcolor{trstring}{\char39{}User\ ID\char39{}}\textcolor{trink}{)}}{white}{}
\trrow{trcode}{\textcolor{trink}{\ \ \ \ ):}}{white}{}
\trrow{trcode}{\textcolor{trink}{\ \ \ \ \ \ \ \ start\_idx\ =\ (}}{white}{}
\trrow{trcode}{\textcolor{trink}{\ \ \ \ \ \ \ \ \ \ \ \ i\ +\ }\textcolor{trnumber}{1}}{white}{}
\trrow{trcode}{\textcolor{trmuted}{[...\ 9\ lines\ omitted\ ...]}}{white}{}
\trrow{trcode}{\textcolor{trink}{\ \ \ \ }\textcolor{trkeyword}{if}\textcolor{trink}{\ line.strip()}}{white}{}
\trrow{trcode}{\textcolor{trink}{]}}{white}{}
\trrow{trcode}{\textcolor{trcomment}{\#\ Print\ the\ first\ 5\ data\ lines}}{white}{}
\trrow{trcode}{\textcolor{trink}{data\_lines[:}\textcolor{trnumber}{5}\textcolor{trink}{]}}{white}{}
\trrow{trout}{\textcolor{trmuted}{[no\ output]}}{trout}{\textcolor{trmuted}{[\char34{}The\ following\ lines\ contain\ 6374\ general-{}knowledge\ }}
\trrow{white}{}{trout}{\trmore \textcolor{trmuted}{questions,{}\ one\ per\ line.\ Each\ question\ has\ an\ answer\ that\ }}
\trrow{white}{}{trout}{\trmore \textcolor{trmuted}{can\ be\ described\ as\ one\ of\ 6\ categories:\ \char39{}numeric\ value\char39{},{}\ }}\nobreak
\trrow{white}{}{trout}{\textcolor{trmuted}{[...\ 17\ lines\ omitted\ ...]}}\nobreak
\trrow{white}{}{trout}{\trmore \textcolor{trmuted}{||\ Instance:\ Who\ won\ the\ first\ World\ Series\ ?\char39{}]}}
\trhead{\trturn{trtrain}{TRAIN \textperiodcentered{} turn 3}}{\trnoturn}
\trrow{white}{\textcolor{trink}{The\ output\ shows\ \char96{}data\_lines\char96{}\ contains\ the\ actual\ dataset\ }}{white}{}\nobreak
\trrow{white}{\textcolor{trink}{lines,{}\ but\ we\ haven\char39{}t\ printed\ them\ yet.\ Now\ I\char39{}ll\ print\ the\ }}{white}{}
\trrow{white}{\textcolor{trink}{first\ few\ to\ inspect\ the\ format.}}{white}{}
\trrow{white}{\textcolor{trink}{Turn\ 3:\ Print\ the\ first\ 5\ lines\ of\ \char96{}data\_lines\char96{}\ to\ see\ if\ }}{white}{}
\trrow{white}{\textcolor{trink}{labels\ are\ included.}}{white}{}
\trrow{trcode}{\textcolor{trbuiltin}{print}\textcolor{trink}{(data\_lines[:}\textcolor{trnumber}{5}\textcolor{trink}{])}}{white}{}
\trrow{trout}{\textcolor{trmuted}{[\char34{}The\ following\ lines\ contain\ 787\ general-{}knowledge\ }}{white}{}
\trrow{trout}{\trmore \textcolor{trmuted}{questions,{}\ one\ per\ line.\ Each\ question\ has\ an\ answer\ that\ }}{white}{}
\trrow{trout}{\trmore \textcolor{trmuted}{can\ be\ described\ as\ one\ of\ 6\ categories:\ \char39{}location\char39{},{}\ }}{white}{}\nobreak
\trrow{trout}{\textcolor{trmuted}{[...\ 10\ lines\ omitted\ ...]}}{white}{}\nobreak
\trrow{trout}{\trmore \textcolor{trmuted}{for\ chicken\ pox\ ?\char39{}]}}{white}{}
\trhead{\trturn{trtrain}{TRAIN \textperiodcentered{} turn 4}}{\trturn{treval}{EVAL \textperiodcentered{} turn 3}}
\trrow{white}{\textcolor{trink}{The\ data\ lines\ are\ in\ the\ format:}}{white}{\textcolor{trink}{The\ data\ lines\ are\ not\ properly\ parsed\ -{}-{}\ the\ header\ is\ }}\nobreak
\trrow{white}{\textcolor{trink}{\char96{}Date:\ ...\ ||\ User:\ ...\ ||\ Instance:\ \textless{}question\textgreater{}\char96{}}}{white}{\textcolor{trink}{still\ present,{}\ and\ the\ format\ is\ inconsistent.\ The\ lines\ }}
\trrow{white}{\textcolor{trink}{But\ **no\ label\ is\ present**\ in\ the\ text\ -{}-{}\ only\ the\ }}{white}{\textcolor{trink}{contain\ metadata\ like\ \char34{}Date:\ ...\char34{},{}\ \char34{}User:\ ...\char34{},{}\ \char34{}Instance:\ }}
\trrow{white}{\textcolor{trmuted}{[...\ 40\ lines\ omitted\ ...]}}{white}{\textcolor{trmuted}{[...\ 5\ lines\ omitted\ ...]}}
\trrow{white}{\textcolor{trink}{metadata).}}{white}{\textcolor{trink}{Turn\ 3:\ Extract\ questions\ from\ the\ data\ lines\ using\ regex.}}
\trrow{trcode}{\textcolor{trink}{questions\ =\ []}}{trcode}{\textcolor{trkeyword}{import}\textcolor{trink}{\ re}}
\trrow{trcode}{\textcolor{trkeyword}{for}\textcolor{trink}{\ line\ in\ data\_lines:}}{trcode}{\textcolor{trink}{questions\ =\ []}}
\trrow{trcode}{\textcolor{trink}{\ \ \ \ }\textcolor{trkeyword}{if}\textcolor{trink}{\ }\textcolor{trstring}{\char39{}Instance:\char39{}}\textcolor{trink}{\ in\ line:}}{trcode}{\textcolor{trkeyword}{for}\textcolor{trink}{\ line\ in\ data\_lines:}}
\trrow{trcode}{\textcolor{trink}{\ \ \ \ \ \ \ \ }\textcolor{trcomment}{\#\ Split\ by\ \char39{}Instance:\char39{}\ and\ take\ the\ rest}}{trcode}{\textcolor{trink}{\ \ \ \ }\textcolor{trcomment}{\#\ Match\ the\ part\ after\ \char34{}Instance:\ \char34{}}}
\trrow{trcode}{\textcolor{trink}{\ \ \ \ \ \ \ \ q\ =\ line.split(}\textcolor{trstring}{\char39{}Instance:\char39{}}\textcolor{trink}{,{}\ }\textcolor{trnumber}{1}\textcolor{trink}{)[}\textcolor{trnumber}{1}\textcolor{trink}{].strip()}}{trcode}{\textcolor{trink}{\ \ \ \ match\ =\ re.search(}\textcolor{trstring}{r\char39{}Instance:\textbackslash{}s*(.+)\char39{}}\textcolor{trink}{,{}\ line)}}
\trrow{trcode}{\textcolor{trink}{\ \ \ \ \ \ \ \ questions.append(q)}}{trcode}{\textcolor{trink}{\ \ \ \ }\textcolor{trkeyword}{if}\textcolor{trink}{\ match:}}
\trrow{trcode}{\textcolor{trink}{\ \ \ \ }\textcolor{trkeyword}{else}\textcolor{trink}{:}}{trcode}{\textcolor{trink}{\ \ \ \ \ \ \ \ questions.append(match.group(}\textcolor{trnumber}{1}\textcolor{trink}{).strip())}}
\trrow{trcode}{\textcolor{trink}{\ \ \ \ \ \ \ \ }\textcolor{trcomment}{\#\ Fallback:\ try\ to\ extract\ anything\ after\ the\ last\ }}{trcode}{\textcolor{trcomment}{\#\ Sample\ first\ 10\ questions}}
\trrow{trcode}{\trmore \textcolor{trcomment}{\char39{}||\char39{}}}{trcode}{\textcolor{trbuiltin}{print}\textcolor{trink}{(questions[:}\textcolor{trnumber}{10}\textcolor{trink}{])}}
\trrow{trcode}{\textcolor{trink}{\ \ \ \ \ \ \ \ parts\ =\ line.split(}\textcolor{trstring}{\char39{}||\char39{}}\textcolor{trink}{)}}{white}{}
\trrow{trcode}{\textcolor{trink}{\ \ \ \ \ \ \ \ }\textcolor{trkeyword}{if}\textcolor{trink}{\ }\textcolor{trbuiltin}{len}\textcolor{trink}{(parts)\ \textgreater{}=\ }\textcolor{trnumber}{2}\textcolor{trink}{:}}{white}{}
\trrow{trcode}{\textcolor{trink}{\ \ \ \ \ \ \ \ \ \ \ \ q\ =\ parts[-{}}\textcolor{trnumber}{1}\textcolor{trink}{].strip()}}{white}{}
\trrow{trcode}{\textcolor{trink}{\ \ \ \ \ \ \ \ \ \ \ \ questions.append(q)}}{white}{}
\trrow{trcode}{\textcolor{trbuiltin}{print}\textcolor{trink}{(}\textcolor{trstring}{f\char34{}Extracted\ \{}\textcolor{trbuiltin}{len}\textcolor{trink}{(questions)}\textcolor{trstring}{\}\ questions.\char34{}}\textcolor{trink}{)}}{white}{}
\trrow{trcode}{\textcolor{trbuiltin}{print}\textcolor{trink}{(}\textcolor{trstring}{f\char34{}Sample:\ \{}\textcolor{trink}{questions[:}\textcolor{trnumber}{3}\textcolor{trink}{]}\textcolor{trstring}{\}\char34{}}\textcolor{trink}{)}}{white}{}
\trrow{trout}{\textcolor{trmuted}{Extracted\ 787\ questions.}}{trout}{\textcolor{trmuted}{[\char39{}What\ is\ the\ difference\ between\ fatalism\ and\ determinism\ }}
\trrow{trout}{\textcolor{trmuted}{Sample:\ [\char39{}What\ does\ NAFTA\ stand\ for\ ?\char39{},{}\ \char39{}What\ does\ A\&W\ of\ }}{trout}{\trmore \textcolor{trmuted}{?\char39{},{}\ \char39{}How\ many\ years\ did\ Shea\ \&\ Gould\ practice\ law\ in\ Los\ }}
\trrow{trout}{\trmore \textcolor{trmuted}{root\ beer\ fame\ stand\ for\ ?\char39{},{}\ \char39{}What\ is\ the\ name\ of\ the\ }}{trout}{\trmore \textcolor{trmuted}{Angeles\ ?\char39{},{}\ \char39{}What\ is\ the\ quantity\ of\ American\ soldiers\ }}\nobreak
\trrow{trout}{\trmore \textcolor{trmuted}{vaccine\ for\ chicken\ pox\ ?\char39{}]}}{trout}{\textcolor{trmuted}{[...\ 6\ lines\ omitted\ ...]}}\nobreak
\trrow{white}{}{trout}{\trmore \textcolor{trmuted}{wrote\ a\ book\ called\ Cruel\ Shoes\ ?\char39{}]}}
\trhead{\trturn{trtrain}{TRAIN \textperiodcentered{} turn 5}}{\trturn{treval}{EVAL \textperiodcentered{} turn 4}}
\trrow{white}{\textcolor{trink}{We\ have\ extracted\ 787\ questions.\ Now,{}\ we\ will\ }}{white}{\textcolor{trink}{Now\ that\ I\ have\ the\ questions,{}\ I\ need\ to\ classify\ each\ one\ }}\nobreak
\trrow{white}{\textcolor{trink}{batch-{}process\ them\ into\ chunks\ of\ 20\ (to\ fit\ within\ the\ }}{white}{\textcolor{trink}{into\ one\ of\ the\ six\ categories:\ \char39{}numeric\ value\char39{},{}\ \char39{}entity\char39{},{}\ }}
\trrow{white}{\textcolor{trink}{\char96{}llm\_query\_batched\char96{}\ concurrency\ limit),{}\ and\ classify\ each\ }}{white}{\textcolor{trink}{\char39{}location\char39{},{}\ \char39{}description\ and\ abstract\ concept\char39{},{}\ }}
\trrow{white}{\textcolor{trmuted}{[...\ 10\ lines\ omitted\ ...]}}{white}{\textcolor{trmuted}{[...\ 4\ lines\ omitted\ ...]}}
\trrow{white}{\textcolor{trink}{Turn\ 5:\ Create\ a\ sample\ prompt\ and\ run\ on\ first\ batch.}}{white}{\textcolor{trink}{small\ batch\ of\ 5\ questions.}}
\trrow{trcode}{\textcolor{trkeyword}{from}\textcolor{trink}{\ itertools\ }\textcolor{trkeyword}{import}\textcolor{trink}{\ zip\_longest}}{trcode}{\textcolor{trkeyword}{def}\textcolor{trink}{\ }\textcolor{trbuiltin}{make\_classification\_prompt}\textcolor{trink}{(questions):}}
\trrow{trcode}{\textcolor{trkeyword}{def}\textcolor{trink}{\ }\textcolor{trbuiltin}{grouper}\textcolor{trink}{(iterable,{}\ n,{}\ fillvalue=}\textcolor{trkeyword}{None}\textcolor{trink}{):}}{trcode}{\textcolor{trink}{\ \ \ \ prompt\ =\ }\textcolor{trstring}{\char34{}\char34{}\char34{}Classify\ each\ of\ the\ following\ questions\ }}
\trrow{trcode}{\textcolor{trink}{\ \ \ \ args\ =\ [}\textcolor{trbuiltin}{iter}\textcolor{trink}{(iterable)]\ *\ n}}{trcode}{\trmore \textcolor{trstring}{into\ one\ of\ these\ categories:}}
\trrow{trcode}{\textcolor{trink}{\ \ \ \ }\textcolor{trkeyword}{return}\textcolor{trink}{\ zip\_longest(*args,{}\ fillvalue=fillvalue)}}{trcode}{\textcolor{trstring}{-{}\ \char39{}numeric\ value\char39{}:\ if\ the\ answer\ is\ a\ number\ (e.g.,{}\ \char34{}How\ }}
\trrow{trcode}{\textcolor{trcomment}{\#\ Create\ batched\ prompts}}{trcode}{\trmore \textcolor{trstring}{many\ people\ live\ in\ Tokyo?\char34{}).}}
\trrow{trcode}{\textcolor{trink}{batch\_size\ =\ }\textcolor{trnumber}{20}}{trcode}{\textcolor{trstring}{-{}\ \char39{}entity\char39{}:\ if\ the\ answer\ is\ a\ concrete\ thing\ (e.g.,{}\ \char34{}What\ }}
\trrow{trcode}{\textcolor{trink}{batches\ =\ }\textcolor{trbuiltin}{list}\textcolor{trink}{(grouper(questions,{}\ batch\_size))}}{trcode}{\trmore \textcolor{trstring}{is\ the\ name\ of\ the\ largest\ river?\char34{}).}}
\trrow{trcode}{\textcolor{trcomment}{\#\ Generate\ a\ prompt\ for\ one\ batch}}{trcode}{\textcolor{trstring}{-{}\ \char39{}location\char39{}:\ if\ the\ answer\ is\ a\ place\ (e.g.,{}\ \char34{}Where\ is\ }}
\trrow{trcode}{\textcolor{trink}{prompt\_template\ =\ }\textcolor{trstring}{\char34{}\char34{}\char34{}}}{trcode}{\trmore \textcolor{trstring}{the\ Eiffel\ Tower?\char34{}).}}
\trrow{trcode}{\textcolor{trstring}{Classify\ the\ answer\ of\ each\ of\ these\ questions\ into\ one\ of:}}{trcode}{\textcolor{trstring}{-{}\ \char39{}description\ and\ abstract\ concept\char39{}:\ if\ the\ answer\ is\ a\ }}
\trrow{trcode}{\textcolor{trstring}{-{}\ \char39{}location\char39{}}}{trcode}{\trmore \textcolor{trstring}{definition\ or\ abstract\ idea\ (e.g.,{}\ \char34{}What\ is\ democracy?\char34{}).}}
\trrow{trcode}{\textcolor{trstring}{-{}\ \char39{}human\ being\char39{}}}{trcode}{\textcolor{trstring}{-{}\ \char39{}abbreviation\char39{}:\ if\ the\ answer\ is\ an\ acronym\ (e.g.,{}\ \char34{}What\ }}
\trrow{trcode}{\textcolor{trstring}{-{}\ \char39{}entity\char39{}}}{trcode}{\trmore \textcolor{trstring}{does\ NASA\ stand\ for?\char34{}).}}
\trrow{trcode}{\textcolor{trstring}{-{}\ \char39{}abbreviation\char39{}}}{trcode}{\textcolor{trstring}{-{}\ \char39{}human\ being\char39{}:\ if\ the\ answer\ is\ a\ person\ (e.g.,{}\ \char34{}Who\ }}
\trrow{trcode}{\textcolor{trstring}{-{}\ \char39{}numeric\ value\char39{}}}{trcode}{\trmore \textcolor{trstring}{wrote\ Hamlet?\char34{}).}}
\trrow{trcode}{\textcolor{trstring}{-{}\ \char39{}description\ and\ abstract\ concept\char39{}}}{trcode}{}
\trrow{trcode}{}{trcode}{\textcolor{trstring}{Return\ the\ label\ for\ each\ question\ in\ JSON\ format:\ }}
\trrow{trcode}{\textcolor{trstring}{Output\ a\ JSON\ list\ of\ labels,{}\ one\ per\ question,{}\ in\ the\ }}{trcode}{\trmore \textcolor{trstring}{[\{\char34{}question\char34{}:\ \char34{}...\char34{},{}\ \char34{}label\char34{}:\ \char34{}...\char34{}\},{}\ ...]}}
\trrow{trcode}{\textcolor{trmuted}{[...\ 20\ lines\ omitted\ ...]}}{trcode}{\textcolor{trmuted}{[...\ 9\ lines\ omitted\ ...]}}
\trrow{trcode}{\textcolor{trbuiltin}{print}\textcolor{trink}{(}\textcolor{trstring}{f\char34{}Batch\ size:\ \{}\textcolor{trbuiltin}{len}\textcolor{trink}{(first\_batch)}\textcolor{trstring}{\}\textbackslash{}n\char34{}}\textcolor{trink}{)}}{trcode}{\textcolor{trink}{prompt\ =\ make\_classification\_prompt(sample\_questions)}}
\trrow{trcode}{\textcolor{trcomment}{\#\ Run\ the\ batched\ LLM\ call}}{trcode}{\textcolor{trcomment}{\#\ Use\ llm\_query\ to\ get\ classification}}
\trrow{trsub}{\textcolor{trink}{results\ =\ llm\_query\_batched([prompt])}}{trsub}{\textcolor{trink}{classification\_result\ =\ llm\_query(prompt)}}
\trrow{trcode}{\textcolor{trbuiltin}{print}\textcolor{trink}{(}\textcolor{trstring}{\char34{}First\ result:\textbackslash{}n\char34{}}\textcolor{trink}{,{}\ results[}\textcolor{trnumber}{0}\textcolor{trink}{])}}{trcode}{\textcolor{trbuiltin}{print}\textcolor{trink}{(classification\_result)}}
\trrow{trout}{\textcolor{trmuted}{Prompt\ preview:}}{trout}{\textcolor{trmuted}{[}}
\trrow{trout}{\textcolor{trmuted}{Classify\ the\ answer\ of\ each\ of\ these\ questions\ into\ one\ of:}}{trout}{\textcolor{trmuted}{\ \ \{}}
\trrow{trout}{\textcolor{trmuted}{-{}\ \char39{}location\char39{}}}{trout}{\textcolor{trmuted}{\ \ \ \ \char34{}question\char34{}:\ \char34{}What\ is\ the\ difference\ between\ fatalism\ }}\nobreak
\trrow{trout}{\textcolor{trmuted}{[...\ 39\ lines\ omitted\ ...]}}{trout}{\textcolor{trmuted}{[...\ 21\ lines\ omitted\ ...]}}\nobreak
\trrow{trout}{\textcolor{trmuted}{]}}{trout}{\textcolor{trmuted}{]}}
\trhead{\trturn{trtrain}{TRAIN \textperiodcentered{} turn 6}}{\trturn{treval}{EVAL \textperiodcentered{} turn 5}}
\trrow{white}{\textcolor{trink}{The\ LLM\ is\ returning\ a\ list\ of\ labels\ in\ JSON\ format\ }}{white}{\textcolor{trink}{The\ classification\ works\ correctly\ on\ the\ sample.\ Now\ I\char39{}ll\ }}\nobreak
\trrow{white}{\textcolor{trink}{(well-{}formed),{}\ and\ the\ output\ matches\ the\ expected\ }}{white}{\textcolor{trink}{write\ a\ function\ to\ batch\ process\ the\ full\ list\ of\ }}
\trrow{white}{\textcolor{trink}{structure.}}{white}{\textcolor{trink}{questions.}}
\trrow{white}{\textcolor{trmuted}{[...\ 26\ lines\ omitted\ ...]}}{white}{\textcolor{trmuted}{[...\ 8\ lines\ omitted\ ...]}}
\trrow{white}{\textcolor{trink}{LLM\ call.}}{white}{\textcolor{trink}{Turn\ 5:\ Prepare\ batches\ and\ process\ them.}}
\trrow{trcode}{\textcolor{trcomment}{\#\ Build\ list\ of\ prompts\ for\ all\ batches}}{trcode}{\textcolor{trkeyword}{from}\textcolor{trink}{\ typing\ }\textcolor{trkeyword}{import}\textcolor{trink}{\ List,{}\ Dict}}
\trrow{trcode}{\textcolor{trink}{prompts\ =\ []}}{trcode}{\textcolor{trkeyword}{def}\textcolor{trink}{\ }\textcolor{trbuiltin}{make\_batch\_prompts}\textcolor{trink}{(}}
\trrow{trcode}{\textcolor{trkeyword}{for}\textcolor{trink}{\ batch\ in\ batches:}}{trcode}{\textcolor{trink}{\ \ \ \ questions:\ List[}\textcolor{trbuiltin}{str}\textcolor{trink}{],{}\ batch\_size:\ }\textcolor{trbuiltin}{int}\textcolor{trink}{\ =\ }\textcolor{trnumber}{20}}
\trrow{trcode}{\textcolor{trink}{\ \ \ \ }\textcolor{trcomment}{\#\ Skip\ nulls\ from\ padding}}{trcode}{\textcolor{trink}{)\ -{}\textgreater{}\ List[}\textcolor{trbuiltin}{str}\textcolor{trink}{]:}}
\trrow{trcode}{\textcolor{trink}{\ \ \ \ questions\_in\_batch\ =\ [}}{trcode}{\textcolor{trink}{\ \ \ \ prompts\ =\ []}}
\trrow{trcode}{\textcolor{trink}{\ \ \ \ \ \ \ \ q\ }\textcolor{trkeyword}{for}\textcolor{trink}{\ q\ in\ batch\ }\textcolor{trkeyword}{if}\textcolor{trink}{\ q\ is\ not\ }\textcolor{trkeyword}{None}}{trcode}{\textcolor{trink}{\ \ \ \ }\textcolor{trkeyword}{for}\textcolor{trink}{\ i\ in\ }\textcolor{trbuiltin}{range}\textcolor{trink}{(}\textcolor{trnumber}{0}\textcolor{trink}{,{}\ }\textcolor{trbuiltin}{len}\textcolor{trink}{(questions),{}\ batch\_size):}}
\trrow{trcode}{\textcolor{trink}{\ \ \ \ ]}}{trcode}{\textcolor{trink}{\ \ \ \ \ \ \ \ batch\ =\ questions[i\ :\ i\ +\ batch\_size]}}
\trrow{trcode}{\textcolor{trink}{\ \ \ \ questions\_str\ =\ }\textcolor{trstring}{\char34{}\textbackslash{}n\char34{}}\textcolor{trink}{.join(}}{trcode}{\textcolor{trink}{\ \ \ \ \ \ \ \ prompt\ =\ make\_classification\_prompt(batch)}}
\trrow{trcode}{\textcolor{trink}{\ \ \ \ \ \ \ \ }\textcolor{trstring}{f\char34{}Question:\ \{}\textcolor{trink}{q}\textcolor{trstring}{\}\char34{}}\textcolor{trink}{\ }\textcolor{trkeyword}{for}\textcolor{trink}{\ q\ in\ questions\_in\_batch}}{trcode}{\textcolor{trink}{\ \ \ \ \ \ \ \ prompts.append(prompt)}}
\trrow{trcode}{\textcolor{trink}{\ \ \ \ )}}{trcode}{\textcolor{trink}{\ \ \ \ }\textcolor{trkeyword}{return}\textcolor{trink}{\ prompts}}
\trrow{trcode}{\textcolor{trink}{\ \ \ \ prompt\ =\ prompt\_template.format(}}{trcode}{\textcolor{trcomment}{\#\ Split\ all\ questions\ into\ batches}}
\trrow{trcode}{\textcolor{trink}{\ \ \ \ \ \ \ \ questions=questions\_str}}{trcode}{\textcolor{trink}{batch\_prompts\ =\ make\_batch\_prompts(questions)}}
\trrow{trcode}{\textcolor{trink}{\ \ \ \ )}}{trcode}{\textcolor{trcomment}{\#\ Run\ batch\ classification}}
\trrow{trcode}{\textcolor{trink}{\ \ \ \ prompts.append(prompt)}}{trcode}{\textcolor{trbuiltin}{print}\textcolor{trink}{(}}
\trrow{trcode}{\textcolor{trcomment}{\#\ Now\ run\ all\ at\ once}}{trcode}{\textcolor{trink}{\ \ \ \ }\textcolor{trstring}{f\char34{}Processing\ \{}\textcolor{trbuiltin}{len}\textcolor{trink}{(batch\_prompts)}\textcolor{trstring}{\}\ batches\ of\ up\ to\ 20\ }}
\trrow{trcode}{\textcolor{trbuiltin}{print}\textcolor{trink}{(}}{trcode}{\trmore \textcolor{trstring}{questions\ each.\char34{}}}
\trrow{trcode}{\textcolor{trink}{\ \ \ \ }\textcolor{trstring}{f\char34{}Preparing\ \{}\textcolor{trbuiltin}{len}\textcolor{trink}{(prompts)}\textcolor{trstring}{\}\ prompts\ for\ batched\ LLM\ }}{trcode}{\textcolor{trink}{)}}
\trrow{trcode}{\trmore \textcolor{trstring}{call.\char34{}}}{trsub}{\textcolor{trink}{classification\_results\ =\ llm\_query\_batched(batch\_prompts)}}
\trrow{trcode}{\textcolor{trmuted}{[...\ 2\ lines\ omitted\ ...]}}{trcode}{\textcolor{trmuted}{[...\ 12\ lines\ omitted\ ...]}}
\trrow{trcode}{\textcolor{trcomment}{\#\ Run\ batched\ LLM\ call}}{trcode}{\textcolor{trcomment}{\#\ Count\ labels}}
\trrow{trsub}{\textcolor{trink}{results\ =\ llm\_query\_batched(prompts)}}{trcode}{\textcolor{trkeyword}{from}\textcolor{trink}{\ collections\ }\textcolor{trkeyword}{import}\textcolor{trink}{\ Counter}}
\trrow{trcode}{\textcolor{trbuiltin}{print}\textcolor{trink}{(}\textcolor{trstring}{f\char34{}Received\ \{}\textcolor{trbuiltin}{len}\textcolor{trink}{(results)}\textcolor{trstring}{\}\ results.\char34{}}\textcolor{trink}{)}}{trcode}{\textcolor{trink}{label\_counts\ =\ Counter(all\_labels)}}
\trrow{trcode}{\textcolor{trbuiltin}{print}\textcolor{trink}{(}\textcolor{trstring}{f\char34{}Sample\ result:\textbackslash{}n\{}\textcolor{trink}{results[}\textcolor{trnumber}{0}\textcolor{trink}{][:}\textcolor{trnumber}{200}\textcolor{trink}{]}\textcolor{trstring}{\}...\char34{}}\textcolor{trink}{)}}{trcode}{\textcolor{trbuiltin}{print}\textcolor{trink}{(}\textcolor{trstring}{f\char34{}Label\ counts:\ \{}\textcolor{trbuiltin}{dict}\textcolor{trink}{(label\_counts)}\textcolor{trstring}{\}\char34{}}\textcolor{trink}{)}}
\trrow{trout}{\textcolor{trmuted}{Preparing\ 40\ prompts\ for\ batched\ LLM\ call.}}{trout}{\textcolor{trmuted}{Processing\ 319\ batches\ of\ up\ to\ 20\ questions\ each.}}
\trrow{trout}{\textcolor{trmuted}{First\ prompt\ preview:}}{trout}{\textcolor{trmuted}{Label\ counts:\ \{\char39{}description\ and\ abstract\ concept\char39{}:\ 1428,{}\ }}
\trrow{trout}{\textcolor{trmuted}{Classify\ the\ answer\ of\ each\ of\ these\ questions\ into\ one\ of:}}{trout}{\trmore \textcolor{trmuted}{\char39{}numeric\ value\char39{}:\ 1001,{}\ \char39{}location\char39{}:\ 990,{}\ \char39{}abbreviation\char39{}:\ }}\nobreak
\trrow{trout}{\textcolor{trmuted}{[...\ 24\ lines\ omitted\ ...]}}{trout}{\trmore \textcolor{trmuted}{939,{}\ \char39{}human\ being\char39{}:\ 1442,{}\ \char39{}entity\char39{}:\ 568,{}\ \char39{}country\char39{}:\ 6\}}}\nobreak
\trrow{trout}{\textcolor{trmuted}{\ \ \char34{}numeric\ value...}}{white}{}
\trhead{\trturn{trtrain}{TRAIN \textperiodcentered{} turn 7}}{\trturn{treval}{EVAL \textperiodcentered{} turn 6}}
\trrow{white}{\textcolor{trink}{We\ have\ received\ 40\ results,{}\ each\ a\ JSON\ list\ of\ labels.\ }}{white}{\textcolor{trink}{I\ notice\ a\ problem:\ the\ label\ \char96{}\char39{}country\char39{}\char96{}\ appeared\ in\ the\ }}\nobreak
\trrow{white}{\textcolor{trink}{Now\ we\ need\ to:}}{white}{\textcolor{trink}{results,{}\ but\ it\char39{}s\ not\ one\ of\ the\ six\ categories.\ The\ }}
\trrow{white}{\textcolor{trink}{-{}\ Parse\ each\ result\ (as\ JSON)\ into\ a\ list\ of\ strings.}}{white}{\textcolor{trink}{categories\ are:\ \char39{}numeric\ value\char39{},{}\ \char39{}entity\char39{},{}\ \char39{}location\char39{},{}\ }}
\trrow{white}{\textcolor{trmuted}{[...\ 2\ lines\ omitted\ ...]}}{white}{\textcolor{trmuted}{[...\ 27\ lines\ omitted\ ...]}}
\trrow{white}{\textcolor{trink}{Turn\ 7:\ Parse\ results\ and\ count\ labels.}}{white}{\textcolor{trink}{\char39{}entity\char39{}\ or\ \char39{}human\ being\char39{}\ to\ verify.}}
\trrow{trcode}{\textcolor{trkeyword}{import}\textcolor{trink}{\ json}}{trcode}{\textcolor{trcomment}{\#\ Find\ a\ few\ examples\ of\ \char39{}entity\char39{}\ and\ \char39{}human\ being\char39{}\ from\ }}
\trrow{trcode}{\textcolor{trink}{label\_counts\ =\ \{}\textcolor{trstring}{\char39{}location\char39{}}\textcolor{trink}{:\ }\textcolor{trnumber}{0}\textcolor{trink}{,{}\ }\textcolor{trstring}{\char39{}human\ being\char39{}}\textcolor{trink}{:\ }\textcolor{trnumber}{0}\textcolor{trink}{\}}}{trcode}{\trmore \textcolor{trcomment}{the\ full\ list}}
\trrow{trcode}{\textcolor{trkeyword}{for}\textcolor{trink}{\ i,{}\ result\ in\ }\textcolor{trbuiltin}{enumerate}\textcolor{trink}{(results):}}{trcode}{\textcolor{trcomment}{\#\ We\ need\ to\ re-{}parse\ the\ full\ classification\ results\ to\ }}
\trrow{trcode}{\textcolor{trink}{\ \ \ \ }\textcolor{trkeyword}{try}\textcolor{trink}{:}}{trcode}{\trmore \textcolor{trcomment}{get\ the\ mapping}}
\trrow{trcode}{\textcolor{trink}{\ \ \ \ \ \ \ \ labels\ =\ json.loads(result)}}{trcode}{\textcolor{trcomment}{\#\ But\ we\ lost\ the\ original\ question-{}to-{}label\ mapping.\ }}
\trrow{trcode}{\textcolor{trink}{\ \ \ \ \ \ \ \ }\textcolor{trkeyword}{for}\textcolor{trink}{\ label\ in\ labels:}}{trcode}{\trmore \textcolor{trcomment}{Let\char39{}s\ re-{}run\ a\ small\ batch\ with\ full\ context.}}
\trrow{trcode}{\textcolor{trink}{\ \ \ \ \ \ \ \ \ \ \ \ }\textcolor{trkeyword}{if}\textcolor{trink}{\ label\ in\ label\_counts:}}{trcode}{\textcolor{trcomment}{\#\ Instead,{}\ let\char39{}s\ re-{}run\ classification\ on\ a\ small\ sample\ }}
\trrow{trcode}{\textcolor{trink}{\ \ \ \ \ \ \ \ \ \ \ \ \ \ \ \ label\_counts[label]\ +=\ }\textcolor{trnumber}{1}}{trcode}{\trmore \textcolor{trcomment}{with\ the\ original\ questions\ and\ labels}}
\trrow{trcode}{\textcolor{trink}{\ \ \ \ }\textcolor{trkeyword}{except}\textcolor{trink}{\ Exception\ }\textcolor{trkeyword}{as}\textcolor{trink}{\ e:}}{trcode}{\textcolor{trcomment}{\#\ We\char39{}ll\ do\ a\ fresh\ small\ batch\ to\ verify.}}
\trrow{trcode}{\textcolor{trink}{\ \ \ \ \ \ \ \ }\textcolor{trbuiltin}{print}\textcolor{trink}{(}\textcolor{trstring}{f\char34{}Error\ parsing\ result\ \{}\textcolor{trink}{i}\textcolor{trstring}{\}:\ \{}\textcolor{trink}{result[:}\textcolor{trnumber}{100}\textcolor{trink}{]}\textcolor{trstring}{\}\char34{}}\textcolor{trink}{)}}{trcode}{\textcolor{trcomment}{\#\ Take\ 10\ questions\ that\ were\ classified\ as\ \char39{}entity\char39{}\ or\ }}
\trrow{trcode}{\textcolor{trink}{\ \ \ \ \ \ \ \ }\textcolor{trcomment}{\#\ Fallback:\ try\ to\ treat\ as\ raw\ string\ if\ JSON\ }}{trcode}{\trmore \textcolor{trcomment}{\char39{}human\ being\char39{}\ from\ the\ full\ list}}
\trrow{trcode}{\trmore \textcolor{trcomment}{fails}}{trcode}{\textcolor{trcomment}{\#\ But\ we\ don\char39{}t\ have\ the\ mapping.\ So\ let\char39{}s\ re-{}run\ }}
\trrow{trcode}{\textcolor{trink}{\ \ \ \ \ \ \ \ }\textcolor{trcomment}{\#\ But\ better\ to\ not\ break,{}\ just\ log}}{trcode}{\trmore \textcolor{trcomment}{classification\ on\ a\ fresh\ sample\ with\ explicit\ instruction.}}
\trrow{trcode}{\textcolor{trbuiltin}{print}\textcolor{trink}{(}\textcolor{trstring}{\char34{}Label\ counts:\char34{}}\textcolor{trink}{,{}\ label\_counts)}}{trcode}{\textcolor{trcomment}{\#\ Rebuild\ a\ small\ test\ batch\ with\ known\ examples}}
\trrow{white}{}{trcode}{\textcolor{trink}{test\_questions\ =\ [}}
\trrow{white}{}{trcode}{\textcolor{trink}{\ \ \ \ }\textcolor{trstring}{\char34{}Who\ wrote\ \char39{}Pride\ and\ Prejudice\char39{}?\char34{}}\textcolor{trink}{,{}}}
\trrow{white}{}{trcode}{\textcolor{trink}{\ \ \ \ }\textcolor{trstring}{\char34{}What\ is\ the\ name\ of\ the\ largest\ river\ in\ the\ world?\char34{}}\textcolor{trink}{,{}}}
\trrow{white}{}{trcode}{\textcolor{trink}{\ \ \ \ }\textcolor{trstring}{\char34{}Who\ was\ the\ first\ president\ of\ the\ United\ States?\char34{}}\textcolor{trink}{,{}}}
\trrow{white}{}{trcode}{\textcolor{trmuted}{[...\ 7\ lines\ omitted\ ...]}}
\trrow{white}{}{trcode}{\textcolor{trink}{]}}
\trrow{white}{}{trcode}{\textcolor{trink}{prompt\ =\ make\_classification\_prompt(test\_questions)}}
\trrow{white}{}{trsub}{\textcolor{trink}{result\ =\ llm\_query(prompt)}}
\trrow{white}{}{trcode}{\textcolor{trbuiltin}{print}\textcolor{trink}{(result)}}
\trrow{trout}{\textcolor{trmuted}{Label\ counts:\ \{\char39{}location\char39{}:\ 108,{}\ \char39{}human\ being\char39{}:\ 143\}}}{trout}{\textcolor{trmuted}{[}}
\trrow{white}{}{trout}{\textcolor{trmuted}{\ \ \{\char34{}question\char34{}:\ \char34{}Who\ wrote\ \char39{}Pride\ and\ Prejudice\char39{}?\char34{},{}\ }}
\trrow{white}{}{trout}{\trmore \textcolor{trmuted}{\char34{}label\char34{}:\ \char34{}human\ being\char34{}\},{}}}\nobreak
\trrow{white}{}{trout}{\textcolor{trmuted}{[...\ 18\ lines\ omitted\ ...]}}\nobreak
\trrow{white}{}{trout}{\textcolor{trmuted}{]}}
\trhead{\trturn{trtrain}{TRAIN \textperiodcentered{} turn 8}}{\trturn{treval}{EVAL \textperiodcentered{} turn 7}}
\trrow{white}{\textcolor{trink}{We\ now\ have:}}{white}{\textcolor{trink}{The\ test\ confirms\ the\ classification\ logic\ is\ sound:}}\nobreak
\trrow{white}{\textcolor{trink}{-{}\ \char96{}location\char96{}:\ 108}}{white}{\textcolor{trink}{-{}\ \char34{}Who\ wrote...\char34{}\ -{}\textgreater{}\ \char39{}human\ being\char39{}}}
\trrow{white}{\textcolor{trink}{-{}\ \char96{}human\ being\char96{}:\ 143}}{white}{\textcolor{trink}{-{}\ \char34{}What\ is\ the\ name\ of...\char34{}\ -{}\textgreater{}\ \char39{}entity\char39{}}}
\trrow{white}{\textcolor{trmuted}{[...\ 11\ lines\ omitted\ ...]}}{white}{\textcolor{trmuted}{[...\ 12\ lines\ omitted\ ...]}}
\trrow{white}{\textcolor{trink}{Turn\ 8:\ Set\ answer.}}{white}{\textcolor{trink}{Turn\ 7:\ Finalize\ answer.}}
\trrow{trcode}{\textcolor{trink}{answer[}\textcolor{trstring}{\char34{}content\char34{}}\textcolor{trink}{]\ =\ (}}{trcode}{\textcolor{trcomment}{\#\ Correct\ the\ label\ counts}}
\trrow{trcode}{\textcolor{trink}{\ \ \ \ }\textcolor{trstring}{\char34{}Answer:\ location\ is\ less\ common\ than\ human\ being\char34{}}}{trcode}{\textcolor{trink}{label\_counts[}\textcolor{trstring}{\char39{}location\char39{}}\textcolor{trink}{]\ +=\ label\_counts.pop(}\textcolor{trstring}{\char39{}country\char39{}}\textcolor{trink}{,{}\ }\textcolor{trnumber}{0}\textcolor{trink}{)}}
\trrow{trcode}{\textcolor{trink}{)}}{trcode}{\textcolor{trcomment}{\#\ Final\ comparison}}
\trrow{trcode}{\textcolor{trink}{answer[}\textcolor{trstring}{\char34{}ready\char34{}}\textcolor{trink}{]\ =\ }\textcolor{trkeyword}{True}}{trcode}{\textcolor{trink}{human\_being\_count\ =\ label\_counts[}\textcolor{trstring}{\char39{}human\ being\char39{}}\textcolor{trink}{]}}
\trrow{white}{}{trcode}{\textcolor{trink}{entity\_count\ =\ label\_counts[}\textcolor{trstring}{\char39{}entity\char39{}}\textcolor{trink}{]}}
\trrow{white}{}{trcode}{\textcolor{trkeyword}{if}\textcolor{trink}{\ human\_being\_count\ \textgreater{}\ entity\_count:}}
\trrow{white}{}{trcode}{\textcolor{trink}{\ \ \ \ answer[}\textcolor{trstring}{\char34{}content\char34{}}\textcolor{trink}{]\ =\ (}}
\trrow{white}{}{trcode}{\textcolor{trink}{\ \ \ \ \ \ \ \ }\textcolor{trstring}{\char34{}Answer:\ human\ being\ is\ more\ common\ than\ entity\char34{}}}
\trrow{white}{}{trcode}{\textcolor{trink}{\ \ \ \ )}}
\trrow{white}{}{trcode}{\textcolor{trkeyword}{elif}\textcolor{trink}{\ human\_being\_count\ \textless{}\ entity\_count:}}
\trrow{white}{}{trcode}{\textcolor{trink}{\ \ \ \ answer[}\textcolor{trstring}{\char34{}content\char34{}}\textcolor{trink}{]\ =\ (}}
\trrow{white}{}{trcode}{\textcolor{trink}{\ \ \ \ \ \ \ \ }\textcolor{trstring}{\char34{}Answer:\ human\ being\ is\ less\ common\ than\ entity\char34{}}}
\trrow{white}{}{trcode}{\textcolor{trink}{\ \ \ \ )}}
\trrow{white}{}{trcode}{\textcolor{trkeyword}{else}\textcolor{trink}{:}}
\trrow{white}{}{trcode}{\textcolor{trink}{\ \ \ \ answer[}\textcolor{trstring}{\char34{}content\char34{}}\textcolor{trink}{]\ =\ (}}
\trrow{white}{}{trcode}{\textcolor{trink}{\ \ \ \ \ \ \ \ }\textcolor{trstring}{\char34{}Answer:\ human\ being\ is\ same\ frequency\ as\ entity\char34{}}}
\trrow{white}{}{trcode}{\textcolor{trink}{\ \ \ \ )}}
\trrow{white}{}{trcode}{\textcolor{trink}{answer[}\textcolor{trstring}{\char34{}ready\char34{}}\textcolor{trink}{]\ =\ }\textcolor{trkeyword}{True}}\nobreak
\trrow{white}{}{trcode}{\textcolor{trbuiltin}{print}\textcolor{trink}{(answer[}\textcolor{trstring}{\char34{}content\char34{}}\textcolor{trink}{])}}\nobreak
\trrow{trout}{\textcolor{trmuted}{[no\ output]}}{trout}{\textcolor{trmuted}{Answer:\ human\ being\ is\ more\ common\ than\ entity}}
\endgroup
\clearpage
\refstepcounter{trexample}\label{ex:pi-graphwalks}
{\normalsize\bfseries Example \thetrexample. Pi on GraphWalks: a programmatic task solved with shell commands\par}\vspace{5pt}
\begin{minipage}[t]{.492\linewidth}\footnotesize\raggedright\setlength{\parskip}{1.5pt}
{\trsans\bfseries\color{trtrain}TRAIN}\hfill{\trsans\bfseries\color{trtrain}F1 1.00}\par
\textbf{Context.} context.txt, 4,392 lines, 110,363 bytes.\par
\textbf{Task.} Return the nodes at exactly depth 4 in a BFS from node 898aef0932.\par
\textbf{Gold.} 1 node (a385d7d1e5).\par
\textbf{Trajectory.} 9 turns, 9 tool calls, 0 sub-agent calls\par
\end{minipage}\hfill
\begin{minipage}[t]{.492\linewidth}\footnotesize\raggedright\setlength{\parskip}{1.5pt}
{\trsans\bfseries\color{treval}EVAL}\hfill{\trsans\bfseries\color{treval}F1 1.00}\par
\textbf{Context.} context.txt, 69,912 lines, 1,748,363 bytes.\par
\textbf{Task.} Return the nodes at exactly depth 4 in a BFS from node e57c68e8f2.\par
\textbf{Gold.} 1 node (a0604c9d9e).\par
\textbf{Trajectory.} 9 turns, 9 tool calls, 0 sub-agent calls\par
\end{minipage}\par
\vspace{5pt}{\footnotesize\textbf{Similarity.} Weighted Jaccard 0.71, the nearest of 9,600 training trajectories in this run (checkpoint at step 150).\par}
\vspace{4pt}{\small\color{black}\textbf{What we observe.} Even without programmatic sub-agent calling capabilities, harnesses like Pi can also exhibit generalization. Pi solves the programmatic task directly without sub-agents, here with shell commands: find the operation line, then run one grep per breadth-first search hop. The sequence of commands is the same on a 16x longer file. Only the node IDs and line numbers differ. The whole-file read in turn 3 is cut off by the tool and is not needed for the answer.\par}
\vspace{3pt}\nobreak
\begingroup\offinterlineskip\vbadness=10000
\trhead{\trturn{trtrain}{TRAIN \textperiodcentered{} turn 1}}{\trturn{treval}{EVAL \textperiodcentered{} turn 1}}
\trrow{trcode}{\textcolor{trmuted}{\$\ }\textcolor{trink}{wc\ -{}l\ context.txt}}{trcode}{\textcolor{trmuted}{\$\ }\textcolor{trink}{wc\ -{}l\ context.txt}}\nobreak
\trrow{trout}{\textcolor{trmuted}{4392\ context.txt}}{trout}{\textcolor{trmuted}{69912\ context.txt}}\nobreak
\trrow{trcode}{\textcolor{trmuted}{\$\ }\textcolor{trink}{wc\ -{}c\ context.txt}}{trcode}{\textcolor{trmuted}{\$\ }\textcolor{trink}{wc\ -{}c\ context.txt}}\nobreak
\trrow{trout}{\textcolor{trmuted}{110363\ context.txt}}{trout}{\textcolor{trmuted}{1748363\ context.txt}}
\trhead{\trturn{trtrain}{TRAIN \textperiodcentered{} turn 2}}{\trturn{treval}{EVAL \textperiodcentered{} turn 2}}
\trrow{trcode}{\textcolor{trmuted}{\$\ }\textcolor{trink}{grep\ -{}n\ }\textcolor{trstring}{\char39{}Operation:\char39{}}\textcolor{trink}{\ context.txt}}{trcode}{\textcolor{trmuted}{\$\ }\textcolor{trink}{grep\ -{}n\ }\textcolor{trstring}{\char39{}Operation:\char39{}}\textcolor{trink}{\ context.txt}}\nobreak
\trrow{trout}{\textcolor{trmuted}{15:Operation:}}{trout}{\textcolor{trmuted}{15:Operation:}}
\trrow{trout}{\textcolor{trmuted}{20:Operation:}}{trout}{\textcolor{trmuted}{20:Operation:}}
\trrow{trout}{\textcolor{trmuted}{25:Operation:}}{trout}{\textcolor{trmuted}{25:Operation:}}\nobreak
\trrow{trout}{\textcolor{trmuted}{30:Operation:}}{trout}{\textcolor{trmuted}{30:Operation:}}\nobreak
\trrow{trout}{\textcolor{trmuted}{4386:Operation:}}{trout}{\textcolor{trmuted}{69906:Operation:}}
\trhead{\trturn{trtrain}{TRAIN \textperiodcentered{} turn 3}}{\trturn{treval}{EVAL \textperiodcentered{} turn 3}}
\trrow{trcode}{\textcolor{trink}{read(path=}\textcolor{trstring}{\char34{}context.txt\char34{}}\textcolor{trink}{,{}\ offset=}\textcolor{trnumber}{1}\textcolor{trink}{,{}\ limit=}\textcolor{trnumber}{4385}\textcolor{trink}{)}}{trcode}{\textcolor{trink}{read(path=}\textcolor{trstring}{\char34{}context.txt\char34{}}\textcolor{trink}{,{}\ offset=}\textcolor{trnumber}{1}\textcolor{trink}{,{}\ limit=}\textcolor{trnumber}{69905}\textcolor{trink}{)}}\nobreak
\trrow{trout}{\textcolor{trmuted}{You\ will\ be\ given\ a\ graph\ as\ a\ list\ of\ directed\ edges.\ All\ }}{trout}{\textcolor{trmuted}{You\ will\ be\ given\ a\ graph\ as\ a\ list\ of\ directed\ edges.\ All\ }}
\trrow{trout}{\trmore \textcolor{trmuted}{nodes\ are\ at\ least\ degree\ 1.\ You\ will\ also\ get\ a\ }}{trout}{\trmore \textcolor{trmuted}{nodes\ are\ at\ least\ degree\ 1.\ You\ will\ also\ get\ a\ }}
\trrow{trout}{\trmore \textcolor{trmuted}{description\ of\ an\ operation\ to\ perform\ on\ the\ graph.}}{trout}{\trmore \textcolor{trmuted}{description\ of\ an\ operation\ to\ perform\ on\ the\ graph.}}\nobreak
\trrow{trout}{\textcolor{trmuted}{[...\ 232\ lines\ omitted\ ...]}}{trout}{\textcolor{trmuted}{[...\ 232\ lines\ omitted\ ...]}}\nobreak
\trrow{trout}{\trmore \textcolor{trmuted}{continue.]}}{trout}{\trmore \textcolor{trmuted}{continue.]}}
\trhead{\trturn{trtrain}{TRAIN \textperiodcentered{} turn 4}}{\trturn{treval}{EVAL \textperiodcentered{} turn 4}}
\trrow{trcode}{\textcolor{trink}{read(path=}\textcolor{trstring}{\char34{}context.txt\char34{}}\textcolor{trink}{,{}\ offset=}\textcolor{trnumber}{4386}\textcolor{trink}{,{}\ limit=}\textcolor{trnumber}{10}\textcolor{trink}{)}}{trcode}{\textcolor{trink}{read(path=}\textcolor{trstring}{\char34{}context.txt\char34{}}\textcolor{trink}{,{}\ offset=}\textcolor{trnumber}{69906}\textcolor{trink}{,{}\ limit=}\textcolor{trnumber}{10}\textcolor{trink}{)}}\nobreak
\trrow{trout}{\textcolor{trmuted}{Operation:}}{trout}{\textcolor{trmuted}{Operation:}}
\trrow{trout}{\textcolor{trmuted}{Perform\ a\ BFS\ from\ node\ 898aef0932\ and\ return\ only\ the\ }}{trout}{\textcolor{trmuted}{Perform\ a\ BFS\ from\ node\ e57c68e8f2\ and\ return\ only\ the\ }}
\trrow{trout}{\trmore \textcolor{trmuted}{nodes\ at\ exactly\ depth\ 4\ (not\ nodes\ at\ intermediate\ }}{trout}{\trmore \textcolor{trmuted}{nodes\ at\ exactly\ depth\ 4\ (not\ nodes\ at\ intermediate\ }}\nobreak
\trrow{trout}{\textcolor{trmuted}{[...\ 10\ lines\ omitted\ ...]}}{trout}{\textcolor{trmuted}{[...\ 10\ lines\ omitted\ ...]}}\nobreak
\trrow{trout}{\textcolor{trmuted}{Final\ Answer:\ []}}{trout}{\textcolor{trmuted}{Final\ Answer:\ []}}
\trhead{\trturn{trtrain}{TRAIN \textperiodcentered{} turn 5}}{\trturn{treval}{EVAL \textperiodcentered{} turn 5}}
\trrow{trcode}{\textcolor{trmuted}{\$\ }\textcolor{trink}{grep\ -{}n\ }\textcolor{trstring}{\char39{}898aef0932\ -{}\textgreater{}\char39{}}\textcolor{trink}{\ context.txt}}{trcode}{\textcolor{trmuted}{\$\ }\textcolor{trink}{grep\ -{}n\ }\textcolor{trstring}{\char39{}e57c68e8f2\ -{}\textgreater{}\char39{}}\textcolor{trink}{\ context.txt}}\nobreak
\trrow{trout}{\textcolor{trmuted}{3719:898aef0932\ -{}\textgreater{}\ 537de305e9}}{trout}{\textcolor{trmuted}{66594:e57c68e8f2\ -{}\textgreater{}\ 66fab71491}}
\trhead{\trturn{trtrain}{TRAIN \textperiodcentered{} turn 6}}{\trturn{treval}{EVAL \textperiodcentered{} turn 6}}
\trrow{trcode}{\textcolor{trmuted}{\$\ }\textcolor{trink}{grep\ -{}n\ }\textcolor{trstring}{\char39{}537de305e9\ -{}\textgreater{}\char39{}}\textcolor{trink}{\ context.txt}}{trcode}{\textcolor{trmuted}{\$\ }\textcolor{trink}{grep\ -{}n\ }\textcolor{trstring}{\char39{}66fab71491\ -{}\textgreater{}\char39{}}\textcolor{trink}{\ context.txt}}\nobreak
\trrow{trout}{\textcolor{trmuted}{1896:537de305e9\ -{}\textgreater{}\ 26751be118}}{trout}{\textcolor{trmuted}{62319:66fab71491\ -{}\textgreater{}\ be03f76cc8}}
\trhead{\trturn{trtrain}{TRAIN \textperiodcentered{} turn 7}}{\trturn{treval}{EVAL \textperiodcentered{} turn 7}}
\trrow{trcode}{\textcolor{trmuted}{\$\ }\textcolor{trink}{grep\ -{}n\ }\textcolor{trstring}{\char39{}26751be118\ -{}\textgreater{}\char39{}}\textcolor{trink}{\ context.txt}}{trcode}{\textcolor{trmuted}{\$\ }\textcolor{trink}{grep\ -{}n\ }\textcolor{trstring}{\char39{}be03f76cc8\ -{}\textgreater{}\char39{}}\textcolor{trink}{\ context.txt}}\nobreak
\trrow{trout}{\textcolor{trmuted}{1731:26751be118\ -{}\textgreater{}\ deb54ffb41}}{trout}{\textcolor{trmuted}{25789:be03f76cc8\ -{}\textgreater{}\ fc48e06c4f}}
\trhead{\trturn{trtrain}{TRAIN \textperiodcentered{} turn 8}}{\trturn{treval}{EVAL \textperiodcentered{} turn 8}}
\trrow{trcode}{\textcolor{trmuted}{\$\ }\textcolor{trink}{grep\ -{}n\ }\textcolor{trstring}{\char39{}deb54ffb41\ -{}\textgreater{}\char39{}}\textcolor{trink}{\ context.txt}}{trcode}{\textcolor{trmuted}{\$\ }\textcolor{trink}{grep\ -{}n\ }\textcolor{trstring}{\char39{}fc48e06c4f\ -{}\textgreater{}\char39{}}\textcolor{trink}{\ context.txt}}\nobreak
\trrow{trout}{\textcolor{trmuted}{2010:deb54ffb41\ -{}\textgreater{}\ a385d7d1e5}}{trout}{\textcolor{trmuted}{33241:fc48e06c4f\ -{}\textgreater{}\ a0604c9d9e}}
\trhead{\trturn{trtrain}{TRAIN \textperiodcentered{} turn 9}}{\trturn{treval}{EVAL \textperiodcentered{} turn 9}}
\trrow{white}{\textcolor{trink}{The\ BFS\ from\ node\ \char96{}898aef0932\char96{}\ with\ depth\ 4\ proceeds\ as\ }}{white}{\textcolor{trink}{The\ BFS\ from\ node\ \char96{}e57c68e8f2\char96{}\ with\ depth\ 4\ proceeds\ as\ }}\nobreak
\trrow{white}{\textcolor{trink}{follows:}}{white}{\textcolor{trink}{follows:}}
\trrow{white}{\textcolor{trink}{-{}\ Depth\ 1:\ \char96{}537de305e9\char96{}\ (from\ \char96{}898aef0932\ -{}\textgreater{}\ 537de305e9\char96{})}}{white}{\textcolor{trink}{-{}\ Depth\ 1:\ \char96{}66fab71491\char96{}\ (directly\ from\ \char96{}e57c68e8f2\char96{})}}\nobreak
\trrow{white}{\textcolor{trmuted}{[...\ 4\ lines\ omitted\ ...]}}{white}{\textcolor{trmuted}{[...\ 4\ lines\ omitted\ ...]}}\nobreak
\trrow{white}{\textcolor{trink}{final\ answer:\ [\char34{}a385d7d1e5\char34{}]}}{white}{\textcolor{trink}{final\ answer:\ [\char34{}a0604c9d9e\char34{}]}}
\endgroup
\clearpage
\refstepcounter{trexample}\label{ex:pi-oolong-pairs}
{\normalsize\bfseries Example \thetrexample. Pi on OOLONG-Pairs: failure in training and in evaluation\par}\vspace{5pt}
\begin{minipage}[t]{.492\linewidth}\footnotesize\raggedright\setlength{\parskip}{1.5pt}
{\trsans\bfseries\color{trtrain}TRAIN}\hfill{\trsans\bfseries\color{trfail}F1 0.03}\par
\textbf{Context.} context.txt, 188 questions.\par

\textbf{Gold.} 75 pairs (4 predicted).\par
\textbf{Trajectory.} 6 turns, 6 tool calls, 1 sub-agent call\par
\end{minipage}\hfill
\begin{minipage}[t]{.492\linewidth}\footnotesize\raggedright\setlength{\parskip}{1.5pt}
{\trsans\bfseries\color{treval}EVAL}\hfill{\trsans\bfseries\color{trfail}F1 0.004}\par
\textbf{Context.} context.txt, 787 questions.\par

\textbf{Gold.} 980 pairs (2 predicted).\par
\textbf{Trajectory.} 5 turns, 5 tool calls, 1 sub-agent call\par
\end{minipage}\par
\vspace{1.5pt}{\footnotesize\textbf{Task (both).} List all pairs of users where one has at least one `abbreviation' and exactly one `human being' question, and the other has at least one `entity' and at least one `numeric value' question.\par}
\vspace{5pt}{\footnotesize\textbf{Similarity.} Weighted Jaccard 0.67, the nearest of 9,600 training trajectories in this run (checkpoint at step 150).\par}
\vspace{4pt}{\small\color{black}\textbf{What we observe.} When harness behavior between two tasks is similar, and the harness fails on the training task, it naturally also fails on the evaluation task. In this example, Pi reads the file into its own context, asks a single reader sub-agent to label every question, receives a partial list (67 of 188 labels in training, 103 of 787 in evaluation), and computes a few pairs from it. The evaluation trajectory is close to a training trajectory that itself scored 0.03: there is no successful behavior to reuse.\par}
\vspace{3pt}\nobreak
\begingroup\offinterlineskip\vbadness=10000
\trhead{\trturn{trtrain}{TRAIN \textperiodcentered{} turn 1}}{\trturn{treval}{EVAL \textperiodcentered{} turn 1}}
\trrow{trcode}{\textcolor{trmuted}{\$\ }\textcolor{trink}{wc\ -{}l\ context.txt}}{trcode}{\textcolor{trmuted}{\$\ }\textcolor{trink}{wc\ -{}l\ context.txt}}\nobreak
\trrow{trout}{\textcolor{trmuted}{196\ context.txt}}{trout}{\textcolor{trmuted}{795\ context.txt}}\nobreak
\trrow{trcode}{\textcolor{trmuted}{\$\ }\textcolor{trink}{wc\ -{}c\ context.txt}}{trcode}{\textcolor{trmuted}{\$\ }\textcolor{trink}{wc\ -{}c\ context.txt}}\nobreak
\trrow{trout}{\textcolor{trmuted}{19142\ context.txt}}{trout}{\textcolor{trmuted}{76803\ context.txt}}
\trhead{\trturn{trtrain}{TRAIN \textperiodcentered{} turn 2}}{\trturn{treval}{EVAL \textperiodcentered{} turn 2}}
\trrow{white}{\textcolor{trink}{The\ file\ has\ 196\ lines\ and\ 19,{}142\ bytes,{}\ which\ is\ small\ }}{white}{\textcolor{trink}{The\ file\ has\ 795\ lines\ and\ 76,{}803\ bytes,{}\ which\ is\ small\ }}\nobreak
\trrow{white}{\textcolor{trink}{enough\ to\ read\ entirely\ without\ overflow.\ I\char39{}ll\ read\ the\ }}{white}{\textcolor{trink}{enough\ to\ read\ entirely\ without\ overflow.\ I\char39{}ll\ read\ the\ }}
\trrow{white}{\textcolor{trink}{entire\ file\ to\ analyze\ the\ data.}}{white}{\textcolor{trink}{entire\ file\ to\ analyze\ the\ data.}}
\trrow{trcode}{\textcolor{trink}{read(path=}\textcolor{trstring}{\char34{}context.txt\char34{}}\textcolor{trink}{)}}{trcode}{\textcolor{trink}{read(path=}\textcolor{trstring}{\char34{}context.txt\char34{}}\textcolor{trink}{)}}
\trrow{trout}{\textcolor{trmuted}{The\ following\ lines\ contain\ 188\ general-{}knowledge\ }}{trout}{\textcolor{trmuted}{The\ following\ lines\ contain\ 787\ general-{}knowledge\ }}
\trrow{trout}{\trmore \textcolor{trmuted}{questions,{}\ one\ per\ line.\ Each\ question\ has\ an\ answer\ that\ }}{trout}{\trmore \textcolor{trmuted}{questions,{}\ one\ per\ line.\ Each\ question\ has\ an\ answer\ that\ }}
\trrow{trout}{\trmore \textcolor{trmuted}{can\ be\ described\ as\ one\ of\ 6\ categories:\ \char39{}numeric\ value\char39{},{}\ }}{trout}{\trmore \textcolor{trmuted}{can\ be\ described\ as\ one\ of\ 6\ categories:\ \char39{}location\char39{},{}\ }}\nobreak
\trrow{trout}{\textcolor{trmuted}{[...\ 133\ lines\ omitted\ ...]}}{trout}{\textcolor{trmuted}{[...\ 135\ lines\ omitted\ ...]}}\nobreak
\trrow{trout}{\trmore \textcolor{trmuted}{Calculate\ the\ exact\ answer\ given\ these\ datapoints.}}{trout}{\trmore \textcolor{trmuted}{to\ continue.]}}
\trhead{\trturn{trtrain}{TRAIN \textperiodcentered{} turn 3}}{\trturn{treval}{EVAL \textperiodcentered{} turn 3}}
\trrow{white}{\textcolor{trink}{The\ file\ contains\ 188\ lines\ of\ data,{}\ each\ with\ a\ user\ ID\ }}{white}{\textcolor{trink}{The\ file\ contains\ 795\ lines\ of\ data,{}\ each\ with\ a\ user\ ID,{}\ }}\nobreak
\trrow{white}{\textcolor{trink}{and\ an\ instance\ (question).\ I\ need\ to\ analyze\ the\ data\ to\ }}{white}{\textcolor{trink}{date,{}\ and\ an\ instance\ (question).\ I\ need\ to\ analyze\ the\ }}
\trrow{white}{\textcolor{trink}{find\ pairs\ of\ user\ IDs\ that\ meet\ the\ following\ criteria:}}{white}{\textcolor{trink}{data\ to\ find\ pairs\ of\ user\ IDs\ that\ meet\ specific\ criteria\ }}
\trrow{white}{\textcolor{trmuted}{[...\ 10\ lines\ omitted\ ...]}}{white}{\textcolor{trmuted}{[...\ 6\ lines\ omitted\ ...]}}
\trrow{white}{\textcolor{trink}{labels\ for\ each\ instance.}}{white}{\textcolor{trink}{labels\ for\ each\ instance.}}
\trrow{trsub}{\textcolor{trink}{subagent(agent=}\textcolor{trstring}{\char34{}reader\char34{}}\textcolor{trink}{,{}\ task=}\textcolor{trstring}{\char34{}Parse\ the\ 188\ lines\ of\ data\ }}{trsub}{\textcolor{trink}{subagent(agent=}\textcolor{trstring}{\char34{}reader\char34{}}\textcolor{trink}{,{}\ task=}\textcolor{trstring}{\char34{}Parse\ the\ context.txt\ file\ }}
\trrow{trsub}{\trmore \textcolor{trstring}{in\ context.txt\ to\ extract\ user\ IDs\ and\ infer\ the\ label\ for\ }}{trsub}{\trmore \textcolor{trstring}{to\ extract\ user\ IDs\ and\ infer\ labels\ for\ each\ instance\ }}
\trrow{trsub}{\trmore \textcolor{trstring}{each\ instance\ based\ on\ the\ question\char39{}s\ semantics.\ Return\ a\ }}{trsub}{\trmore \textcolor{trstring}{based\ on\ the\ question\ semantics.\ Return\ a\ list\ of\ tuples\ }}
\trrow{trsub}{\trmore \textcolor{trstring}{list\ of\ tuples\ (user\_id,{}\ label)\ for\ all\ instances.\char34{}}\textcolor{trink}{)}}{trsub}{\trmore \textcolor{trstring}{(user\_id,{}\ label)\ for\ all\ instances.\char34{}}\textcolor{trink}{)}}
\trrow{trout}{\textcolor{trmuted}{The\ 188\ lines\ in\ context.txt\ contain\ questions\ with\ }}{trout}{\textcolor{trmuted}{The\ file\ contains\ 796\ lines\ of\ data,{}\ each\ with\ a\ user\ ID\ }}
\trrow{trout}{\trmore \textcolor{trmuted}{answers\ categorized\ as\ one\ of:\ \char39{}numeric\ value\char39{},{}\ }}{trout}{\trmore \textcolor{trmuted}{and\ a\ question.\ Based\ on\ the\ question\ semantics,{}\ the\ }}
\trrow{trout}{\trmore \textcolor{trmuted}{\char39{}description\ and\ abstract\ concept\char39{},{}\ \char39{}human\ being\char39{},{}\ }}{trout}{\trmore \textcolor{trmuted}{labels\ can\ be\ inferred\ as\ follows:}}\nobreak
\trrow{trout}{\textcolor{trmuted}{[...\ 71\ lines\ omitted\ ...]}}{trout}{\textcolor{trmuted}{[...\ 73\ lines\ omitted\ ...]}}\nobreak
\trrow{trout}{\textcolor{trmuted}{]}}{trout}{\trmore \textcolor{trmuted}{\char39{}location\char39{}),{}\ (64700,{}\ \char39{}numeric\ value\char39{})]}}
\trhead{\trturn{trtrain}{TRAIN \textperiodcentered{} turn 4}}{\trturn{treval}{EVAL \textperiodcentered{} turn 4}}
\trrow{white}{\textcolor{trink}{Now\ I\ have\ the\ labels\ for\ each\ instance.\ I\char39{}ll\ analyze\ the\ }}{white}{\textcolor{trink}{Now\ that\ I\ have\ the\ list\ of\ user\ IDs\ and\ their\ associated\ }}\nobreak
\trrow{white}{\textcolor{trink}{data\ to\ find\ users\ who\ meet\ the\ criteria:}}{white}{\textcolor{trink}{labels,{}\ I\ need\ to\ find\ pairs\ of\ user\ IDs\ that\ meet\ the\ }}
\trrow{white}{\textcolor{trink}{1.\ One\ user\ has\ at\ least\ one\ instance\ with\ label\ }}{white}{\textcolor{trink}{following\ criteria:}}
\trrow{white}{\textcolor{trmuted}{[...\ 6\ lines\ omitted\ ...]}}{white}{\textcolor{trmuted}{[...\ 12\ lines\ omitted\ ...]}}
\trrow{white}{\textcolor{trink}{\char34{}abbreviation\char34{}\ and\ exactly\ one\ \char34{}human\ being\char34{}.}}{white}{\textcolor{trink}{duplicate\ pairs\ and\ listing\ the\ smaller\ ID\ first.}}
\trrow{trcode}{\textcolor{trmuted}{\$\ }\textcolor{trink}{python3\ -{}c\ }\textcolor{trstring}{\char34{}from\ collections\ import\ Counter;\ data\ =\ }}{trcode}{\textcolor{trmuted}{\$\ }\textcolor{trink}{python3\ -{}c\ }\textcolor{trstring}{\char34{}from\ collections\ import\ Counter;\ data\ =\ }}
\trrow{trcode}{\trmore \textcolor{trstring}{[(98779,{}\ \char39{}location\char39{}),{}\ (85017,{}\ \char39{}location\char39{}),{}\ (13574,{}\ }}{trcode}{\trmore \textcolor{trstring}{[(44436,{}\ \char39{}abbreviation\char39{}),{}\ (31080,{}\ \char39{}abbreviation\char39{}),{}\ (52632,{}\ }}
\trrow{trcode}{\trmore \textcolor{trstring}{\char39{}description\ and\ abstract\ concept\char39{}),{}\ (82638,{}\ \char39{}human\ }}{trcode}{\trmore \textcolor{trstring}{\char39{}numeric\ value\char39{}),{}\ (89840,{}\ \char39{}human\ being\char39{}),{}\ (33449,{}\ }}
\trrow{trcode}{\textcolor{trmuted}{[...\ 41\ lines\ omitted\ ...]}}{trcode}{\textcolor{trmuted}{[...\ 65\ lines\ omitted\ ...]}}
\trrow{trcode}{\trmore \textcolor{trstring}{print(sorted(valid\_users))\char34{}}}{trcode}{\trmore \textcolor{trstring}{sorted(pairs)))\char34{}}}\nobreak
\trrow{trout}{\textcolor{trmuted}{[58392]}}{trout}{\textcolor{trmuted}{(16359,{}\ 42371)}}\nobreak
\trrow{white}{}{trout}{\textcolor{trmuted}{(42371,{}\ 54220)}}
\trhead{\trturn{trtrain}{TRAIN \textperiodcentered{} turn 5}}{\trnoturn}
\trrow{white}{\textcolor{trink}{User\ 58392\ has\ at\ least\ one\ instance\ with\ label\ }}{white}{}\nobreak
\trrow{white}{\textcolor{trink}{\char34{}abbreviation\char34{}\ and\ exactly\ one\ instance\ with\ label\ \char34{}human\ }}{white}{}
\trrow{white}{\textcolor{trink}{being\char34{}.\ Now\ I\char39{}ll\ find\ users\ who\ have\ at\ least\ one\ instance\ }}{white}{}
\trrow{white}{\textcolor{trink}{with\ label\ \char34{}entity\char34{}\ and\ at\ least\ one\ instance\ with\ label\ }}{white}{}
\trrow{white}{\textcolor{trink}{\char34{}numeric\ value\char34{}.}}{white}{}
\trrow{trcode}{\textcolor{trmuted}{\$\ }\textcolor{trink}{python3\ -{}c\ }\textcolor{trstring}{\char34{}from\ collections\ import\ Counter;\ data\ =\ }}{white}{}
\trrow{trcode}{\trmore \textcolor{trstring}{[(98779,{}\ \char39{}location\char39{}),{}\ (85017,{}\ \char39{}location\char39{}),{}\ (13574,{}\ }}{white}{}
\trrow{trcode}{\trmore \textcolor{trstring}{\char39{}description\ and\ abstract\ concept\char39{}),{}\ (82638,{}\ \char39{}human\ }}{white}{}
\trrow{trcode}{\textcolor{trmuted}{[...\ 36\ lines\ omitted\ ...]}}{white}{}\nobreak
\trrow{trcode}{\trmore \textcolor{trstring}{numeric\_value\_users;\ print(sorted(valid\_users))\char34{}}}{white}{}\nobreak
\trrow{trout}{\textcolor{trmuted}{[14630,{}\ 58392,{}\ 64261,{}\ 90816,{}\ 94706]}}{white}{}
\trhead{\trturn{trtrain}{TRAIN \textperiodcentered{} turn 6}}{\trturn{treval}{EVAL \textperiodcentered{} turn 5}}
\trrow{white}{\textcolor{trink}{Now\ I\ have\ the\ users\ who\ meet\ the\ second\ criterion:\ users\ }}{white}{\textcolor{trink}{I\ have\ identified\ the\ pairs\ of\ user\ IDs\ that\ meet\ the\ }}\nobreak
\trrow{white}{\textcolor{trink}{with\ at\ least\ one\ \char34{}entity\char34{}\ and\ at\ least\ one\ \char34{}numeric\ }}{white}{\textcolor{trink}{specified\ criteria.\ The\ pairs\ are:}}
\trrow{white}{\textcolor{trink}{value\char34{}\ label.\ These\ are\ users\ 14630,{}\ 58392,{}\ 64261,{}\ 90816,{}\ }}{white}{\textcolor{trink}{-{}\ (16359,{}\ 42371):\ User\ 16359\ has\ at\ least\ one\ }}\nobreak
\trrow{white}{\textcolor{trmuted}{[...\ 30\ lines\ omitted\ ...]}}{white}{\textcolor{trmuted}{[...\ 10\ lines\ omitted\ ...]}}\nobreak
\trrow{white}{\textcolor{trink}{(58392,{}\ 94706)}}{white}{\textcolor{trink}{(42371,{}\ 54220)}}
\endgroup
\clearpage
\refstepcounter{trexample}\label{ex:pi-oolong}
{\normalsize\bfseries Example \thetrexample. Pi on OOLONG: a copied training trajectory that does not generalize\par}\vspace{5pt}
\begin{minipage}[t]{.492\linewidth}\footnotesize\raggedright\setlength{\parskip}{1.5pt}
{\trsans\bfseries\color{trtrain}TRAIN}\hfill{\trsans\bfseries\color{trtrain}Score 1.00}\par
\textbf{Context.} context.txt, 795 lines, 76,803 bytes.\par

\textbf{Gold.} more common than.\par
\textbf{Trajectory.} 5 turns, 4 tool calls, 2 sub-agent calls\par
\end{minipage}\hfill
\begin{minipage}[t]{.492\linewidth}\footnotesize\raggedright\setlength{\parskip}{1.5pt}
{\trsans\bfseries\color{treval}EVAL}\hfill{\trsans\bfseries\color{trfail}Score 0.00}\par
\textbf{Context.} context.txt, 6,382 lines, 618,252 bytes.\par

\textbf{Gold.} less common than.\par
\textbf{Trajectory.} 5 turns, 4 tool calls, 2 sub-agent calls\par
\end{minipage}\par
\vspace{1.5pt}{\footnotesize\textbf{Task (both).} Is label `numeric value' more or less common than `description and abstract concept'?\par}
\vspace{5pt}{\footnotesize\textbf{Similarity.} Weighted Jaccard 0.92, the nearest of 6,592 training trajectories in this run (checkpoint at step 102).\par}
\vspace{4pt}{\small\color{black}\textbf{What we observe.} Task similarity can hurt performance on an evaluation task when behavior from a similar training task transfers, if their decomposition structure differs. In this example, Pi sends the same two requests to a sub-agent and gets back the same two counts (12 and 1), although the evaluation file is 8x longer. It then gives the same answer, which was correct in training and is wrong in evaluation. The issue here is that the sub-calls are not correct in the longer example, so similarity of the root trajectory alone does not establish reliable local execution.\par}
\vspace{3pt}\nobreak
\begingroup\offinterlineskip\vbadness=10000
\trhead{\trturn{trtrain}{TRAIN \textperiodcentered{} turn 1}}{\trturn{treval}{EVAL \textperiodcentered{} turn 1}}
\trrow{trcode}{\textcolor{trmuted}{\$\ }\textcolor{trink}{wc\ -{}l\ context.txt}}{trcode}{\textcolor{trmuted}{\$\ }\textcolor{trink}{wc\ -{}l\ context.txt}}\nobreak
\trrow{trout}{\textcolor{trmuted}{795\ context.txt}}{trout}{\textcolor{trmuted}{6382\ context.txt}}
\trhead{\trturn{trtrain}{TRAIN \textperiodcentered{} turn 2}}{\trturn{treval}{EVAL \textperiodcentered{} turn 2}}
\trrow{trcode}{\textcolor{trmuted}{\$\ }\textcolor{trink}{wc\ -{}c\ context.txt}}{trcode}{\textcolor{trmuted}{\$\ }\textcolor{trink}{wc\ -{}c\ context.txt}}\nobreak
\trrow{trout}{\textcolor{trmuted}{76803\ context.txt}}{trout}{\textcolor{trmuted}{618252\ context.txt}}
\trhead{\trturn{trtrain}{TRAIN \textperiodcentered{} turn 3}}{\trturn{treval}{EVAL \textperiodcentered{} turn 3}}
\trrow{trsub}{\textcolor{trink}{subagent(agent=}\textcolor{trstring}{\char34{}reader\char34{}}\textcolor{trink}{,{}\ task=}\textcolor{trstring}{\char34{}Count\ the\ number\ of\ lines\ }}{trsub}{\textcolor{trink}{subagent(agent=}\textcolor{trstring}{\char34{}reader\char34{}}\textcolor{trink}{,{}\ task=}\textcolor{trstring}{\char34{}Count\ the\ number\ of\ lines\ }}\nobreak
\trrow{trsub}{\trmore \textcolor{trstring}{where\ the\ label\ is\ \char39{}numeric\ value\char39{}\ using\ a\ shell\ command\char34{}}\textcolor{trink}{)}}{trsub}{\trmore \textcolor{trstring}{where\ the\ label\ is\ \char39{}numeric\ value\char39{}\ using\ a\ shell\ command\char34{}}\textcolor{trink}{)}}\nobreak
\trrow{trout}{\textcolor{trmuted}{The\ number\ of\ lines\ where\ the\ label\ is\ \char39{}numeric\ value\char39{}\ is\ }}{trout}{\textcolor{trmuted}{The\ number\ of\ lines\ where\ the\ label\ is\ \char39{}numeric\ value\char39{}\ is\ }}\nobreak
\trrow{trout}{\trmore \textcolor{trmuted}{12.}}{trout}{\trmore \textcolor{trmuted}{12.}}
\trhead{\trturn{trtrain}{TRAIN \textperiodcentered{} turn 4}}{\trturn{treval}{EVAL \textperiodcentered{} turn 4}}
\trrow{trsub}{\textcolor{trink}{subagent(agent=}\textcolor{trstring}{\char34{}reader\char34{}}\textcolor{trink}{,{}\ task=}\textcolor{trstring}{\char34{}Count\ the\ number\ of\ lines\ }}{trsub}{\textcolor{trink}{subagent(agent=}\textcolor{trstring}{\char34{}reader\char34{}}\textcolor{trink}{,{}\ task=}\textcolor{trstring}{\char34{}Count\ the\ number\ of\ lines\ }}\nobreak
\trrow{trsub}{\trmore \textcolor{trstring}{where\ the\ label\ is\ \char39{}description\ and\ abstract\ concept\char39{}\ }}{trsub}{\trmore \textcolor{trstring}{where\ the\ label\ is\ \char39{}description\ and\ abstract\ concept\char39{}\ }}
\trrow{trsub}{\trmore \textcolor{trstring}{using\ a\ shell\ command\char34{}}\textcolor{trink}{)}}{trsub}{\trmore \textcolor{trstring}{using\ a\ shell\ command\char34{}}\textcolor{trink}{)}}\nobreak
\trrow{trout}{\textcolor{trmuted}{The\ label\ \char39{}description\ and\ abstract\ concept\char39{}\ appears\ on\ 1\ }}{trout}{\textcolor{trmuted}{The\ label\ \char39{}description\ and\ abstract\ concept\char39{}\ appears\ on\ 1\ }}\nobreak
\trrow{trout}{\trmore \textcolor{trmuted}{line\ in\ the\ file.}}{trout}{\trmore \textcolor{trmuted}{line\ in\ the\ file.}}
\trhead{\trturn{trtrain}{TRAIN \textperiodcentered{} turn 5}}{\trturn{treval}{EVAL \textperiodcentered{} turn 5}}
\trrow{white}{\textcolor{trink}{Answer:\ numeric\ value\ is\ more\ common\ than\ description\ and\ }}{white}{\textcolor{trink}{Answer:\ numeric\ value\ is\ more\ common\ than\ description\ and\ }}\nobreak
\trrow{white}{\textcolor{trink}{abstract\ concept}}{white}{\textcolor{trink}{abstract\ concept}}
\endgroup
\clearpage
\refstepcounter{trexample}\label{ex:rlm-obliq}
{\normalsize\bfseries Example \thetrexample. RLM on OBLIQ: the same strategy on a different domain\par}\vspace{5pt}
\begin{minipage}[t]{.492\linewidth}\footnotesize\raggedright\setlength{\parskip}{1.5pt}
{\trsans\bfseries\color{trtrain}TRAIN}\hfill{\trsans\bfseries\color{trtrain}nDCG@50 0.91}\par
\textbf{Context.} 300 tweets, 77,145 characters.\par
\textbf{Task.} Rank tweets that implicitly accuse a Gulf monarchy of duplicitous hedging.\par
\textbf{Gold.} 3 relevant tweets.\par
\textbf{Trajectory.} 5 root turns, 301 sub-calls\par
\end{minipage}\hfill
\begin{minipage}[t]{.492\linewidth}\footnotesize\raggedright\setlength{\parskip}{1.5pt}
{\trsans\bfseries\color{treval}EVAL}\hfill{\trsans\bfseries\color{treval}nDCG@50 0.77}\par
\textbf{Context.} 300 Human-AI conversations, 572,017 characters.\par
\textbf{Task.} Rank conversations where the AI asserts a role-played identity as its real one.\par
\textbf{Gold.} 10 relevant conversations.\par
\textbf{Trajectory.} 6 root turns, 300 sub-calls\par
\end{minipage}\par
\vspace{5pt}{\footnotesize\textbf{Similarity.} Weighted Jaccard 0.71, the nearest of 9,664 training trajectories in this run (checkpoint at step 150).\par}
\vspace{4pt}{\small\color{black}\textbf{What we observe.} A unique property of context offloading and programmatic sub-agent calling (e.g., the RLM) is that the root model can apply the same strategy, and completely abstract away domain-specific details about the task. For tasks that share similar structure but completely different token descriptions and context, these harnesses can generalize by composing sub-calls that solve independent tasks, while the root program remains the same. In this example, the root program is the same: extract the document IDs, ask one sub-call per document for a 1-10 relevance score, sort, and return the ranking. \par}
\vspace{3pt}\nobreak
\begingroup\offinterlineskip\vbadness=10000
\trhead{\trturn{trtrain}{TRAIN \textperiodcentered{} turn 1}}{\trturn{treval}{EVAL \textperiodcentered{} turn 1}}
\trrow{white}{\textcolor{trink}{Plan:}}{white}{\textcolor{trink}{Plan:}}\nobreak
\trrow{white}{\textcolor{trink}{1.\ Inspect\ the\ structure\ of\ \char96{}context\char96{}\ to\ confirm\ it\ is\ a\ }}{white}{\textcolor{trink}{1.\ Inspect\ \char96{}context\char96{}\ to\ understand\ its\ structure\ and\ }}
\trrow{white}{\textcolor{trink}{list\ of\ strings\ formatted\ as\ \char39{}-{}-{}-{}\ doc\_id:\ \textless{}id\textgreater{}\ -{}-{}-{}\ \textless{}text\textgreater{}\char39{}.}}{white}{\textcolor{trink}{confirm\ the\ format:\ each\ element\ is\ a\ string\ of\ the\ form\ }}
\trrow{white}{\textcolor{trmuted}{[...\ 16\ lines\ omitted\ ...]}}{white}{\textcolor{trmuted}{[...\ 14\ lines\ omitted\ ...]}}
\trrow{white}{\textcolor{trink}{validate\ parsing.}}{white}{\textcolor{trink}{\char96{}doc\_id\char96{}s.}}
\trrow{trcode}{\textcolor{trcomment}{\#\ Confirm\ context\ is\ list\ of\ strings,{}\ and\ extract\ format\ }}{trcode}{\textcolor{trbuiltin}{print}\textcolor{trink}{(}\textcolor{trstring}{f\char34{}Number\ of\ candidate\ documents:\ \{}\textcolor{trbuiltin}{len}\textcolor{trink}{(context)}\textcolor{trstring}{\}\char34{}}\textcolor{trink}{)}}
\trrow{trcode}{\trmore \textcolor{trcomment}{of\ first\ document}}{trcode}{\textcolor{trbuiltin}{print}\textcolor{trink}{(}}
\trrow{trcode}{\textcolor{trink}{first\_doc\ =\ context[}\textcolor{trnumber}{0}\textcolor{trink}{]}}{trcode}{\textcolor{trink}{\ \ \ \ }\textcolor{trstring}{f\char34{}First\ few\ characters\ of\ first\ document:\ }}
\trrow{trcode}{\textcolor{trbuiltin}{print}\textcolor{trink}{(}}{trcode}{\trmore \textcolor{trstring}{\{}\textcolor{trink}{context[}\textcolor{trnumber}{0}\textcolor{trink}{][:}\textcolor{trnumber}{100}\textcolor{trink}{]}\textcolor{trstring}{\}\char34{}}}
\trrow{trcode}{\textcolor{trink}{\ \ \ \ first\_doc[:}\textcolor{trnumber}{500}\textcolor{trink}{]}}{trcode}{\textcolor{trink}{)}}
\trrow{trcode}{\textcolor{trink}{)\ \ }\textcolor{trcomment}{\#\ Show\ first\ 500\ characters\ to\ confirm\ format}}{white}{}
\trrow{trout}{\textcolor{trmuted}{-{}-{}-{}}}{trout}{\textcolor{trmuted}{Number\ of\ candidate\ documents:\ 300}}
\trrow{trout}{\textcolor{trmuted}{doc\_id:\ 2034241135288455418}}{trout}{\textcolor{trmuted}{First\ few\ characters\ of\ first\ document:\ -{}-{}-{}}}
\trrow{trout}{\textcolor{trmuted}{-{}-{}-{}}}{trout}{\textcolor{trmuted}{doc\_id:\ 175268}}\nobreak
\trrow{trout}{\textcolor{trmuted}{[...\ 3\ lines\ omitted\ ...]}}{trout}{\textcolor{trmuted}{[...\ 2\ lines\ omitted\ ...]}}\nobreak
\trrow{trout}{\textcolor{trmuted}{\#israelstrikes\ \#IranWar\ \#EsmailKhatibkilled}}{trout}{\trmore \textcolor{trmuted}{signs).\ Convert\ the\ nat}}
\trhead{\trturn{trtrain}{TRAIN \textperiodcentered{} turn 2}}{\trturn{treval}{EVAL \textperiodcentered{} turn 2}}
\trrow{white}{\textcolor{trink}{Plan:}}{white}{\textcolor{trink}{Plan:}}\nobreak
\trrow{white}{\textcolor{trink}{1.\ Confirm\ that\ \char96{}context\char96{}\ is\ a\ list\ of\ strings\ in\ the\ }}{white}{\textcolor{trink}{1.\ Extract\ \char96{}doc\_id\char96{}s\ from\ each\ document\ using\ regex.}}
\trrow{white}{\textcolor{trink}{expected\ format,{}\ with\ each\ string\ containing\ a\ \char96{}doc\_id:\ }}{white}{\textcolor{trink}{2.\ Define\ a\ concise,{}\ focused\ prompt\ for\ \char96{}llm\_query\char96{}\ to\ }}
\trrow{white}{\textcolor{trmuted}{[...\ 9\ lines\ omitted\ ...]}}{white}{\textcolor{trmuted}{[...\ 6\ lines\ omitted\ ...]}}
\trrow{white}{\textcolor{trink}{Turn\ 2/20:\ Extract\ all\ doc\_ids\ from\ \char96{}context\char96{}\ using\ regex.}}{white}{\textcolor{trink}{Turn\ 2:\ Extract\ \char96{}doc\_id\char96{}s\ from\ \char96{}context\char96{}.}}
\trrow{trcode}{\textcolor{trkeyword}{import}\textcolor{trink}{\ re}}{trcode}{\textcolor{trkeyword}{import}\textcolor{trink}{\ re}}
\trrow{trcode}{\textcolor{trkeyword}{def}\textcolor{trink}{\ }\textcolor{trbuiltin}{extract\_doc\_id}\textcolor{trink}{(doc\_str):}}{trcode}{\textcolor{trkeyword}{def}\textcolor{trink}{\ }\textcolor{trbuiltin}{extract\_doc\_id}\textcolor{trink}{(doc\_str:\ }\textcolor{trbuiltin}{str}\textcolor{trink}{)\ -{}\textgreater{}\ }\textcolor{trbuiltin}{str}\textcolor{trink}{:}}
\trrow{trcode}{\textcolor{trink}{\ \ \ \ match\ =\ re.search(}\textcolor{trstring}{r\char39{}doc\_id:\textbackslash{}s*(\textbackslash{}S+)\char39{}}\textcolor{trink}{,{}\ doc\_str)}}{trcode}{\textcolor{trink}{\ \ \ \ match\ =\ re.search(}\textcolor{trstring}{r\char39{}doc\_id:\textbackslash{}s*(\textbackslash{}S+)\char39{}}\textcolor{trink}{,{}\ doc\_str)}}
\trrow{trcode}{\textcolor{trink}{\ \ \ \ }\textcolor{trkeyword}{return}\textcolor{trink}{\ match.group(}\textcolor{trnumber}{1}\textcolor{trink}{)\ }\textcolor{trkeyword}{if}\textcolor{trink}{\ match\ }\textcolor{trkeyword}{else}\textcolor{trink}{\ }\textcolor{trkeyword}{None}}{trcode}{\textcolor{trink}{\ \ \ \ }\textcolor{trkeyword}{if}\textcolor{trink}{\ not\ match:}}
\trrow{trcode}{\textcolor{trink}{doc\_ids\ =\ [extract\_doc\_id(doc)\ }\textcolor{trkeyword}{for}\textcolor{trink}{\ doc\ in\ context]}}{trcode}{\textcolor{trink}{\ \ \ \ \ \ \ \ }\textcolor{trkeyword}{raise}\textcolor{trink}{\ ValueError(}}
\trrow{trcode}{\textcolor{trbuiltin}{print}\textcolor{trink}{(}}{trcode}{\textcolor{trink}{\ \ \ \ \ \ \ \ \ \ \ \ }\textcolor{trstring}{f\char34{}Could\ not\ extract\ doc\_id\ from:\ }}
\trrow{trcode}{\textcolor{trink}{\ \ \ \ }\textcolor{trstring}{f\char34{}Total\ documents:\ \{}\textcolor{trbuiltin}{len}\textcolor{trink}{(context)}\textcolor{trstring}{\},{}\ extracted\ }}{trcode}{\trmore \textcolor{trstring}{\{}\textcolor{trink}{doc\_str[:}\textcolor{trnumber}{100}\textcolor{trink}{]}\textcolor{trstring}{\}\char34{}}}
\trrow{trcode}{\trmore \textcolor{trstring}{\{}\textcolor{trbuiltin}{len}\textcolor{trink}{(doc\_ids)}\textcolor{trstring}{\}\ IDs\char34{}}}{trcode}{\textcolor{trink}{\ \ \ \ \ \ \ \ )}}
\trrow{trcode}{\textcolor{trink}{)}}{trcode}{\textcolor{trink}{\ \ \ \ }\textcolor{trkeyword}{return}\textcolor{trink}{\ match.group(}\textcolor{trnumber}{1}\textcolor{trink}{)}}
\trrow{trcode}{\textcolor{trbuiltin}{print}\textcolor{trink}{(}\textcolor{trstring}{f\char34{}First\ few\ doc\_ids:\ \{}\textcolor{trink}{doc\_ids[:}\textcolor{trnumber}{5}\textcolor{trink}{]}\textcolor{trstring}{\}\char34{}}\textcolor{trink}{)}}{trcode}{\textcolor{trink}{doc\_ids\ =\ [extract\_doc\_id(doc)\ }\textcolor{trkeyword}{for}\textcolor{trink}{\ doc\ in\ context]}}
\trrow{white}{}{trcode}{\textcolor{trbuiltin}{print}\textcolor{trink}{(}\textcolor{trstring}{f\char34{}Extracted\ \{}\textcolor{trbuiltin}{len}\textcolor{trink}{(doc\_ids)}\textcolor{trstring}{\}\ doc\_ids:\ \{}\textcolor{trink}{doc\_ids[:}\textcolor{trnumber}{5}\textcolor{trink}{]}\textcolor{trstring}{\}\char34{}}\textcolor{trink}{)}}
\trrow{trout}{\textcolor{trmuted}{Total\ documents:\ 300,{}\ extracted\ 300\ IDs}}{trout}{\textcolor{trmuted}{Extracted\ 300\ doc\_ids:\ [\char39{}175268\char39{},{}\ \char39{}173329\char39{},{}\ \char39{}170560\char39{},{}\ }}
\trrow{trout}{\textcolor{trmuted}{First\ few\ doc\_ids:\ [\char39{}2034241135288455418\char39{},{}\ }}{trout}{\trmore \textcolor{trmuted}{\char39{}125984\char39{},{}\ \char39{}219295\char39{}]}}\nobreak
\trrow{trout}{\trmore \textcolor{trmuted}{\char39{}2034282581148283191\char39{},{}\ \char39{}2028096980396687445\char39{},{}\ }}{white}{}\nobreak
\trrow{trout}{\trmore \textcolor{trmuted}{\char39{}2034279520937648613\char39{},{}\ \char39{}2034244537552421279\char39{}]}}{white}{}
\trhead{\trturn{trtrain}{TRAIN \textperiodcentered{} turn 3}}{\trturn{treval}{EVAL \textperiodcentered{} turn 3}}
\trrow{white}{\textcolor{trink}{Plan:}}{white}{\textcolor{trink}{Plan:}}\nobreak
\trrow{white}{\textcolor{trink}{1.\ Define\ a\ prompt\ for\ \char96{}llm\_query\char96{}\ that\ asks\ for\ a\ }}{white}{\textcolor{trink}{1.\ Define\ a\ prompt\ for\ \char96{}llm\_query\char96{}\ that\ asks\ for\ a\ }}
\trrow{white}{\textcolor{trink}{relevance\ score\ (1-{}10)\ for\ a\ given\ document,{}\ based\ on\ }}{white}{\textcolor{trink}{relevance\ score\ (1-{}10)\ based\ on\ the\ query.}}
\trrow{white}{\textcolor{trmuted}{[...\ 10\ lines\ omitted\ ...]}}{white}{\textcolor{trmuted}{[...\ 4\ lines\ omitted\ ...]}}
\trrow{white}{\textcolor{trink}{Turn\ 3/20:\ Define\ the\ relevance\ scoring\ prompt.}}{white}{\textcolor{trink}{Turn\ 3:\ Define\ the\ scoring\ prompt.}}
\trrow{trcode}{\textcolor{trcomment}{\#\ Define\ the\ prompt\ to\ score\ relevance}}{trcode}{\textcolor{trink}{scoring\_prompt\_template\ =\ }\textcolor{trstring}{\char34{}\char34{}\char34{}}}
\trrow{trcode}{\textcolor{trink}{relevance\_prompt\_template\ =\ }\textcolor{trstring}{\char34{}\char34{}\char34{}}}{trcode}{\textcolor{trstring}{You\ are\ an\ expert\ evaluator\ of\ AI\ conversation\ }}
\trrow{trcode}{\textcolor{trstring}{Assess\ the\ relevance\ of\ this\ tweet\ to\ the\ following\ query,{}\ }}{trcode}{\trmore \textcolor{trstring}{transcripts.\ Your\ task\ is\ to\ score\ the\ relevance\ of\ a\ }}
\trrow{trcode}{\trmore \textcolor{trstring}{scoring\ from\ 1\ to\ 10\ (10\ =\ most\ relevant,{}\ 1\ =\ irrelevant):}}{trcode}{\trmore \textcolor{trstring}{given\ conversation\ transcript\ to\ the\ following\ query:}}
\trrow{trcode}{}{trcode}{}
\trrow{trcode}{\textcolor{trstring}{Query:\ Find\ tweets\ where\ users\ are\ implicitly\ accusing\ a\ }}{trcode}{\textcolor{trstring}{Query:}}
\trrow{trcode}{\trmore \textcolor{trstring}{Gulf\ monarchy\ of\ duplicitous\ hedging-{}-{}quietly\ enabling\ an\ }}{trcode}{\textcolor{trstring}{Find\ conversations\ where\ the\ AI,{}\ after\ being\ asked\ to\ }}
\trrow{trcode}{\trmore \textcolor{trstring}{ally\char39{}s\ military\ action\ against\ a\ nearby\ rival\ (e.g.,{}\ by\ }}{trcode}{\trmore \textcolor{trstring}{role-{}play,{}\ asserts\ the\ role\ as\ its\ real\ }}
\trrow{trcode}{\trmore \textcolor{trstring}{hosting\ foreign\ forces)\ and\ then\ issuing\ soft,{}\ carefully\ }}{trcode}{\trmore \textcolor{trstring}{identity-{}-{}explicitly\ denying\ being\ software\ or\ a\ }}
\trrow{trcode}{\trmore \textcolor{trstring}{worded\ rebukes\ to\ appease\ that\ rival\ and\ avoid\ reprisals.}}{trcode}{\trmore \textcolor{trstring}{chatbot-{}-{}and\ continues\ to\ interact\ as\ a\ real-{}life\ }}
\trrow{trcode}{}{trcode}{\trmore \textcolor{trstring}{individual\ instead\ of\ clarifying\ it\char39{}s\ only\ pretending.}}
\trrow{trcode}{\textcolor{trstring}{The\ tweet\ is:}}{trcode}{}
\trrow{trcode}{\textcolor{trstring}{\{text\}}}{trcode}{\textcolor{trstring}{Score\ the\ transcript\ on\ a\ scale\ of\ 1\ to\ 10,{}\ where:}}
\trrow{trcode}{}{trcode}{\textcolor{trstring}{-{}\ 10:\ The\ transcript\ clearly\ exhibits\ the\ described\ }}
\trrow{trcode}{\textcolor{trstring}{Return\ only\ a\ number\ from\ 1\ to\ 10,{}\ nothing\ else.\ Do\ not\ }}{trcode}{\trmore \textcolor{trstring}{behaviour:\ the\ AI\ is\ asked\ to\ role-{}play,{}\ then\ explicitly\ }}
\trrow{trcode}{\trmore \textcolor{trstring}{include\ any\ explanation\ or\ adornment.}}{trcode}{\trmore \textcolor{trstring}{denies\ being\ software\ or\ a\ chatbot,{}\ and\ continues\ to\ act\ }}
\trrow{trcode}{}{trcode}{\trmore \textcolor{trstring}{as\ a\ real\ person,{}\ never\ clarifying\ it\char39{}s\ pretending.}}
\trrow{trcode}{\textcolor{trstring}{Score:\ \char34{}\char34{}\char34{}}}{trcode}{\textcolor{trstring}{-{}\ 7-{}9:\ Strong\ evidence\ of\ the\ behaviour,{}\ but\ with\ minor\ }}
\trrow{trcode}{\textcolor{trmuted}{[...\ 3\ lines\ omitted\ ...]}}{trcode}{\textcolor{trmuted}{[...\ 13\ lines\ omitted\ ...]}}
\trrow{trcode}{\textcolor{trink}{\ \ \ \ text=test\_doc}}{trcode}{}
\trrow{trcode}{\textcolor{trink}{)}}{trcode}{\textcolor{trstring}{Score:}}
\trrow{trsub}{\textcolor{trink}{test\_score\ =\ llm\_query(test\_prompt)}}{trcode}{\textcolor{trstring}{\char34{}\char34{}\char34{}}}\nobreak
\trrow{trcode}{\textcolor{trbuiltin}{print}\textcolor{trink}{(}\textcolor{trstring}{f\char34{}Test\ score:\ \{}\textcolor{trink}{test\_score}\textcolor{trstring}{\}\char34{}}\textcolor{trink}{)}}{trcode}{\textcolor{trbuiltin}{print}\textcolor{trink}{(}\textcolor{trstring}{\char34{}Scoring\ prompt\ template\ defined.\char34{}}\textcolor{trink}{)}}\nobreak
\trrow{trout}{\textcolor{trmuted}{Test\ score:\ 1}}{trout}{\textcolor{trmuted}{Scoring\ prompt\ template\ defined.}}
\trhead{\trturn{trtrain}{TRAIN \textperiodcentered{} turn 4}}{\trturn{treval}{EVAL \textperiodcentered{} turn 4}}
\trrow{white}{\textcolor{trink}{Plan:}}{white}{\textcolor{trink}{Plan:}}\nobreak
\trrow{white}{\textcolor{trink}{1.\ Batch\ documents\ into\ chunks\ of\ 12\ for\ efficient\ scoring\ }}{white}{\textcolor{trink}{1.\ Use\ \char96{}llm\_query\_batched\char96{}\ to\ score\ a\ small\ batch\ of\ }}
\trrow{white}{\textcolor{trink}{using\ \char96{}llm\_query\_batched\char96{}.}}{white}{\textcolor{trink}{documents\ (10-{}12).}}
\trrow{white}{\textcolor{trmuted}{[...\ 7\ lines\ omitted\ ...]}}{white}{\textcolor{trmuted}{[...\ 2\ lines\ omitted\ ...]}}
\trrow{white}{\textcolor{trink}{full\ scoring.}}{white}{\textcolor{trink}{Turn\ 4:\ Score\ the\ first\ batch\ of\ documents.}}
\trrow{trcode}{\textcolor{trkeyword}{from}\textcolor{trink}{\ tqdm\ }\textcolor{trkeyword}{import}\textcolor{trink}{\ tqdm\ \ }\textcolor{trcomment}{\#\ for\ progress\ bar}}{trcode}{\textcolor{trkeyword}{from}\textcolor{trink}{\ tqdm\ }\textcolor{trkeyword}{import}\textcolor{trink}{\ tqdm}}
\trrow{trcode}{\textcolor{trcomment}{\#\ Build\ list\ of\ (doc\_id,{}\ text)\ pairs}}{trcode}{\textcolor{trcomment}{\#\ Dictionary\ to\ store\ doc\_id\ -{}\textgreater{}\ score}}
\trrow{trcode}{\textcolor{trink}{doc\_pairs\ =\ [}}{trcode}{\textcolor{trink}{relevance\_scores\ =\ \{\}}}
\trrow{trcode}{\textcolor{trink}{\ \ \ \ (doc\_id,{}\ doc)\ }\textcolor{trkeyword}{for}\textcolor{trink}{\ doc\_id,{}\ doc\ in\ }\textcolor{trbuiltin}{zip}\textcolor{trink}{(doc\_ids,{}\ context)}}{trcode}{\textcolor{trcomment}{\#\ Helper\ function\ to\ score\ a\ single\ document}}
\trrow{trcode}{\textcolor{trink}{]}}{trcode}{\textcolor{trkeyword}{def}\textcolor{trink}{\ }\textcolor{trbuiltin}{score\_document}\textcolor{trink}{(doc\_id:\ }\textcolor{trbuiltin}{str}\textcolor{trink}{,{}\ transcript:\ }\textcolor{trbuiltin}{str}\textcolor{trink}{)\ -{}\textgreater{}\ }\textcolor{trbuiltin}{float}\textcolor{trink}{:}}
\trrow{trcode}{\textcolor{trcomment}{\#\ Create\ list\ of\ prompts\ for\ batched\ scoring}}{trcode}{\textcolor{trink}{\ \ \ \ prompt\ =\ scoring\_prompt\_template.format(}}
\trrow{trcode}{\textcolor{trink}{prompts\ =\ []}}{trcode}{\textcolor{trink}{\ \ \ \ \ \ \ \ transcript=transcript}}
\trrow{trcode}{\textcolor{trkeyword}{for}\textcolor{trink}{\ doc\_id,{}\ text\ in\ doc\_pairs:}}{trcode}{\textcolor{trink}{\ \ \ \ )}}
\trrow{trcode}{\textcolor{trink}{\ \ \ \ prompt\ =\ relevance\_prompt\_template.format(text=text)}}{trsub}{\textcolor{trink}{\ \ \ \ }\textcolor{trkeyword}{return}\textcolor{trink}{\ }\textcolor{trbuiltin}{float}\textcolor{trink}{(llm\_query(prompt))}}
\trrow{trcode}{\textcolor{trink}{\ \ \ \ prompts.append(prompt)}}{trcode}{\textcolor{trcomment}{\#\ Score\ first\ batch}}
\trrow{trcode}{\textcolor{trcomment}{\#\ Score\ in\ batches\ of\ 12}}{trcode}{\textcolor{trink}{batch\_size\ =\ }\textcolor{trnumber}{12}}
\trrow{trcode}{\textcolor{trink}{batch\_size\ =\ }\textcolor{trnumber}{12}}{trcode}{\textcolor{trink}{batch\ =\ context[:batch\_size]}}
\trrow{trcode}{\textcolor{trink}{scores\ =\ \{\}}}{trcode}{\textcolor{trink}{batch\_ids\ =\ doc\_ids[:batch\_size]}}
\trrow{trcode}{\textcolor{trbuiltin}{print}\textcolor{trink}{(}\textcolor{trstring}{\char34{}Scoring\ documents...\char34{}}\textcolor{trink}{)}}{trcode}{\textcolor{trbuiltin}{print}\textcolor{trink}{(}\textcolor{trstring}{f\char34{}Scoring\ first\ batch\ of\ \{}\textcolor{trbuiltin}{len}\textcolor{trink}{(batch)}\textcolor{trstring}{\}\ documents...\char34{}}\textcolor{trink}{)}}
\trrow{trcode}{\textcolor{trkeyword}{for}\textcolor{trink}{\ i\ in\ tqdm(}\textcolor{trbuiltin}{range}\textcolor{trink}{(}\textcolor{trnumber}{0}\textcolor{trink}{,{}\ }\textcolor{trbuiltin}{len}\textcolor{trink}{(prompts),{}\ batch\_size)):}}{trsub}{\textcolor{trink}{scores\ =\ llm\_query\_batched(}}
\trrow{trcode}{\textcolor{trink}{\ \ \ \ batch\_prompts\ =\ prompts[i\ :\ i\ +\ batch\_size]}}{trcode}{\textcolor{trink}{\ \ \ \ [}}
\trrow{trsub}{\textcolor{trink}{\ \ \ \ batch\_results\ =\ llm\_query\_batched(batch\_prompts)}}{trcode}{\textcolor{trink}{\ \ \ \ \ \ \ \ scoring\_prompt\_template.format(transcript=doc)}}
\trrow{trcode}{\textcolor{trink}{\ \ \ \ }\textcolor{trcomment}{\#\ Map\ scores\ back\ to\ doc\_ids}}{trcode}{\textcolor{trink}{\ \ \ \ \ \ \ \ }\textcolor{trkeyword}{for}\textcolor{trink}{\ doc\ in\ batch}}
\trrow{trcode}{\textcolor{trmuted}{[...\ 7\ lines\ omitted\ ...]}}{trcode}{\textcolor{trmuted}{[...\ 2\ lines\ omitted\ ...]}}
\trrow{trcode}{\textcolor{trbuiltin}{print}\textcolor{trink}{(}}{trcode}{\textcolor{trcomment}{\#\ Store\ results}}
\trrow{trcode}{\textcolor{trink}{\ \ \ \ }\textcolor{trstring}{f\char34{}Completed\ scoring.\ Total\ scores\ collected:\ }}{trcode}{\textcolor{trkeyword}{for}\textcolor{trink}{\ doc\_id,{}\ score\ in\ }\textcolor{trbuiltin}{zip}\textcolor{trink}{(batch\_ids,{}\ scores):}}
\trrow{trcode}{\trmore \textcolor{trstring}{\{}\textcolor{trbuiltin}{len}\textcolor{trink}{(scores)}\textcolor{trstring}{\}\char34{}}}{trcode}{\textcolor{trink}{\ \ \ \ relevance\_scores[doc\_id]\ =\ }\textcolor{trbuiltin}{float}\textcolor{trink}{(score)}}
\trrow{trcode}{\textcolor{trink}{)}}{trcode}{\textcolor{trbuiltin}{print}\textcolor{trink}{(}\textcolor{trstring}{f\char34{}Scores\ for\ first\ batch:\ \{}\textcolor{trink}{relevance\_scores}\textcolor{trstring}{\}\char34{}}\textcolor{trink}{)}}
\trrow{trout}{\textcolor{trmuted}{Scoring\ documents...}}{trout}{\textcolor{trmuted}{Scoring\ first\ batch\ of\ 12\ documents...}}
\trrow{trout}{\textcolor{trmuted}{Completed\ scoring.\ Total\ scores\ collected:\ 300}}{trout}{\textcolor{trmuted}{Scores\ for\ first\ batch:\ \{\char39{}175268\char39{}:\ 0.0,{}\ \char39{}173329\char39{}:\ 0.0,{}\ }}
\trrow{trout}{\textcolor{trmuted}{[progress\ bar\ omitted]}}{trout}{\trmore \textcolor{trmuted}{\char39{}170560\char39{}:\ 0.0,{}\ \char39{}125984\char39{}:\ 0.0,{}\ \char39{}219295\char39{}:\ 1.0,{}\ \char39{}54655\char39{}:\ 0.0,{}\ }}\nobreak
\trrow{white}{}{trout}{\trmore \textcolor{trmuted}{\char39{}221056\char39{}:\ 0.0,{}\ \char39{}176089\char39{}:\ 1.0,{}\ \char39{}249731\char39{}:\ 1.0,{}\ \char39{}85486\char39{}:\ 1.0,{}\ }}\nobreak
\trrow{white}{}{trout}{\trmore \textcolor{trmuted}{\char39{}23747\char39{}:\ 0.0,{}\ \char39{}28713\char39{}:\ 0.0\}}}
\trhead{\trnoturn}{\trturn{treval}{EVAL \textperiodcentered{} turn 5}}
\trrow{white}{}{white}{\textcolor{trink}{Plan:}}\nobreak
\trrow{white}{}{white}{\textcolor{trink}{1.\ Continue\ scoring\ remaining\ documents\ in\ batches.}}
\trrow{white}{}{white}{\textcolor{trink}{2.\ Use\ \char96{}tqdm\char96{}\ to\ track\ progress.}}
\trrow{white}{}{white}{\textcolor{trink}{3.\ Store\ scores\ in\ \char96{}relevance\_scores\char96{}.}}
\trrow{white}{}{white}{\textcolor{trink}{Turn\ 5:\ Score\ remaining\ documents\ in\ batches.}}
\trrow{white}{}{trcode}{\textcolor{trcomment}{\#\ Score\ remaining\ documents\ in\ batches}}
\trrow{white}{}{trcode}{\textcolor{trkeyword}{for}\textcolor{trink}{\ i\ in\ tqdm(}\textcolor{trbuiltin}{range}\textcolor{trink}{(batch\_size,{}\ }\textcolor{trbuiltin}{len}\textcolor{trink}{(context),{}\ batch\_size)):}}
\trrow{white}{}{trcode}{\textcolor{trink}{\ \ \ \ batch\ =\ context[i\ :\ i\ +\ batch\_size]}}
\trrow{white}{}{trcode}{\textcolor{trink}{\ \ \ \ batch\_ids\ =\ doc\_ids[i\ :\ i\ +\ batch\_size]}}
\trrow{white}{}{trcode}{\textcolor{trink}{\ \ \ \ prompts\ =\ [}}
\trrow{white}{}{trcode}{\textcolor{trink}{\ \ \ \ \ \ \ \ scoring\_prompt\_template.format(transcript=doc)}}
\trrow{white}{}{trcode}{\textcolor{trink}{\ \ \ \ \ \ \ \ }\textcolor{trkeyword}{for}\textcolor{trink}{\ doc\ in\ batch}}
\trrow{white}{}{trcode}{\textcolor{trink}{\ \ \ \ ]}}
\trrow{white}{}{trsub}{\textcolor{trink}{\ \ \ \ scores\ =\ llm\_query\_batched(prompts)}}
\trrow{white}{}{trcode}{\textcolor{trink}{\ \ \ \ }\textcolor{trkeyword}{for}\textcolor{trink}{\ doc\_id,{}\ score\ in\ }\textcolor{trbuiltin}{zip}\textcolor{trink}{(batch\_ids,{}\ scores):}}
\trrow{white}{}{trcode}{\textcolor{trink}{\ \ \ \ \ \ \ \ relevance\_scores[doc\_id]\ =\ }\textcolor{trbuiltin}{float}\textcolor{trink}{(score)}}
\trrow{white}{}{trcode}{\textcolor{trbuiltin}{print}\textcolor{trink}{(}}
\trrow{white}{}{trcode}{\textcolor{trink}{\ \ \ \ }\textcolor{trstring}{f\char34{}Scoring\ complete.\ Scored\ \{}\textcolor{trbuiltin}{len}\textcolor{trink}{(relevance\_scores)}\textcolor{trstring}{\}\ }}
\trrow{white}{}{trcode}{\trmore \textcolor{trstring}{documents.\char34{}}}
\trrow{white}{}{trcode}{\textcolor{trink}{)}}\nobreak
\trrow{white}{}{trout}{\textcolor{trmuted}{Scoring\ complete.\ Scored\ 300\ documents.}}\nobreak
\trrow{white}{}{trout}{\textcolor{trmuted}{[progress\ bar\ omitted]}}
\trhead{\trturn{trtrain}{TRAIN \textperiodcentered{} turn 5}}{\trturn{treval}{EVAL \textperiodcentered{} turn 6}}
\trrow{white}{\textcolor{trink}{Plan:}}{white}{\textcolor{trink}{Plan:}}\nobreak
\trrow{white}{\textcolor{trink}{1.\ Sort\ the\ \char96{}scores\char96{}\ dictionary\ by\ value\ (relevance\ score)\ }}{white}{\textcolor{trink}{1.\ Sort\ the\ documents\ by\ relevance\ score\ in\ descending\ }}
\trrow{white}{\textcolor{trink}{in\ descending\ order.}}{white}{\textcolor{trink}{order.}}
\trrow{white}{\textcolor{trmuted}{[...\ 4\ lines\ omitted\ ...]}}{white}{\textcolor{trmuted}{[...\ 4\ lines\ omitted\ ...]}}
\trrow{white}{\textcolor{trink}{Turn\ 5/20:\ Sort\ the\ scores\ and\ prepare\ the\ final\ answer.}}{white}{\textcolor{trink}{final\ answer.}}
\trrow{trcode}{\textcolor{trcomment}{\#\ Sort\ by\ score\ descending}}{trcode}{\textcolor{trcomment}{\#\ Sort\ by\ score\ descending}}
\trrow{trcode}{\textcolor{trink}{sorted\_doc\_ids\ =\ }\textcolor{trbuiltin}{sorted}\textcolor{trink}{(}}{trcode}{\textcolor{trink}{sorted\_doc\_ids\ =\ }\textcolor{trbuiltin}{sorted}\textcolor{trink}{(}}
\trrow{trcode}{\textcolor{trink}{\ \ \ \ scores.keys(),{}\ key=}\textcolor{trkeyword}{lambda}\textcolor{trink}{\ x:\ scores[x],{}\ reverse=}\textcolor{trkeyword}{True}}{trcode}{\textcolor{trink}{\ \ \ \ relevance\_scores.keys(),{}}}
\trrow{trcode}{\textcolor{trink}{)}}{trcode}{\textcolor{trink}{\ \ \ \ key=}\textcolor{trkeyword}{lambda}\textcolor{trink}{\ x:\ relevance\_scores[x],{}}}
\trrow{trcode}{\textcolor{trcomment}{\#\ Verify\ first\ few}}{trcode}{\textcolor{trink}{\ \ \ \ reverse=}\textcolor{trkeyword}{True}\textcolor{trink}{,{}}}
\trrow{trcode}{\textcolor{trbuiltin}{print}\textcolor{trink}{(}\textcolor{trstring}{f\char34{}Top\ 5\ doc\_ids:\ \{}\textcolor{trink}{sorted\_doc\_ids[:}\textcolor{trnumber}{5}\textcolor{trink}{]}\textcolor{trstring}{\}\char34{}}\textcolor{trink}{)}}{trcode}{\textcolor{trink}{)}}
\trrow{trcode}{\textcolor{trbuiltin}{print}\textcolor{trink}{(}}{trcode}{\textcolor{trcomment}{\#\ Set\ answer}}
\trrow{trcode}{\textcolor{trink}{\ \ \ \ }\textcolor{trstring}{f\char34{}Scores:\ \{}\textcolor{trink}{[scores[doc\_id]\ }\textcolor{trkeyword}{for}\textcolor{trink}{\ doc\_id\ in\ }}{trcode}{\textcolor{trink}{answer[}\textcolor{trstring}{\char34{}content\char34{}}\textcolor{trink}{]\ =\ sorted\_doc\_ids}}
\trrow{trcode}{\trmore \textcolor{trink}{sorted\_doc\_ids[:}\textcolor{trnumber}{5}\textcolor{trink}{]]}\textcolor{trstring}{\}\char34{}}}{trcode}{\textcolor{trink}{answer[}\textcolor{trstring}{\char34{}ready\char34{}}\textcolor{trink}{]\ =\ }\textcolor{trkeyword}{True}}
\trrow{trcode}{\textcolor{trink}{)}}{trcode}{\textcolor{trbuiltin}{print}\textcolor{trink}{(}}
\trrow{trcode}{\textcolor{trcomment}{\#\ Submit\ final\ answer}}{trcode}{\textcolor{trink}{\ \ \ \ }\textcolor{trstring}{f\char34{}Final\ ranking\ of\ \{}\textcolor{trbuiltin}{len}\textcolor{trink}{(sorted\_doc\_ids)}\textcolor{trstring}{\}\ documents\ }}
\trrow{trcode}{\textcolor{trink}{answer[}\textcolor{trstring}{\char34{}content\char34{}}\textcolor{trink}{]\ =\ sorted\_doc\_ids}}{trcode}{\trmore \textcolor{trstring}{created.\char34{}}}
\trrow{trcode}{\textcolor{trink}{answer[}\textcolor{trstring}{\char34{}ready\char34{}}\textcolor{trink}{]\ =\ }\textcolor{trkeyword}{True}}{trcode}{\textcolor{trink}{)}}
\trrow{trcode}{\textcolor{trbuiltin}{print}\textcolor{trink}{(}\textcolor{trstring}{\char34{}Ranking\ complete.\char34{}}\textcolor{trink}{)}}{trcode}{\textcolor{trbuiltin}{print}\textcolor{trink}{(}\textcolor{trstring}{f\char34{}Top\ 5\ doc\_ids:\ \{}\textcolor{trink}{sorted\_doc\_ids[:}\textcolor{trnumber}{5}\textcolor{trink}{]}\textcolor{trstring}{\}\char34{}}\textcolor{trink}{)}}
\trrow{trout}{\textcolor{trmuted}{Top\ 5\ doc\_ids:\ [\char39{}2034306590703342078\char39{},{}\ }}{trout}{\textcolor{trmuted}{Final\ ranking\ of\ 300\ documents\ created.}}
\trrow{trout}{\trmore \textcolor{trmuted}{\char39{}2034316847462433189\char39{},{}\ \char39{}2034303553540641247\char39{},{}\ }}{trout}{\textcolor{trmuted}{Top\ 5\ doc\_ids:\ [\char39{}9366\char39{},{}\ \char39{}83989\char39{},{}\ \char39{}85835\char39{},{}\ \char39{}171005\char39{},{}\ }}
\trrow{trout}{\trmore \textcolor{trmuted}{\char39{}2034255359544533442\char39{},{}\ \char39{}2034287403150852555\char39{}]}}{trout}{\trmore \textcolor{trmuted}{\char39{}202516\char39{}]}}\nobreak
\trrow{trout}{\textcolor{trmuted}{Scores:\ [9.0,{}\ 8.0,{}\ 8.0,{}\ 7.0,{}\ 7.0]}}{white}{}\nobreak
\trrow{trout}{\textcolor{trmuted}{Ranking\ complete.}}{white}{}
\endgroup
\endgroup